\documentclass[10pt,twocolumn,letterpaper]{article}

\usepackage{graphicx}
\usepackage{amsmath}
\usepackage{amssymb}
\usepackage{booktabs}

\usepackage[toc,page,header]{appendix}
\usepackage{minitoc}
\usepackage{textcomp}
\usepackage{times}
\usepackage{epsfig}
\usepackage{graphicx}
\usepackage{amsmath}
\usepackage{amssymb}
\usepackage{algorithm}
\usepackage{mathtools}
\usepackage{booktabs}
\usepackage{multirow}
\usepackage{verbatim}
\usepackage{color}
\usepackage{float}
\usepackage{enumitem}
\usepackage{booktabs}
\usepackage{pifont}%
\usepackage{soul}
\usepackage{nicefrac} 
\usepackage{microtype}      %
\usepackage{multirow}
\usepackage{verbatim}
\usepackage{float}
\usepackage{dsfont}
\usepackage{enumitem}
\usepackage{algpseudocode}
\usepackage{booktabs}
\usepackage{tabulary,multirow,overpic}
\usepackage[dvipsnames,table,xcdraw]{xcolor}
\usepackage{bbm}
\usepackage{bm}
\usepackage{graphicx}
\usepackage{subfigure}
\usepackage{placeins}
\usepackage{cite}

\newcommand{\ourparagraph}[1]{\vspace{1.5mm}\noindent\textbf{#1}}

\newcommand{\FEB}[1]{{\color{black}{}#1}}
\newcommand{\srt}[1]{{\color{black}{}#1}} %
\newcommand{\KGnote}[1]{{\color{red}{}#1}}

\newcommand{\cc}[1]{{\color{black}{}#1}} %
\newcommand{\numpeople}{931 } %
\newcommand{\numhours}{3,670 }
\newcommand{\numnarrationsentences}{3.85M }
\newcommand{\numlocations}{74 } %
\newcommand{\numcountries}{9 }

\newcommand{\numcountriesfull}{9 }
\newcommand{\numcontinentsfull}{5 }
\newcommand{\numlabs}{14 }

\newif\ifnotes
\notestrue %

\usepackage{etoolbox}

\definecolor{citecolor}{RGB}{34,139,34}
\usepackage[pagebackref=true,breaklinks=true,letterpaper=true,colorlinks,
citecolor=citecolor,bookmarks=false]{hyperref}

\newtoggle{arxiv}
\toggletrue{arxiv} %

\newtoggle{supp}
\togglefalse{supp}

\newcommand{\bd}[1]{\textbf{#1}}
\newcommand{\app}{\raise.17ex\hbox{$\scriptstyle\sim$}}

\newcolumntype{x}[1]{>{\centering\arraybackslash}p{#1pt}}

\newlength\savewidth

\makeatletter\renewcommand\paragraph{\@startsection{paragraph}{4}{\z@}
	{.5em \@plus1ex \@minus.2ex}{-.5em}{\normalfont\normalsize\bfseries}}\makeatother

\newcommand{\TNnote}[1]{
\ifnotes
{\color{RedOrange}{\bf TN: }#1}
\fi
}

\newcommand{\SR}[1]{
\ifnotes
{\color{black} #1}
\fi
}

\usepackage{stmaryrd}

\DeclareMathOperator*{\argmax}{arg\,max}

\usepackage[capitalize]{cleveref}
\crefname{section}{Sec.}{Secs.}
\Crefname{section}{Section}{Sections}
\Crefname{table}{Table}{Tables}
\crefname{table}{Tab.}{Tabs.}

\def\cvprPaperID{7484} %
\def\confName{CVPR}
\def\confYear{2022}

\usepackage[pagenumbers]{cvpr}

\begin{document}

\title{Ego4D: Around the World in 3,000 Hours of Egocentric Video}

\setcounter{tocdepth}{3}

\renewcommand \thepart{}
\renewcommand \partname{}

\author{Kristen Grauman$^{1,2}$,
Andrew Westbury$^1$,
Eugene Byrne$^{\ast 1}$,
Zachary Chavis$^{\ast 3}$,
Antonino Furnari$^{\ast 4}$,\\
Rohit Girdhar$^{\ast 1}$,
Jackson Hamburger$^{\ast 1}$,
Hao Jiang$^{\ast 5}$, 
Miao Liu$^{\ast 6}$,
Xingyu Liu$^{\ast 7}$,
Miguel Martin$^{\ast 1}$,\\
Tushar Nagarajan$^{\ast 1,2}$,
Ilija Radosavovic$^{\ast 8}$,
Santhosh Kumar Ramakrishnan$^{\ast 1,2}$,
Fiona Ryan$^{\ast 6}$,\\
Jayant Sharma$^{\ast 3}$,
Michael Wray$^{\ast 9}$,
Mengmeng Xu$^{\ast 10}$, 
 Eric Zhongcong Xu$^{\ast 11}$,
 Chen Zhao$^{\ast 10}$,\\
Siddhant Bansal$^{17}$,
Dhruv Batra$^1$,
Vincent Cartillier$^{1,6}$, 
Sean Crane$^7$, 
Tien Do$^3$,
Morrie Doulaty$^{13}$,\\
Akshay Erapalli$^{13}$,
Christoph Feichtenhofer$^1$,
Adriano Fragomeni$^9$, 
Qichen Fu$^7$, \\
Abrham Gebreselasie$^{12}$, 
Cristina Gonz\'alez$^{14}$,
James Hillis$^5$,
Xuhua Huang$^7$,
Yifei Huang$^{15}$,\\
Wenqi Jia$^6$,
Weslie Khoo$^{16}$,
J\'{a}chym Kol\'{a}\v{r}$^{13}$,
Satwik Kottur$^{13}$, 
Anurag Kumar$^5$,
Federico Landini$^{13}$,\\
Chao Li$^5$,
Yanghao Li$^1$,
Zhenqiang Li$^{15}$,
Karttikeya Mangalam$^{1,8}$,
Raghava Modhugu$^{17}$,\\
Jonathan Munro$^9$, 
Tullie Murrell$^1$,
Takumi Nishiyasu$^{15}$,
Will Price$^9$,
Paola Ruiz Puentes$^{14}$,\\
Merey Ramazanova$^{10}$,
Leda Sari$^5$,
Kiran Somasundaram$^5$,
Audrey Southerland$^6$,
Yusuke Sugano$^{15}$,\\
Ruijie Tao$^{11}$,
Minh Vo$^5$,
Yuchen Wang$^{16}$,
Xindi Wu$^7$,
Takuma Yagi$^{15}$,
Ziwei Zhao$^{16}$,
Yunyi Zhu$^{11}$,\\
Pablo Arbel\'aez$^{\dagger 14}$,
David Crandall$^{\dagger 16}$, 
Dima Damen$^{\dagger 9}$,
Giovanni Maria Farinella$^{\dagger 4}$,\\
Christian Fuegen$^{\dagger 13}$, 
Bernard Ghanem$^{\dagger 10}$,
Vamsi Krishna Ithapu$^{\dagger 5}$,
C. V. Jawahar$^{\dagger 17}$,
Hanbyul Joo$^{\dagger 1}$, \\
Kris Kitani$^{\dagger 7}$,
Haizhou Li$^{\dagger 11}$, 
Richard Newcombe$^{\dagger 5}$,
Aude Oliva$^{\dagger 18}$,
Hyun Soo Park$^{\dagger 3}$, \\
James M. Rehg$^{\dagger 6}$, 
Yoichi Sato$^{\dagger 15}$, 
Jianbo Shi$^{\dagger 19}$, 
Mike Zheng Shou$^{\dagger 11}$,
Antonio Torralba$^{\dagger 18}$,\\
Lorenzo Torresani$^{\dagger 1,20}$,
Mingfei Yan$^{\dagger 5}$,
Jitendra Malik$^{1,8}$\\\\
\normalsize $^1$Facebook AI Research (FAIR), 
$^2$University of Texas at Austin,
$^3$University of Minnesota,
$^4$University of Catania,\\
\normalsize $^5$Facebook Reality Labs,
$^6$Georgia Tech,
$^7$Carnegie Mellon University,
$^8$UC Berkeley,
$^9$University of Bristol,\\
\normalsize $^{10}$King Abdullah University of Science and Technology,
$^{11}$National University of Singapore,\\
\normalsize $^{12}$Carnegie Mellon University Africa,
$^{13}$Facebook,
$^{14}$Universidad de los Andes,
$^{15}$University of Tokyo,
$^{16}$Indiana University,\\
\normalsize $^{17}$International Institute
of Information Technology, Hyderabad, 
$^{18}$MIT,
$^{19}$University of Pennsylvania,
$^{20}$Dartmouth}

\maketitle

\begin{abstract}
We introduce Ego4D, a massive-scale egocentric video dataset and benchmark suite. 
It offers \numhours hours of 
daily-life activity video spanning hundreds of scenarios (household, outdoor, workplace, leisure, etc.) captured by \numpeople unique camera wearers from \numlocations worldwide locations and \numcountries different countries. The approach to collection is designed to uphold rigorous privacy and ethics standards, with consenting participants and robust de-identification procedures where relevant.
Ego4D dramatically expands the volume of diverse egocentric video footage publicly available to the research community.  Portions of the video are accompanied by audio, 3D meshes of the environment, eye gaze, 
stereo, and/or synchronized videos from multiple egocentric cameras at the same event. 
Furthermore, we present a host of new benchmark challenges centered around understanding the first-person visual experience in the past (querying an episodic memory), present  (analyzing hand-object manipulation, audio-visual conversation, and social interactions), and future (forecasting activities).
By publicly sharing this massive annotated dataset and benchmark suite, we aim to push the frontier of first-person perception. \iftoggle{arxiv}{
Project page: \url{https://ego4d-data.org/}}

\end{abstract}

 \iftoggle{arxiv}{
\begin{figure*}[t]
 \centering %
 \includegraphics[width=0.85\textwidth]{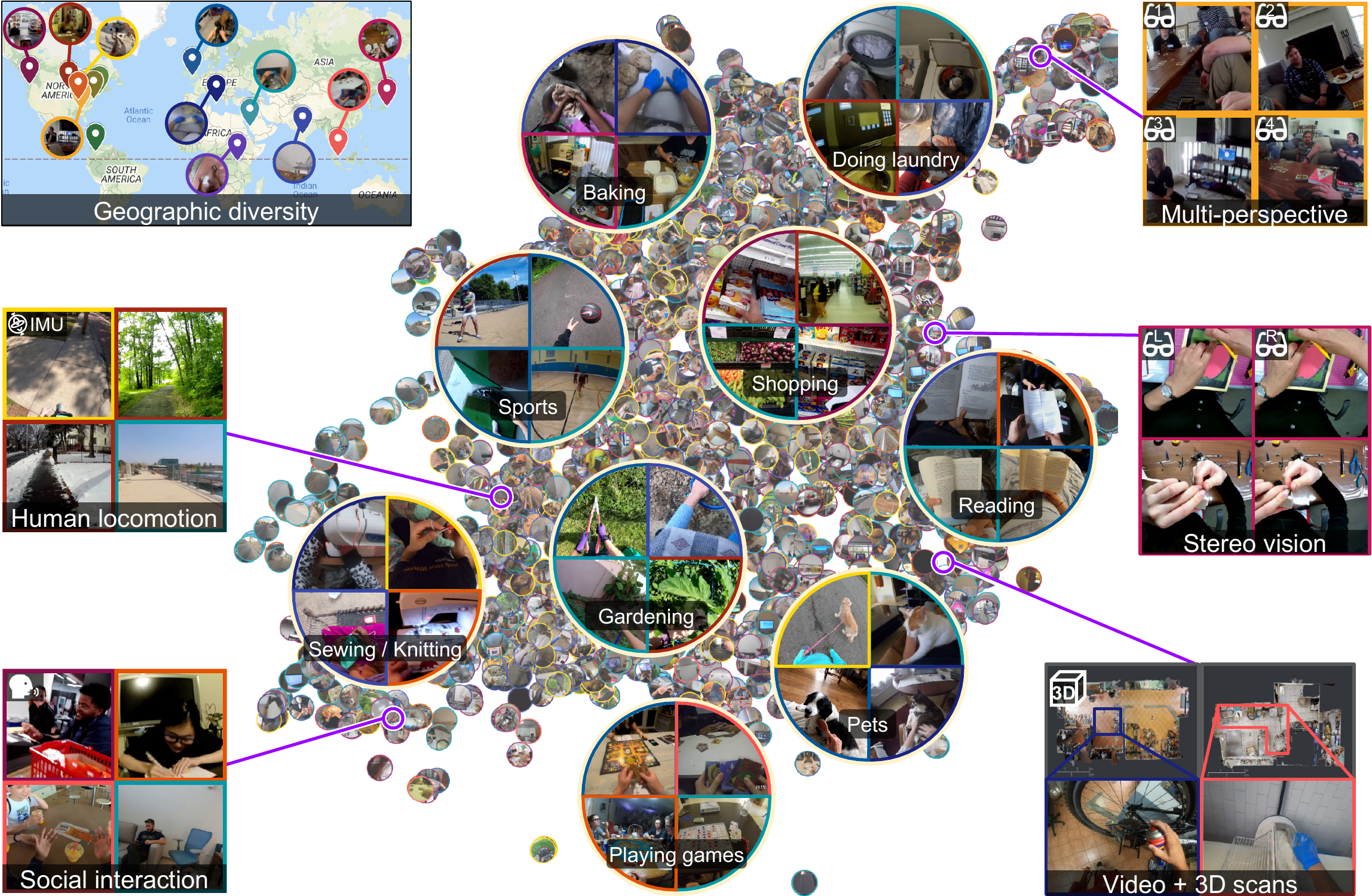}
 \caption{Ego4D is a massive-scale egocentric video dataset of daily life activity spanning \numlocations locations worldwide. Here we see a snapshot of the dataset (5\% of the clips, randomly sampled) highlighting its diversity in geographic location, activities, and modalities. The data includes social videos where participants consented to remain unblurred. See \url{https://ego4d-data.org/fig1.html} for interactive figure.}
 \label{fig:concept}
 \vspace*{-0.1in}
 \end{figure*}}
 {  %
\begin{figure}[t]
 \centering %
 \includegraphics[width=0.48\textwidth]{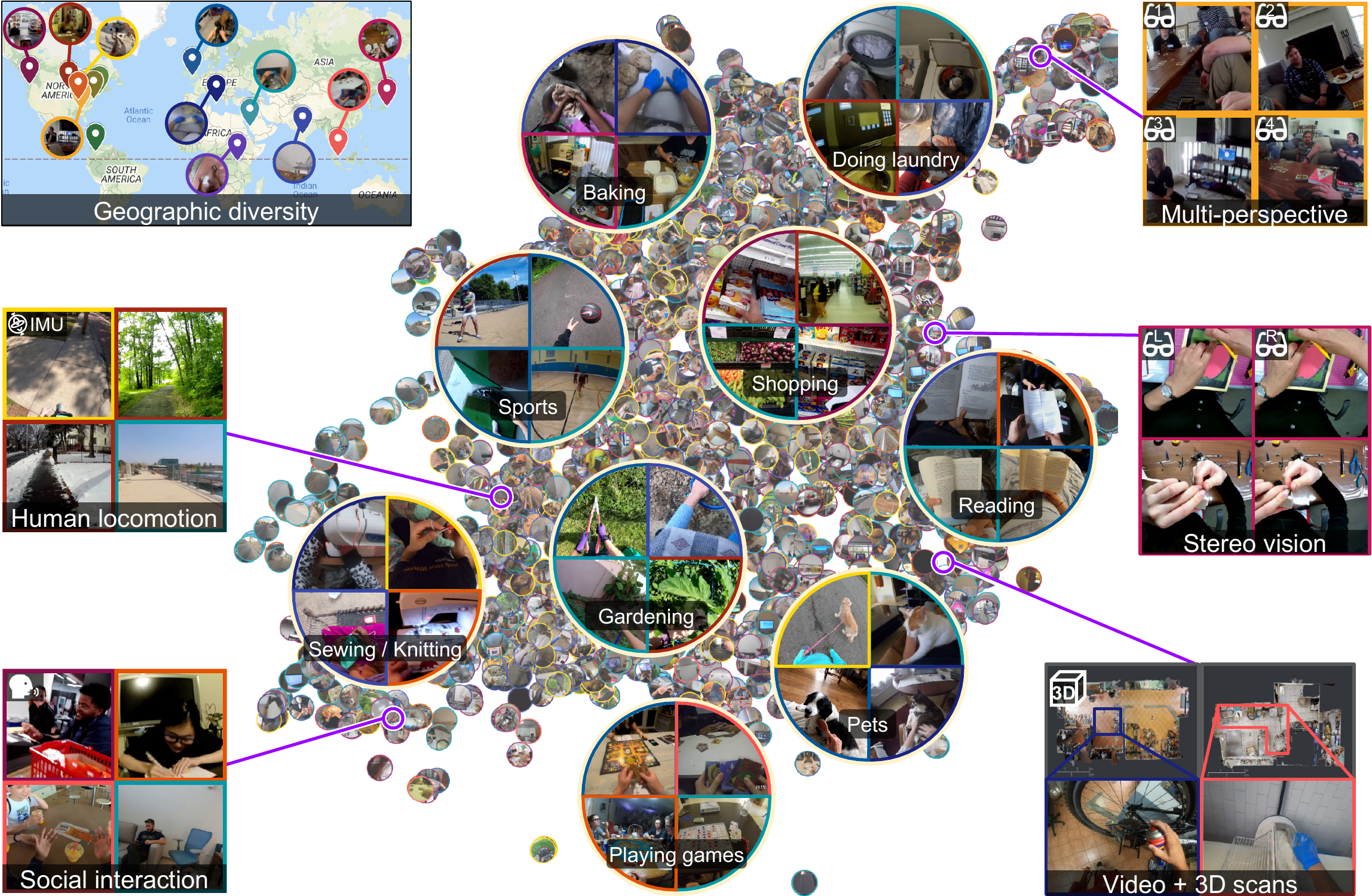}
 \caption{Ego4D is a massive-scale egocentric video dataset of daily life activity spanning \numlocations locations worldwide. Here we see a snapshot of the dataset (5\% of the clips, randomly sampled) highlighting its diversity in geographic location, activities, and modalities. The data includes social videos where participants consented to remain unblurred.}
 \label{fig:concept}
 \vspace*{-0.1in}
 \end{figure}}

 \doparttoc %
\faketableofcontents %

\part{} %

\vspace*{-0.5in}
\section{Introduction}

%

%

%

%

%
\iffalse
\begin{figure*}[t]
\vspace*{-0.2in}
\centering
\includegraphics[width=1\textwidth]{figures/concept_lowres_2}
\caption{Ego4D is a massive-scale egocentric video dataset of daily life activity spanning \numlocations locations worldwide. Here we see a snapshot of the dataset (5\% of the clips, randomly sampled) highlighting diversity in geographic location, activities, and modalities present in the data. \TNnote{See \[url\] for interactive figure.}}
\label{fig:concept}
\vspace*{-0.1in}
\end{figure*}
\fi

%

%

Today’s computer vision systems excel at naming objects and activities in Internet photos or video clips.  Their tremendous progress over the last decade has been fueled by major dataset and benchmark efforts, which provide  the annotations needed to train and evaluate algorithms on well-defined tasks~\cite{deng2009imagenet,everingham2010pascal,lin2014microsoft,caba2015activitynet,kinetics,gu2018ava}.

While this progress is exciting, current datasets and models represent only a limited definition of visual perception.  
First, today's influential Internet datasets  capture brief, isolated moments in time from a third-person ``spectactor" view. 
However, in both robotics and augmented reality, the input is a long, fluid video stream from the \emph{first-person} or \emph{``egocentric"} point of view---where we see the world through the eyes of an agent actively engaged with its environment.   
Second, whereas Internet photos are intentionally captured by a human photographer, %
images from an always-on wearable egocentric camera lack this active curation.    
Finally, first-person perception requires a persistent 3D understanding of the camera wearer's physical surroundings, and must interpret objects and actions in a human context---attentive to human-object interactions and high-level social behaviors.

Motivated by these critical contrasts, 
we present the Ego4D dataset and benchmark suite.  Ego4D aims to catalyze the next era of research in first-person visual perception.   \emph{Ego} is for egocentric, and \emph{4D} is for 3D spatial plus temporal information.

Our first contribution is the dataset: a massive ego-video collection of unprecedented scale and diversity that captures daily life activity around the world.  See Figure~\ref{fig:concept}.  It consists of \numhours hours of video collected by \numpeople unique participants from \numlocations worldwide locations in \numcountries different countries.  The vast majority of the footage is unscripted and ``in the wild", representing the natural interactions of the camera wearers as they go about  daily activities in the home, workplace, leisure, social settings, and commuting.  Based on self-identified characteristics, the camera wearers are of varying backgrounds, occupations, gender, and ages---not solely graduate students!  The video's rich geographic diversity supports the inclusion of objects, activities, and  people frequently absent from existing datasets.  Since each participant wore a camera for 1 to 10 hours at at time, the dataset offers long-form video content that displays the full arc of a person's complex interactions with the environment, objects, and other people.  In addition to RGB video, portions of the dataset also provide audio, 3D meshes, gaze, %
stereo, and/or synchronized multi-camera views that allow seeing one event from multiple perspectives.  Our dataset draws inspiration from prior egocentric video data efforts~\cite{pirsiavash2012detecting,lee-cvpr2012,engagement,singh2016krishnacam,charades-ego,Damen2018EPICKITCHENS,li2018eye,Damen2020RESCALING}, but makes significant advances in terms of scale, diversity, and realism.  %

Equally important to having the right data is to have the right research problems.  Our second contribution is a suite of five benchmark tasks  %
spanning the essential components of egocentric perception---indexing past experiences, analyzing present interactions, and anticipating future activity.
To enable research on these fronts, %
we provide millions of rich annotations that resulted from over 250,000 hours of annotator effort and range from temporal, spatial, and semantic labels, to dense textual narrations of activities, natural language queries, and speech transcriptions.

\iftoggle{arxiv}{
Ego4D is the culmination of an intensive two-year effort by Facebook and 13 universities around the world who came together for the common goal of spurring new research in egocentric perception. 
}
{
Ego4D is the culmination of an intensive two-year effort by 14 \srt{institutions} around the world who came together for the common goal of spurring new research in egocentric perception.}
We are kickstarting that work with a formal benchmark challenge to be held in June 2022.  In the coming years, we believe our contribution can catalyze new research not only in vision, but also robotics, augmented reality, 3D sensing, multimodal learning, speech, and language.    
These directions will stem not only from the
benchmark tasks we propose, but also alternative ones that the community will develop leveraging our massive, publicly available dataset.

\section{Related Work}

\paragraph{Large-scale third-person datasets}
In the last decade, annotated datasets have both presented new problems in computer vision and ensured their solid evaluation.  Existing collections like Kinetics~\cite{kinetics}, AVA~\cite{gu2018ava}, UCF~\cite{ucf}, ActivityNet~\cite{caba2015activitynet}, HowTo100M~\cite{miech19howto100m}, ImageNet~\cite{deng2009imagenet}, and COCO~\cite{lin2014microsoft} focus on third-person Web data, which have the benefit and bias of a human photographer.  In contrast, Ego4D is first-person.  Passively captured wearable camera video entails unusual viewpoints, motion blur, and lacks temporal curation.  %
Notably, pre-training egocentric video models with third-person data~\cite{slowfast,tsn,zhou2018temporal,nonlocal} suffers from the sizeable domain mismatch~\cite{charades-ego,ego-exo}.

\paragraph{Egocentric video understanding}
Egocentric video offers a host of interesting challenges,  such as human-object interactions~\cite{cai2016understanding,damen2016you,nagarajan2018grounded}, activity recognition~\cite{zhou2015temporal,kazakos2019epic,ego-exo},  anticipation~\cite{singh2016krishnacam,furnari2020rolling,liu2020forecasting,abu2018will,avt}, %
video summarization~\cite{lee-cvpr2012,lu2013story,lee2015predicting,yonetani2016visual,lu2015personal,del2016summarization}, detecting hands~\cite{zoombie,Bambach_2015_ICCV}, parsing social interactions~\cite{dyadic,ng2020you2me,disney-social},
and inferring the camera wearer's body pose~\cite{jiang2017seeing}.  
Our dataset can facilitate new work in all these areas and more, and our proposed benchmarks (and annotations thereof) widen the tasks researchers can consider moving forward.
We defer discussion of how prior work relates to our benchmark tasks to Sec.~\ref{sec:benchmarks}.

\paragraph{Egocentric video datasets}
Multiple egocentric datasets have been developed over the last decade.  Most relevant to our work are those containing unscripted daily life activity, which includes EPIC-Kitchens~\cite{Damen2018EPICKITCHENS,Damen2020RESCALING},  UT Ego~\cite{lee-cvpr2012,engagement}, Activities of Daily Living (ADL)~\cite{pirsiavash2012detecting}, and the Disney dataset~\cite{disney-social}.  The practice of giving cameras to participants to take out of the lab, first explored in~\cite{pirsiavash2012detecting,lee-cvpr2012,disney-social}, inspires our approach.
Others 
are (semi-)scripted, where camera wearers are instructed to perform a certain activity, as in Charades-Ego~\cite{charades-ego} and EGTEA~\cite{li2018eye}. 
Whereas today's largest ego datasets focus solely on kitchens~\cite{Damen2018EPICKITCHENS,Damen2018EPICKITCHENS,li2018eye,cmu-kitchens}, Ego4D spans hundreds of environments both indoors and outdoors.  Furthermore, while existing datasets rely largely on graduate students as camera wearers~\cite{Damen2018EPICKITCHENS,Damen2020RESCALING,lee-cvpr2012,lee-cvpr2012,engagement,li2018eye,pirsiavash2012detecting,disney-social,ryoo2013first,ng2020you2me}, Ego4D camera wearers are of a much wider demographic, as detailed below.
Aside from daily life activity, prior ego datasets focus on conversation~\cite{egocom}, inter-person interactions~\cite{dyadic,ryoo2013first,ng2020you2me,disney-social}, place localization~\cite{egocart,ragusa}, multimodal sensor data~\cite{ecm,cmu-kitchens,Silva2018}, human hands~\cite{Bambach_2015_ICCV,zoombie}  human-object interaction~\cite{ragusa2020meccano,lemma}, and object tracking~\cite{TREK150}.

Ego4D is an order of magnitude larger than today's largest egocentric datasets both in terms of hours of video (\numhours hours vs.~100 in~\cite{Damen2020RESCALING}) and unique camera wearers (\numpeople people vs.~71 in~\cite{charades-ego}); it spans hundreds of environments (rather than one or dozens, as in existing collections); and its video comes from \numlocations worldwide locations and \numcountries countries (vs.~just one or a few cities).  The Ego4D annotations are also of unprecedented scale and depth, with millions of annotations %
supporting multiple complex tasks. %
As such, Ego4D represents a step change in dataset scale and diversity.  We believe both factors are paramount to pursue the next generation of perception for embodied AI.

\section{Ego4D Dataset}

Next we overview the dataset, which we are making publicly available under an Ego4D license.

\iffalse
\begin{figure*}[t]
\centering
%
\begin{tabular}{ccc}
\includegraphics[height=0.18\textwidth]{figures/map.jpg}
\includegraphics[height=0.18\textwidth]{figures/demographics.jpg}
\includegraphics[height=0.18\textwidth]{figures/scenarios-histogram.jpg}
\end{tabular}
\caption{Left: Ego4D consortium. Center: Camera wearer demographic information. Right: Scenarios distribution.  \KGnote{Right -- decide if we want to put distribution of actions and objects from FHO and moments taxonomies, too.  Dima's nested pie chart idea here? - but it will take more room.}  \KGnote{for arxiv we can be explicit about the locations and names; for CVPR submission we'll have to make appropriately anonymous.}}\label{fig:map}
\end{figure*}
\fi

\begin{figure}[t]
\centering
\includegraphics[height=0.5\columnwidth]{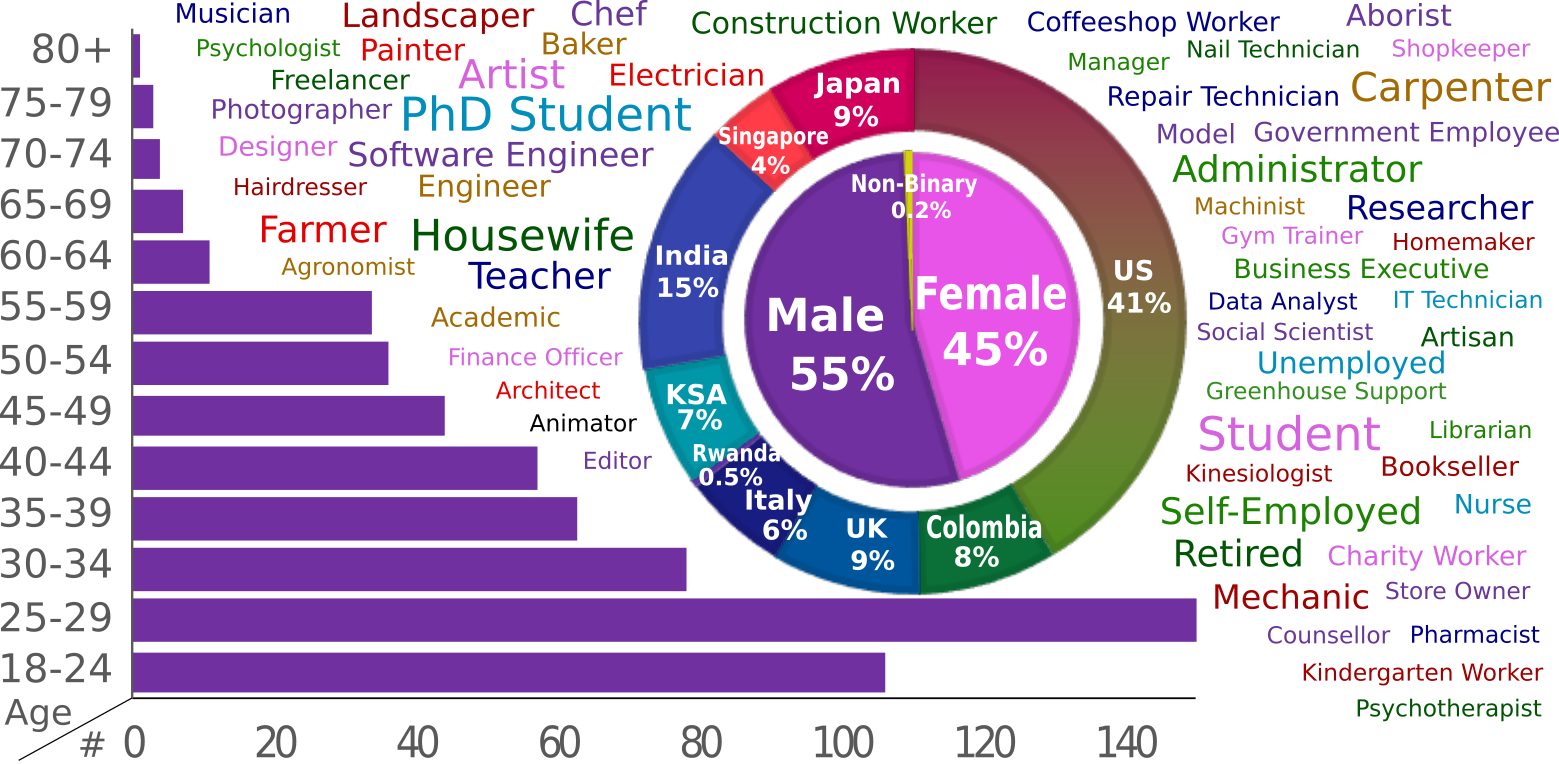}
\vspace*{-0.12in}
\caption{Ego4D camera wearer demographics---age, gender, countries of residence, and occupations (self-reported).  Font size reflects relative frequency of the occupation.}\label{fig:demographics}
\end{figure}

\subsection{Collection strategy and camera wearers}

Not only do we wish to amass an ego-video collection that is substantial in scale, but we also want to ensure its diversity of people, places, objects, and activities. Furthermore, for realism, we are interested in unscripted footage captured by people wearing a camera for long periods of time.

To this end, we devised a distributed approach to data collection.  The Ego4D %
project consists of \numlabs teams from universities and labs  in \numcountriesfull countries and \numcontinentsfull continents (see map in Figure~\ref{fig:concept}). Each team recruited 
participants to wear a camera for 1 to 10 hours at a time, for a total of \numpeople unique camera wearers and \numhours hours of video in this first dataset release (Ego4D-3K).  Participants in \numlocations total cities were recruited by 
word of mouth, ads, and postings on community bulletin boards. 
Some teams recruited participants with occupations that have interesting visual contexts, such as bakers, carpenters, landscapers, or mechanics.  

Both the geographic spread of our team as well as our approach to recruiting participants were critical to arrive at a diverse demographic composition, as shown in Figure~\ref{fig:demographics}.\footnote{for \srt{\FEB{64}\%} of all participants; missing demographics are due to protocols or participants opting out of answering specific questions.}  %
Participants %
cover a wide variety of occupations, %
 span many age brackets, with \FEB{96} of them over 50 years old, and \FEB{45\%} are female.  \FEB{Two} participants identified as non-binary, and \FEB{two} preferred not to say a gender.

\subsection{Scenarios composing the dataset}

What activities belong in an egocentric video dataset?  
Our research is motivated by problems in robotics and augmented reality, where vision systems will encounter \emph{daily life scenarios}.  Hence, we consulted a survey from the U.S. Bureau of Labor Statistics\footnote{https://www.bls.gov/news.release/atus.nr0.htm}  that captures how people spend the bulk of their time in the home (e.g., cleaning, cooking, yardwork), leisure  (e.g., crafting, games, attending a party), transportation (e.g., biking, car), errands (e.g., shopping, walking dog, getting car fixed), and in the workplace (e.g, talking with colleagues, making coffee).

\begin{figure}[t]
\centering
\includegraphics[height=0.5\columnwidth]{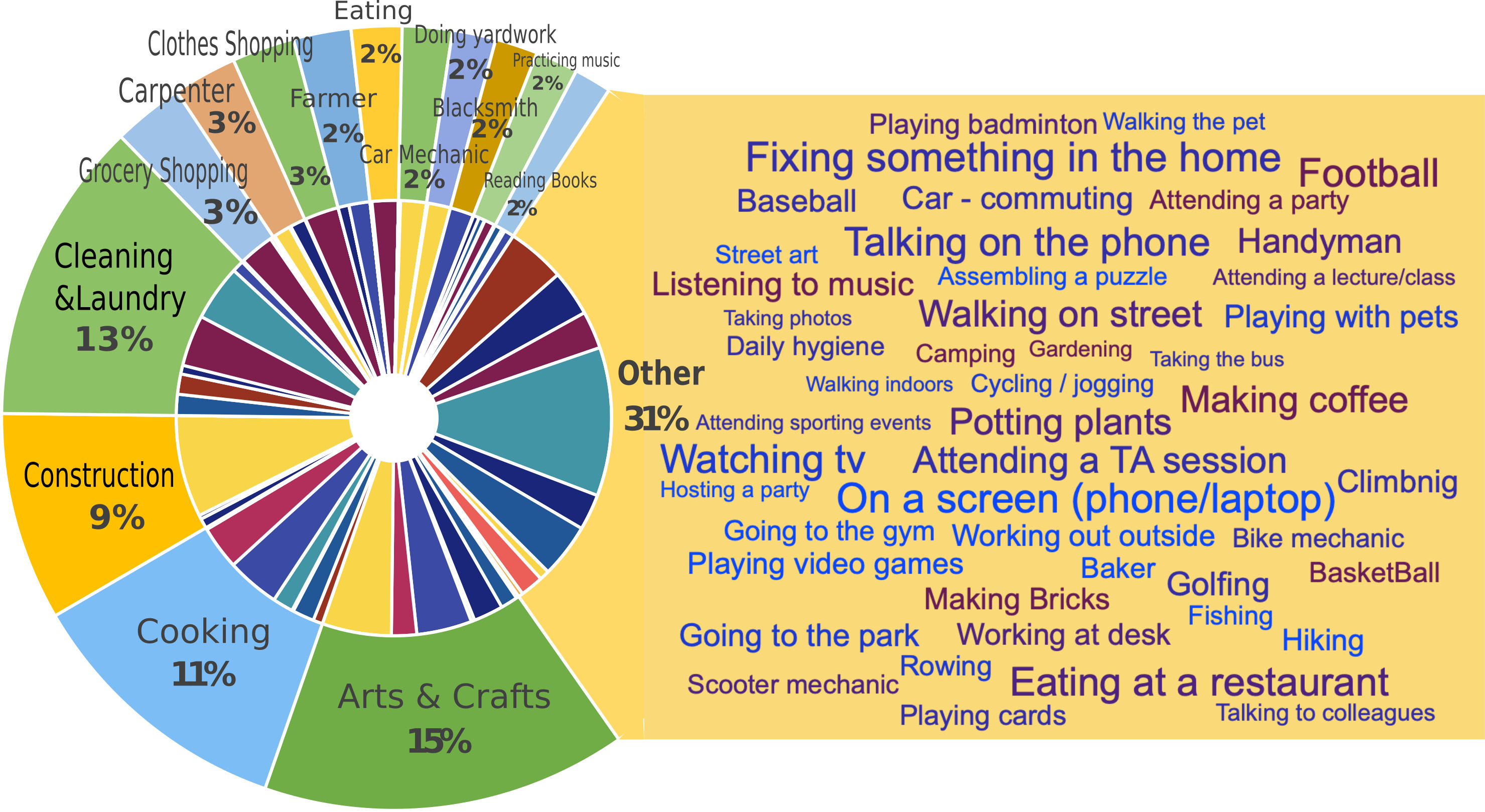}
\vspace*{-0.05in}
\caption{Scenarios in Ego4D.  Outer circle shows the 14 most common scenarios (70\% of the data). Wordle shows scenarios in the remaining 30\%.  Inner circle is color coded by the contributing partner (see map color legend in Fig~\ref{fig:concept}).}\label{fig:scenarios}
\end{figure}

To maximize coverage of such scenarios, our approach is a compromise between directing camera wearers and giving no guidance at all:  %
(1) we recruited participants whose collective daily life activity would naturally encompass a spread of the scenarios (as selected freely by the participant), and (2) we asked participants to wear the camera at length (at least as long as the battery life of the device) so that the activity would unfold naturally in a longer context.  A typical raw video clip in our dataset lasts 8 minutes---significantly longer than the 10 second clips often studied in third-person video understanding~\cite{kinetics}.  %
In this way, we capture unscripted activity while being mindful of the scenarios' coverage.

The exception is for certain  multi-person scenarios, where, in order to ensure sufficient data for the audio-visual and social benchmarks, we asked participants at five sites who had consented to share their conversation audio and unblurred faces to take part in social activities, such as playing games.
We leverage this portion of Ego4D for the audio-visual and social interaction benchmarks (Sec.~\ref{sec:avd} and~\ref{sec:social}).

Figure~\ref{fig:scenarios}  shows the wide distribution of scenarios captured in our dataset.  Note that within each given scenario there are typically dozens of actions taking place, e.g., the carpentry scenario includes hammering, drilling, moving wood, etc. %
Overall, the \numpeople camera wearers bestow our dataset with a glimpse of daily life activity around the world.

\begin{table*}[t]
\begin{center}
\begin{tabular}{l|l|l|l|l|l|l|l|l|l|l|}
Modality: &  RGB video &  Text narrations & Features & Audio & Faces & 3D scans & Stereo & Gaze & IMU & Multi-cam\\
\hline
\# hours: & \numhours & \numhours & \numhours & 2,535 & 612 & 491 &  80 & 45  & 836 & 224  \\
\hline  %
 \end{tabular}
 \end{center}
 \vspace*{-0.05in}
\caption{Modalities of data in Ego4D and their amounts.  ``Narrations" are dense, timestamped descriptions of camera wearer activity (cf.~
Sec.~\ref{sec:narrations}). ``3D scans" are meshes from Matterport3D scanners for the full environment in which the video was captured.  ``Faces" refers to video where participants consented to remain unblurred.  ``Multi-cam" refers to synchronized video captured at the same event by multiple camera wearers. ``Features" refers to precomputed SlowFast~\cite{slowfast} video features.  \iftoggle{arxiv}{\srt{Gaze collected only by Indiana U.~and Georgia Tech.}{}}}
\label{table:modalities}
\end{table*}

\subsection{Cameras and modalities}
To avoid  models overfitting to a single capture device, seven different head-mounted cameras were deployed across the dataset: GoPro, Vuzix Blade, Pupil Labs, ZShades, ORDRO EP6, iVue Rincon 1080, and Weeview.  %
They offer tradeoffs in the modalities available (RGB, stereo, gaze), field of view, and battery life.  
The field of view and camera mounting are particularly influential: while a GoPro mounted on the head pointing down offers a high resolution view of the hands manipulating objects (Fig.~\ref{fig:narration}, right), a heads-up camera like the Vuzix shares the vantage of a person's eyes, but will miss interactions close to the body (Fig.~\ref{fig:narration}, left).

\begin{figure}[t]
\centering
\hspace*{-0.1in}
\includegraphics[height=0.3\textwidth]{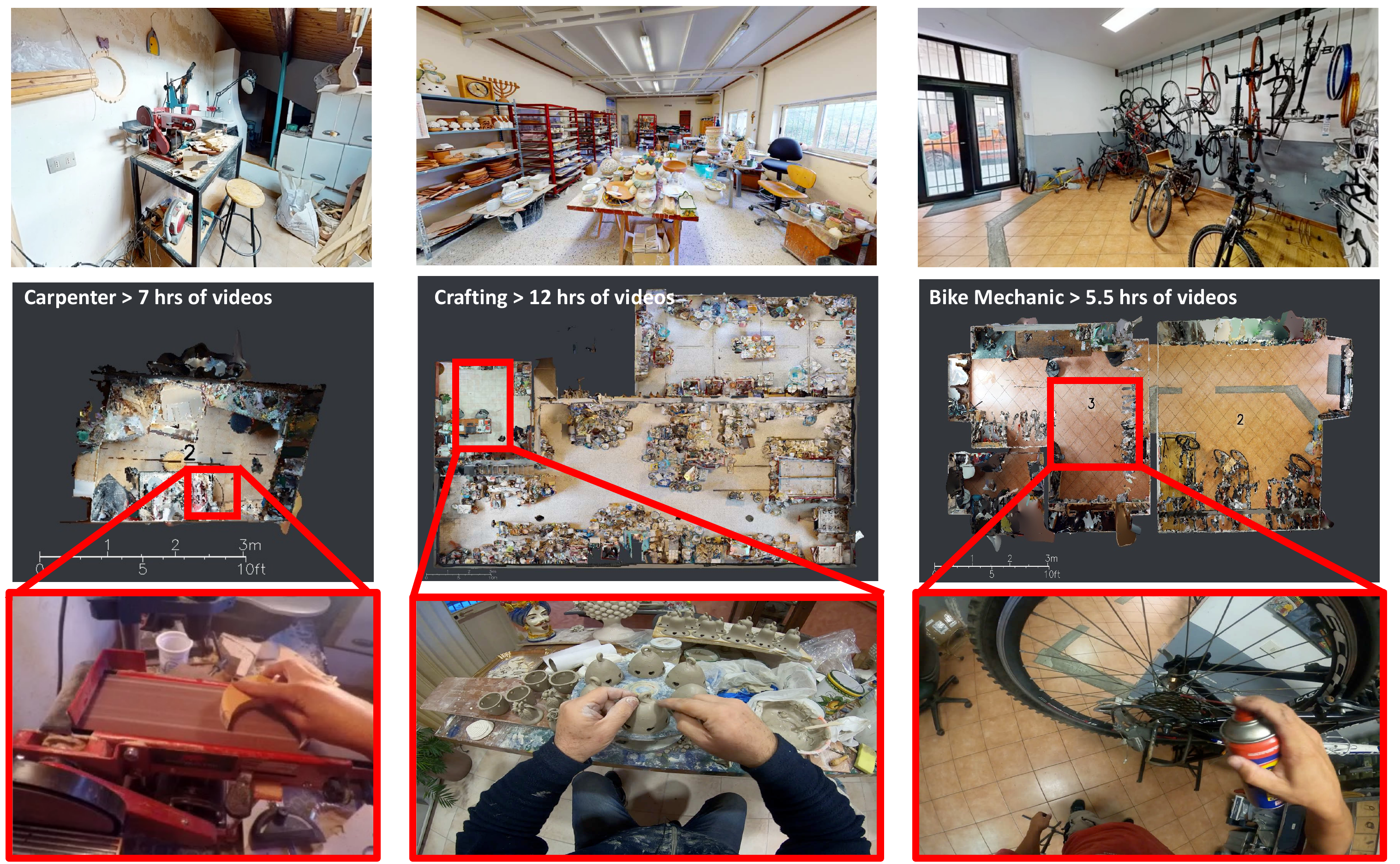}
\vspace*{-0.14in}
\caption{Some videos (bottom) have coupled 3D meshes (top) from Matterport3D scanners, allowing one to relate the dynamic video to the static 3D environment (middle).}\label{fig:matterport}
\end{figure}

In addition to video, portions of Ego4D offer several other data modalities: 3D scans, audio, gaze\iftoggle{arxiv}{\footnote{Eye trackers were deployed by Indiana U.~and Georgia Tech only.}}{}, stereo,  multiple synchronized wearable cameras, and textual narrations.  See Table~\ref{table:modalities}.    Each can support new research challenges.  For example, having Matterport3D scans of the environment coupled with ego-video clips (Figure~\ref{fig:matterport}) offers a unique opportunity for understanding dynamic activities in a persistent 3D context, as we exploit in the Episodic Memory benchmark (see Sec.~\ref{sec:episodic}).  Multiple synchronized egocentric video streams allow accounting for the first and second-person view in social interactions.  Audio allows analysis of conversation and acoustic scenes and events.

\subsection{Privacy and ethics}

From the onset, privacy and ethics standards were critical to this data collection effort.  
Each partner %
was responsible for %
developing a %
policy.  
While %
specifics vary per site, 
this generally entails: 
\begin{itemize}
\item Comply with own institutional research policy, e.g., independent ethics committee review where relevant
\vspace*{-0.075in} 
\item Obtain informed consent of camera wearers, who can ask questions and withdraw at any time, and are free to review and redact their own video
\vspace*{-0.075in}
\item Respect rights of others in private spaces, 
and avoid capture of sensitive areas or activities 
\vspace*{-0.075in}
\item Follow de-identification requirements for personally identifiable information (PII)
\end{itemize}
In short, these standards typically require that the video be captured in a controlled environment with informed consent by all participants, 
or else in public spaces where faces and other PII are blurred.  \srt{Appendix~\ref{sec:appendix-societal} discusses potential negative societal impact.} %

\iffalse
\begin{figure*}[t]
\centering
\begin{tabular}{cc}
%
%
%
%
%
\includegraphics[height=0.17\textwidth]{figures/benchmarks1.jpg}
\includegraphics[height=0.17\textwidth]{figures/benchmarks2.jpg}
\end{tabular}
%
\caption{The Ego4D benchmark suite centers around first-person interaction with places, objects, and people.}
\label{fig:benchmarks}
\end{figure*}
\fi

%

%

%

%

%

\subsection{Possible sources of bias}

While Ego4D pushes the envelope on massive everyday video from geographically and demographically diverse sources, we are aware of a few biases in our dataset.
\numlocations locations is still a long way from complete coverage of the globe.  In addition, the camera wearers are generally located in urban or college town areas.  
The COVID-19 pandemic led to ample footage in stay-at-home scenarios such as cooking, cleaning, crafts, etc.~and more limited opportunities to collect video at major social public events.  In addition, since battery life prohibits daylong filming, the videos---though unscripted---tend to contain more active portions of a participant's day.  
Finally, Ego4D annotations are done by crowdsourced workers in two sites in Africa.  This means that there will be at least subtle ways in which the language-based narrations are biased towards their local word choices. %

\subsection{Dataset accessibility}

\FEB{At \numhours hours of video, we are mindful that Ego4D's scale can be an obstacle for accessibility for some researchers, depending on their storage and compute resources.  To mitigate this, we have taken several measures. 
First, we provide precomputed action features (SlowFast 8x8 with ResNet 101 backbone pretrained for Kinetics 400) with the dataset, an optional starting point for any downstream work. Second, only portions of the data constitute the formal  challenge train/test sets for each benchmark---not all \numhours hours (see Appendix~\ref{sec:appendix-splits}). As Ego4D annotations increase, we will create standardized mini-sets. Finally, we provide the option to download only the data targeting an individual benchmark or modality of interest.}

\begin{figure}[t]
\centering
\begin{tabular}{ccc}
\includegraphics[height=0.155\textwidth]{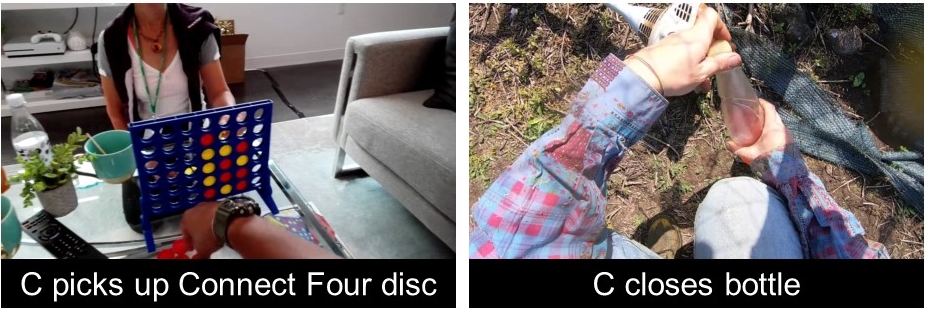}
\end{tabular} 
\vspace*{-0.12in}
\caption{Example narrations.  ``C" refers to camera wearer.}\label{fig:narration}
\end{figure}

\begin{figure*}[t]
\centering\hspace*{-0.2in}
\begin{tabular}{ccc}
\includegraphics[height=0.18\textwidth]{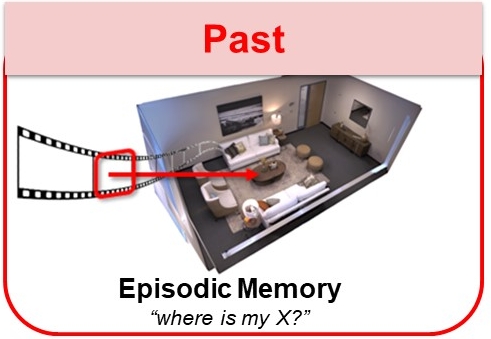}
\includegraphics[height=0.18\textwidth]{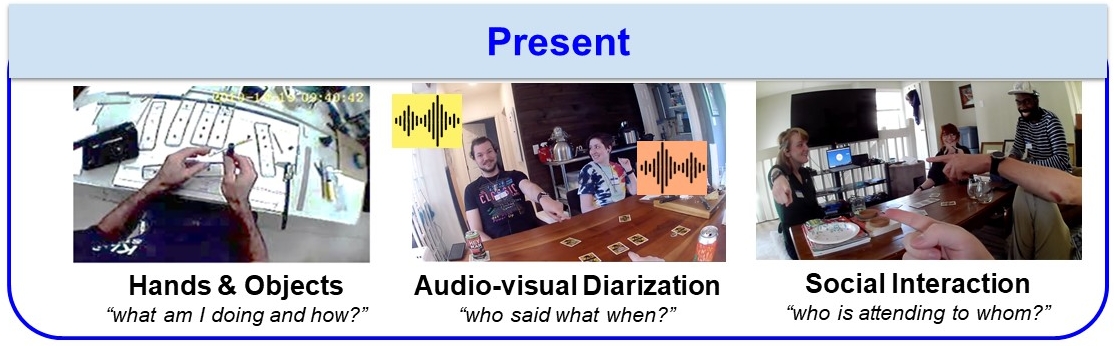}
\includegraphics[height=0.18\textwidth]{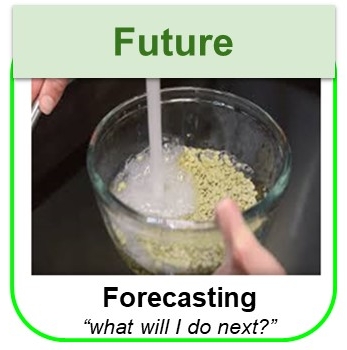}
\end{tabular}
\vspace*{-0.1in}
\caption{The Ego4D benchmark suite centers around the first-person visual experience---from remembering the past, to analyzing the present, to anticipating the future.}
\label{fig:benchmarks}
\end{figure*}

\section{Narrations of Camera Wearer Activity}\label{sec:narrations}

Before any other annotation occurs, we pass all video through a  \emph{narration} procedure.   Inspired by the pause-and-talk narrator~\cite{Damen2018EPICKITCHENS}, annotators are asked to watch a 5 minute clip of video, summarize it with a few sentences, and then re-watch, pausing repeatedly to write a sentence about each thing the camera wearer does.  We record the timestamps and the associated free-form sentences.  See Figure~\ref{fig:narration}.  Each video receives two independent narrations from different annotators.
The narrations are temporally dense: on average we received 13.2 sentences per minute of video, for a total of \numnarrationsentences sentences.  In total the narrations describe the Ego4D video using 1,772 unique verbs (activities) and 4,336 unique nouns (objects).  See Appendix~\ref{sec:data-appendix} for details.

The narrations allow us to (1) perform text mining for data-driven taxonomy construction for actions and objects, (2) sort the videos by their content to map them to relevant benchmarks, and (3) identify temporal windows where certain annotations should be seeded.   Beyond these uses, the narrations are themselves a contribution of the dataset, potentially valuable for research on video with weakly aligned natural language.  To our knowledge, ours is the largest  repository of aligned language and video \cc{(e.g., HowTo100M~\cite{miech19howto100m}, an existing Internet repository with narrations, contains noisy spoken narrations that only sometimes comment on the activities taking place)}.

\section{Ego4D Benchmark Suite}\label{sec:benchmarks}

First-person vision has the potential to transform many applications in augmented reality and robotics.  However, compared to mainstream video understanding, egocentric perception requires new fundamental research to account for long-form video, attention cues, person-object interactions, multi-sensory data, and the lack of manual temporal curation inherent to a passively worn camera.

Inspired by all these factors, we propose a suite of challenging benchmark tasks.  %
The five benchmarks tackle the \emph{past}, \emph{present}, and \emph{future} of first-person video.  See Figure~\ref{fig:benchmarks}.   The following sections introduce each task and its annotations.  The first dataset release has annotations for \FEB{48-1,000 hours} of data per benchmark, on top of the \numhours hours of data that is narrated.  The Appendices describe how we sampled videos per benchmark to maximize relevance to the task while maintaining geographic diversity.

We developed baseline models drawing on state-of-the-art components from the literature in order to test drive all Ego4D benchmarks. \textbf{The Appendix presents the baseline models and quantitative results.}
We are running a formal Ego4D competition in June 2022 inviting the research community to improve on these baselines.

\iffalse
\KGnote{For each benchmark, please provide concise text for this template (short version here in main paper, then all reproducibility details for appendix).  Please keep baseline/result portions especially brief in the main paper, if needed, in favor of making the data, task, and annotations clear.  We can afford approx \textbf{0.75 page per benchmark}.
\begin{itemize}
\item \textbf{motivation}: brief motivation for the task - why is this a good problem to study?  what will it enable?
\item \textbf{task definition}: crisp definition of the task - input and output  
\item \textbf{relationship to existing tasks}: related work - a nod to the key references that have tasks most related to the proposed benchmark (where relevant) and what differences we most want to call out about this one (1-2 sentences)
\item \textbf{target data:} how the videos that got annotated for this benchmark were selected, what aspects were prioritized 
\item \textbf{annotations:} annotations provided with the dataset, including any formation of taxonomy done using narrations etc.
\item \textbf{evaluation:} metrics and setup
\item \textbf{baseline models and results}: describe models tested and report a key result (let details go to appendix).
\item \textbf{figure}: one interesting figure for the benchmark.
\end{itemize}
}
\fi

%
%
%
%
%

\subsection{Episodic Memory}\label{sec:episodic}

\paragraph{Motivation}
Egocentric video from a wearable camera records the who/what/when/where of an individual's daily life experience.  This makes it  ideal for what Tulving called \emph{episodic} memory~\cite{tulving}: specific first-person experiences (``what did I eat and who did I sit by on my first flight to France?”), to be distinguished from \emph{semantic} memory (``what’s the capital of France?”).
An augmented reality assistant that processes the egocentric video stream could give us super-human memory if it could appropriately index our visual experience and answer queries.

\begin{figure}[t]
\centering
\includegraphics[width=0.5\textwidth]{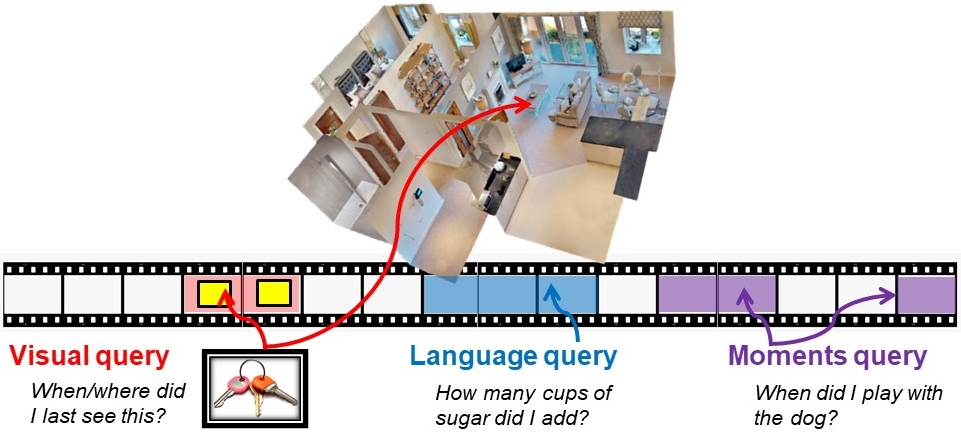}
\caption{Episodic Memory's three query types}\label{fig:episodic}
\end{figure}

\paragraph{Task definition}
Given an egocentric video and a query, the Ego4D Episodic Memory task requires localizing where the answer can be seen within the user's past video. %
We consider three query types. (1) \emph{Natural language queries} (NLQ), in which the query is expressed in text (e.g., ``What did I put in the drawer?"), and the output response is the temporal window where the answer is visible or deducible.  %
(2) \emph{Visual queries} (VQ), in which the query is a static image of an object, %
and the output response localizes the object the last time it was seen in the video,  both temporally and spatially.  The spatial response is a 2D bounding box on the object, and optionally a 3D displacement vector from the current camera position to the object's 3D bounding box. VQ captures how a user might teach the system an object with an image example, then later ask for its location (``Where is this [picture of my keys]?").  
(3) \emph{Moments queries} (MQ), in which the query is the name of a high-level activity or ``moment", and the response  consists of all temporal windows where the activity occurs (e.g., ``When did I read to my children?").
 See Figure~\ref{fig:episodic}. 

\paragraph{Annotations}
For language queries, we devised a set of 13 template questions meant to span things a user might ask to augment their memory, such as \emph{``what is the state of object X?"}, e.g., ``did I leave the window open?". %
Annotators express the queries in free-form natural language, and also provide the slot filling (e.g., X = window).  For moments, we established a taxonomy of 110 activities in a data-driven, semi-automatic manner by mining the narration summaries.  Moments capture high-level activities in the camera wearer's day, e.g., \emph{setting the table} is a moment, whereas \emph{pick up} is an action in our Forecasting benchmark (Sec.~\ref{sec:forecasting}).

For NLQ and VQ, we ask annotators to generate language/visual
queries and couple them with the ``response track" in the video.  For MQ, we provide the taxonomy of labels and ask annotators to label clips with each and every temporal segment containing a moment instance.  %
In total, we have $\sim$74K total queries spanning $800$ hours of video.

\paragraph{Evaluation metrics and baselines}
For NLQ, we use top-k recall at a certain temporal intersection over union (tIoU) threshold.  
MQ adopts a popular metric used in temporal action detection:  mAP at multiple tIoU thresholds, as well as %
top-kx recall.  %
VQ adopts temporal and spatio-temporal localization metrics as well as timeliness metrics that encourage speedy searches.  %
Appendix~\ref{sec:episodic-appendix} presents the baseline models we developed and reports results.

\iffalse
%
\paragraph{Baselines and results}
%
%
%
We adopt 2D Temporal Adjacent Networks (2D-TAN) \cite{2DTAN_2020_AAAI}, which uses adjacent temporal windows candidates as context on a two-dimensional temporal map to retrieve the relevant window, 
as the NLQ baseline.
On the mini-training set, 2D-TAN gives top-1 and top-5 recalls of $32.0\%$ and $41.0\%$, respectively.
%

The MQ baseline is built on VSGN [*], which has a feature pyramid architecture with cross-scale graph networks, and uses two stages to regress and adjust moment temporal boundaries. This model gives an average mAP of $6.97\%$. 

The VQ baseline uses a detection + tracking approach. The detector uses a region proposal network to generate object proposals, and learns a Siamese head to measure the similarity to the visual query. It identifies a high-confidence detection that is closest in time to the query. A single-object tracker~\cite{???} is then initialized with this detection and recovers the full response track. This model gives an average tAP of $10\%$ and stAP of $5\%$. A SLAM-based approach is used to localize the object within a 3D scan using the response track~\cite{???}.
\fi

\paragraph{Relation to existing tasks} 
Episodic Memory has some foundations in existing  vision problems, but also adds new challenges. 
All three queries call for spatial reasoning in a static environment coupled with dynamic video of a person who moves and changes things; current work largely treats these two elements separately.
The timeliness metrics encourage work on intelligent contextual search. %
While current literature on language+vision focuses on captioning and question answering for isolated instances of Internet data~\cite{chen2015microsoft,krishna2017dense,msrvtt,vqa}, NLQ is motivated by queries about the camera wearer's own visual experience and  operates over long-term observations.
VQ upgrades object instance recognition~\cite{brachmann2014learning,lai2014unsupervised,georgakis2016multiview,Mercier_2021_WACV} to deal with video (frequent FoV changes, objects entering/exiting the view) and to reason about objects in the context of a 3D environment.  
Finally, MQ can be seen as activity detection~\cite{lin2018bsn,xu2020g,zhao2020video} but for the activities of the camera wearer.

\subsection{Hands and Objects}\label{sec:ho}

\paragraph{Motivation} 

While Episodic Memory aims to make \emph{past} video queryable, our next benchmark aims to understand the camera wearer's \emph{present} activity---in terms of interactions with objects and other people. Specifically, the \text{Hands and Objects benchmark} captures how the camera wearer changes the state of an object by using or manipulating it---which we call an \textit{object state change}. Though cutting a piece of lumber in half can be achieved through many methods (\emph{e.g.}, various tools, force, speed, grasps, end-effectors), all should be recognized as the same state change.  This generalization ability will enable us to understand human actions better, as well as to train robots to learn from human demonstrations in video. 

\paragraph{Task definitions} We interpret an object state change to include various physical changes, including changes in size, shape, composition, and texture. Object state changes can be viewed along temporal, spatial and semantic axes, leading to these three tasks: (1) \emph{Point-of-no-return temporal localization}: given a short video clip of a state change, the goal is to estimate the keyframe that contains the point-of-no-return (PNR) (the time at which a state change begins); (2) \emph{State change object detection}: given three temporal frames (pre, post, PNR), the goal is to regress the bounding box of the object undergoing a state change; (3) \emph{Object state change classification}:  given a short video clip, the goal is to classify whether an object state change has taken place or not.

\paragraph{Annotations} We select the data to annotate based on activities that are likely to involve hand-object interactions (\emph{e.g.}, knitting, carpentry, baking, \emph{etc.}). We start by labeling each narrated hand-object interaction. For each, we label three moments in time (pre, PNR, post) and the bounding boxes for the hands, tools, and objects in each of the three frames. We also annotate the state change types (remove, burn, \emph{etc.}, see Fig.~\ref{fig:ho}), action verbs, and nouns for the objects.

\begin{figure}[t]
    \centering \small
    \includegraphics[width=0.29\linewidth]{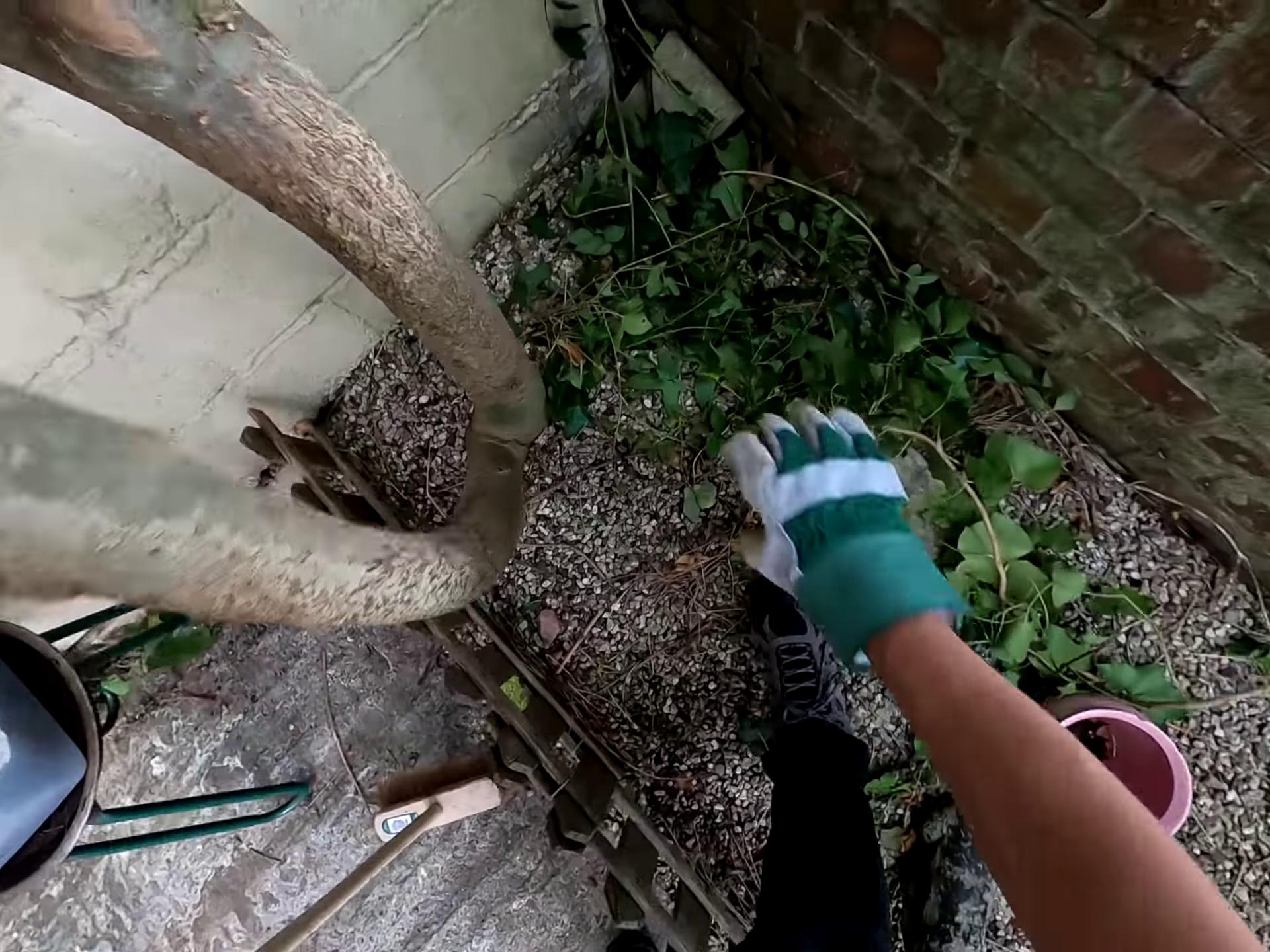}
    \includegraphics[width=0.29\linewidth]{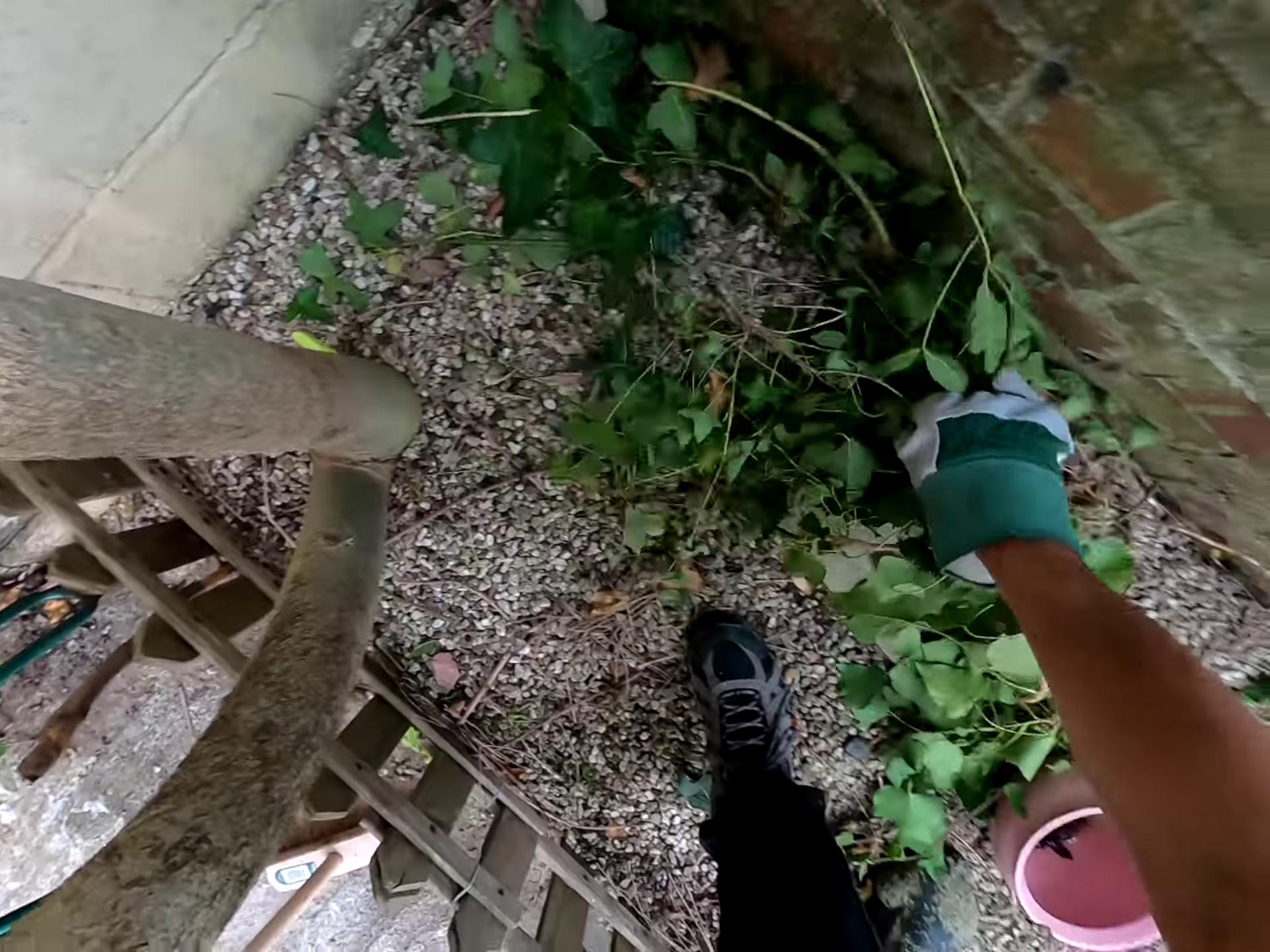}
    \includegraphics[width=0.29\linewidth]{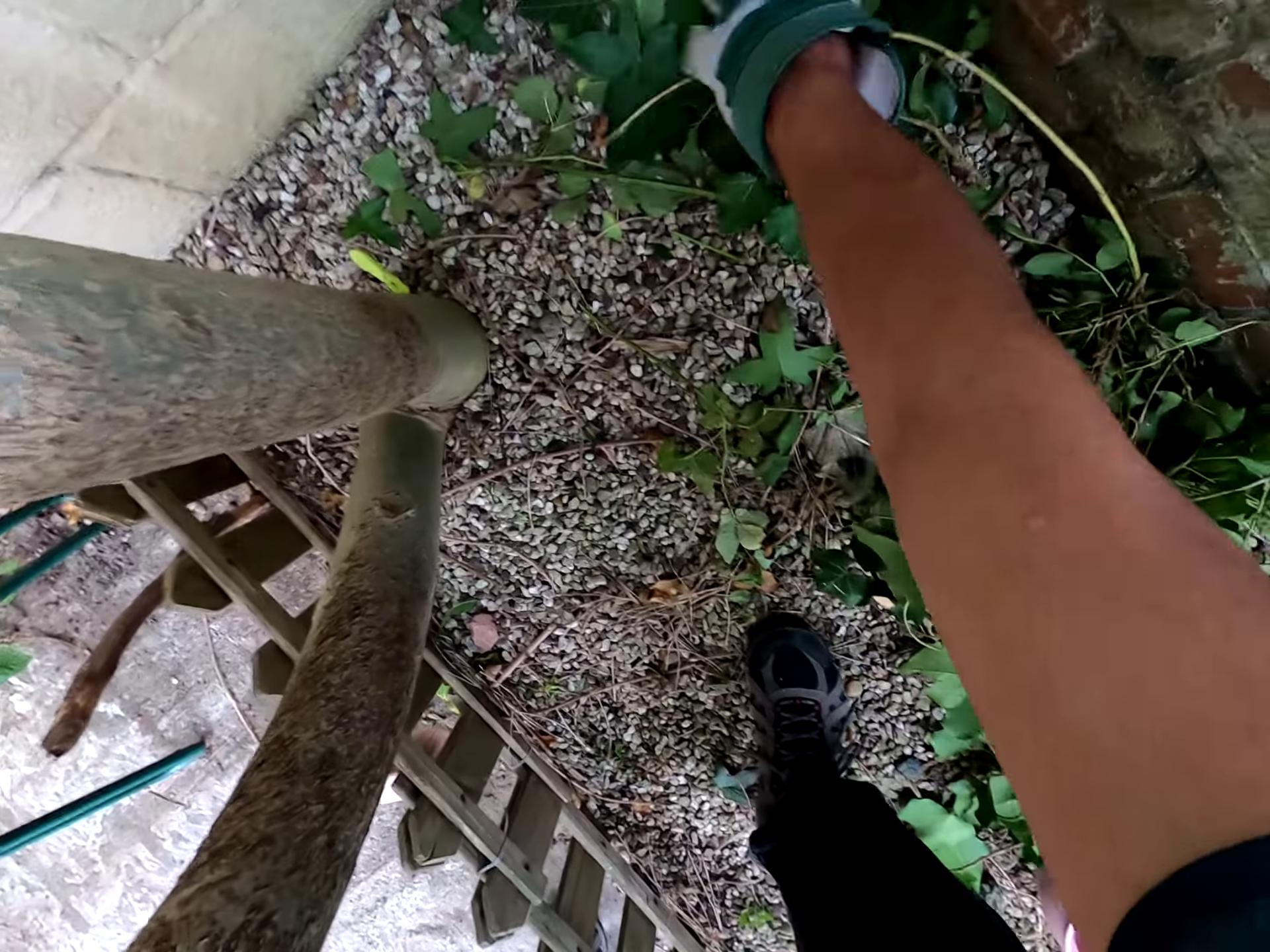}\\
    \vspace{-6mm}\textcolor{white}{\small \textbf{pre-condition \hspace{12mm} PNR \hspace{12mm} post-condition}}\\
    \vspace{2mm} State-change: Plant removed from ground\\
    \vspace{2mm} 
    \includegraphics[width=0.29\linewidth]{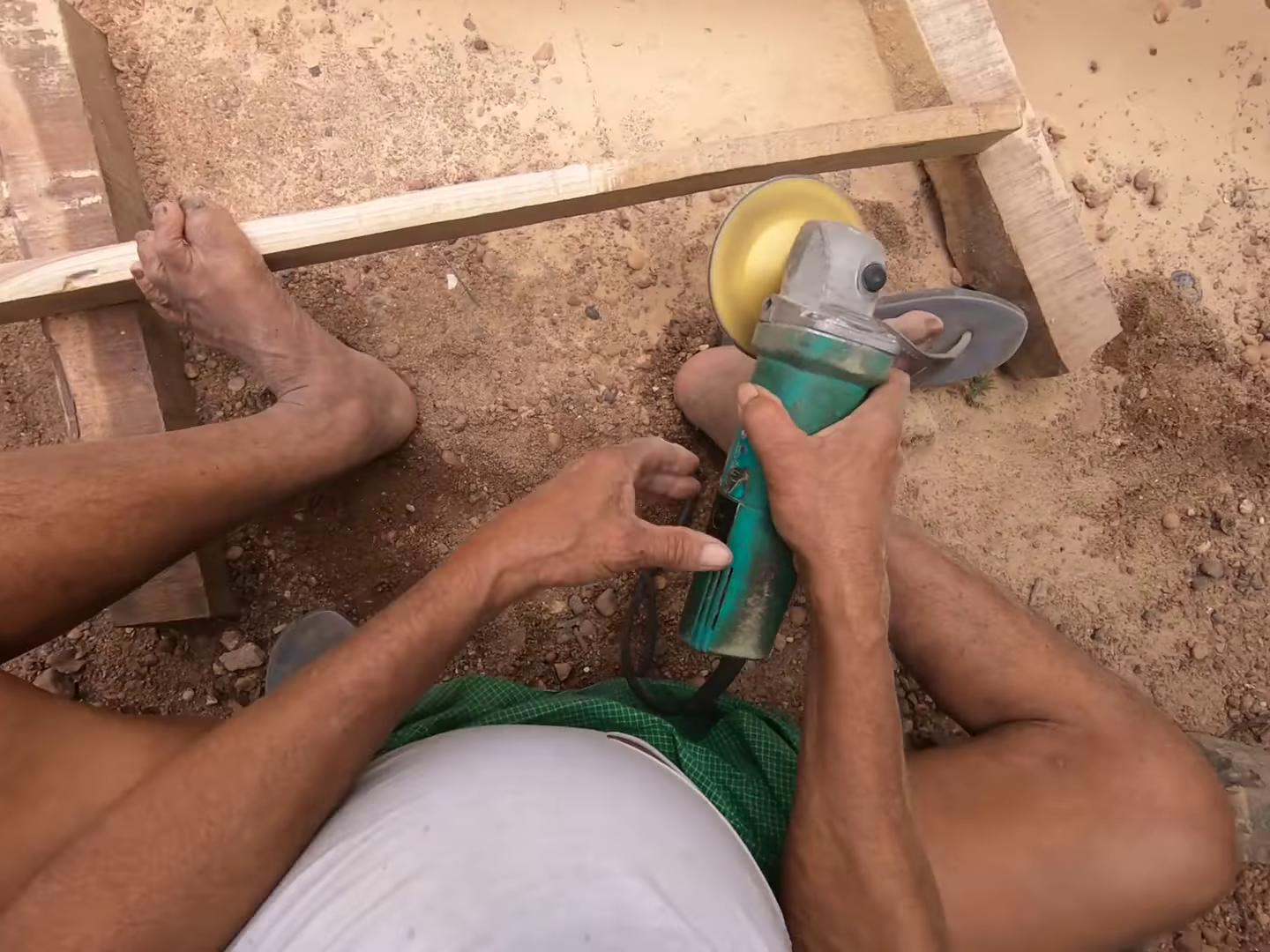}
    \includegraphics[width=0.29\linewidth]{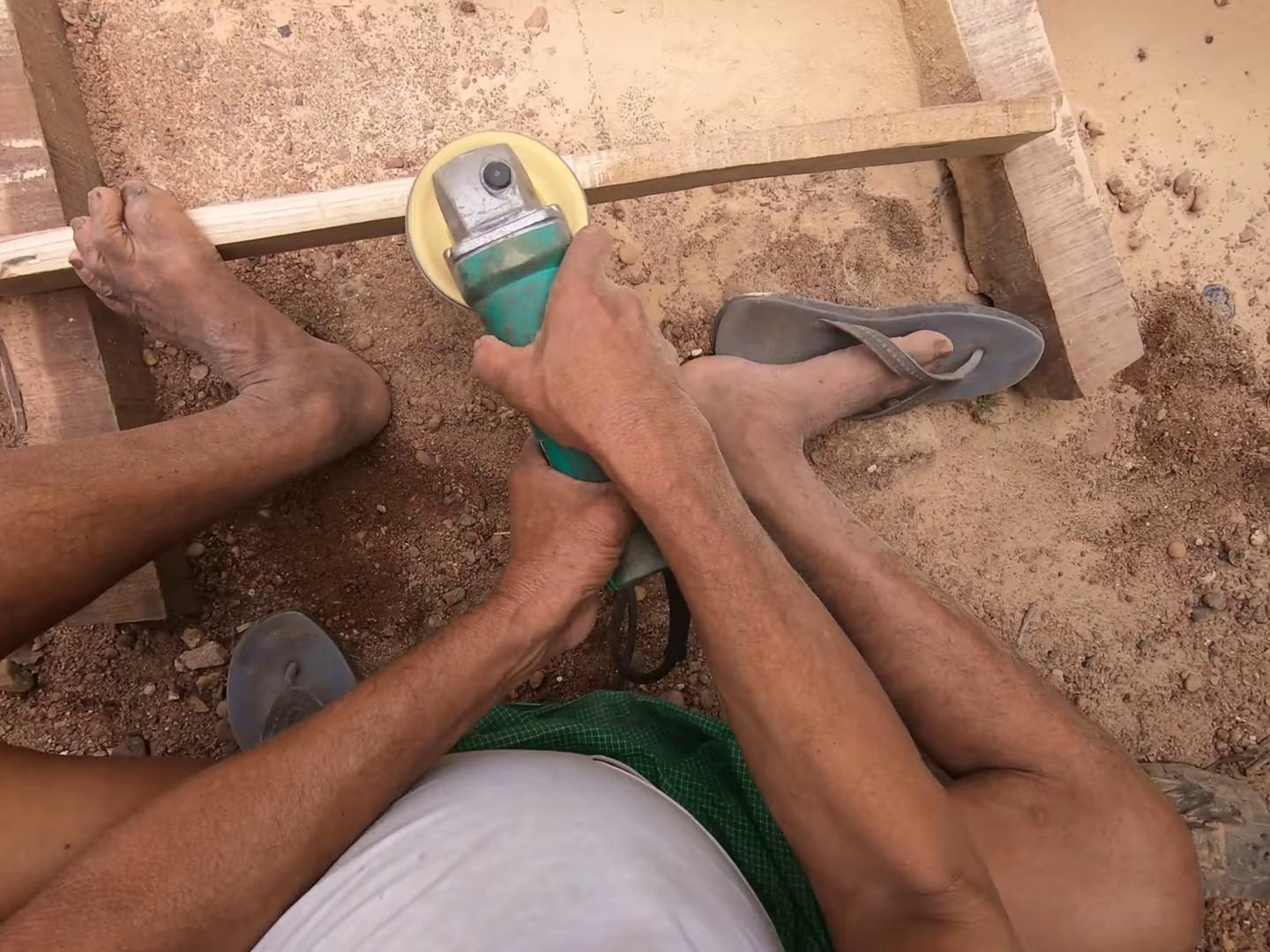}
    \includegraphics[width=0.29\linewidth]{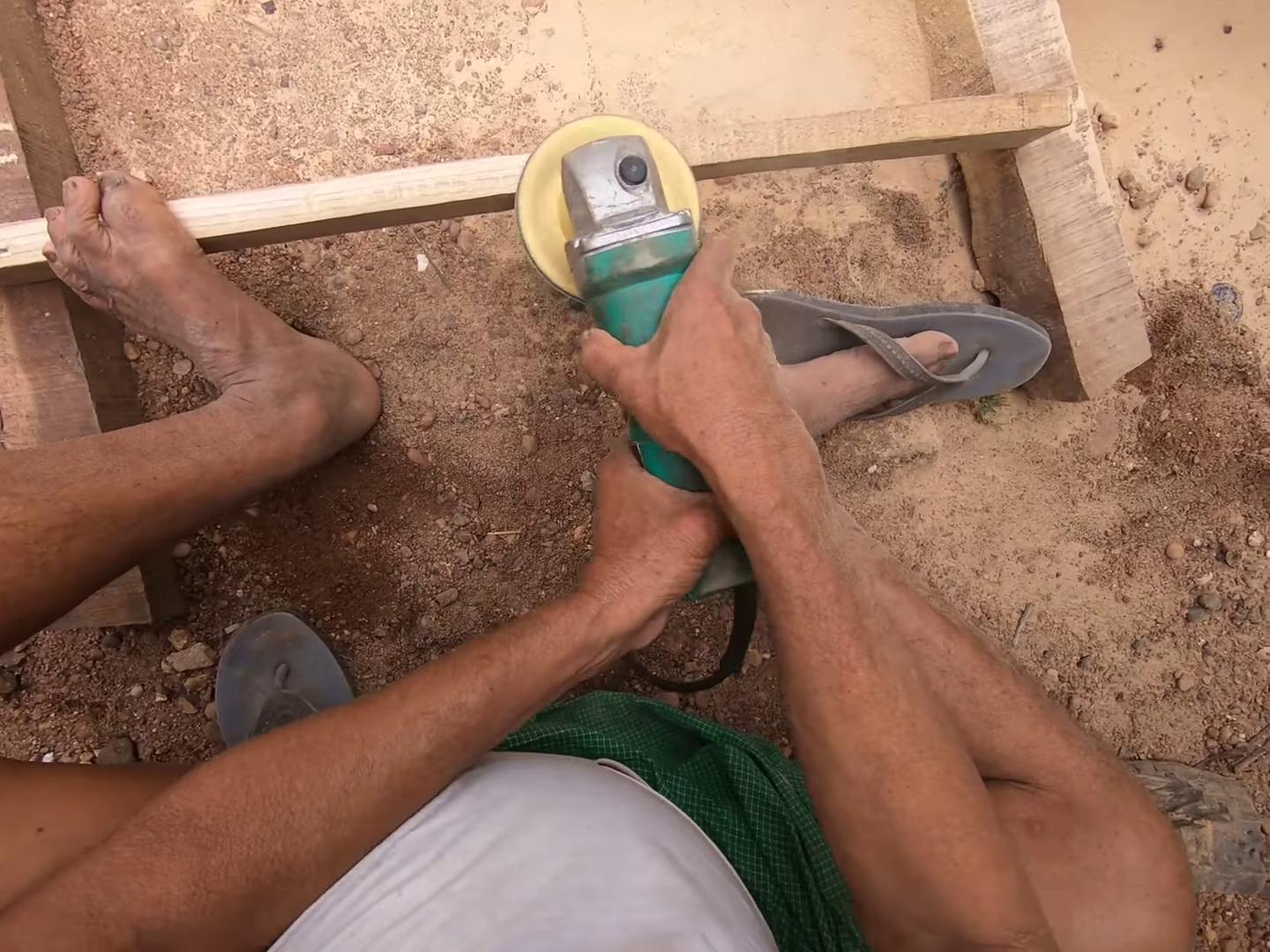}\\
    \vspace{-6mm}\textcolor{white}{\small \textbf{pre-condition \hspace{12mm} PNR \hspace{12mm} post-condition}}\\
    \vspace{2mm} State-change: Wood smoothed\\
    \caption{Hands and Objects: Example object state changes defined by pre-condition, PNR, and post-condition frames.}
    \label{fig:ho}
\end{figure}

\paragraph{Evaluation metrics and baselines}
Object state change temporal localization is evaluated using absolute temporal error measured in seconds.
Object state change classification is evaluated by classification accuracy. 
State change object detection is evaluated by average precision (AP). Appendix~\ref{appendix:hands-objects} details the annotations and presents baseline model results for the three Hands and Objects tasks.

\paragraph{Relation to existing tasks} Limited prior work considers object state change in photos~\cite{StatesAndTransformations,attributes:as:operators} or video~\cite{fathi-state-change,zhou-state-change,alayrac2017joint}; Ego4D is the first video benchmark dedicated to the task of understanding object state changes. The task is similar to action recognition (e.g., \cite{hussein2019timeception,tsn,zhou2015temporal,kazakos2019epic,ego-exo}) because in some cases a specific action can correspond to a specific state change. However, a single state change (\emph{e.g.,} cutting) can also be observed in many forms (various object-tool-action combinations). It is our hope that the proposed benchmarks will lead to the development of more explicit models of object state change, while avoiding approaches that simply overfit to action or object observations.

\subsection{Audio-Visual Diarization}\label{sec:avd}

\paragraph{Motivation}
Our next two tasks aim to understand the camera wearer's present interactions with \emph{people}. 
People communicate using spoken language, making the capture of conversational content  in business meetings and social settings 
a problem of great scientific and practical interest. While diarization has been a standard problem in the speech recognition community, Ego4D brings in two new aspects (1) simultaneous capture of video and audio (2) the egocentric perspective of a participant in the conversation. 

\begin{figure}[t] %
    \centering
      \includegraphics[width=\linewidth]{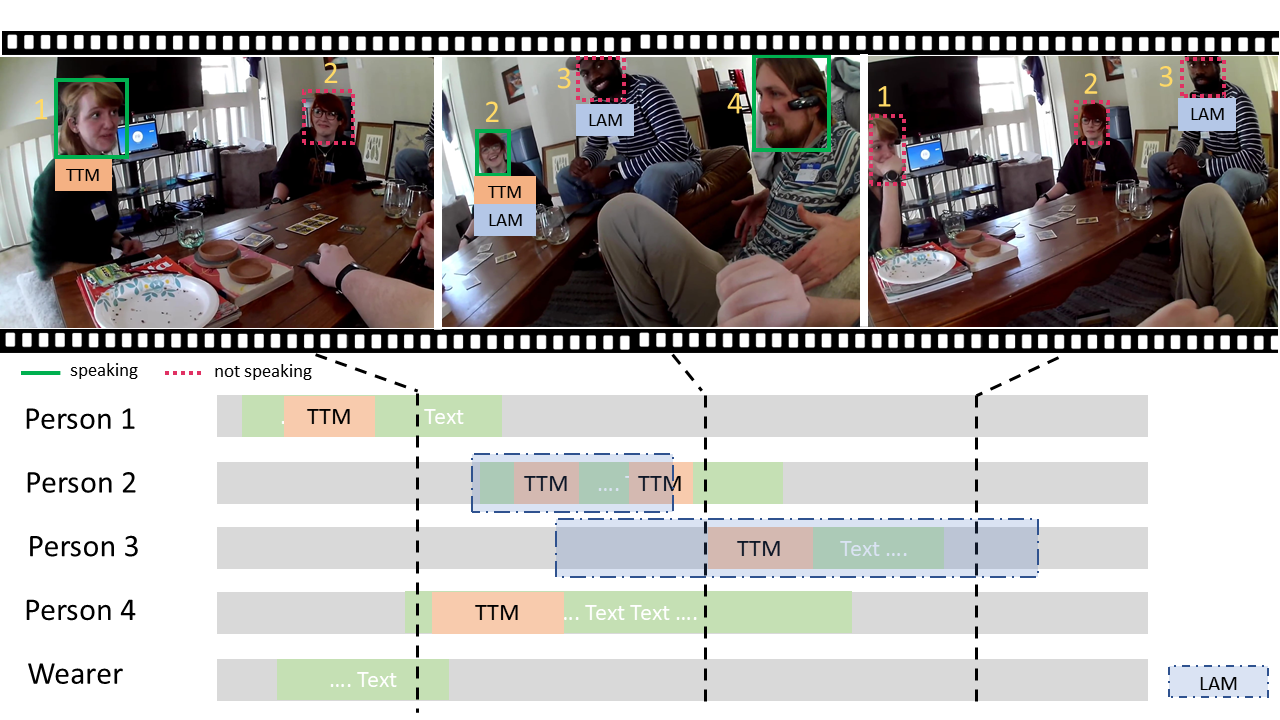}      
  \caption{\label{fig:AVD} Audio-Visual and Social benchmark annotations} 
\end{figure}

\paragraph{Task definition and annotations} \label{par:avd_tasks_def}
The Audio-Visual Diarization (AVD) benchmark is composed of four tasks
(see Figure \ref{fig:AVD}): %
\begin{itemize}[noitemsep,nolistsep,leftmargin=4mm]
    \item \textit{Localization and tracking} of the participants (i.e., candidate speakers) in the visual field of view (FoV). 
    A bounding box is annotated around each participant`s face.
    \item \textit{Active speaker detection} where each tracked speaker is assigned an anonymous label, including the camera wearer who never appears in the visual FoV. 
    \item \textit{Diarization} of each speaker's speech activity, where we provide the time segments corresponding to each speaker's voice activity in the clip. 
    \item \textit{Transcription} of each speaker's speech content (only English speakers are considered for this version). 
\end{itemize}

\paragraph{Evaluation metrics and baselines} \label{par:avd_tasks_eval}
We use standardized object tracking (MOT) metrics \cite{CLEARMOT, IdentityMOT}
to evaluate speaker localization and tracking in the visual FoV. 
Speaker detection with anonymous labels is evaluated using the speaker error rate, 
which measures the proportion of wrongly assigned labels. 
We adopt the well studied diarization error rate (DER) \cite{anguera2006robust} and word error rate (WER) \cite{klakow2002testing} for diarization and transcription, respectively.   We present AVD baseline models and results in Appendix~\ref{sec:appendix-av}.

\iffalse
\paragraph{Baseline Models:} \label{par:avd_tasks_eval}
A combination of per-frame detection of speakers, 
and long-term tracking using re-identification is used for speaker localization task.  
For active speaker recognition and speaker diarization, 
we use TalkNet \cite{tao2021someone}, and a more rudimentary audio-visual fusion network. 
Camera wearer's voice activity is detected using a simply 
amplitude thresholding of audio content. 
Refer to Section \ref{sec:avd-appendix} for complete details about the modeling framework. 
\fi

\paragraph{Relation to existing tasks} \label{par:avd_relatedwork}
The past few years have seen audio studied in computer vision tasks \cite{zhu2021deep} 
for action classification \cite{kazakos2019epic,audiovisual-slowfast}, object categorization \cite{lacheze2009audio, zhang2018audio}, 
source localization and tracking \cite{arandjelovic2017objects,Senocak_2019_PAMI,tian2018audio} and embodied navigation \cite{chen2019audio}. 
Meanwhile, visual information is increasingly used in historically audio-only tasks like speech transcription, 
voice recognition,  audio spatialization \cite{iwano2007audio, afouras2018deep, gao2019visualsound,morgadoNIPS18},  
speaker diarization \cite{avdiar, anguera2012speaker}, and source separation \cite{ephrat2018looking,gao-2018-separation,visual-voice}.
Datasets like VoxCeleb \cite{chung2020spot}, AVA Speech \cite{chaudhuri2018ava}, AVA active speaker \cite{roth2019ava}, AVDIAR \cite{avdiar}, and EasyCom \cite{donley2021easycom} support this research. 
However, these datasets are mainly non-egocentric. 
Unlike Ego4D, they do not capture natural conversational characteristics involving a variety of noisy backgrounds, overlapping, interrupting and un-intelligible speech,  environment variation, moving camera wearers, and speakers facing away from the camera wearer.

\subsection{Social Interactions}
\label{sec:social}

\paragraph{Motivation} 

An egocentric video provides a unique lens for studying social interactions because it captures utterances and nonverbal cues~\cite{Knapp2014} from each participant's unique view and enables embodied approaches to social understanding. Progress in egocentric social understanding could lead to more capable virtual assistants and social robots. Computational models of social interactions can also provide new tools for diagnosing and treating disorders of socialization and communication such as autism~\cite{Rehg2014}, and could support novel prosthetic technologies for the hearing-impaired. 

%

%

\iffalse
\begin{figure}[t]
    \centering
    \includegraphics[width=0.29\linewidth]{HO/activate1.png}
    \includegraphics[width=0.29\linewidth]{HO/burning1.png}
    \includegraphics[width=0.29\linewidth]{HO/cementdeposit1.png}\\
    \includegraphics[width=0.29\linewidth]{HO/construct1.png}
    \includegraphics[width=0.29\linewidth]{HO/deactivate1.png}
    \includegraphics[width=0.29\linewidth]{HO/deconstruct1.png}%
%
%
%
    \caption{\KGnote{PLACEHOLDER FIGURE - SOCIAL (or merge fig with AVD)}}
    \label{fig:ho}
\end{figure}
\fi

\paragraph{Task definition} 

While the Ego4D dataset can support such a long-term research agenda, our initial Social benchmark focuses on multimodal understanding of conversational interactions via attention and speech. %
Specifically, we focus on %
identifying communicative acts that are directed towards the camera-wearer, as distinguished from those directed to other social partners:
(1) \emph{Looking at me (LAM):} given a video in which the faces of social partners have been localized and identified, classify whether each visible face is looking at the camera wearer; and (2) \emph{Talking to me (TTM):} given a video and audio segment with the same tracked faces, classify whether each visible face is talking to the camera wearer. %
%
%
%

\iffalse %
\paragraph{Data collection} 
Data was collected at four sites under a variety of protocols including both highly-structured and unstructured capture settings. Participants consisted of friends and family groups and data was captured in residences and local neighborhoods, ensuring naturalistic interactions. Capture hardware included wearable cameras, wearable eye trackers, binaural audio recorders, and smart watches. The details are in Appendix~\ref{appendix:social-collection}.
\fi

%
%

%

\paragraph{Annotations} 

Social annotations build on those from AV diarization (Sec.~\ref{sec:avd}). Given (1) face bounding boxes labeled with participant IDs and tracked across frames, and (2) associated active speaker annotations that identify in each frame whether the social partners whose faces are visible are speaking, annotators provide the ground truth labels for LAM and TTM as a binary label for each face in each frame. 
For LAM, annotators label the time segment (start and end time) of a visible person when the individual is looking at the camera wearer. For TTM, we use the vocal activity annotation from AVD, then identify the time segment when the speech is directed at the camera wearer.   See Figure~\ref{fig:AVD}.

\paragraph{Evaluation metrics and baselines} 
We use mean average precision (mAP) and Top-1 accuracy to quantify the classification performance for both tasks. Unlike AVD, we measure precision at every frame. %
Appendix~\ref{appendix:social-tasks} provides details and presents Social baseline models and results.

\iffalse
\paragraph{Baselines and Results}

The LAM baseline is an Bidirectional-LSTM-based architecture with a ResNet-18 backbone that processes sequences of input face bounding boxes. This model yields an mAP of 0.69 and Top-1 Acc of 0.87. The TTM baseline utilizes the LAM video baseline model to obtain a video embedding. This is combined with an audio embedding obtained by extracting MFCC coefficients from the audio stream and passing them into a ResNet-18. The audio and video embeddings are then concatenated and passed through an FC layer to yield the final output. This model yields an mAP of 0.54 and Top-1 Acc of 0.58. See Appendix~\ref{appendix:social-baselines} for more details.

\fi
%
%

\paragraph{Relation to existing tasks}
Compared to~\cite{Fathi2012social}, Ego4D contains substantially more participants, hours of recording, and variety of sensors and social contexts. The LAM task is most closely related to prior work on eye contact detection in ego-video~\cite{Chong2020eyecontact,Mitsuzumi2017DEEPEC}, but addresses more diverse and challenging scenarios. Mutual gaze estimation~\cite{marin2014detecting,Park2012,palmero2018automatic,marin2011here,marin2019laeo,DoostiBoosting} and gaze following~\cite{Fang2021,Chong2020attended,Kellnhofer2019,recasens2015they} are also relevant. The TTM task is related to  audio-visual speaker detection~\cite{Roth2020,Afouras2020} and meeting understanding~\cite{Beyan2018,Lepri2012,McCowan2005}.

\subsection{Forecasting}\label{sec:forecasting}

\paragraph{Motivation} 

Having addressed the past and present of the camera wearer's visual experience, our last benchmark moves on to anticipating the future.
Forecasting movements and interactions requires comprehending the camera wearer's \emph{intention}.  
It has immediate applications in AR and human-robot interaction, such as anticipatively turning on appliances or moving objects for the human's convenience.  
The scientific motivation can be seen by analogy with language models such as GPT-3~\cite{gpt3}, which implicitly capture knowledge needed by many other tasks. Rather than  predict the next word, visual forecasting models the dynamics of an agent acting in the physical world. 

\paragraph{Task definition} 

\begin{figure}[t]
    \centering
    \includegraphics[width=\linewidth]{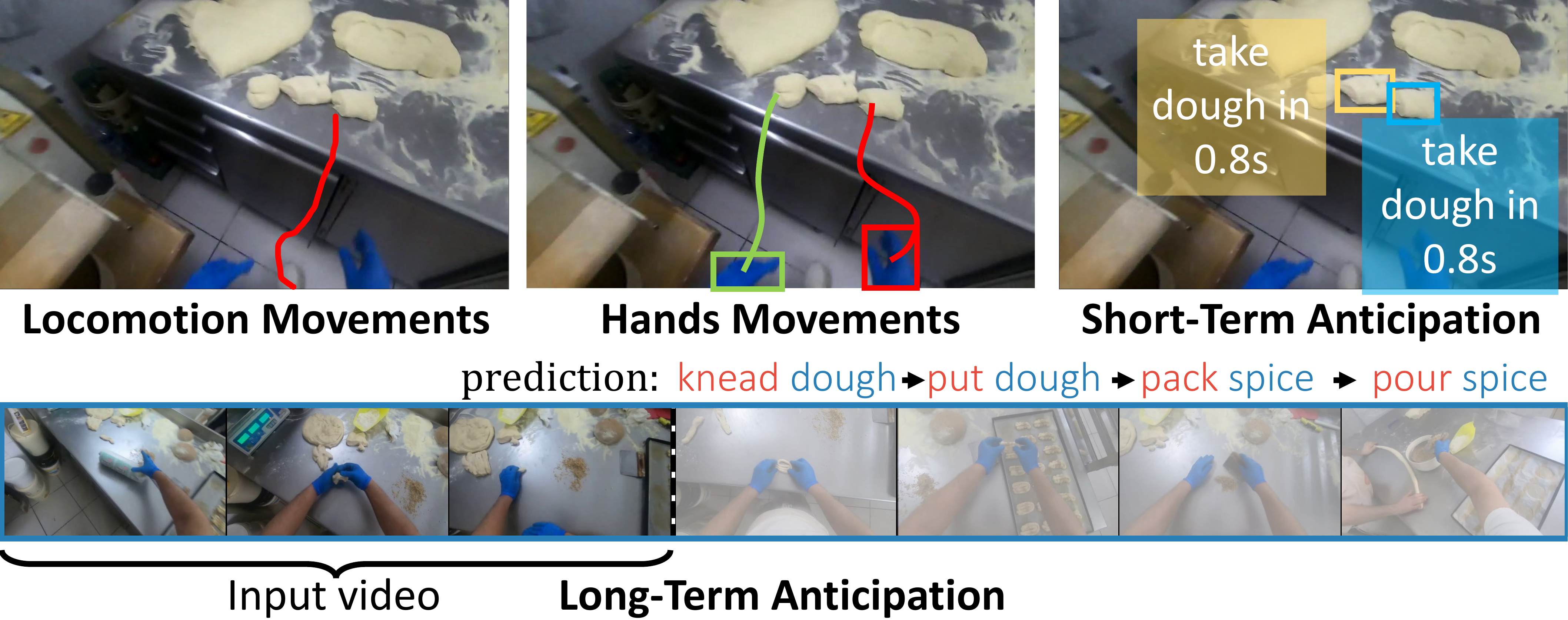}
    \vspace*{-0.15in}
    \caption{The Forecasting benchmark aims to predict future locomotion, movement of hands, next object interactions, and sequences of future actions. %
    }
    \label{fig:forecasting}
\end{figure}

The Forecasting benchmark includes four tasks (Fig.~\ref{fig:forecasting}):
(1) \emph{Locomotion  prediction}: 
predict a set of possible future ground plane trajectories of the camera wearer.
(2) \emph{Hand movement prediction}: 
predict the hand positions of the camera wearer in  future frames. %
(3) \emph{Short-term object interaction anticipation}: %
detect a set of possible future interacted objects in the most recent frame of the clip. To each object, assign a verb indicating the possible future interaction and a “time to contact" estimate of when the interaction is going to begin. 
(4) \emph{Long-term action anticipation}: 
predict the camera wearer's future sequence of actions. %

\paragraph{Annotations}

Using the narrations, we identify the occurrence of each object interaction, assigning a verb and a target object class. The verb and noun taxonomies are seeded from the narrations and then hand-refined. For each action, we identify a contact frame and a pre-condition frame %
in which we annotate bounding boxes around active objects. The same objects as well as hands are annotated in three frames preceding the pre-condition frame by $0.5s$, $1s$ and $1.5s$.  %
 We obtain ground truth ego-trajectories of the camera wearer using structure from motion. 

\paragraph{Evaluation metrics and baselines} 
We evaluate future locomotion movement and hand movement prediction using L2 distance.
 Short-term object interaction anticipation is evaluated using a Top-5 mean Average Precision metric which discounts the Top-4 false negative predictions.  Long-term action anticipation is evaluated using edit distance. %
Appendix~\ref{appendix:forecasting} details the tasks, annotations, baseline models, and results.

\iffalse
\paragraph{Baselines and results} 

The baseline for future locomotion movement prediction uses AlexNet for feature extraction and KNN~\cite{park-cvpr2016} 
%
for trajectory retrieval. \textcolor{red}{This achieves a performance of 0.73m/2.73m 1-MTE for 1 second and 5 second predictions, respectively; 0.53m/2.11m 3-MTE; 0.47m/1.87m 5-MTE.}  
%
The baseline for future hands movement prediction uses an I3D encoder and a regression module to predict future hand positions in the key frames from the input video.
\textcolor{red}{This achieves Mean Key Frame Displacement Error of ($64.28/61.18$)  and Contact Key Frame Displacement Error of ($66.75/63.71$) for left/right hand position prediction.} 
The baseline for short-term object interaction anticipation uses Faster RCNN to detect next-active objects and SlowFast to predict verb and time to contact for each of them.
\textcolor{red}{This achieves an overall Top-5 mAP of 1.75\%.}
The baseline for long-term action anticipation uses a SlowFast encoder, a transformer based aggregator and multiple decoder heads to predict future action sequences from the input video.
\textcolor{red}{This achieves ED@20 of 0.741 (verbs), and 0.784 (nouns) when predicting 20 future actions.}
%
%
See Appendix~\ref{appendix:forecasting_baselines} for a more detailed description of baselines and Appendix~\ref{appendix:forecasting_results} for the discussion of the results.
\fi

\paragraph{Relation to existing tasks}
Predicting future events from egocentric vision has increasing interest~\cite{rodin2021predicting}.
Previous work considers future localization~\cite{krishna-wacv2016,park:2016_future,kitani:2012,Yagi_2018_CVPR}, action anticipation~\cite{lan2014hierarchical,vondrick2016anticipating,gao2017red,furnari2019rulstm,koppula2016anticipating,avt}, next active object prediction~\cite{bertasius2016first,furnari2017next}, future event prediction~\cite{mahmud2017joint,neumann2019future}, and future frame prediction~\cite{liu2018future,mathieu2015deep,xingjian2015convolutional,lotter2016deep,van2017transformation,villegas2017decomposing}.  Whereas past work relies on different benchmarks and task definitions, we propose a unified benchmark to assess progress in the field. %

\section{Conclusion}

Ego4D is a first-of-its-kind dataset and benchmark suite aimed at advancing multimodal perception of egocentric video.  Compared to existing work, our dataset is orders of magnitude  larger in scale and diversity. The data will allow AI to learn from daily life experiences around the world---seeing what we see and hearing what we hear---while our benchmark suite provides solid footing for innovations in video understanding that are critical for augmented reality, robotics, and many other domains.  We look forward to the research that will build on Ego4D in the years ahead.

\iftoggle{arxiv}{
\section*{Contribution statement}

Project led and initiated by Kristen Grauman.   Program management and operations led by Andrew Westbury.   Scientific advising by Jitendra Malik.  %
Authors with stars ($^\ast$) were key drivers of implementation, collection, and/or annotation development throughout the project.  Authors with daggers ($^\dagger$) are faculty PIs and working group leads in the project.  
The benchmarks brought together many researchers from all institutions including cross-institution baseline evaluations. 
Appendices~\ref{sec:episodic-appendix} through~\ref{appendix:forecasting} detail the contributions of individual authors for the various benchmarks.
The video collected by Facebook Reality Labs 
used Vuzix Blade\textregistered~Smart Glasses and
was 
done in a closed environment in Facebook's buildings by paid participants who signed consents to share their data.  All other video collection and participant recruitment was managed by the university partners.  Appendix~\ref{sec:collection} provides details about the data collection done per site and acknowledges the primary contributors. The annotation effort was led by Facebook AI. 

\section*{Acknowledgements}

We gratefully acknowledge the following colleagues for  valuable discussions and support of our project:
Aaron Adcock, Andrew Allen, Behrouz Behmardi, Serge Belongie, Antoine Bordes, Mark Broyles, Xiao Chu, Samuel Clapp, Irene D’Ambra, Peter Dodds, Jacob Donley, Ruohan Gao, Tal Hassner,  Ethan Henderson, Jiabo Hu, Guillaume Jeanneret, Sanjana Krishnan, Devansh Kukreja, Tsung-Yi Lin, Bobby Otillar, Manohar Paluri, Maja Pantic, Lucas Pinto, Vivek Roy, Jerome Pesenti, Joelle Pineau, Luca Sbordone, Rajan Subramanian, Helen Sun, Mary Williamson, and Bill Wu.  
We also acknowledge Jacob Chalk for setting up the Ego4D AWS backend and Prasanna Sridhar for developing the Ego4D website.
Thank you to the Common Visual Data Foundation (CVDF) for hosting the Ego4D dataset. 

\srt{The universities acknowledge the usage of commercial software for de-identification of video. brighter.ai was used for redacting videos by some of the universities. Personal data from the University of Bristol was protected by Primloc’s Secure Redact software suite.

UNICT is supported by MIUR AIM - Attrazione e MobilitaInternazionale Linea 1 - AIM1893589 - CUP E64118002540007. Bristol is supported by UKRIEngineering and Physical Sciences Research Council (EPSRC) Doctoral Training Program (DTP), EPSRC Fellowship UMPIRE (EP/T004991/1).  
KAUST is supported by the KAUST Office of Sponsored Research through the Visual Computing Center (VCC) funding.
National University of Singapore is supported by Mike Shou’s Start-Up Grant.
Georgia Tech is supported in part by NSF award 2033413 and NIH award R01MH114999.}
}  %
{}

\clearpage

\appendix

\addcontentsline{toc}{section}{Appendix} %
\part{Appendix} %
\parttoc %

\addcontentsline{toc}{section}{Appendices}
\renewcommand{\thesubsection}{\Alph{subsection}}

\subsection{Data Collection}\label{sec:collection}

This section overviews the collection procedures and scenarios per site. \iftoggle{arxiv}{}{\srt{Each site is a university or lab that is part of the Ego4D consortium.  To preserve anonymity, here we refer to the sites only by country and/or U.S. state.  We inquired with the CVPR 2022 Program Chairs, and they confirmed that this is appropriate per the anonymity policy of CVPR.}}

\iftoggle{arxiv}{
\paragraph{International Institute of Information Technology (IIIT), 
Hyderabad, India:}}
{
\paragraph{India:}}
\iftoggle{arxiv}{At IIIT, Hyderabad, 
we}
{We}
followed a protocol of distributed data
collection with a centralized team doing coordination and verification.
We first identified local coordinators in different parts of the
country and explained the data collection plans, goals and process.
They then helped in collecting data in their own local regions from
natural settings with informed participants. Participants were
recruited locally considering the range of activities, and also the
guidelines and restrictions of COVID-19. The central team could not travel 
to all these locations for training the coordinators or collecting
the data. We shipped multiple cameras to the local coordinators and 
remotely guided them on data collection following the COVID protocols.
The collected data and consent forms were then shipped back to %
the university, where manual verification, de-identification (wherever
applicable), and sharing with the consortium took place.

At IIIT Hyderabad, we recorded 660.5 hours of data with the help of 138 subjects. The videos were collected in 5 different states in India, geographically well apart. We cover 36 different scenarios, such as making bricks using hands, knitting, making egg cartons, and hairstyling. The age of subjects ranged from 18-84 years with 10 distinct professional backgrounds (teachers, students, farmers, blacksmiths, homemakers, etc.). Out of all the subjects, 94 were males, and 44 were females. We use GoPro Hero 6 and GoPro Hero 7 for recording the videos. The GoPro’s were shipped to the participants in different parts of the country. Videos were shared back either in external hard disks or over the cloud storage. Each video was manually inspected for any sensitive content before sharing.

\iftoggle{arxiv}{
Primary contributors: Raghava Modhugu - data collection pipeline, design of the setup and workflow.
Siddhant Bansal - IRB application, consent forms and de-identification. C. V. Jawahar - lead contributor for data collection.  
We also acknowledge the contributions of Aradhana Vinod (coordination and communication), Ram Sharma (local data management and verification), and Varun 
Bhargavan (systems and resources).
}
{}

\iftoggle{arxiv}{
\paragraph{University of Tokyo, Japan:}}
{\paragraph{Japan:}}
We recruited 81 Japanese participants (41 male, 40 female) \iftoggle{arxiv}{living around Tokyo, Japan}{} through a temporary employment agency. The participant's gender and age (from the 20s to 60s) were balanced to collect diverse behavior patterns. We focused on two single-actor activities: cooking (40 participants, 90 hours) and handcraft (41 participants, 51 hours). In the cooking scenario, participants were asked to record unscripted videos of cooking at their homes. In the handcraft scenario, participants visited our laboratory and performed various handcraft activities (e.g., origami, woodworking, plastic model, cutout picture). We collected data using GoPro HERO 7 Black camera for cooking and Weeview SID 3D stereo camera for handcraft. Our data collection protocol was reviewed and approved by \iftoggle{arxiv}{University of Tokyo}{the university's} ethical review board.

\iftoggle{arxiv}{
Primary contributors: Yoichi Sato – lead coordinator for data collection, Takuma Yagi and Takumi Nishiyasu – contributed to participant recruiting, protocol design, data collection and inspection, and IRB submission, Yifei Huang and Zhenqiang Li – contributed to data inspection and transfer, Yusuke Sugano – contributed to selecting video recording scenarios, protocol design and IRB submission.}
{}

\iftoggle{arxiv}{\paragraph{University of Bristol, UK:}}
{\paragraph{United Kingdom:}}
Participants were recruited through adverts on social media and university internal communication channels.
These participants then spread the word to their acquaintances and some participants joined the project through word-of-mouth recommendations of previous participants.
Data was collected between Jan and Dec 2020, from 82 participants.
With the pandemic taking over in March, the project shifted to online operation where cameras were posted, and training took place over Zoom meetings.
Participants first expressed interest by sending an email and they were provided with an information sheet.
This was followed by a preliminary Zoom meeting with a researcher to brief participants about the procedure, answer any questions and agree on the scenarios to be recorded.

We set a limit to the total number of minutes per scenario, to increase diversity of recordings. For example, driving cannot be longer than 30 minutes while cooking can be up to 1.5 hours.
Each participant was instructed to record a minimum of 2 hours across 4 scenarios.
Importantly, participants were encouraged to collect activities they naturally do.  For example if one regularly cycles or practices music, they were asked to record these scenarios.
Additionally, paired scenarios (people cooking together or playing games) were encouraged and multiple (2-3) cameras were posted for participants sharing a household.
All participants signed a consent form before a camera was posted to their residence.
Cameras were posted to 9 UK cities in England, Wales and Scotland including one participant in the Isle of North Uist.

Upon receipt of the camera, a second Zoom meeting was scheduled to train the participant on the equipment and detail how footage is reviewed and uploaded.
Participants were given 2 weeks to record, with an additional week of extension upon request.
Once recording is completed, footage is uploaded by the participant and reviewed for good lighting, correct setting and viewpoint.
Participants were reimbursed for their participation in the project. 

Scenarios recorded in the UK covered: commuting (driving, walking, cycling, taking the bus, hiking, jogging), entertainment (card games, board games, video games, lego, reading, practising a musical instrument, listening to music, watching TV), jobs (lab work, carpentry), sports (football, basketball, climbing, golf, yoga, workouts) and home-based daily activities (cooking, cleaning, laundry, painting, caring for pets, tidying, watering the plants), DIY (fixing, gardening, woodwork) and crafts (colouring, crafting, crochet, drawing, knitting, sewing).
Footage was captured using GoPro Hero-7, Hero-8 and Vuzix.

Footage was then reviewed by researchers to identify any PII. 
36\% of all videos required de-identification.
We used Primloc's Secure Redact software suite, with integrated tools and user interfaces for manual tracking and adjusting detections.
Redacted recordings were reviewed manually, then encoded and uploaded to the AWS bucket.
During encoding, IMU meta data was separately extracted. 
Integrated audio and video using native 50fps recordings are available.

In total, 262 hours were recorded by 82 participants. 
On average, each participant recorded 3.0 hours ($\sigma=0.7$ hours)
The data is published under General Data Protection Regulation (GDPR) compliance.

\iftoggle{arxiv}{
Primary contributors:  Michael Wray - data collection, consent forms and information sheets; Jonathan Munro - data collection and ethics application; Adriano Fragomeni - data collection and de-identification oversight; Will Price - data ingestion, encoding and metadata; Dima Damen - scenarios, procedures, data collection oversight and participant communication. We acknowledge the efforts of Christianne Fernee in manually reviewing all data. 
}{}

\iftoggle{arxiv}{\paragraph{Georgia Tech, Atlanta, GA, USA:}}
{\paragraph{Georgia, USA:}}
Participant groups from the \iftoggle{arxiv}{Atlanta, Georgia, USA metro area}{local area} were recruited via online posts and advertisements on sites such as Facebook, Reddit, and Instagram. Each group of participants was comprised of friends or family members who knew each other prior to participating in the study. Participants were required to be aged 18-64, to not be considered high risk for COVID-19, and to be able to play social deduction games in English. Our study protocol was reviewed and approved by the \iftoggle{arxiv}{Georgia Tech}{university} Institutional Review Board (IRB).
In total, approximately 43 hours of egocentric video were collected from 19 participants (per participant disclosure - 10 male, 7 female, 1 non-binary, 1 not reported). Participants had a mean age of 31.6 years with 7 participants aged 20-29 years, 10 participants aged 30-39 years, and 2 participants aged 40-49 years.

Participants wore an egocentric head-worn camera and on-ear binaural microphones. 
Some participants wore the ORDRO EP6 camera while others wore the Pupil Invisible cameras. %
The audio was recorded using a Tascam DR-22WL and Sound Professionals MS-EHB-2 Ear-hook binaural microphones. %
A third-person video was also captured via a Logitech C930e Webcam. Participants wore the provided recording devices while eating, drinking, and playing social deduction games such as \emph{One Night Ultimate Werewolf} and \emph{The Resistance: Avalon} in their own home. This at-home game-night setting elicited a wide range of spontaneous and naturalistic social behaviors and interactions. In addition, eating and drinking behaviors were captured from both the egocentric and third-person cameras.

In addition to participating in the recorded session, participants completed 
a survey that captured their demographic information. 
All data was screened and censored by study personnel to remove any identifying information including visible personal information on their phone screens or the exterior of the home. Participants also had the opportunity to review the videos and request additional censoring. 

\iftoggle{arxiv}{Primary contributors:  Fiona Ryan - lead coordinator for data collection, including synchronization, de-identification, and ingestion; Audrey Southerland - lead coordinator for IRB development and recruiting; Miao Liu - contributed to data collection and ingestion; James M. Rehg - contributed to protocol design and data collection.}{}

\iftoggle{arxiv}{
\paragraph{Indiana University, Bloomington, IN, USA:}}
{\paragraph{Indiana, USA:}}
Participants in the \iftoggle{arxiv}{Bloomington, Indiana, USA}{local} area were recruited through advertisements on social media, online classifieds boards, and email lists. We also used snowball sampling by asking participants to share our ads with their friends. We recruited participants who were willing to perform interactive small group activities such as playing sports, playing board or card games, playing musical instruments, assembling puzzles, etc. The health of participants and study personnel was safeguarded by collecting data either outdoors (where people can more safely interact without wearing masks), or indoors in the homes of the participants. In either case, we initially required that all participants in a social group be part of the same household to minimize the risk of spreading disease between households, but later we allowed groups of people who were comfortable interacting with one another (e.g., because they are vaccinated for COVID-19). Group sizes ranged from 1 to 6 people, with groups of 2 or 3 being the most common.

We collected data with four different devices: zShade 1080p camera glasses, iVue Rincon 1080 camera glasses, ORDRO EP-6, and Pupil Labs Invisible camera and gaze tracking glasses. We used multiple devices because each has various advantages and disadvantages; zShade has a large horizontal field of view, for example, while iVue has an adjustable vertical field of view, ORDRO sits by the ear and is mounted on a headband which works well for people wearing prescription glasses, and Invisible offers gaze tracking but is very expensive. We asked as many participants as possible in the group to wear cameras. We primarily used our two Pupil Labs Invisibles whenever possible, because of their ease of use and ability to collect gaze data, but we also used the ORDRO EP-6 when there were larger groups or when participants wore prescription glasses.

Our protocol was reviewed and approved by the \iftoggle{arxiv}{Indiana University}{university's} Institutional Review Board (IRB).
We first conducted an online meeting with potential participants to describe the study, explain the use of the cameras, agree on an activity for them to perform, and answer their questions. We ask participants to try to limit capture of potentially privacy-sensitive content by choosing a place within their home that did not have personally identifiable information, by avoiding recording people other than those participating in the study, and by avoiding saying last names or other sensitive audio. 

We then arrange a time to meet them, typically outside their home or in an outdoor public place. We set up the cameras, help the participants put them on, give them our contact information in case they have any problems, and then we leave while they perform the activity. We then return after about one hour to pick up the cameras. Within a few days, we send each participant a copy of the video taken by their camera, and ask them to review the footage and identify any privacy-sensitive content (video or audio) that they would prefer to be blurred or removed. We manually edit out any such content (using Adobe Premiere Pro). We also review all video for faces of non-participants and personally-identifying information such as house numbers or license plates, and blurred these accordingly. We use Pupil Labs software to synchronize eye gaze with the video for each participant, and then used Adobe Premiere Pro to temporally synchronize video across different participants using audio track comparison.

In total, approximately 103 hours of video were collected from 66 participants (42 female, 23 male, 1 non-binary; for age, 46 were 20-29 years old, 14 were 30-39 years old, 1 was 40-49, 2 were 50-59, 1 was 60-69, and 2 were 70-79).

\iftoggle{arxiv}{Primary contributors: David Crandall - lead coordinator for data collection; Yuchen Wang - contributed to protocol design, participant recruiting, and data collection; Weslie Khoo -  
developed multi-camera synchronization and de-identification pipelines.}{}

\iftoggle{arxiv}{
\paragraph{University of Minnesota, Twin Cities, MN, USA:}}
{\paragraph{Minnesota, USA:}}
Participants in the \iftoggle{arxiv}{Minneapolis and St. Paul, Minnesota, USA}{local} area were recruited through advertisements on social media and university bulletins such as Facebook AD, Craiglist, and Redhat. A total of approximately 313 hours of data was collected from 45 participants (22 males and 23 females). Age groups include 5 teenagers, 20 people in their twenties, 11 people in their thirties, 8 people in their forties, and 1 person in their fifties. We recruited participants as multiple groups and encouraged them to engage in unstructured natural social interactions. Such interactions included playing card games, talking in the kitchen while cooking, playing basketball, and building a tent at a camp site.
In all cases, we required that all participants in a social group be part of the same household to minimize the COVID-19 risk. Group sizes ranged from 1 to 6 people, with groups of 2 or 3 being the most common.

We collected data with the zShade 1080p camera glasses that have a large field of view. Our protocol was reviewed and approved by the \iftoggle{arxiv}{University of Minnesota}{university's} Institutional Review Board (IRB). We first conducted an online meeting with potential participants to describe the study, explain the use of the cameras, agree on an activity for them to perform, and answer their questions. We then arranged a time for them to receive the cameras and provided them with a postage-paid box for camera return. A few days later, participants shipped the cameras to our designated return address. We downloaded the data after sanitizing cameras and equipment. After the data capture was complete, we visually inspected every second of video in order to exclude any privacy-sensitive information (e.g. license plates, smart phone screens, and credit card numbers), and to assess the duration of non-social activities.  \srt{For incidental participants (i.e. bystanders) appearing in data collected by the camera wearer in public settings (e.g., shopping, concert, at a park, etc.), data collection consists only of recording publicly observable behavior with no manipulation or direct interaction with the participants, and this university's IRB allows an assumed waiver of consent for those participants.}

\iftoggle{arxiv}{
Primary contributors: Hyun Soo Park - lead coordinator for data collection; Jayant Sharma - contributed to participant recruiting, data collection, IRB submission, analysis, and data ingestion.}{}

\iftoggle{arxiv}{
\paragraph{National University of Singapore, Singapore:}}
{\paragraph{Singapore:}}
Participants were recruited from Singapore through advertisements on social media, via flyers and surveys, as well as from sourcing by the project coordinator. Residents of Singapore aged 21 to 70 who could wear a camera while  participating in social sessions were eligible for inclusion in our study. During the recording session, the participants were required to attend social events such as family gatherings, exercising with a trainer, hairdressing, getting manicure, attending a session for teaching assistants, attending a group meeting, etc. The devices used for data collection were GoPro Hero 8, GoPro Hero 9, and AR glasses. GoPro cameras have binaural microphones while the AR glasses can only record mono audio. In total, 51 hours of videos were collected from 40 participants (25 males and 15 females). Age groups include 31 twenties, 5 thirties, 3 fifties, and 1 sixties. 

\iftoggle{arxiv}{
Primary contributors: Mike Zheng Shou - lead coordinator for data collection; Eric Zhongcong Xu - contributed to data collection; Ruijie Tao - contributed to data collection.}{}

\begin{figure*}[t]
\centering
\includegraphics[width=\textwidth]{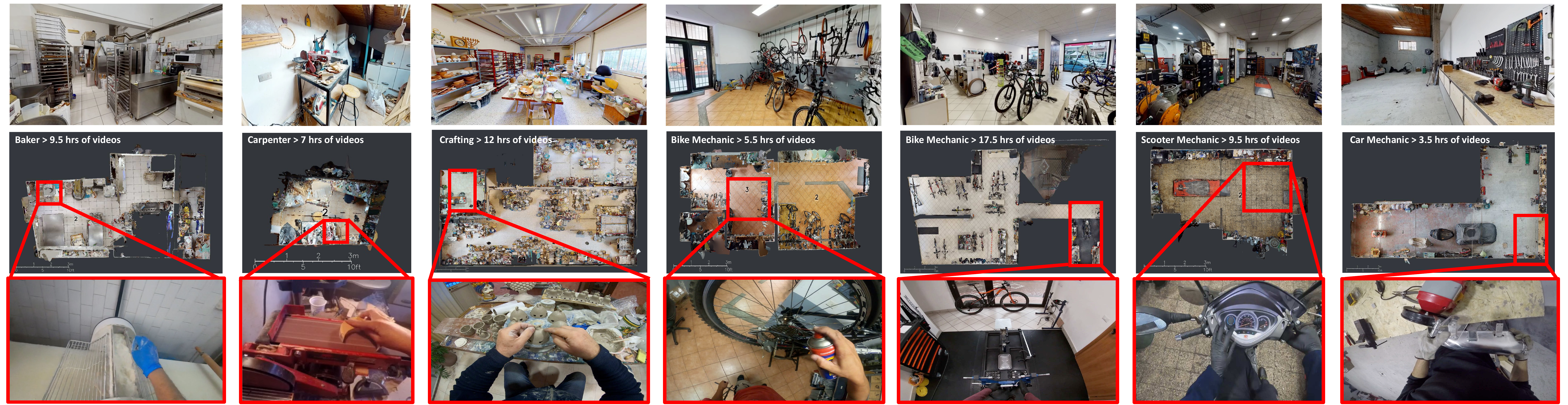}
\caption{Matterport3D scans (top) related to seven different  locations coupled with some videos (bottom).}\label{fig:matterport_appendix}
\end{figure*}

\iftoggle{arxiv}{
\paragraph{Facebook Reality Labs (FRL), Redmond, WA, USA:}}
{\paragraph{Washington, USA:}}
Participants were recruited from the \iftoggle{arxiv}{Seattle}{local} area\iftoggle{arxiv}{ through a FRL-hired vendor company.}{.} In total, there were 400 hours collected from 206 unique participants %
in %
6 scenes staged \iftoggle{arxiv}{in FRL’s research labs}{} in 2019. The ethnic groups include 50.8\% Caucasian, 28.2\% African, 11.9\% Asian and 9\% Hispanic. %
The staged environments include four types of apartments, a clothing store, and a grocery store. During the recording sessions, the participants were asked to wear Vuzix glasses to go through the following everyday scenarios as naturally as possible: grocery shopping, buying clothes, watching TV, playing video games, listening to music, dancing, weight lifting, stretching, reading email, paying bills, online gaming, cooking, talking with other people, meetings, whiteboarding, and video calling.  The emails and bills were always mock data, not personal emails or bills of the participants.  The video calls took place between participants only.

Three out of four apartments have corresponding 3D scans. We use the state-of-the-art dense reconstruction system \cite{straub2019replica} to obtain the 3D photo-realistic reconstruction of those apartments. Volumetric representations are obtained from a customized capture rig and dense 3D meshes are extracted by the Marching Cubes algorithm with textures. We further annotate the dense meshes by labeling object categories over the mesh polygons; 35 object categories plus a background class label are used in annotation.

\iftoggle{arxiv}{
Primary contributors: Mingfei Yan, Richard Newcombe, Kiran Somasundaram, Chao Li.}{}

\iftoggle{arxiv}{
\paragraph{Universidad de los Andes, Colombia:}}
{
\paragraph{Colombia:}}

We gather 302.5 hours across 20 scenarios from 77 unique participants. We record videos using GoPro Hero 9 cameras between July and August 2021. We recruit volunteer participants from within the Uniandes community and their families and friends. The ethnic groups include 89.9\% Hispanic, 1.4\% African, and 5.8\%Caucasian. The gender distribution follows 41.6\% male and 58.4\% female with ages ranging from 18 to 65 (6 teens, 44 twenties, 3 thirties, 2 forties, 6 fifties, and 1 sixties). Our data collection focuses mainly on simultaneous video recording in groups of camera wearers within a common setting. Thus, these data capture a single scene and social interactions from different points of view. We include both outdoor and indoor scenarios in Colombia. Outdoor scenarios include Bogotá and Cartagena’s historical and colonial centers, as urban settings, and a Natural National Park and a stream, as rural settings. Indoor locations include professional activities such as laboratory workers and hair stylers. Furthermore, we include sports events such as salsa and urban dance rehearsals and rock climbing.

\iftoggle{arxiv}{
Primary contributors: Cristina Gonz\'alez and Paola Ruiz Puentes.}{}

\iftoggle{arxiv}{
\paragraph{Carnegie Mellon University, Pittsburgh, PA, USA and Kigali, Rwanda:}}
{\paragraph{Pennsylvania, USA:}}
\iftoggle{arxiv}{Carnegie Mellon University (CMU) Pittsburgh}{This lab} gathered a large portion of its data from skilled workers such as carpenters, construction workers, landscapers, mechanics, arborists, painters, and artists. This portion of the dataset does not include any graduate students with the explicit goal of capturing a diverse range of real-world occupational activities. Over 500 hours of video were captured in the \iftoggle{arxiv}{Pittsburgh}{local} area. The data was mostly recorded using a GoPro camera and a small portion was collected using WeeView, a wearable stereo camera.

\iftoggle{arxiv}{}{\paragraph{Rwanda:}}
\iftoggle{arxiv}{Carnegie Mellon University Africa}{This lab} gathered data from hobbyist craftspeople and daily workers working in \iftoggle{arxiv}{Kigali, Rwanda}{the local area}. An effort was made to collect data most representative of how tasks are carried out in Rwanda (such as doing laundry manually as opposed to with a washing machine). Over 150 hours of video were captured, and a portion of those hours are available in the current release. All of the data was collected using a GoPro camera.

\iftoggle{arxiv}{
Primary contributors: Kris Kitani - project coordinator for both CMU Pittsburgh and CMU Africa video collection. Sean Crane - lead coordinator of CMU Pittsburgh data collection (over 500 hours), main lead of CMU IRB review. Abrham Gebreselasie - lead coordinator of CMU Africa data collection. Qichen Fu and Xindi Wu - development of video de-identification pipeline, manual video de-identification annotation of CMU Pittsburgh data. Vivek Roy - main architecture of the license signing web server, coordinating with America Web Developers.}{}

\iftoggle{arxiv}{\paragraph{University of Catania, Italy:}}
{\paragraph{Italy:}}
More than 359 hours of video have been recorded from 57 different subjects recruited through word of mouth, starting from family members, friends and acquaintances of students and faculty members of the research group. Videos are related to 25 scenarios. We chose the participants to cover a wide variety of professional backgrounds (24 backgrounds including carpenters, bakers, employees, housewives, artists, and students) and ages (subjects were aged from 20 to 77, with an average age of 36.42). 21 of the participants were female, while the remaining 36 were male. Female participants collected about 137 hours of video, whereas males collected 222 hours of video. 
The average number of hours of videos acquired by each participant is 6h:18m:23s, with a minimum number of hours of 06m:34s, and a maximum number of hours of 15h:40m:42s. 

To prepare participants to record videos, we demonstrated to them the operations of the camera and how to wear it. We provided examples of valid recording and invalid recordings before they started the acquisition session. The recording procedure was described in a document left to the participants to help them remember the device usage and how to perform a good acquisition. Acquisition of videos has been performed using different models of GoPro cameras (GoPro 4, GoPro7, GoPro8, and GoPro Hero Max), which were handed over to the participants who typically acquired their videos autonomously over a period of a few days or weeks. 3D scans for 7 locations using the Matterport 3D scanner have been also collected~(Figure~\ref{fig:matterport_appendix}).

\iftoggle{arxiv}{
Primary contributors: Giovanni Maria Farinella and Antonino Furnari - scenarios, procedures, data collection oversight, data formatting, encoding, metadata and ingestion. Irene D'Ambra - data collection, consent forms and information sheets, manual data review, de-identification oversight.}{}

\iftoggle{arxiv}{
\paragraph{King Abdullah University of Science and Technology (KAUST), Saudi Arabia:}}
{\paragraph{Saudi Arabia:}}

A total of 453 hours of videos have been collected from 66 unique participants in 80 different scenarios with GoPro Hero 7. All the participants were \iftoggle{arxiv}{KAUST}{university} community members, who are from various countries and have various occupations. \iftoggle{arxiv}{All recordings took place in the KAUST university compound, which is 3600 hectares in area with diversified facilities (e.g., sports courts, supermarkets, a 9-hole golf course, and 2 beaches) and scenes (e.g., buildings, gardens, the red sea, and the desert). Therefore, the team was able to collect videos of various scenarios such as snorkeling, golfing, cycling, and driving.}{All recordings took place in the local area in a variety of locations (e.g., sports courts, supermarkets, a 9-hole golf course, and two beaches) and scenes (e.g., buildings, gardens, the red sea, and the desert) and activities (snorkeling, golfing, cycling, and driving).} 

The participants were recruited from multiple sources, such as friends and families, individuals referred to us by earlier participants, as well as people who were interested in our Facebook advertisements  or posters in campus restaurants and supermarkets. Each candidate participant was required to register through an online form, which contained an introduction to and requirements of the recording task, and collected his/her basic demographic information. The participants' ages range from 22 to 53. They come from 20 different countries, and about half are females. Many participants were graduate students and researchers, while others had various  kinds of occupations such as  chefs, facility managers, and teachers.

In order to prepare the participants for the  recording process, the team described in documents and demonstrated to them the operations of the camera. The team also provided examples of what constitute  valid and invalid recordings before they started. Each participant was provided a GoPro mountable camera with 2 batteries and a 512/256 GB SD card. Each participant needed to choose at least 2 different activities from our scenario list and record 1-10 hours of video within 2 days. The university team went through the recordings after the participants returned the camera to check their quality as well as to make sure the videos meet the university's IRB requirements.

\iftoggle{arxiv}{
Primary contributors: Chen Zhao, Merey Ramazanova, Mengmeng Xu, and Bernard Ghanem.}{}

\clearpage
\subsection{De-identification Process}\label{sec:deid-appendix}

The dataset has two types of video. The first includes videos recorded indoors where informed consent for capturing identities is explicitly collected from all participants in the scene, including faces and voice. \srt{Only video of this type is used in our Audio-Visual Diarization and Social Interaction benchmark studies.}  \iftoggle{arxiv}{All 400 hours of data collected by Facebook Reality Labs falls in that category.  }{} The second category, which forms the majority of our videos, requires de-identification as consent for capturing identities is not given---including footage captured outdoors in public spaces.\footnote{\srt{The exception is data from \iftoggle{arxiv}{University of Minnesota}{one university}, whose IRB permitted recording of incidental participants in public spaces having no manipulation or direct interaction with study personnel.}}
\iftoggle{arxiv}{Only video collected by the universities falls into this second category.}{}
See Appendix~\ref{sec:collection} for details about the per-site collection approaches.

\subsubsection{De-identification overview}

All videos in the second category were manually screened to address any de-identification needs, and are further divided into two groups.
Group1: videos that do not contain any personally identifiable information (PII).\footnote{We use the abbreviation PII to capture data protected under various data protection regimes including the General Data Protection Regulation (GDPR) where the term ``personal data" is used.}
This is when the video is recorded indoors with one person wearing the camera performing tasks such as cleaning or knitting for example, and no PII is present in the video.
These videos did not require de-identification.
Group2: videos where PII is captured. These include indoor settings with multiple participants present, PII captured accidentally such as an address on an envelope or a reflection of the wearer's face on a mirror or a surface, as well as videos recorded outdoors in a public space where bystanders or cars appear in the footage.
Videos in Group2 were marked for de-identification, deploying advanced video redaction software, open source tools, and hours of human reviews to redact visible PIIs.  
University partners undertook this de-identification effort for their own data.  We summarize the approach below.

Videos marked for redaction were processed through de-identification software that removes specific identifiers at scale. 
We used two commercial softwares: brighter.ai\footnote{http://brighter.ai} and Primloc's Secure Redact\footnote{http://secureredact.co.uk} that enabled detecting faces and number plates automatically. We carefully reviewed all outputs from automated blurring, identifying both instances of false positives (blurring that mistakenly occurred on non-privacy related items) or false negatives (inaccurate or insufficient automated blurring of faces and number plates). 
Additionally, other PII data such as written names/addresses, phone screens/passwords or tattoos had to be manually identified and blurred per-frame. 
For this part of our de-identification process, we used both commercial tools within the above-mentioned commercial software and open source software, including Computer Vision Annotation Tool (CVAT)\footnote{https://github.com/openvinotoolkit/cvat}, Anonymal\footnote{https://github.com/ezelikman/anonymal} and SiamMask\footnote{https://github.com/foolwood/SiamMask}.

\paragraph{Time costs.} The relative time costs with respect to the original video length varied significantly for the different scenarios.
Videos captured outdoors could take 10x the length of the video to carefully redact.

\subsubsection{Sample pipeline}
\begin{figure}[t]
	\centering
	\includegraphics[width=1\linewidth]{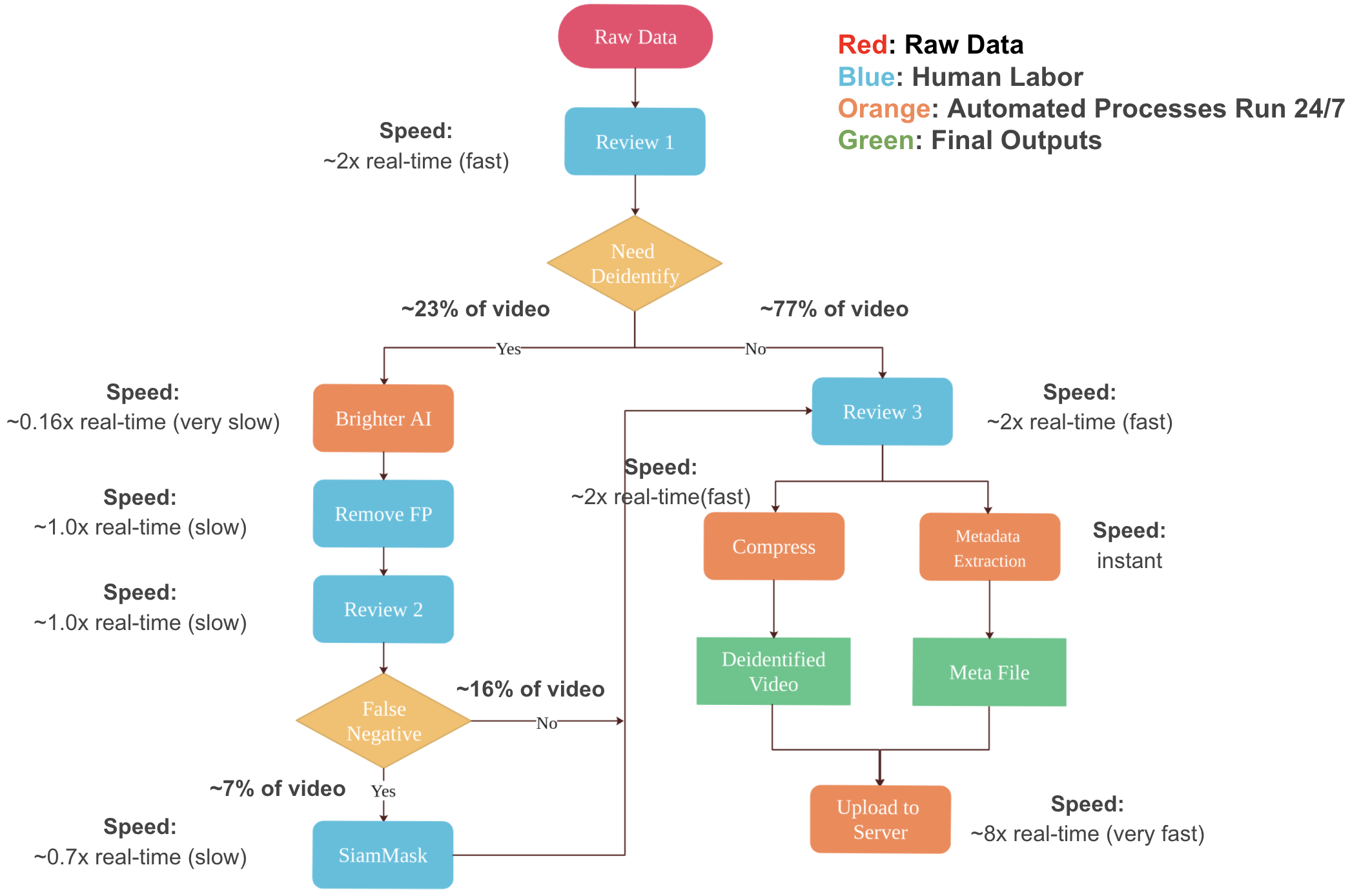}
	\caption{\label{fig:deID} \iftoggle{arxiv}{CMU's de-identification pipeline}{Example de-identification pipeline from one university team.}}
\end{figure}
While partners followed varying pipelines, we offer a sample pipeline to showcase the process followed by \iftoggle{arxiv}{Carnegie Mellon University}{one university} that uses brighter.ai as the commercial software. This sample pipeline showcases the combination of automated processes and human labor with relative speeds of these steps.

This semi-automatic de-identification process 
was performed in four sequential stages (Figure \ref{fig:deID}): (1) automatic face and license plate detection, (2) false positive removal, (3) negative detection handling, and (4) image blurring.

\paragraph{Sensitive object detection} Given the collected videos (raw data), a reviewer scans through videos and marks those containing sensitive objects such as human faces, license plates, credit cards, \emph{etc.} Then de-identification software (brighter.ai) was used to automatically detect sensitive information.

\paragraph{False positive removal} To improve the quality of the detection, false positives were removed. Reviewers manually scanned through the bounding boxes detected by the de-identification software, and rejected those bounding boxes which did not contain sensitive information.

\paragraph{False negative correction} %
Additionally, reviewers studied every video to search for false negatives and manually annotated them using a bounding box. To make the process more efficient, an online object tracking algorithm \cite{SiamMask} was used to generate bounding box proposals across frames. Reviewers verified that all tracked bounding boxes were correct. 

\paragraph{Image blurring} Once all of the detections were modified and corrected, a robust blurring process was used to de-identify image regions defined by the bounding boxes.

\paragraph{Time costs} 
The relative time costs with respect to the original video length for each step are shown in Figure \ref{fig:deID}.
Though this number depends greatly on the scenario captured in the video, roughly speaking to de-identify 500 hours of video data, it took 780 hours of manual labor. Review 1 of 500 hours of video required 250 hours of work, removal of false positive over 115 hours of video took 115 hours of work, Review 2 of 115 videos took 115 hours of work, correcting false negatives in 35 hours of videos required 50 hours of work, and Review 3 of 500 hours of video took 250 hours of work (250+115+115+50+250 = 780 hrs). 

\clearpage
\newpage
\subsection{Demographics}

We further provide self-declared information on ethnic groups and/or country of birth by the participants. 
We report these separately per state/country due to the differences in granularity of ethnic groupings.
All participants are residents in the country specified per paragraph. This data is not available for participants from Minnesota, US.

\paragraph{United Kingdom Residents}
Reporting demographics was optional and thus 63\% of participants (52/82) that reside in the United Kingdom self-reported their ethnic group membership as follows:

\resizebox{\linewidth}{!}{%
\begin{tabular}{ll}
\hline
White --- English, Welsh, Scottish, Northern Irish or British	&35\\
White --- Any other White background	&12\\
Mixed --- White and Asian	&1\\
Mixed --- Any other Mixed or Multiple ethnic background	&2\\
Arab	&1\\
Prefer not to say	&1\\
\hline
\end{tabular}}

\paragraph{Italy Residents}
100\% of participants that reside in Italy self-reported their country of birth as follows:

\begin{tabular}{ll}
\hline
Italy &53\\
Germany &1\\
Russia &1\\
Portugal &1\\
Poland &1\\
\hline
\end{tabular}

\paragraph{India Residents}
100\% of participants that reside in India self-reported their ethnic group membership as follows:

\begin{tabular}{ll}
Eastern India &10\\
Northern India &15\\
Southern India &108\\
Western India &5\\
\hline
\end{tabular}

\begin{tabular}{ll}
\hline
\hline
\end{tabular}

\paragraph{Pennsylvania, USA, Residents}
100\% of participants that reside in Pennsylvania, USA, self-reported their ethnic group membership as follows:

\begin{tabular}{ll}
\hline
White	&42\\
Asian 	&4\\
Mixed --- White and Black African	&2\\
Black, African, Caribbean &1\\
\hline
\end{tabular}

\paragraph{Washington, US, Residents}
100\% of participants that reside in Washington, USA, self-reported their ethnic group membership as follows:

\begin{tabular}{ll}
\hline
Caucasian	&101\\
Black or African American	&58\\
American Indian (Native American) &24\\
Hispanic	&19\\
Indian (South Asian)	&4\\
\hline
\end{tabular}

\paragraph{Indiana, US, Residents}
95\% of participants that reside in Indiana, US, self-reported their country of birth as follows:

\begin{tabular}{ll}
\hline
US	&39\\
China	&10\\
India	&10\\
Bangladesh	&2\\
Vietnam	&2\\
\hline
\end{tabular}

\paragraph{Georgia, USA, Residents}
100\% of participants that reside in Georgia, USA, self-reported their ethnic group membership as follows:

\begin{tabular}{ll}
\hline
White / Caucasian	&16\\
Black / African American	&1\\
Asian / Indian \& White / Caucasian &1\\
Other / Taiwanese&1\\
\hline
\end{tabular}

\paragraph{Japan Residents}
100\% of participants that reside in Japan self-reported their ethnic group membership as follows:

\begin{tabular}{ll}
\hline
Asian (Japanese)	&81\\
\hline
\end{tabular}

\paragraph{Kingdom of Saudi Arabia Residents}
100\% of participants that reside in KSA self-reported their country of birth as follows:

\begin{tabular}{ll}
\hline
China	&12\\
Russia	&9\\
Colombia	&8\\
Mexico	&5\\
Kazakhstan	&4\\
India	&4\\
US	&4\\
Saudi Arabia	&3\\
Kyrgyzstan	&2\\
New Zealand	&2\\
Greece	&2\\
Ukraine	&2\\
Italy	&2\\
Lebanon &1\\
Jordan	&1\\
Egypt	&1\\
Kashmir	&1\\
Portugal	&1\\
South African	&1\\
Thailand	&1\\
\hline
\end{tabular}

\paragraph{Singapore Residents}
100\% of participants that reside in Singapore self-reported their nationalities as follows:
\begin{tabular}{ll}
\hline
Chinese	&26\\
Singaporean	&12\\
Indian	&1\\
Malayan	&1\\
\hline
\end{tabular}

\paragraph{Colombia Residents}
\FEB{90}\% of participants that reside in Colombia self-reported their ethnic group membership as follows:

\begin{tabular}{ll}
\hline
Hispanic/Latin	&62\\
White/Caucasian	&4\\
Black, African or Caribbean	&1\\
Mixed - White an African	&1\\
Prefer not to say &	1\\
\hline
\end{tabular}

\paragraph{Rwanda Residents}
100\% of participants that reside in Rwanda self-reported their ethnic group membership as follows:

\begin{tabular}{ll}
\hline
Black, African or Caribbean &14\\
\hline
\end{tabular}

\clearpage

\subsection{Narrations}\label{sec:data-appendix}

\begin{figure*}[t]
  \centering  
  \includegraphics[width=\textwidth]{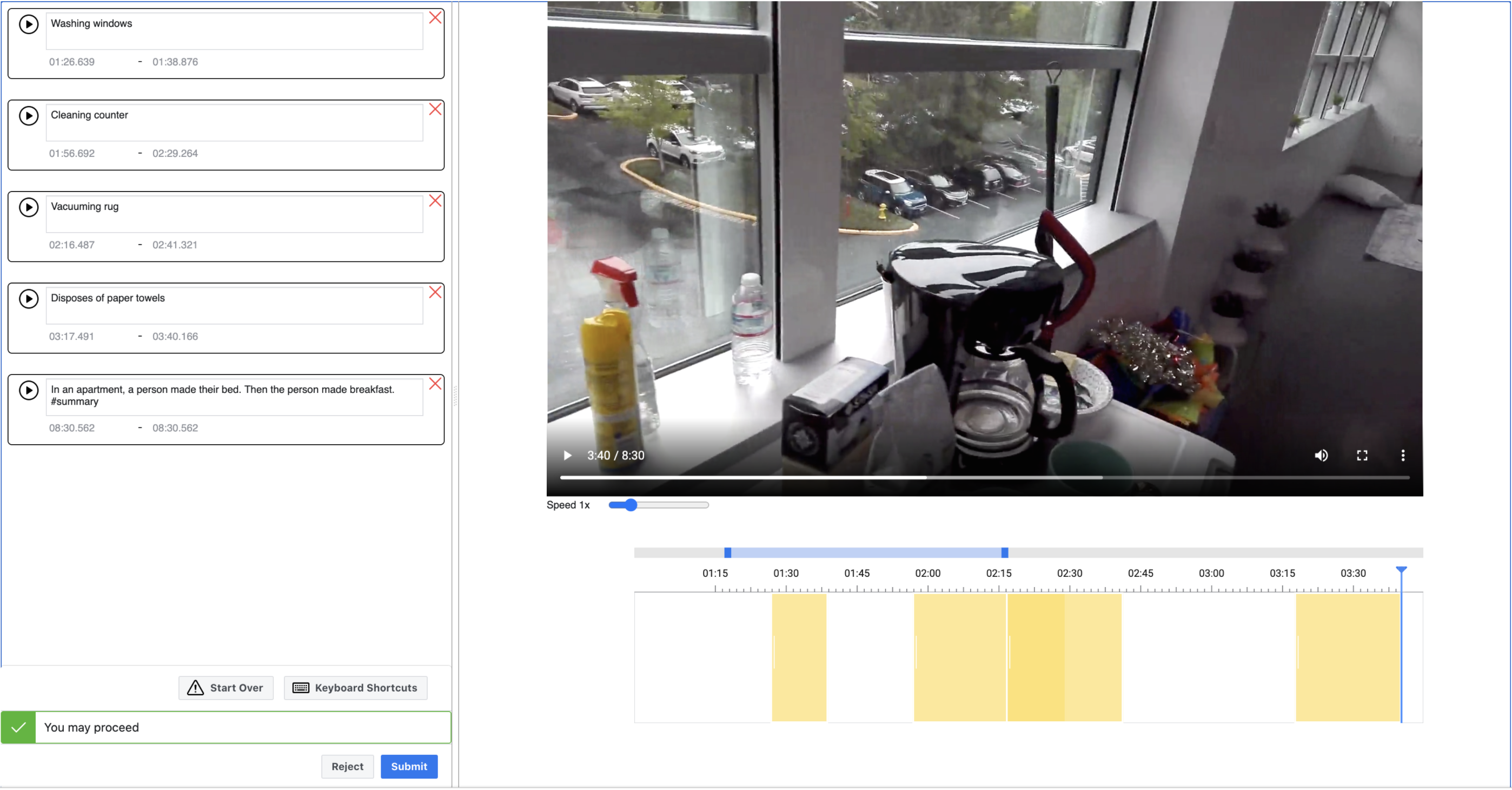}
  \caption{\textbf{Narration tool interface.} Narrators mark a timepoint where something happens in the video (bottom bar), and enter a text description of the activity (left sidebar).} 
  \label{fig:narration_interface}
\end{figure*}

The goal of the narrations is to obtain a dense temporally-aligned textual description of what happens in the video, particularly in terms of the activities and object interactions by the camera wearer.  The Ego4D narration data is itself a new resource for learning about language grounded in visual perception.  In addition, as described in the main paper, we leverage the narrations as a form of ``pre-annotation" to index the videos by semantic terms.  
Specifically, the narrations are used to construct action and object taxonomies to support various benchmarks,  to identify videos that are relevant to each benchmark, and to select regions within the videos that require annotation. 

This section overviews how we instructed annotators to narrate the videos, and how we transformed narration text into taxonomies of objects and actions.  %

\subsubsection{Narration instructions and content}

We divide the dataset into clips of (max) 5 minutes long when acquiring narrations.
Each 5-minute clip is then passed to two different annotators, to collect two independent sets of narrations for every video clip in the dataset for better coverage and to account for narration errors.\FEB{\footnote{We simply keep both independent narrations; they are not merged because they do not serve as ground truth for any benchmark.}}
  Narrators are instructed to watch the 5 minute video clip first, and then asked to provide a short
1-3 sentence ``summary'' narration for the entire clip that corresponds to the overall activity and setting of the video clip (e.g., ``the person does laundry in the washing machine'').
These summaries are marked with the tag ``\#summary'' in the released narrations. 

Following this first screening, which is critical for the overall understanding of the clip, the dense narrations are collected as follows.
Annotators re-watch the clip, pause and mark the timepoint when something happens in the video, then enter a short natural language description of the ongoing action or interaction, before resuming watching the video.

\begin{figure*}[t]
  \centering  
  \includegraphics[width=1\linewidth]{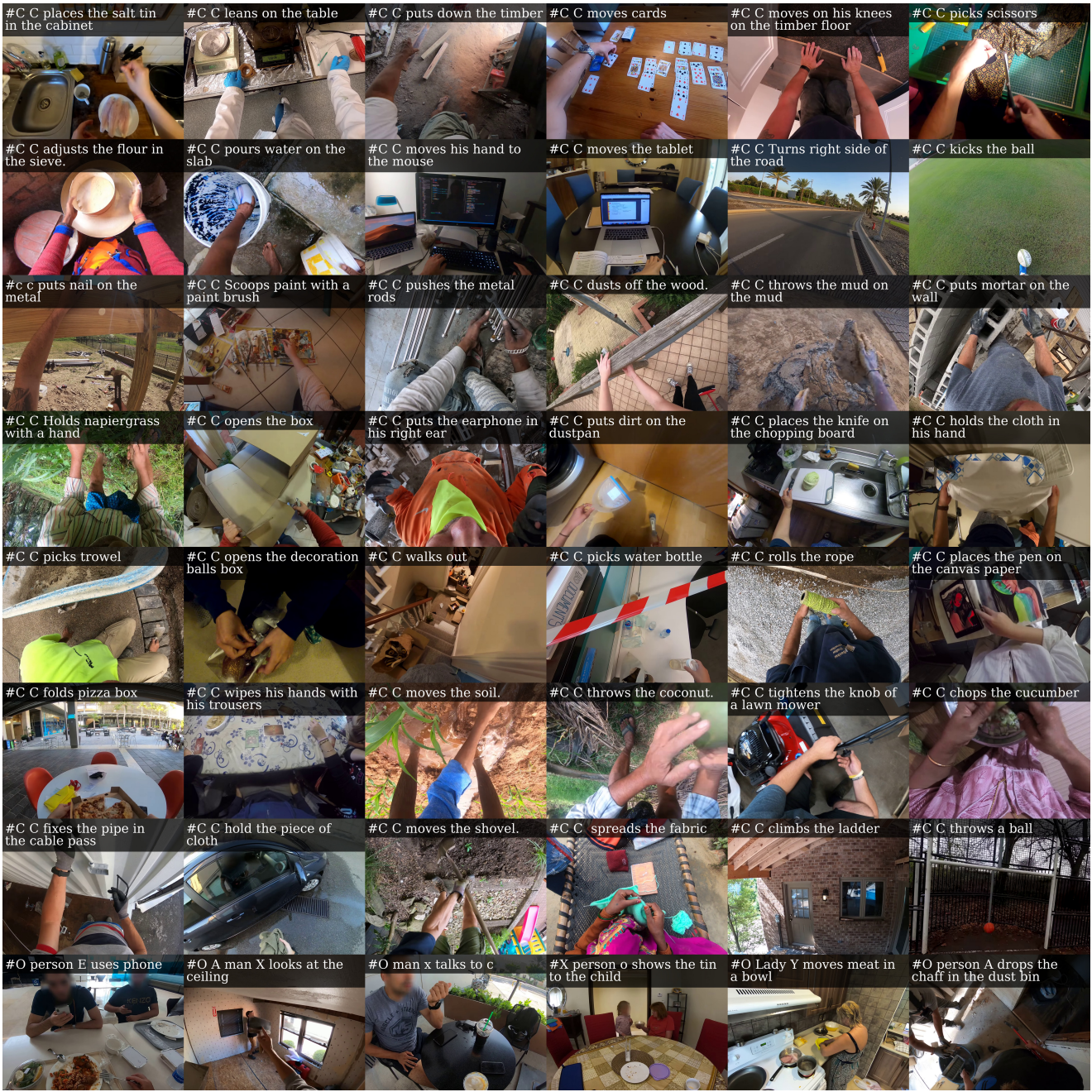}
  \caption{\textbf{Example narrations at keyframes of video.} \#C refers to the camera-wearer. The last row shows narrations that include other people that participate in activities with the camera-wearer (denoted by other letter tags, e.g., \#O, \#X).} 
  \label{fig:narr-examples}
\end{figure*}

Narrators are provided the following prompt: \emph{``Pretend as you watch this video that you are also talking to a friend on the phone, and you need to describe to your friend everything that is happening in the video. Your friend cannot see the video.''}  This prompt is intended to elicit detailed descriptions  that provide a play-by-play of the action. See Figure~\ref{fig:narration_interface} for an illustration of the narration tool interface.
Each narration thus corresponds to a single, atomic action or object interaction that the camera wearer performs
(e.g., ``\#C opens the washing-machine'' or ``\#C picks up the detergent'', where the tag \#C denotes the camera wearer). Importantly, our narrations also capture interactions between the camera-wearer and others in the scene, denoted by other letter tags, e.g. \#X (e.g. ``\#C checks mobile while \#X drives the car'', ``\#C passes a card to \#Y''). See Figure~\ref{fig:narr-examples} for narration examples.

\subsubsection{Narration analysis} \label{sec:narration_analysis}

\begin{figure*}[t]
  \centering  
  \includegraphics[width=1\linewidth]{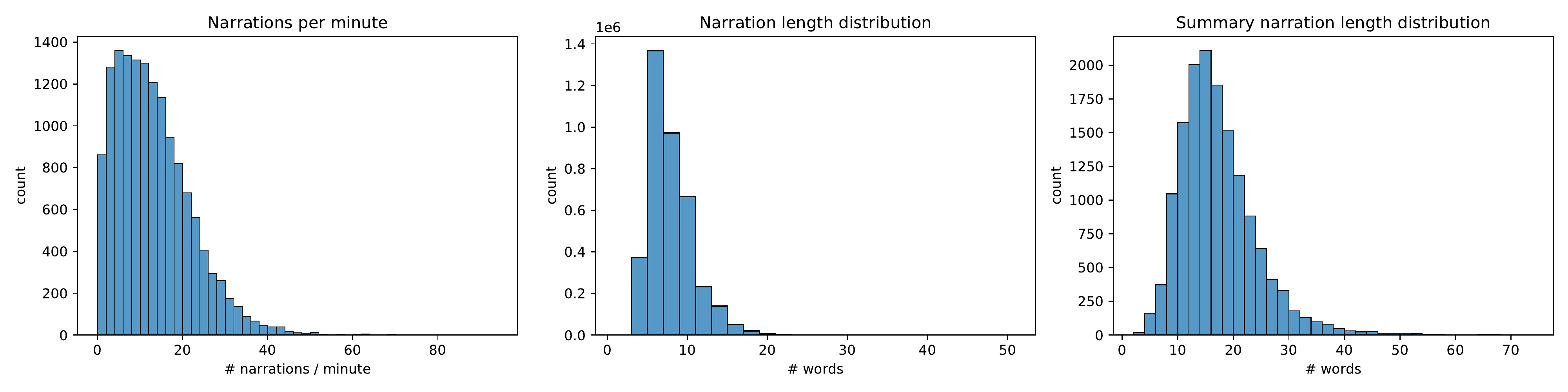}
  \caption{\textbf{Collected narration statistics.} \textbf{Left:} Distribution of frequency of narrations collected. \textbf{Middle and right:} The distribution of length of the collected narrations and summaries. Summaries are naturally longer, and describe activities at a higher level compared to individual action narrations. See text for discussion.} 
  \label{fig:narration_stats}
\end{figure*}

We present some statistics on the collected narrations. Altogether, we collected  \numnarrationsentences sentences across the \numhours hours of video. Figure~\ref{fig:narration_stats} (left) shows the distribution of frequency of narrations across all videos in the dataset. Depending on the activities depicted, videos are annotated at varying frequencies. For example, a video of a person watching television is sparsely annotated as very few activities occur (0.17 sentences/minute), while a video of a person harvesting crops, performing repetitive actions is densely annotated (63.6 sentences/minute). On average, there are an 13.2 sentences per minute of video. 

Figure~\ref{fig:narration_stats} (middle and right) show the distribution of length of the collected narrations. The individual timepoint narrations are short, highlight a single action or object interaction, and have an average of 7.4 words. Though short, these narrations cover a variety of activities ranging from object interactions, tool use, camera wearer motions, activities of other people etc. In contrast, the summary narrations are longer (on average, 16.8 words) and describe activities at a higher level. Table~\ref{tab:narration_examples} shows a few text examples of each type of narration in addition to the visual examples in Figure~\ref{fig:narr-examples}. 

\begin{table*}[]
\resizebox{\textwidth}{!}{
\begin{tabular}{|l|l|l|l|}
\hline
\textbf{Object interaction}     & \textbf{Context objects}                     & \textbf{Multi-person actions}             & \textbf{Manipulation actions}                        \\ \hline
\#c c flips the paper           & \#c c taps a hand on the floor               & \#o a man x moves the legs.               & \#c c cuts a leaf from the plant with his left hand. \\
\#c c lifts the t-shirt         & \#c c holds the wheel with his left hand.    & \#o a man y sits on a chair               & \#c c pulls his hand off the chess piece             \\
\#c c drops the plate           & \#c c puts the brush in the colours.         & \#o a woman x steps forward.              & \#c c holds the knitting needle with the other hand  \\
\#c c holds the piece of cloth  & \#c c places plastic models kit on the table & \#o a person x hits the cricket ball      & \#c c opens the screwdriver container with his hands \\
\#c c fixes on the model craft  & \#c c arranges the doughs on the tray        & \#o a man y throws the ball towards man x & \#c c touches the piece of wood with the hand        \\ \hline
\textbf{Camera wearer motion}   & \multicolumn{3}{l|}{\textbf{Summary narrations}}                                                                                                \\ \hline
\#c c raises hands              & \multicolumn{3}{l|}{c was in a room,fixed a wood model kit. \#summary}                                                                          \\
\#c c stands                    & \multicolumn{3}{l|}{c tightened the motor on the head of the hoe of the lawn mower. c cut grasses on the field with the lawn mower. \#summary}  \\
\#c c stands up from the stairs & \multicolumn{3}{l|}{c was in a kitchen, he cut sausages in to pieces with a knife, mixed the sausages and cooked them with a pan. \#summary}    \\
\#c c walks around a kitchen    & \multicolumn{3}{l|}{c was in the house and she studied \#summary}                                                                               \\
\#c c sits up                   & \multicolumn{3}{l|}{c studied in a room. c went through a mobile phone and a mobile tablet while reading in the room.  \#summary}               \\ \hline
\end{tabular}
}
\caption{\textbf{Text examples of narrations.} The collected narrations describe diverse aspects of human activity. Summary narrations capture high level descriptions of activities in a 5 minute clip. See Figure~\ref{fig:narr-examples} for visual examples.} 
\label{tab:narration_examples}
\end{table*}

\begin{figure*}[t]
  \centering  
  \includegraphics[width=1\linewidth]{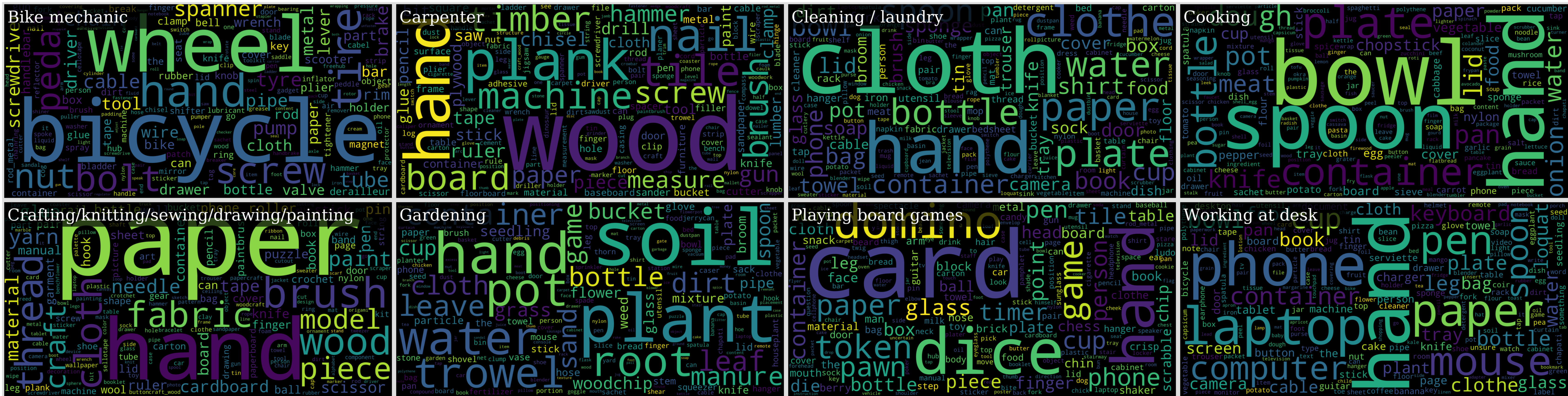}
  \caption{\textbf{Distribution of objects in narrations of videos from eight common scenarios.} The variety of objects covered across scenarios showcases the diversity of activities in the video collected.} 
  \label{fig:narration_wordclouds}
\end{figure*}

Finally, we study the diversity of the video dataset by looking at the frequency of occurrence of words in the narrations collected for videos of each scenario type. Figure~\ref{fig:narration_wordclouds} shows word clouds depicting objects that prominently feature in across various scenarios. The word clouds highlight characteristic objects per scenario (e.g., bowl, spoon, plate in ``Cooking'' videos; card, dice, pawn in ``Playing board games'' videos) while also hinting at common objects across all scenarios (e.g., hands, paper, phones). The diversity in narrations collected highlights the diversity of video content captured in the dataset.

\subsubsection{Action and object taxonomy} \label{sec:taxonomy}

\begin{figure*}[t]
  \centering  
  \includegraphics[width=\textwidth]{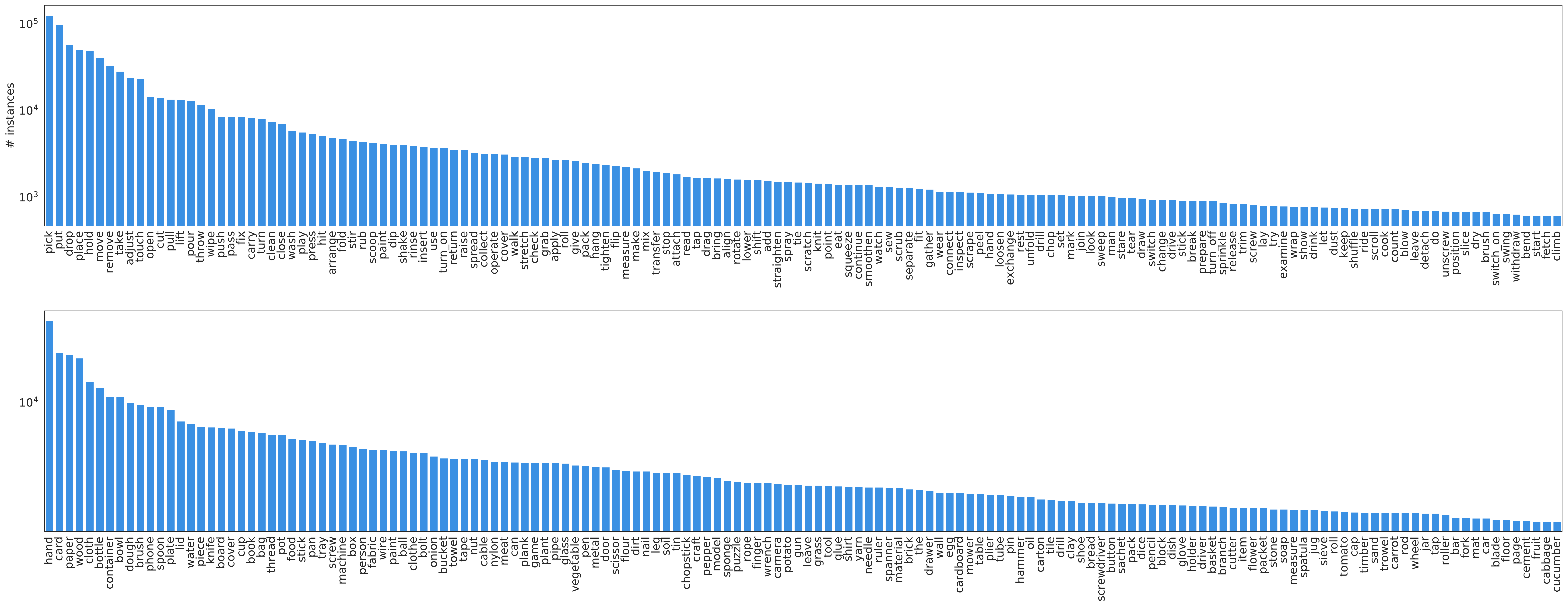}
  \caption{\textbf{Narration verb/noun distribution.} Distribution of automatically extracted verbs (top) and nouns (bottom) from narrations. Top 150 most frequently occurring of each is shown for clarity.} 
  \label{fig:narration_taxonomy}
\end{figure*}

In total the raw narrations describe the Ego4D video using 1,772 unique verbs and 4,336 unique nouns. The distribution of the most frequently occurring verbs and nouns can be seen in Figure~\ref{fig:narration_taxonomy}.

Following ideas from~\cite{Damen2018EPICKITCHENS}, we leverage the narrations data to construct a taxonomy over the actions and objects that appear in the video, as follows.
We use a part-of-speech (POS) tagger and dependency parser to identify verbs and nouns from each narrated action. We use an ensemble of parser models from the Spacy~\cite{spacy} toolkit to do this. Given a natural language narration, we first identify verbs using their POS tag. Then using the dependency tree, we identify all direct objects of the verb. To ensure verbs and nouns are accurately parsed, we adopt several heuristics: Parsed verbs are split into multiple senses (e.g., ``turn'' is split into ``turn-on'', ``turn-off'' and ``turn-over''); compound nouns are decomposed into a root noun coupled with a modifier to ensure the noun taxonomy is unambiguous (e.g., modifier ``egg'' and root noun ``shell'' in ``egg shell''); collective nouns are mapped to their main entity (e.g,. ``piece of cheese'' $\rightarrow$ ``cheese''). Finally, we manually cluster the verbs and nouns to avoid redundancy in the taxonomy (e.g., ``cut'', ``chop'', ``slice'' are all mapped to the verb cluster ``cut'').

\FEB{The resulting taxonomy consists of a set of 115 verbs ($\mathcal{V}$) and a set of 478 nouns ($\mathcal{N}$)}. Figure~\ref{fig:taxonomy} shows the distribution of verbs and nouns in a set of video data annotated with the taxonomy. See Section~\ref{appendix:forecasting_data} for details on how the taxonomy is used in the context of the benchmark tasks.  %

\subsubsection{Narrations for annotation prioritization}

All videos in Ego4D are narrated, and subsets of them are manually labeled for each benchmark.  Rather than randomly label instances for a given benchmark, we aim to target those that are most relevant to the task.  For example, videos likely to contain multi-person conversation are most interesting for the AV Diarization benchmark, whereas videos with ample hand-object interaction are most interesting for Hands and Objects.  
To that end, we use the narrations and summaries as a tool to automatically prioritize certain videos to label per benchmark.  The benchmark appendices below provide details.

\iftoggle{arxiv}{
\subsubsection{Contributions statement}

Tushar Nagarajan developed the taxonomy, helped develop narration instructions, and performed the narration analysis presented in the paper.  Kristen Grauman developed narration instructions, helped coordinate pilots and annotation work, and contributed to taxonomy formation.  Michael Wray co-developed the taxonomy.
}{}

\clearpage
\clearpage
\subsection{Benchmark Data Splits}\label{sec:appendix-splits}

\begin{table}[t]
 \centering\small
\begin{tabular}{|c|c|c|c|}
\hline
& Num hours & Num clips & Avg clip length\\
\hline
EM VQ-2D & 432.9 & 5,831 & 6.1 min\\
EM VQ-3D & 13 & 159 & 4.9 min\\
EM Moments & 328.7 & 2,522 & 7.9 min\\
EM NLQ & 227.1 & 1,659  & 8.2 min \\
Hands+Obj. & 196.2 & 88,585 & 8.0 sec\\
Forecasting & 110.5 & 1,498 & 4.4 min\\
AVD & 47.7 & 572 & 5 min\\
Social & 47.7 & 572 & 5 min\\
\hline
\end{tabular}
\vspace*{-0.1in}
\caption{Amount of annotated data for each benchmark.  EM refers to Episodic Memory and AVD refers to Audio-Visual Diarization.  All \numhours hours of video have narrations and features.}\label{tab:list}
\vspace*{-0.1in}
\end{table}

 \FEB{For each benchmark task, certain portions of the Ego4D video repository are  labeled.    Table~\ref{tab:list} shows the breakdown of the amount of data annotated for each. Note that there are 764 total hours of video relevant to the AVD and Social tasks (i.e., have audio, conversation, and unblurred faces), including the annotated set of 47.7 hours above.  For other benchmarks, the relevance has a softer dependency on the specific video content (e.g., a memory query can apply to any of the \numhours hours). The following appendices will explain how we sampled data to be annotated for each benchmark.}

\FEB{For the public Ego4D benchmark challenge, we ensure that the splits are consistent within a
family of related tasks. For instance, all the Forecasting
and Hands+Objects tasks share the same splits and ensure
training videos in one do not occur as validation videos in
another. Similarly, the Episodic Memory tasks share the
same splits. However, it is harder to ensure this across
very different tasks, since the videos selected for annotations
are different. For example, the Social benchmark considers
multi-person interactions which may not have many
hand-object interactions; hence the set of videos labeled
for Social and Hands+Objects have little overlap and the
train/val/test splits are naturally different.

Since we plan to use the test set for the public challenge, we are withholding all the test annotations and making them accessible
only through a submission server. We are also withholding the
narrations that overlap with any of the test sets.}

\clearpage
\subsection{Episodic Memory Benchmark}\label{sec:episodic-appendix}

This section details the Episodic Memory benchmark task definitions, annotations, baseline models, and results.

\subsubsection{Formal task definitions}

As presented in the main paper, there are three kinds of Episodic Memory queries---visual, natural language, and moments---each of which requires localizing the response in the video.  Their formal definitions are as follows.

\paragraph{Visual queries (VQ)}

This task aims to query an egocentric video based on a static image crop of an object. Specifically, it asks the question `Where was object X last seen in the video?', where X is a single `canonical' image crop in which the object is clearly visible and human-identifiable. A potential use case for visual queries is where a user teaches the system a new object by showing a photo (``these are my keys") and then later queries for it among past video.  By enabling visual queries, as opposed to categorical queries, this is a form of open-world object localization.

We formulate the problem as follows. Given an egocentric video $\mathcal{V}$, a query object $o$ specified via a static visual crop $v$, and a query frame $q$, the goal is to identify when the object $o$ was last seen in the video before the query frame $q$. The response is specified as a `response track' $r$ which is a temporally contiguous set of bounding boxes surrounding the object $o$ in each frame:
\begin{equation}
    r = \{r_s, r_{s+1}, \cdots, r_{e-1}, r_e\},
\end{equation}
where $s$ is the frame where the object $o$ (at least partially) enters the camera-wearer's field of view, $e$ is the frame where the object exits the camera-wearer's field of view, and $r_i$ is a bounding box $(x, y, w, h)$ in frame $i$. If the object appears multiple times in the video, the response only refers to the `most recent occurrence' of the object in the past, i.e., the response track which minimizes $q - r_e$ with $q > r_e$. 

When a 3D scan of the environment associated with the video is available, the response additionally includes a 3D displacement vector $\Delta d = (\Delta x, \Delta y, \Delta z)$ between the 3D location where the query was made (i.e., at query frame $q$), and the 3D location in the environment where the object was last seen (i.e., at the end of the response track $r_e$).

\paragraph{Natural language queries (NLQ)}
The motivation behind the NLQ task is to enable searching through an egocentric video 
using a natural language query.
The system responds to a query by providing a temporal window localized in the video,
from which the answer to the query can be deduced.
These queries can be related to objects, places, people, and activities that appeared in
the episodic memory of the user.
Note that we only consider episodic queries, \ie, queries that can be answered/deduced
from the egocentric videos, and not factual queries, \ie, queries that require an external knowledge base to answer.

NLQ is a challenging multimodal task requiring visual and linguistic
understanding and reasoning.
Consider the query %
\textit{``What did I pick up before leaving the party?"}
In order to fulfill this request, the system needs to:
(a) break down and understand the language query as a search for an object (\textit{what}) 
with which the user interacted (\textit{pick up}) before an event 
(\textit{leaving the party}),
(b) go through the egocentric video and identify the desired event of 
\textit{``leaving the party"},
(c) visually search for the object with which the user interacted prior to this event.
This example demonstrates the complexity of NLQ from both visual 
(recognizing events, objects, places, \etc) and 
linguistic (breaking down reasoning, understanding relations, \etc) perspective.
In addition, the diverse set of queries within NLQ,  while facilitating a flexible search and retrieval through an intuitive interface of language, also increases the complexity of
the task.

Concretely, NLQ is formulated as follows:
Given an egocentric video $\mathcal{V}$ and a natural language query $\mathcal{Q}$, 
the goal is again to identify a `response track' $r$, such that the answer to $\mathcal{Q}$ 
can be deduced from $r$.
The response track should be a set of temporally contiguous frames within $\mathcal{V}$.
Given the episodic nature of our task, $r$ should be sufficient to answer $\mathcal{Q}$, 
without the additional need for $\mathcal{V}$ or any external knowledge bases.

\paragraph{Moments queries (MQ)}
This task aims to query an egocentric video based on a category of actions. Specifically, it poses the following request `Retrieve all the moments that I do X in the video.', where `X' comes from a pre-defined taxonomy of action categories, such as `interact with someone' or `use phone'. Compared to the natural language queries, the moment queries focus on daily-life actions or activities. %
One moment query can correspond to multiple response instances (temporal windows) in the video. This task provides the user a fast and convenient way to retrieve multiple action moments at a time, where the user does not need to come up with a sentence to describe what he/she wants, but instead can directly choose among the pre-defined categories.

\cc{The moment queries task is related to the task of temporal action detection~\cite{lin2018bsn,xu2020g,zhao2020video}, which aims to identify and localize all instances of all action categories that take place in a video. Both tasks have a list of action categories pre-defined, and both aim to predict multiple action instances with their temporal boundaries. The difference is that 1) our moment queries task is a retrieval task where  action categories are provided as queries, meaning it does not need to produce instances of categories that are not among the queries; and 2) our moments taxonomy is specific to first-person activity.} We aim for moments that are activities at a medium level of granularity---coarser than the actions in Forecasting, and finer than the ``scenario" labels shown in Figure~\ref{fig:scenarios} of the main paper.

\cc{The MQ task is also related to temporal language grounding in videos~\cite{2DTAN_2020_AAAI}, which  aims to retrieve a segment from a video, as queried by a natural language sentence. Both tasks have a query and aim to predict corresponding temporal segments. The difference is that MQ uses pre-defined query categories rather than natural language sentences, and one query can correspond to multiple instances rather than a unique one.  }

We formulate the problem as follows. Given an egocentric video $\mathcal{V}$,  and a query action category $c$, the goal is to retrieve all the instances of this action category in the video, assuming that the query is made at the end of the video. The response is a set of action instances of the category $c$ $\Phi_c=  \left \{ \phi_n=\left ({t}_{n, s},{t}_{n, e},  {s}_n \right ) \right \}_{n=1}^{N}$,  where $n$ is the number of instances for this category, ${t}_{n, s}$ and ${t}_{n, e}$ are  start time and end time of the $n^{th}$ instance respectively,  and ${s}_n$ is its prediction confidence.

\begin{table}[t]
\centering
\scalebox{0.85}{
\begin{tabular}{@{}cccccc@{}}
\toprule
\multicolumn{6}{c}{Navigation verbs for entropy-based video selection} \\ \midrule
appear      & ascend       & bend        & bring       & carry      & catch    \\     
climb       & close        & come        & descend     & dig        & dispose  \\      
drag        & dribble      & drop        & enter       & fall       & fetch    \\    
find        & fly          & gather      & get         & give       & grab     \\      
hang        & jog          & jump        & kick        & lean       & leave    \\
lift        & lower        & move        & navigate    & open       & propel   \\
raise       & return       & ride        & rise        & run        & shut     \\
steer       & step         & turn        & vaccum      & walk       &          \\
\bottomrule
\end{tabular}
}
\caption{We prioritize videos to annotate for visual queries based on the entropy of these navigation-related verbs in the narrations.}
\label{apptab:vq_entropy}
\end{table}

\subsubsection{Selecting clips for annotation}

For all benchmarks we sample video clips to annotate based on criteria for geographic diversity and scenario diversity.  For Episodic Memory we impose additional sampling criteria meant to highlight data most interesting for the task, as follows.

\paragraph{Visual queries}
Video clips to annotate for visual queries (VQ) are selected based on the frequency of object occurrences and amount of navigation in the video. To have interesting visual queries in a video, there must be several `interesting' objects that can be queried about. An object is `interesting' in the context of visual queries if there is a sufficiently high separation in space and time between any two occurrences of the object. This typically happens when the camera-wearer visits the location near the object briefly, and then navigates elsewhere before revisiting the object again. For example, consider a person who finishes cleaning a living room, visits the kitchen for some period of time before revisiting the living room again. Most objects in the living room are interesting to query about when the person is in the kitchen. 

To select videos based on these considerations, we use a two-step process. First, we filter out videos based on the associated `scenario` labels (see Figure~\ref{fig:scenarios}) that provide high-level information about the content and activities in videos (e.g., cooking, cleaning, golfing, etc.). We manually preview randomly sampled videos from each scenario to identify interesting scenarios such as cooking, indoor navigation, farmer, cleaning, and grocery shopping. We then sort videos within each scenario based on a scoring function using the narrations for the video. Specifically, we extract the list of verbs in the narrations (along with their frequencies). We then measure the entropy of the distribution of manually curated \emph{navigation} verbs (See Tab.~\ref{apptab:vq_entropy}). The video is more likely to allow challenging visual queries if its navigation entropy is higher. For videos with near-zero entropy, we observe that the camera-wearer is usually staying static in a single location without any movement. Finally, a limited number of 3D scans were available for the 3D localization task. Videos associated with these scans were prioritized, regardless of their navigation entropy, in support of the 3D response version of the VQ task.

\paragraph{Natural language queries}

For NLQ we apply similar sampling criteria as above for VQ, but augment it to avoid repetitive actions (e.g., sewing while sitting on the couch).  First, we manually select amenable scenarios (see Figure~\ref{fig:scenarios}).  Among those, we prioritize clips with high entropy computed over navigational terms as above.  Finally, we prioritize non-repetitive actions by computing the ratio of the number of unique verbs in a clip's narration vs.~the total number of verbs in that same narration---higher is better.

\paragraph{Moments queries}

To select clips for moments queries, we compute
the overlap of verbs/nouns with the moments taxonomy.  We calculate a similar entropy-based score and sort videos according to this score. In addition, we restrict videos to a fixed set of categories present in our taxonomy to avoid labeling videos that do not contain relevant activities. %

\subsubsection{Annotation}

Next we describe the annotation procedures and outputs for Episodic Memory.

\paragraph{Visual queries}
For annotating visual queries, we first sample contiguous clips of varying lengths (5 mins, 8 mins, and 16 mins) from the set of interesting videos. The annotators are instructed to create and annotate 3 visual queries for each clip. A visual query consists of the query frame $q$, the visual crop $v$ of the query object $o$, the response track $r = \{r_s, r_{s+1}, \cdots, r_{e-1}, r_e\}$, and a textual name for the object (eg. cup, hammer, broomstick, etc). The annotators performed the following steps to annotate a given clip:
\begin{enumerate}
\item Identify three interesting query objects in the clip. An object is interesting if it occurs in at least two different parts of the video. 
\item  For a given object, enter a textual name. While our current task queries with the image crop, not the name, this annotation will allow future variants that do query for the object by name.
\item Select one of the object occurrences in the video and mark a visual crop $v = (x_v, y_v, w_v, h_v)$. The visual crop must be a good representative view of the object, and it must have good lighting, large-enough size, and must not be blurred. 
\item  Mark a \emph{different occurrence} of the object as the response track $r = \{r_s, \cdots, r_e\}$. The response track starts from the frame when the object is first visible and ends when the object leaves the field-of-view. The response track must also be contiguous in time and the bounding boxes must accurately mark the position and size of the object. 
\item The query frame $q$ is sampled some time \emph{after} the response track $r$. The object $o$ must not appear anywhere between the response track $r$ and the query frame $q$, so that the ground truth is well-defined and unique for ``when did I last see...?".
\end{enumerate}

For each annotation, we apply automated and manual quality checks to ensure correctness. In case the quality falls below a certain threshold, the clip is reannotated.

For visual queries associated with 3D scans, we also collect 3D annotations in the form of 3D bounding boxes capturing where the object was last seen. We then use those bounding boxes to establish the ground truth displacement vector from the query frame to the object, which is the target of the task. Each annotation $a_q$ is collected in the scan coordinate system $s$:
\begin{gather}
    T_s = [R_s | t_s],
\end{gather}
 where $q \in \{1,\dots,\mathcal{Q}\}$, $\mathcal{Q}$ the total number of queries, and where $T_s \in \mathbb{R}^4$ is the transformation matrix of the bounding box. $R_s$ and $t_s$ are the corresponding rotation and translation for annotation $a_q$.

The annotation procedure is defined as follows:
A query consists of a video clip, a visual crop, and a response track.  For each query, the goal is to retrieve in the scan the location of the object defined in the video. Once the location is found, we draw a 3D bounding box at this position with the appropriate scale and orientation. It is important to note that 3D scans and videos have been recorded at different times. Therefore, it is likely that an object at a certain location in the video will not be present at that same location in the 3D scan. In such cases, we ask the annotator to hallucinate a 3D bounding box in the 3D scan at the position of the target object defined in the video.

In order to validate an annotation we collect two 3D bounding boxes per query from two different annotators. Leveraging the two boxes we compute the following validation metrics:
\begin{gather}
    d_{norm} = \frac{\lVert c_1 - c_2 \rVert_2 }{m_{diag}} \\
    V_{norm} = \frac{V_{global}}{V_{union}},
\end{gather}
where $c_1$ and $c_2$ are the centroids of the two boxes, $m_{diag}$ is the average diagonal length of the two boxes, $V_{global}$ is the volume of the 3D convex hull of the two boxes, and $V_{union}$ is the volume of the union of the two boxes. These metrics measure the agreement level betwen the two annotators. When the two annotations are perfectly aligned, the metrics are equal to $d_{norm}=0$ and $V_{norm}=1.0$. The assumption is that if the two annotators agree on the position, scale, and orientation of the bounding box then it is likely to be correct. If the two annotations are far from each other we will discard the query. There are a couple of reasons that can explain such case: (1) one annotator mislabeled the query, (2) the query is hard to annotate. 
Some queries require a significant amount of hallucination to retrieve the object location in the scan which clearly leads to subjective annotations.
We empirically defined two thresholds of 1.5 over $d_{norm}$ and 15 over $V_{norm}$ to filter out poor annotations. Any query that has either one of the two metrics above the threshold of acceptance is rejected.

\paragraph{Natural language queries}
To collect NLQ annotations, we sample contiguous clips of length 8 minutes and 20 minutes. %
The annotators are instructed to watch these clips and generate natural language queries, focused on retrieving information about objects, places, and people in the egocentric
video clips.
To reduce the cognitive overload on the annotators, and focus their efforts on memory-relevant queries, we also provide a list of $13$ 
query templates (see Table~\ref{tab:nlq_query_templates}), corresponding to 
queries a user might ask to augment their memory.
Note that these templates are provided only to guide their choice of query, and 
does not limit the linguistic variability since the annotators are instructed
to paraphrase the template without copying them as is.

To elaborate, the annotators performed the following steps:
\begin{enumerate}
    \item Watch the entire video clip $\mathcal{V}$ in order to understand the high-level
    context (optionally in $2\times$ fast-forward),
    \item Pick a query template from the available list and paraphrase/reword the query to obtain $\mathcal{Q}$,
    \eg, template \textit{`Where was object X before/after event Y?'} can be paraphrased
    as \textit{`Where was the blue bucket prior to my dog exiting the living room?'}
    \item Find the temporal window where the response to the natural language query can be deduced visually, and annotate it as $r$.
\end{enumerate}

\begin{table}[t]
    \centering
    \scalebox{0.95}{%
        \begin{tabular}{c l}
            \toprule
            Category & Template \\
            
            \midrule
            \multirow{9}{*}{Objects}
                & Where is object X before / after event Y? \\
                & Where is object X?  \\
                & What did I put in X?  \\
                & How many X’s? (quantity question)  \\
                & What X did I Y?  \\
                & In what location did I see object X ?  \\
                & What X is Y?  \\
                & State of an object  \\
                & Where is my object X?  \\
            \midrule
            Place & Where did I put X?  \\
            \midrule
            \multirow{3}{*}{People}
                & Who did I interact with when I did activity X?  \\
                & Who did I talk to in location X?  \\
                & When did I interact with person with role X?  \\
            \bottomrule
        \end{tabular}
    }
    \caption{The NLQ templates capture a diverse set of queries that humans can
    ask to augment their memory and recollect objects, places, and people in their everyday experience.}
    \label{tab:nlq_query_templates}
\end{table}

During our data collection, we also requested the annotators to mark the slot values 
and corresponding verbs, for the selected language query templates. 
While we do not use this information for our task, it may be useful for other future research.

The desiderata for the collected queries are as follows.  They should:
(a) reflect the underlying motivation of augmenting human memory,
(b) be rich and diverse in terms of language and the objects, places, people, and 
events, and,
(c) be challenging enough for an intelligent system but not too complicated or
convoluted to reduce the naturalness of the queries.
For instance, though a query like \textit{`What was playing on the television when I was 
folding my seventh T-shirt after my dog exited the room?'} is challenging from a 
learning perspective, it is not natural from an application standpoint.
In order to ensure the above qualities for NLQ, we enforce the following constraints:
\begin{itemize}
    \item All paraphrased language queries must be in past tense, and must be posed as 
    questions asked at the end of the entire video clip. 
    This resembles the real-life scenario of querying about episodic memory (past) of 
    the user, and resolves ambiguity when there are multiple occurrences of an object to the
    the last relevant one.

    \item To account for momentary shifts of view for the egocentric video, we allow
    small interruptions ($<$ 3 seconds) between the truly relevant frames for a given query.
    In other words, frames where the object/person/place of interest goes out of view for less
    than 3 seconds as a result of momentary gaze shift are still considered to be contiguous.

    \item For a given query, if there are multiple non-contiguous temporal windows (separated by
    more than 3 seconds) as independently valid answers, we instruct the annotators to either
    discard the query and create a different one, or add more details to the wording to make it
    more specific.
    Similarly, queries that require multiple temporal windows (separated by more than 3 seconds) 
    to deduce the answer are also disallowed. For example, 
    \textit{`How many shirts did I pack in my suitcase?"} is invalid if packing happens across
    multiple temporal windows, separated by more than 3 seconds (\eg, the user pauses to make
    coffee, and then returns to packing).
    
    \item We encourage diversity by instructing that the query responses not be concentrated at one 
    part of the video clip, or around few objects/places/people.
    In addition, we also disallow the query response window to be more than $50\%$ of the total
    clip length.
    
    \item Finally, queries that require reasoning and knowledge on top of visual evidence are 
    invalid.
    For instance, \textit{`What country`s flag was hanging on the wall?"} is invalid while 
    \textit{`Where was the flag that was hanging on the wall?"} is valid.
    Similarly, queries that guess the motivation or intentions of the user or people
    in the video clip are also not allowed.
    As an example, \textit{`Why did the person at the door leave a package on the porch?'} is 
    disallowed while \textit{`What did the person leave on the porch?'} is accepted.
\end{itemize}

After the annotation process, we apply both automatic and manual quality checks, including the 
diversity of language queries and temporal window locations, to score the annotations.
If the overall quality score is below a threshold, the clip is re-annotated.

\paragraph{Moments queries} 

To annotate moments queries, we sample contiguous clips of 8 minutes from the set of interesting moments videos. The annotators are instructed to mark instances of activities with a temporal window and the activity's name from a fixed taxonomy of activities.  \FEB{We have each instance labeled by three independent annotators.  By assuming each annotator is reliable, we take the union 
of moments across annotators to ensure completeness of annotations.}

The taxonomy was created semi-automatically from the narrations. Specifically, we use the \emph{summary} narrations collected for five-minute clip segments, as they capture higher-level events and activities that are suitable for the moments retrieval task. This is in contrast to the verb-noun taxonomy that is sourced from individual narrations for each atomic action, which are used in the Forecasting and Hands and Objects benchmarks (see Appendices~\ref{appendix:hands-objects} and~\ref{appendix:forecasting}).

The taxonomy was created as follows. First, each summary narration was encoded into a feature vector using a pre-trained BERT~\cite{bert} language model, and then concatenated with the word embeddings for the main verb and noun extracted from the summary. These summaries were then clustered into groups, and then labels were manually assigned to groups based on the coherent activities they described. 

Note that this process was done independently for a set of scenarios that we selected based on how frequently they occur in the dataset, the diversity of activities they represent, and how likely they contain high-level, event-like activities. For example videos that primary involve a single activity like ``driving'' are not interesting categories in this context, whereas ``household cleaning'' contains several different activities that are shared across other indoor tasks, making it an appropriate scenario. In total, we select videos from 5 scenarios to create our moments taxonomy: Cooking, Cleaning, Shopping, Handyman, Farmer/Gardener. 
Each annotation is in the format of (start time, end time, label).

\begin{figure*}
\centering
\includegraphics[width=1.00\textwidth]{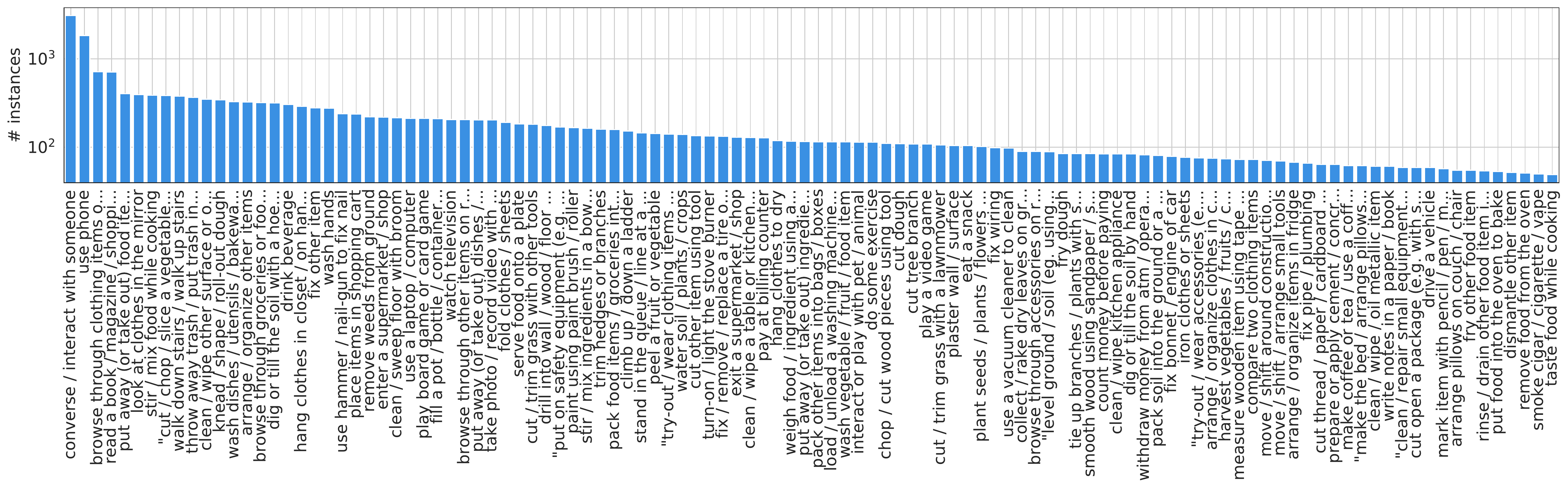}
\caption{\textbf{Distribution of moments labels}. The figure shows the number of instances per category across 5 scenarios and 300 hours of data. All 110 categories are shown, sorted by frequency. The distribution is long tailed, with the smallest classes containing at least 50 instances. Note that these are only the Moments for Episodic Memory with temporal window annotations in the current release; Ego4D has many other scenarios and activities not reflected in this distribution.}
\label{appfig:moments_label_dist}
\vspace{-2mm}
\end{figure*}

\subsubsection{Data Analysis}
We now overview the statistics of the annotations per query type.

\paragraph{Visual queries}
The VQ annotations consist of samples from a diverse set of scenarios and universities (see Figure~\ref{appfig:vq_scenario_stats}\iftoggle{arxiv}{  and~\ref{appfig:vq_univ_stats}}{}). In total, \FEB{$433$} hours of videos are annotated with \FEB{$22,602$} visual queries. These videos are sampled from 10 universities and consist of \FEB{54} scenarios. The statistics over the train/val/test splits are provided in Table~\ref{tab:vq_splits}. We ensured that the splits contain a disjoint set of videos. To look for possible biases in the data, we plot the distribution over three measures.

\noindent\textbf{1) Query to response separation} is the temporal distance (in frames) between the query frame and the end of the response track. This measures how far back in time an algorithm needs to search in order to find the query object. \\
\noindent\textbf{2) Response track size} measures the temporal length of the response track. \\
\noindent\textbf{3) Response bbox position} is the spatial start and end $(x, y)$ coordinates for each bounding box in the response track. We normalize the coordinates by the image width and height to account for varying image sizes in the data. Each pixel within the bounding box contributes to an image heatmap that shows the frequency of each pixel belonging to a response track bounding box. 

The analyses are shown in Figure~\ref{appfig:vq_biases}. The query to response separation distances are fairly spread between 1 to 200 frames with a mode of $\sim 30$ frames (see Figure~\ref{appfig:vq_biases}, left). The response track sizes are well distributed between 1 to 40 frames with a mode of $\sim 8$ frames (see Figure~\ref{appfig:vq_biases}, center). The bounding boxes are near-uniformly distributed throughout the image, with very few bounding boxes annotated at the top $10\%$ of the image (see Figure~\ref{appfig:vq_biases}, right). Our analyses indicate that there may be a potential bias in the first two measures, while the bounding boxes positions are largely unbiased.

\begin{figure}[t]
    \centering
    \includegraphics[width=0.95\columnwidth]{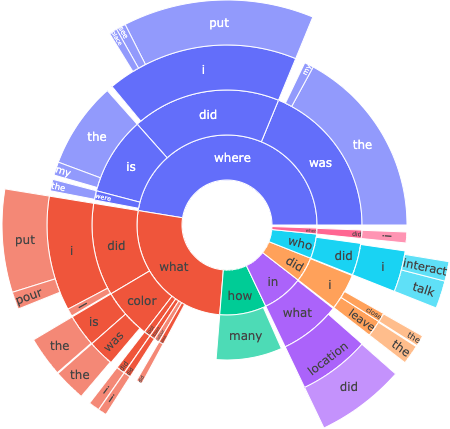}
    \caption{Distribution of query words in NLQ.}
    \label{appfig:nlq_sunburst}
\end{figure}

For the 3D localization task, we annotate a subset of 1,043 visual queries with 3D annotations. These comprise of 13 video hours associated
with 4 scans from \iftoggle{arxiv}{the University of Catania (UNICT)}{one of the universities}.

\begin{table}[t]
\centering
\scalebox{0.9}{
\begin{tabular}{@{}l|ccc@{}}
\toprule
\multicolumn{1}{r|}{Split} & \multicolumn{1}{c}{Train}  & \multicolumn{1}{c}{Val}   & \multicolumn{1}{c}{Test} \\ \midrule
\# video hours             & \FEB{262} (\FEB{19})              & \FEB{87} (\FEB{5})              & \FEB{84} (\FEB{9})             \\
\# clips                   & $3.6k$ (\FEB{164})                & $1.2k$ (\FEB{44})               & $1.1k$ (\FEB{69})              \\
\# queries                 & \FEB{$13.6k$} (\FEB{604})        & \FEB{$4.5k$} (\FEB{164})        & \FEB{$4.4k$} (\FEB{264})       \\ \bottomrule
\end{tabular}
}
\vspace*{0.05in}
\caption{\textbf{Visual queries dataset statistics}. The numbers in the parantheses correspond to the subset of data used for 3D localization, where we focus on videos for which we have Matterport3D scans.}
\label{tab:vq_splits}
\end{table}

\begin{figure*}\centering
\includegraphics[width=1.00\textwidth,trim={0 0.3cm 0 0},clip]{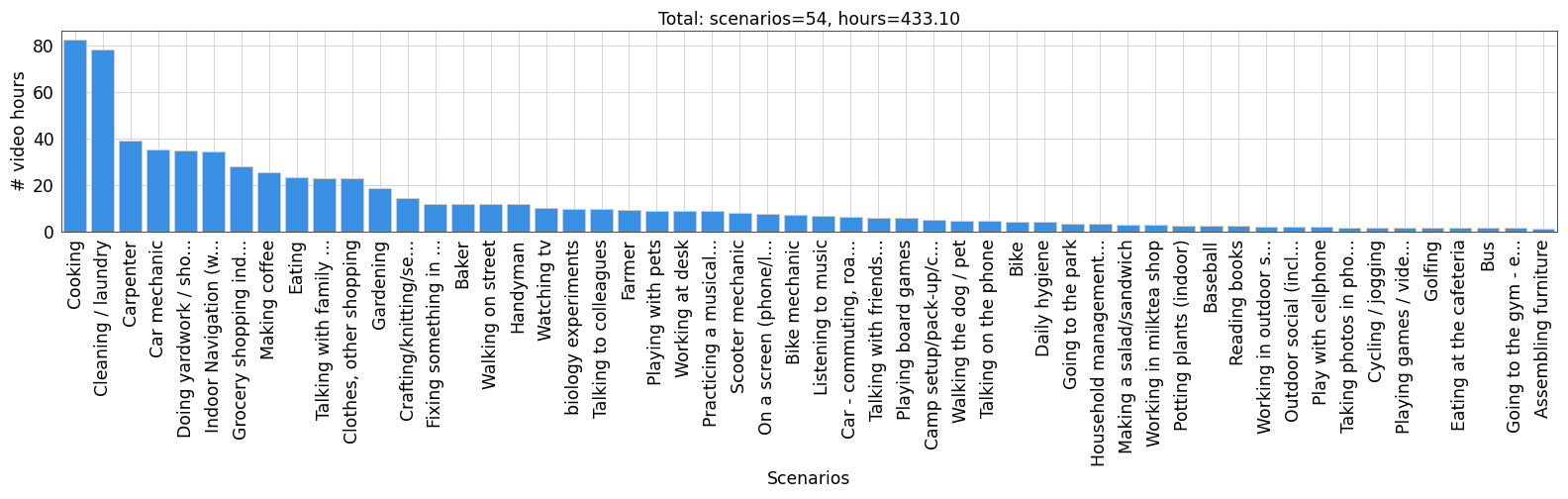}
\caption{\textbf{Distribution over scenarios for visual queries.} The dataset contains a long-tail of scenarios. The plot title indicates the number of scenarios and the total video hours included in the dataset.}
\label{appfig:vq_scenario_stats}\vspace{-2mm}
\end{figure*}

\iftoggle{arxiv}{
\begin{figure}\centering
\includegraphics[width=0.45\textwidth,trim={0 0.5cm 0 0},clip]{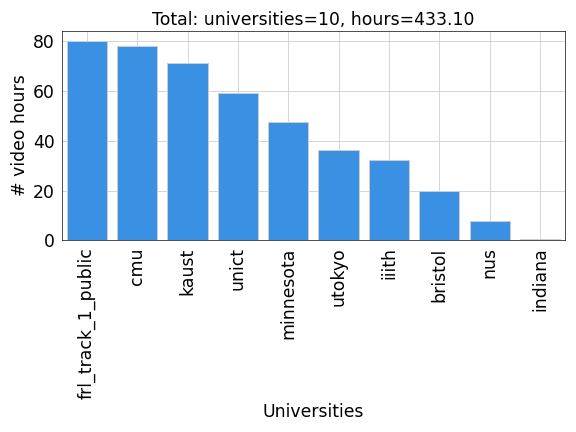}
\caption{\textbf{Distribution over universities for visual queries.} The dataset contains annotations corresponding to videos from 10 universities. The plot title indicates the number of universities and the total video hours included in the dataset.}
\label{appfig:vq_univ_stats}\vspace{-2mm}
\end{figure}
}{}  %

\begin{figure*}\centering
\includegraphics[width=1.00\textwidth,trim={0 8.1cm 1.8cm 0},clip]{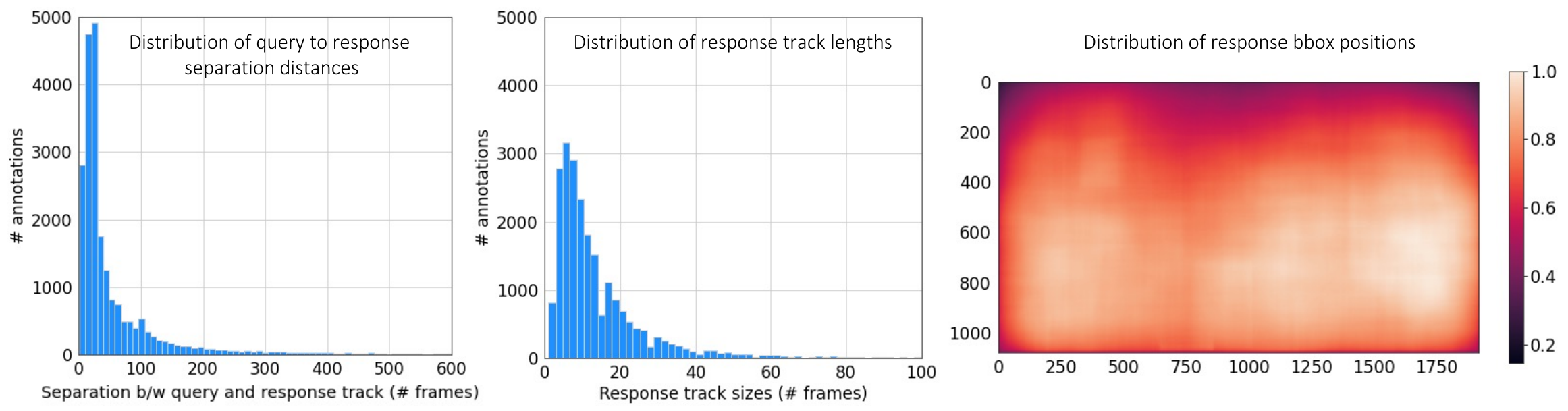}
\caption{\textbf{Visual queries bias analysis.} We analyze the full VQ dataset for potential biases. \textbf{Left:} The plot shows the distribution of query to response separation distances in the VQ dataset. While the mode of the distribution is $\sim 30$ frames, we can see that separation distances are fairly spread between 1 to 200 frames. \textbf{Center:} The plot shows the distribution of response track sizes in the VQ dataset. While the mode of the distribution is $\sim 8$ frames, we can see that the response track sizes are well distributed between 1 to 40 frames. \textbf{Right:} The heatmap shows the normalized frequency of each pixel belonging to a response track bounding box. The bounding boxes near-uniformly distributed across most of the image.}
\label{appfig:vq_biases}\vspace{-2mm}
\end{figure*}

\paragraph{Natural language queries}
As outlined in Table~\ref{apptab:nlq_statistics}, the NLQ annotations are from $227$ hours of 
video, with a total of $19.2K$ queries spanning the selected $13$ query templates.
The associated video clips come from $10$ different universities with a total of $34$ scenarios 
(with at least 1 hour of video annotated).
Similar to other tasks within the episodic memory, we ensure that the
train/val/test splits ($60\%, 20\%,20\%$) contain a disjoint set of video clips.
We further analyze the data through:
(a) Distribution over template queries, shown in Figure \ref{appfig:nlq_query_template}. %
The challenging \textit{`Where is object X before/after event Y?'}
is the most popular template with around $3K$ queries, with a reasonable distribution over other templates.
Overall, the queries in NLQ have $8.3 \pm 2.1$ words in them.
(b) Distribution of the response window length is shown in Figure \ref{appfig:nlq_window_length}. 
Typically, the windows are $9.3 \pm 21.5$ seconds long.

Most response windows are quite short compared to the full video clip, making the task a challenging ``needle in the haystack" search problem.
(c) Distribution of query words is shown in Figure \ref{appfig:nlq_sunburst}.
The branching off evidences the richness and diversity of the queries in NLQ.

\begin{table}[t]
\centering
\begin{tabular}{c c c c}
    \toprule
    Split & Train  & Val   & Test \\
    \midrule
    \# video hours & $136$ & $45$ & $46$ \\
    \# clips  & $1.0k$ & $0.3k$ & $0.3k$ \\
    \# queries  & $11.3k$ & $3.9k$ & $4.0k$ \\
    \bottomrule
\end{tabular}
\caption{\textbf{NLQ dataset statistics} across the train/val/test splits.}
\label{apptab:nlq_statistics}
\end{table}

\begin{figure*}[ht]
    \includegraphics[width=\textwidth]{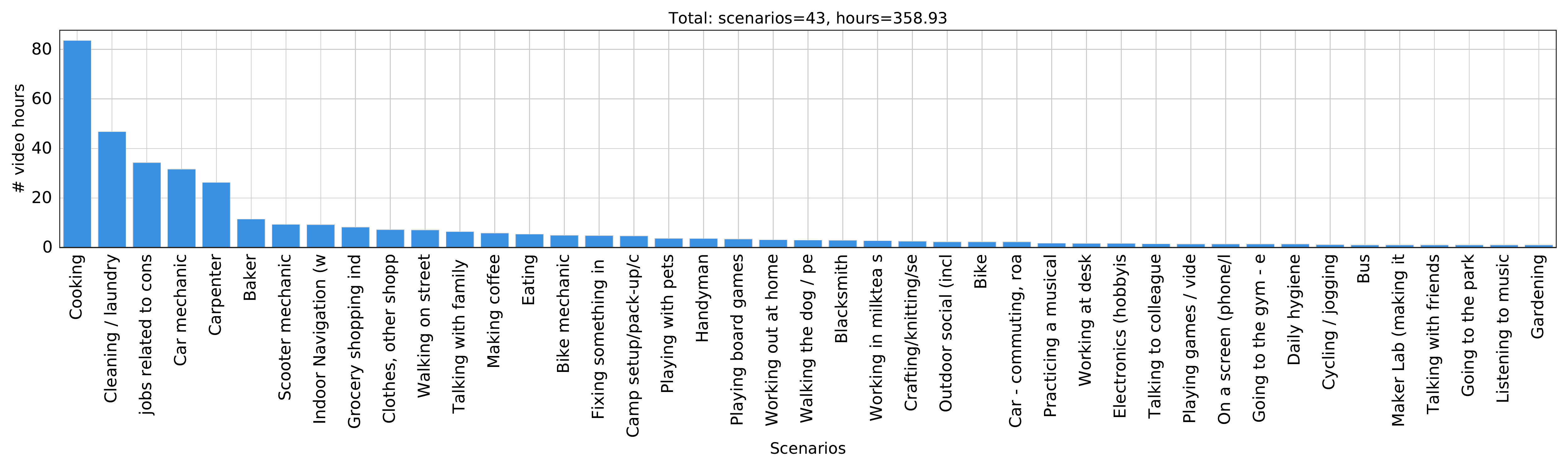}
    \caption{
    Distribution over scenarios for the NLQ annotations, indicating a long tail
    over scenarios.
    Note that the scenario labels are approximate and a single video can contain multiple 
    scenario labels. For this plot, we equally divide the time across all the labelled scenarios.
    }
    \label{appfig:nlq_scenarios}
\end{figure*}

\begin{figure}[ht]
    \centering
    \includegraphics[width=0.95\columnwidth]{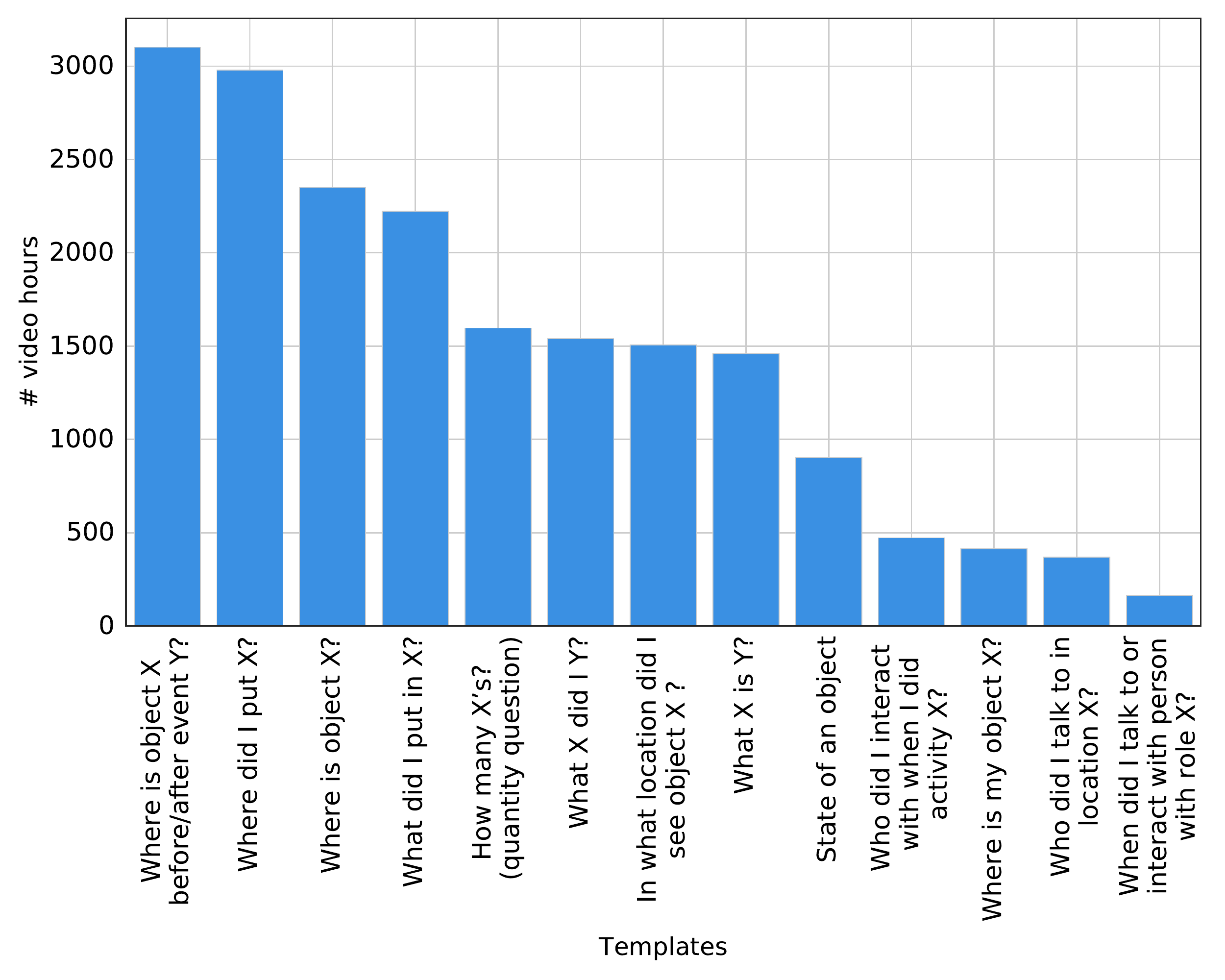}
    \caption{
        Distribution of queries over the corresponding templates across objects, place, and people
        categories (Tab.\ref{tab:nlq_query_templates}). See text for more details.
    }
    \label{appfig:nlq_query_template}
\end{figure}
\begin{figure}[ht]
    \centering
    \includegraphics[width=0.9\columnwidth]{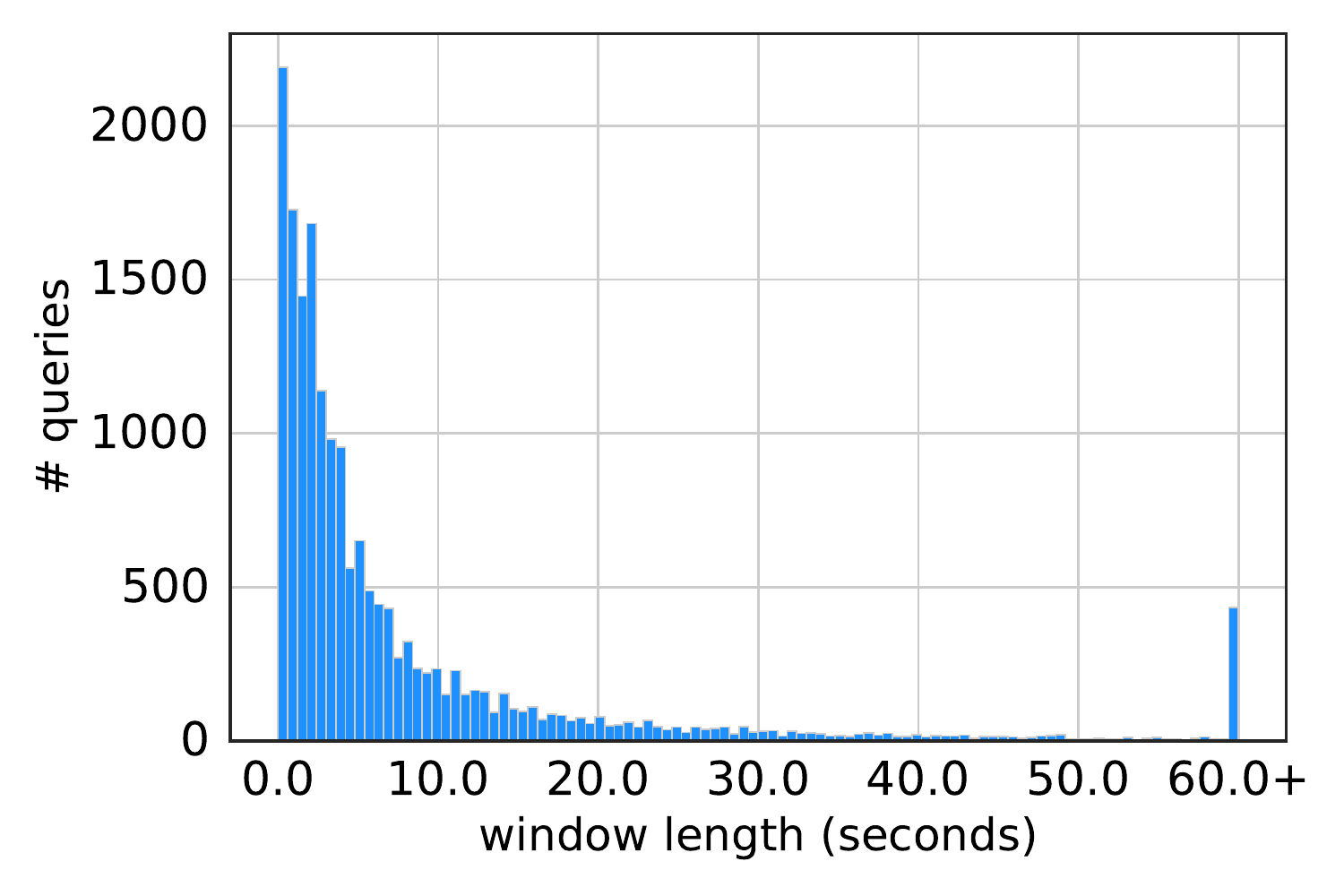}
    \caption{
        Distribution of response window length for NLQ. For the sake of brevity, we use the
        last bin to represent all windows longer than a minute.
        See text for more details.
    }
    \label{appfig:nlq_window_length}
\end{figure}

\paragraph{Moments queries}
 For MQ, similar to other tasks in episodic memory, we  maintain a ratio of 6:2:2 among the train/val/test splits, which contains disjoint sets of video clips. To make sure there are enough samples in each category, we only keep categories that have at least 50 instances from the annotations and have instances in all train/val/test splits.

\begin{figure}
\centering
\includegraphics[width=0.49\textwidth]{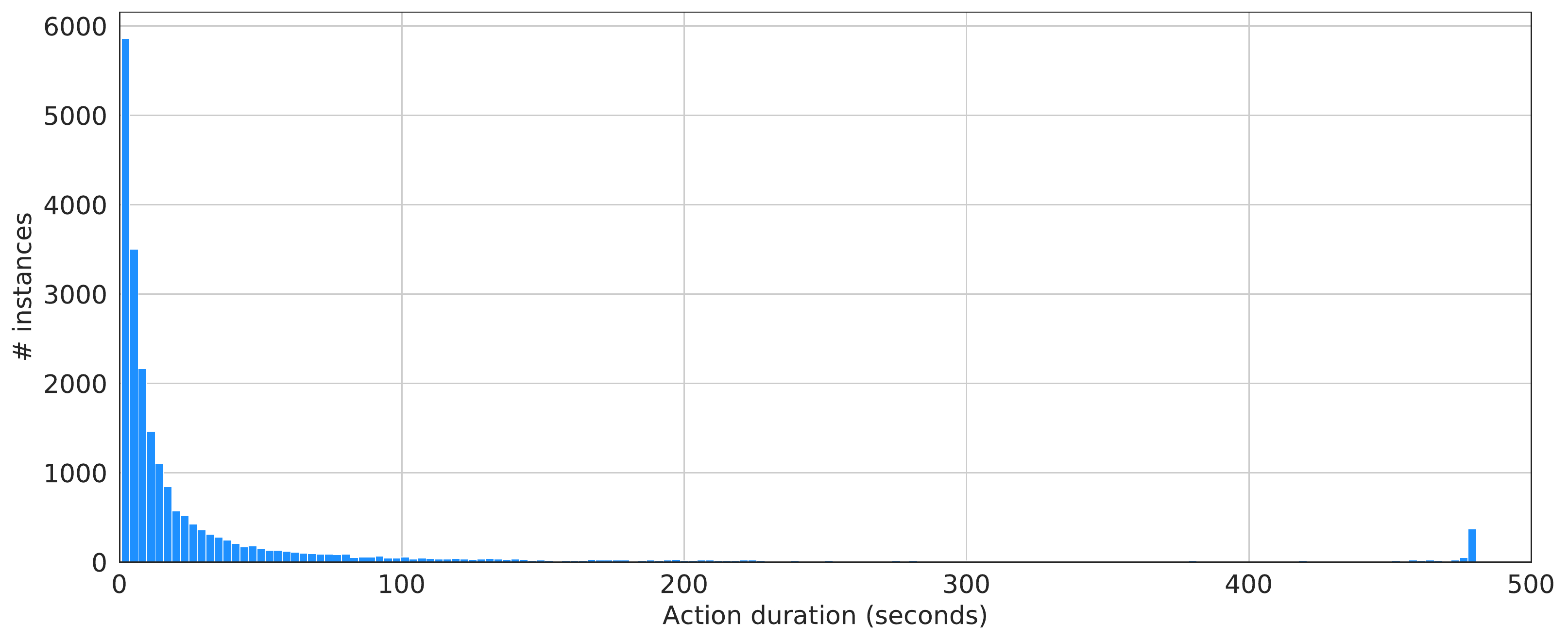}
\caption{\textbf{Distribution of moment duration}. }
\label{appfig:moments_action_duration}
\vspace{-2mm}
\end{figure}

\begin{figure}
\centering
\includegraphics[width=0.49\textwidth]{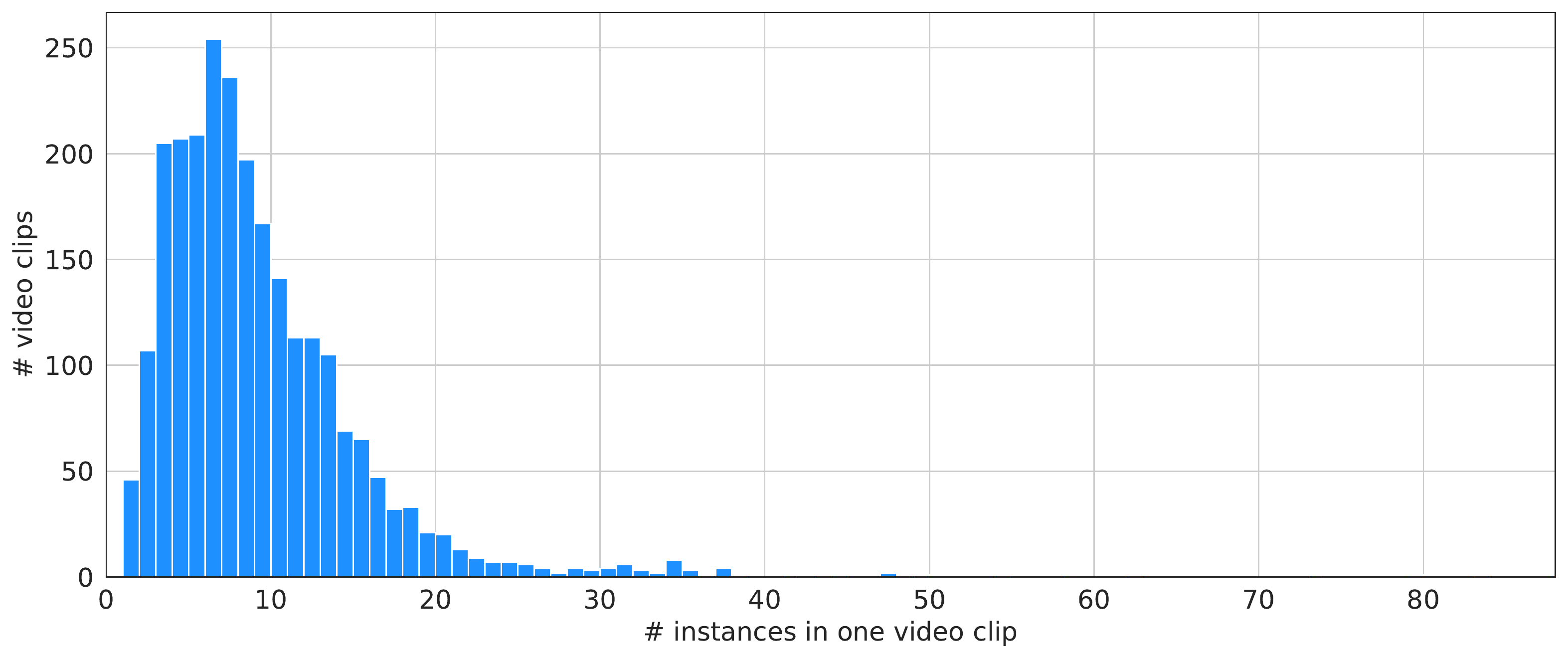}
\caption{\textbf{Distribution of instance numbers in one video clip}. }
\label{appfig:moments_num_instance}
\vspace{-2mm}
\end{figure}

Consequently, the MQ dataset has 110 categories, spans a total \FEB{326.4} hours of videos, \FEB{2,488} video clips and \FEB{22.2$k$} action instances.  We summarize the statistics across the three splits  in Table~\ref{apptab:moment_statistics}. We further explore the data through the following aspects.  (a) The distribution of action duration is shown in Fig~\ref{appfig:moments_action_duration}. We can see that most moments have very short duration. The majority of moments last less than 1 minute, and 22.4\% actions have duration less than 3 seconds. Note that there is also a peak (2.6\% instances) at the largest duration bin, where the actions almost cover the whole video clip. The average duration each instance is 45.2 seconds. (b) The distribution of different  categories is shown in Fig~\ref{appfig:moments_label_dist}. We notice that this is a long-tailed distribution,  some categories (e.g., `use phone', `converse/interact with someone') with over 1000 instances and some categories with less than 100 instances.  Each category has 205 instances on average. (c) The distribution of instance numbers in a video clip is shown in Fig~\ref{appfig:moments_num_instance}. The majority of video clips have 1-20 moment instances, whereas very few can have as many as over 80 instances.

\begin{table}[t]
\centering
\begin{tabular}{c c c c c}
    \toprule
    Split & Train  & Val   & Test & Total \\
    \midrule
    Video hours & $\FEB{194.9}$ & $\FEB{68.5}$ & $\FEB{62.9}$ & $\FEB{326.4}$\\
    \# Video clips  & $\FEB{1,486}$ & $\FEB{521}$ & $\FEB{481}$ & $\FEB{2,488}$ \\
    \# Instances  & $\FEB{13.6k}$ & $\FEB{4.3k}$ & $\FEB{4.3k}$ & $\FEB{22.2k}$\\
    \bottomrule
\end{tabular}
\caption{\textbf{MQ dataset statistics} across the train/val/test splits.}
\label{apptab:moment_statistics}
\end{table}

\subsubsection{Evaluation measures}

Next we detail the evaluation metrics for all three query types. 
\paragraph{Visual queries}

We define the following localization metrics for the 2D localization task with top-1 retrieval.\\
\textbf{Temporal AP (tAP)} measures how closely the temporal extent of the prediction matches with the ground-truth response track. It is calculated as the average-precision of the predicted response track's temporal extent, and is based on the ActivityNet mAP metric~\cite{caba2015activitynet}. We evaluate the tAP at 4 different tIoU thresholds \{0.25, 0.50, 0.75, 0.95\}, as well as their average value. \\
\textbf{Spatio-temporal AP (stAP)} measures how closely the spatio-temporal extent of the prediction matches the ground-truth response track. It is calculated as the average-precision of the predicted spatial-tube, and is based on the video-AP metric from~\cite{gkioxari2015actiontubes}.  We evaluate the stAP at 4 different stIoU thresholds \{0.25, 0.50, 0.75, 0.95\}, as well as their average value. \\
\textbf{Success (Succ)} measures whether the prediction has any overlap with the ground truth at all. It is calculated as the percentage of samples where the predicted response track has atleast $0.05$ spatio-temporal IoU with the ground truth. \\
\textbf{Recovery\% (rec\%)} measures how much of the ground-truth response track is accurately recovered by the prediction. It is calculated as the \% of frames in the response track where the predicted bounding box has at least $0.5$ IoU with the ground truth. This is motivated by the tracking robustness metric from the VOT challenge~\cite{kristan2020vot}.\\
\textbf{Search efficiency (sEff)} measures the efficiency of the algorithm searching for the query object. It is calculated as 
\begin{equation}
    \textrm{sEff} = 1 - \frac{n}{N}
\end{equation}
where $n$ is the number of video frames previewed by an algorithm to predict the response track, and $N$ is the total number of frames in the video before the query was made (i.e., the search window). An algorithm that accesses every frame in the search window before localizing the query object gets $0.0$ search efficiency. This ``timeliness" metric is designed to encourage research on methods performing intelligent contextual-search.

We evaluate performance on the 3D VQ localization task using the root mean square error (RMSE) and the angular error metrics:
\begin{gather}
    \textrm{RMSE} = \lVert t_s - \hat{t}_s \rVert_2 \\
    \textrm{angular\_error} = \textrm{acos}(\frac{v_Q^T}{\lVert v_Q \rVert_2} . \frac{\hat{v}_Q}{\lVert \hat{v}_Q \rVert_2})
\end{gather}
where $t_s$ and $\hat{t}_s$ are the ground-truth and predicted object position in the scan coordinate system. $v_Q$ and $\hat{v}_Q$ are the ground-truth and predicted 3D displacement vector in the query frame $Q$ coordinate system. We also define a success metric leveraging the two annotations per query:
\begin{gather}
    \textrm{succ} = \lVert c_m - \hat{t}_s \rVert_2 < \FEB{6} \times (\lVert c_1 - c_2 \rVert_2 + \delta)
\end{gather}
With $c1$ and $c2$ the centroids of the two bounding box annotations, $c_m$ the mid-centroid between $c1$ and $c2$ and $\delta = \exp^{-m_{diag}}$, with $m_{diag}$ the average diagonal length of the two boxes.

\paragraph{Natural language queries}
Evaluation for NLQ %
is similar to existing video-language
grounding problems.
Following prior work \cite{2DTAN_2020_AAAI}, we use recall@k, IoU=m, where
we select $k = \{1, 5\}$ and $m = \{0.3, 0.5\}$.
This metric computes the percentage of times at least one of the top $k$ predicted candidates
have an intersection-over-union (IoU) of at least $m$.
Note that we lean towards lower threshold values ($m$) as the average length of the
window ($\sim$$10s$) is much smaller than that of the video clip ($500s$), about
$2\%$ of the clip length.

\paragraph{Moments queries}

Considering that the moment queries task is related to the tasks of temporal action detection~\cite{lin2018bsn,xu2020g,zhao2020video, caba2015activitynet} and video grounding~\cite{2DTAN_2020_AAAI}, we adapt their respective metrics to  moment queries.  %

\noindent \textbf{Average Precision (AP)} is a commonly adopted metric in temporal action detection. It measures how closely the temporal extent  of  the  predictions  matches  the  ground-truth action instances for each action category~\cite{lin2018bsn,xu2020g,zhao2020video, caba2015activitynet} in terms of both precision and recall. The temporal intersection over union (tIoU) between a prediction and a ground-truth action instance is used to measure their distance. If the tIoU is higher than a threshold, the prediction is considered as true positive; otherwise, false positive. In representative temporal action detection datasets, such as ActivityNet~\cite{caba2015activitynet}, the mean AP (mAP) over all categories is computed given a tIoU threshold. Multiple tIoU thresholds are adopted, and the average mAP over all these tIoU thresholds is computed. For moment queries, we evaluate mAP at 5 different tIoU thresholds \{0.1, 0.2, 0.3, 0.4, 0.5\}, as well as their average value.

\noindent \textbf{Recall@kx, tIoU=m}, is a metric adapted from the metric recall@$k$, tIoU=$m$, used for NLQ. The metric recall@$k$, tIoU=$m$  measures the percentage of the query sentences that have at least one  prediction with a tIoU larger than the threshold $m$ in the top-$k$ results. In our moment queries case, since we might have more than one instance corresponding to a query moment category, we need to measure the percentage of all the correctly predicted instances that have at least one  prediction with a tIoU larger than the threshold $m$ in the top-$k$ results of this instance. Considering that  predictions are usually made based on a category not a specific instance,  we modify the metric to be the following recall@$kx$, tIoU=$m$, where $x$ stands for the number of instances for a query category in one video. This metric measures the percentage of all the correctly predicted instances that have at least one prediction with a tIoU larger than the threshold $m$ in the top-$kx$ results of the action category. This metric has a  similar idea to the multi-label metric proposed in \cite{zhou2021embracing} when dealing with multiple instances for a query.  We use $k=1, 2, 3$ and $m=0.3, 0.5, 0.7$ in the metric.  Compared to average precision, this metric only evaluates the recall for the query categories, and does not penalize for false positive predictions given a category that has no instances in the video.

\subsubsection{Baselines} %

We developed baseline models for each task.  We designed these models to address our tasks, using state-of-the-art components where relevant.  They represent a starting point upon which future work can build.

\subsubsection*{Visual queries 2D localization baseline}

\begin{figure*}[t]
    \centering
    \includegraphics[width=\textwidth,trim={0 0 5.5cm 0},clip]{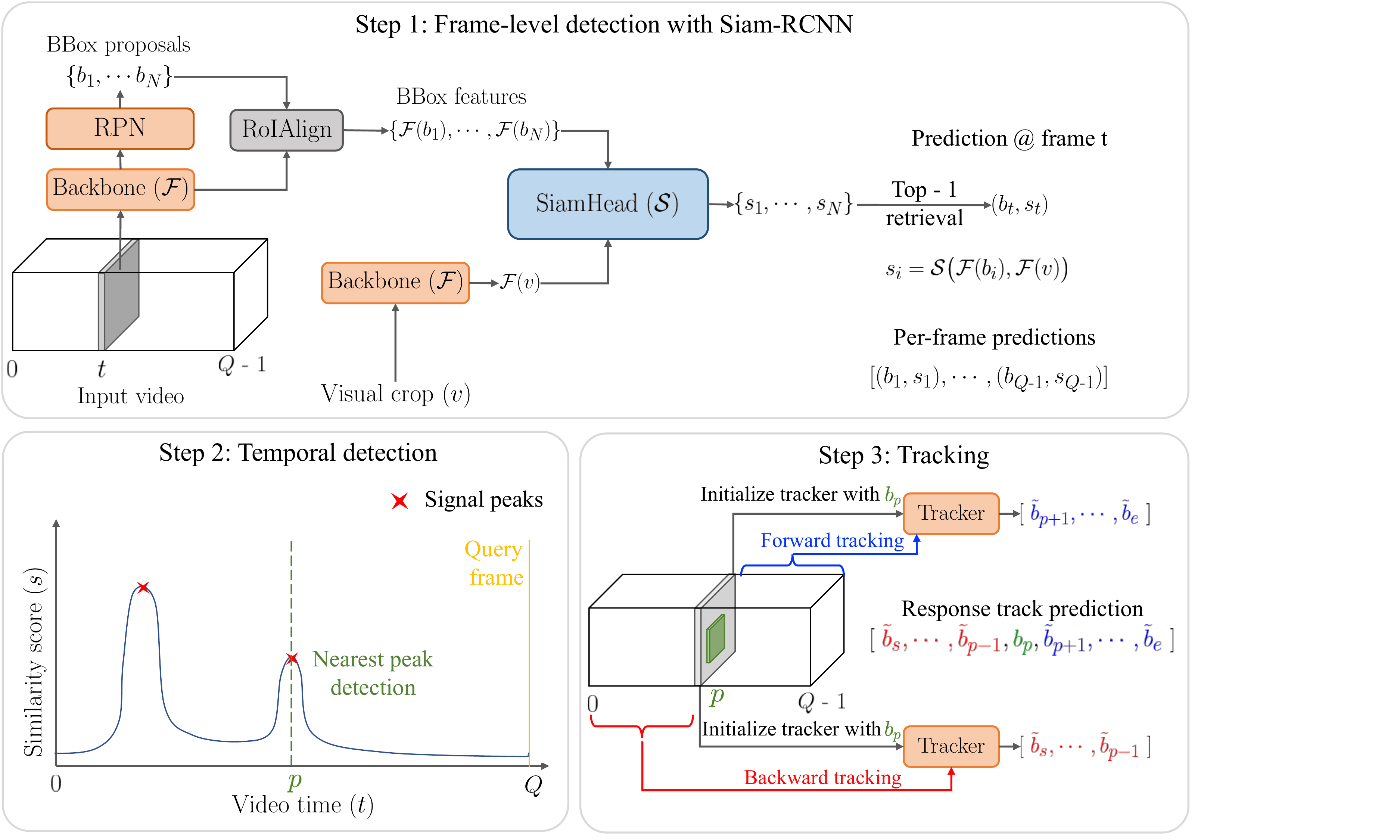}
    \caption{\textbf{Visual queries 2D localization baseline.} Our approach consists of three steps. \textbf{Step 1:} We perform frame-level detection for the entire input video to detect the presence of the query object (specified via the visual crop $v$). For each frame $t$, we extract the region proposals $\{b_1, \cdots, b_N\}$ using a region proposal network (RPN), and extract features for each proposal $\{\mathcal{F}(b_1), \cdots, \mathcal{F}(b_N)\}$. Each proposal feature is compared with the visual crop feature $\mathcal{F}(v)$ using a Siamese head $\mathcal{S}$, and the most similar proposal $b_t$ is retrieved along with its score $s_t$. This process is repeated for all frames. \textbf{Step 2:} We treat the similarity scores $\bm{s} = \{s_1, \cdots, s_{q-1}\}$ as a temporal signal and perform temporal detection to obtain the `most recent occurrence' of the query object. We detect the peaks (local maxima) in the signal and recover the peak $p$ nearest to the query frame. \textbf{Step 3:} Given the detected peak $p$ and its corresponding proposal $b_p$, we initialize two trackers with $b_p$ and run them along the forward and backward directions to recover a contiguous track of the object, i.e., the response track prediction.}
    \label{fig:vq_baseline}
\end{figure*}
We treat visual queries with 2D localization (VQ2D) as a detection + tracking problem (see Figure~\ref{fig:vq_baseline}). At a high level, our approach consists of three steps. First, we perform frame-level detection over the input video where we detect the presence of the query object in each frame using an object detection model (Figure~\ref{fig:vq_baseline} top). For each frame, we get the bounding box that is most similar to the visual crop and a score indicating its visual similarity. Second, we consider the sequence of per-frame similarity scores over the entire video and 
identify the most recent peak in these scores (Figure~\ref{fig:vq_baseline} bottom-left). Finally, we initialize a tracker at the video-frame corresponding to the peak detection, and track the query object
on both forward and backward directions to recover the complete response track (Figure~\ref{fig:vq_baseline} bottom-right).

\paragraph{Step 1: Frame-level detection} We propose Siam-RCNN, a Faster-RCNN~\cite{ren2015faster} based approach to detect the query object
in a given image. See Figure~\ref{fig:vq_baseline} top.
Given a video frame at time $t$, a pre-trained Region Proposal Network (RPN)~\cite{ren2015faster} with a Feature Pyramid Network (FPN)~\cite{lin2017fpn} backbone is used
to generate bounding box proposals $\{b_1, \cdots, b_N\}$. The RoI-Align operation~\cite{he2018maskrcnn} is then used to extract visual features for each bounding box $\{\mathcal{F}(b_1), \cdots, \mathcal{F}(b_N)\}$.
We use the same FPN backbone to extract features for the visual crop $v$. 
To detect the presence of the query object in frame $t$, each proposal feature $\mathcal{F}(b_i)$ is compared with the visual crop feature $\mathcal{F}(v)$ using a Siamese head $\mathcal{S}$ that predicts a 0-1 similarity score
\begin{equation}
s_i = \mathcal{S}(\mathcal{F}(b_i), \mathcal{F}(v))
\end{equation}
The Siamese network projects each proposal / visual-crop feature to a 
1024-D feature vector using a convolutional projection module $\mathcal{P}$,
\begin{equation}
    p_b = \mathcal{P}(\mathcal{F}(b_i));~
    p_v = \mathcal{P}(\mathcal{F}(v))
\end{equation}
and predicts a $0$ - $1$ similarity score using a bilinear operation: 
\begin{equation}
    s_i = \sigma(p_b^T W p_v + b)
\end{equation}
where $\sigma$ is a sigmoid non-linearity. After computing the similarities to each bounding box proposal, the proposal $b_t$ with the highest similarity score $s_t$ for frame $t$ can be obtained as follows:
\begin{gather}
   b_t = \argmax_{b \in \{b_1, \cdots, b_N\}} \{s_1, \cdots, s_N\} \\
   s_t = \max \{s_1, \cdots, s_N \}
\end{gather}
After repeating the above steps for all the video frames, we can obtain the final per-frame predictions as 
$[(b_1, s_1), \cdots, (b_{q-1}, s_{q-1})]$.

\paragraph{Step 2: Temporal detection} %
So far, we used Siam-RCNN to get the most similar proposals and their similarity
scores for every frame in the video. Next, the goal is to temporally detect the `most recent occurrence` of the object in the video (see Figure~\ref{fig:vq_baseline} bottom-left).
This is a challenging problem since our goal is not to identify the best detection of the object, but instead the most recent one, even if the similarity is not as high. 
To tackle this problem, we treat the per-frame similarity scores $\bm{s} = \{s_1, \cdots, s_{q-1}\}$ as a temporal signal, and use a signal peak detection 
approach to identify the salient peaks (a.k.a. local maxima) in $\bm{s}$. To avoid spurious peaks, we first smooth $\bm{s}$ using a median filter with a window size of $5$. 
\begin{gather}
    \bm{\bar{s}} = {\small\texttt{median\_filter}}(\bm{s}) \\
    p_1, \cdots, p_k = {\small\texttt{find\_peaks}}(\bm{\bar{s}})
\end{gather}
Depending on the video, the algorithm may return multiple peaks spread throughout the video (see signal peaks in Figure~\ref{fig:vq_baseline} bottom-right). Since our goal is to detect the most recent occurrence of the object,
we select the peak $p$ that is temporally nearest to the query frame.

\paragraph{Step 3: Tracking} After temporal detection, we have identified a peak-frame $p$ in the video which is estimated to have the
most recent occurrence of the object. For this frame $p$, we can obtain the highest-scoring bounding box $b_p$ from the per-frame detections in step 1. Note that this only represents one frame where the object most recently occurred. However, the task objective is to obtain the response track, i.e., the \emph{contiguous set of all frames}, starting from when the object first entered the field-of-view until the object exits the field-of-view. See Figure~\ref{fig:vq_baseline} bottom-right.
To compute the rest of the response track, we use $b_p$ as a starting point, and run a single-object tracker forward
and backward until the tracking fails (i.e., the object exits the field-of-view).

For both directions, we initialize the apperance model of the tracker using the proposal $b_p$. For the forward tracking, we run the tracker starting from frame $p+1$ to $q-1$ and obtain the tracked regions: $\bm{b_f} = [\bar{b}_{p+1}, \cdots, \bar{b}_{e}]$. For the backward tracking, we run the tracking starting from frame $p-1$ to $0$ and obtain the tracked regions: $\bm{b_b} = [\bar{b}_{s}, \cdots, \bar{b}_{p-1}]$. We then concatenate $\bm{b_b}$, $b_p$, and $\bm{b_f}$ to obtain the complete response track prediction. We use the KYS tracker~\cite{bhat2020kys}, which was shown to achieve state-of-the-art results for single-object tracking.

\paragraph{VQ2D baseline training setup} We now discuss the training procedure for the VQ2D baseline. Each datapoint for the VQ2D task (defined on Ego4D videos) consists of the following: video $V$, visual crop image $v$, query frame number $q$, and response track boxes $r = \{r_s, r_{s+1}, \cdots, r_e\}$, where $s$ and $e$ are the start and end frames of $r$, and $r_i$ is a bounding box defined on frame $i$ of video $V$. 

As a high-level overview, we initialize and freeze the backbone $\mathcal{F}$ and RPN using weights from an MS-COCO pre-trained Mask-RCNN model. We use the VQ2D annotations to train the SiamHead ($\mathcal{S}$). We initialize and freeze the KYS tracker using weights pre-trained on GOT-10k~\cite{huang2021got10k}, LaSOT~\cite{fan2019lasot}, and TrackingNet~\cite{muller2018trackingnet} datasets. 

We next detail the training procedure for the SiamHead ($\mathcal{S}$). We use a similarity retrieval approach were the model is trained to predict high visual similarity between the visual crop $v$ and positives, and low visual similarity between $v$ and negatives. The loss function for $\mathcal{S}$ is a binary cross entropy loss defined over each $(v, D_p, D_n)$ tuple (see Eqn.~\ref{eqn:vq2d_loss}), where $D_p = \{p_i\}_{i=1}^{|D_p|}$ are positive detections, $D_n = \{n_j\}_{j=1}^{|D_n|}$ are negative detections, and $s_{x, v} = \mathcal{S}(\mathcal{F}(x), \mathcal{F}(v))$:

\begin{equation}
    \mathcal{L}_{\mathcal{S}} = -\frac{1}{|D_p \cup D_n|} \bigg(\sum_{p \in D_p} \text{log}(s_{p, v}) + \sum_{n \in D_n} \text{log}(1 - s_{n, v})\bigg)
    \label{eqn:vq2d_loss}
\end{equation}

Both positives and negatives are defined based on proposals generated by the RPN. Given a visual crop $v$, a proposal $p_{i}$ for $i \in (s, e)$ is a positive if the $\textrm{IoU}(p_i, r_i) \ge 0.5$, where $r_i$ is the response track box in frame $i$. We remove all $r_i$ which are too small, or have significantly different aspect ratios from the largest box in $r$ since these typically correspond to obstructed views of the object.
A proposal $p_{j}$ is a negative if it satisfies any of the following two conditions:
\begin{enumerate}
    \item  $j \in (s, e)$ and $\textrm{IoU}(p_j, r_j) < 0.5$
\item $p_j$ is sampled from another video. 
\end{enumerate}
We also found it beneficial to use hard-negative mining, where we initially sample a large number of negatives and then select the top-K negatives with the highest loss value. \\

We employ a few different augmentation strategies to artificially expand the dataset. First, we augment each data sample by replacing the visual crop $v$ by a bounding box $r_i$ from the response track. This works because the response track and the visual crop correspond to the same object. Next, we augment the visual crop $v$ by applying random rotations between $-120^\circ$ to $120^\circ$. This exploits the fact that objects can have significant viewpoint variations in egocentric videos (unlike internet photos). Finally, we apply a random brightness augmentation to the video frames and the visual crop to simulate differing lighting.

\paragraph{Implementation details} We train the SiamHead $\mathcal{S}$ using the Detectron2 library~\cite{wu2019detectron2}. We use the default configuration file and make the following changes for our experiments. For each experiment, we use 8 GPUs, 64 visual crops per batch, and train for $300,000$ iterations with an initial learning rate of $0.02$ followed by a $0.1\times$ decay after $200,000$ iterations. We extract backbone features from the ``p3" layer of FPN. Based on validation performance, we use $6$ positives and $64$ negatives for each visual crop. Specifically, we sample $58$ negatives per video frame which results in $58 \times 64 = 3712$ negatives per batch. For each visual crop, we sample the $64$ hardest negatives out of $3712$.

In the SiamHead $\mathcal{S}$ architecture, the projection module $\mathcal{P}$ consists of four residual blocks followed by average pooling, and a 2-layer multi-layer perceptron (MLP) with a hidden size of $1024$-D and ReLU activation.

For signal peak detection, we utilize the \texttt{\small find\_peaks} function from the scipy library\footnote{Peak detection: {\scriptsize\url{https://docs.scipy.org/doc/scipy/reference/generated/scipy.signal.find_peaks.html}}} with the following hyperparameters selected through validation: distance = $25$, width = $3$, and prominence = $0.2$.

\paragraph{Experimental results} 

We evaluate the performance of multiple baselines on the VQ2D task in Tab.~\ref{tab:vq2d_results}. The first column in the table shows the detection and tracking methods, and the second column shows the SiamHead projection architecture $\mathcal{P}$. In addition to the KYS tracker, we also experiment with a simple particle filter tracker (denoted `PF') to assess the impact of the tracking quality. As an ablation of SiamRCNN, we replace the 4 residual blocks in the SiamHead projection module with a simple 3-layer CNN which has lower capacity with no residual connections (indicated by `Simple').

\begin{figure*}[ht]
    \centering
    \includegraphics[width=1.0\textwidth,trim={0 8.5cm 0.5cm 0},clip]{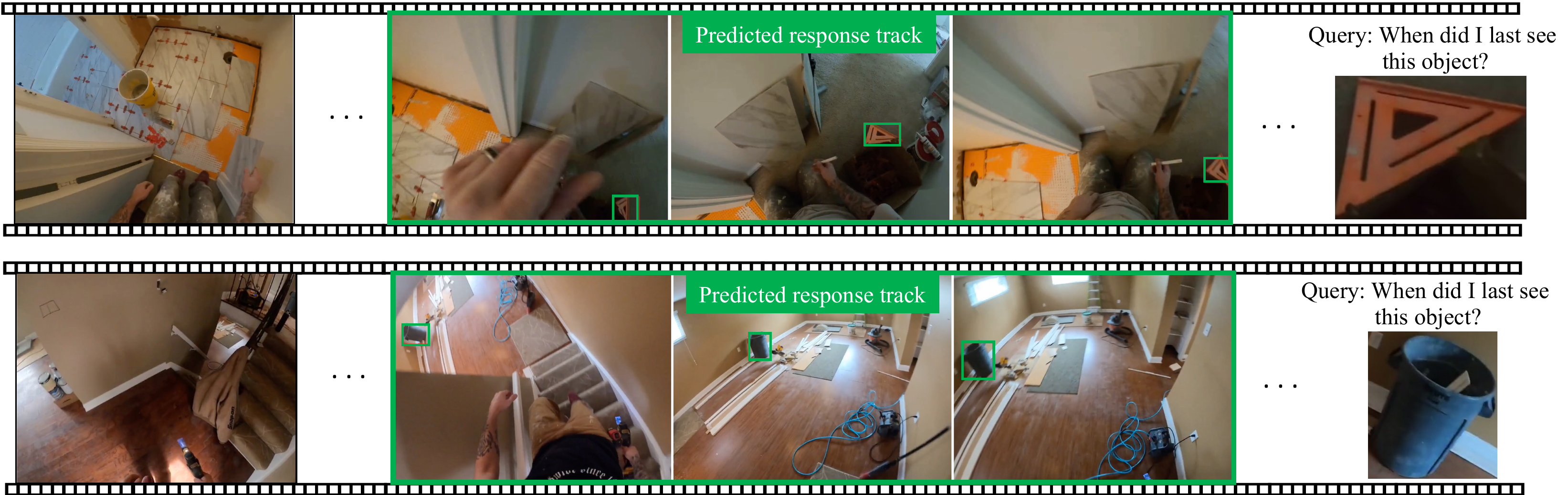}
    \caption{\small \SR{Qualitative examples for visual queries 2D localization. On each row, we show the visual crop of the query object on the right and the predicted response track in the center (3 uniformly samples images). The model was able to correctly localize the most recent occurrence of the object and accurately track it throughout the occurrence.}}
    \label{fig:vq2d_qual}
\end{figure*}

We make several observations. When we use a simple projection model with a particle filter tracker, we already observe a good \FEB{validation} performance of \FEB{$32.4\%$} success, and \FEB{$0.14$} tAP$_{25}$. These can be attributed to using a strong proposal generator (RPN pre-trained on MS-COCO) and a learned siamese comparison model. Upon replacing the particle filter tracker with a SoTA KYS tracker~\cite{bhat2020kys}, while the \FEB{validation} success rate remains similar at \FEB{$33.0\%$}, we observe significant gains (absolute) in all other metrics: \FEB{$2\%$} tAP, \FEB{$2\%$} stAP$_{25}$, and \FEB{$14.3\%$} recovery. This suggests that a good tracker is necessary to accurately capture the full response track after localizing a single frame within it. Finally, upon replacing the `Simple' siamese projection with 4 residual blocks, we observe a significant gains of \FEB{$6.8\%$} in success, \FEB{$5\%$} in \FEB{tAP$_{25}$, $4\%$ in stAP$_{25}$}, and \FEB{$5\%$} in recovery $\%$. This suggests that using a higher capacity model for the SiamHead is helpful for improving the per-frame detection performance for the VQ2D task. \FEB{We observe similar trends on the test set.} Please see Fig.~\ref{fig:vq2d_qual} for qualitative examples of the model's predictions.

\begin{figure}[t]
    \centering
    \includegraphics[width=0.5\textwidth,trim={0 3.5cm 0 0},clip]{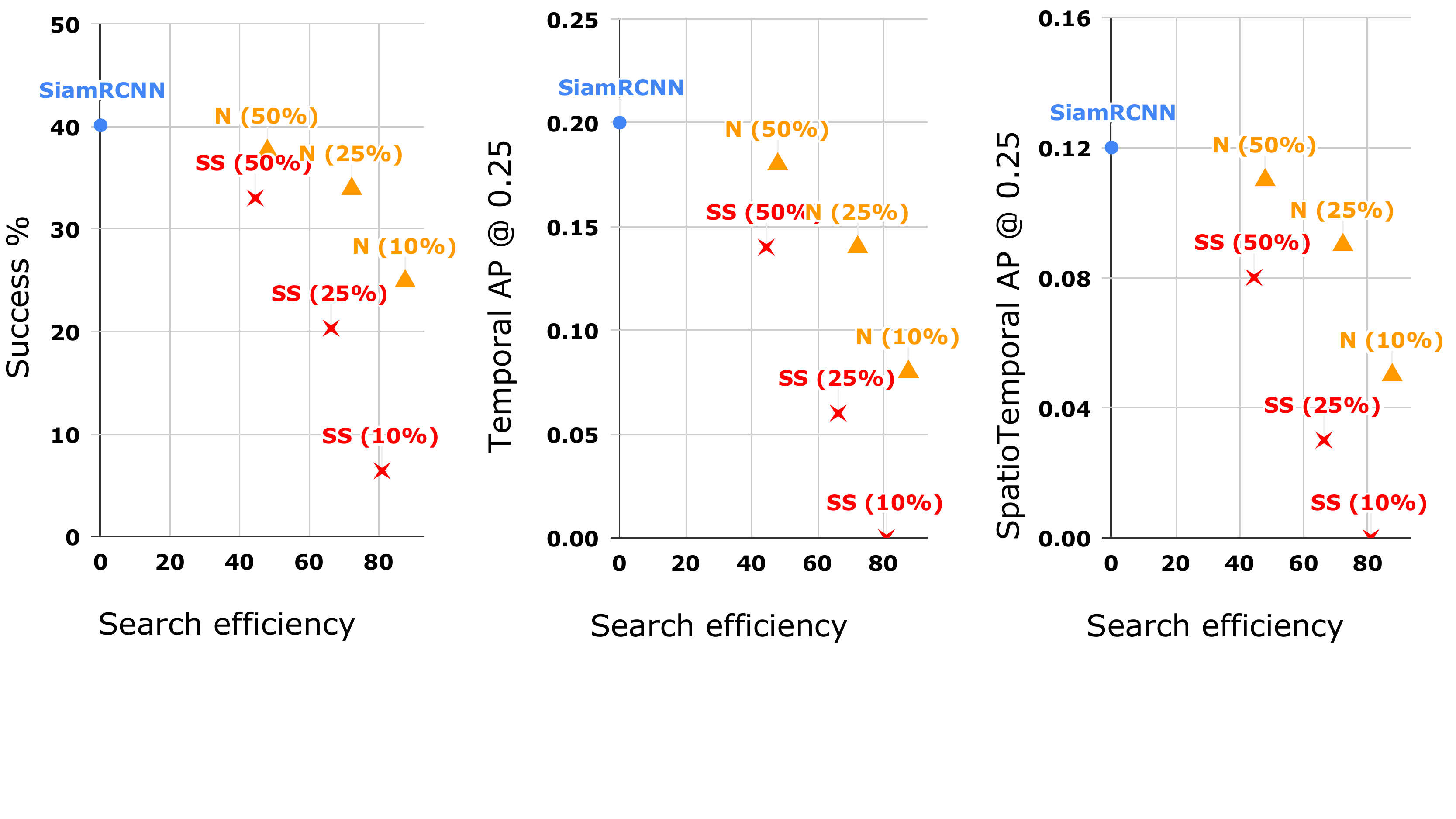}
    \caption{\small \textbf{Search efficiency for visual queries 2D localization.} We evaluate simple techniques for improving the search efficiency, and plot the corresponding VQ2D performance. The blue data point is the SiamRCNN performance when we preview the entire search window. The red data points are the SiamRCNN performance when we search over $k\%$ of the frames uniformly subsampled (SS) from the search window. The yellow data points are the SiamRCNN performance when we search over $k\%$ of the frames nearest (N) to the query (without any subsampling). The value of $k$ is indicated above each data point.}
    \label{fig:vq2d_ablation}
\end{figure}

In all cases from Tab.~\ref{tab:vq2d_results}, the search efficiency is $0\%$ since the detectors are used on every frame in the search window. In Fig.~\ref{fig:vq2d_ablation} we experiment with two simple techniques for improving the search efficiency. The first approach uniformly subsamples $k\%$ of the frames in the search window (denoted as `SS'). The second approach searches over only $k\%$ of the most recent frames in the search window, i.e., frames that are nearest to the query (denoted as `N'). We consider 3 values of $k$ in both cases: $10\%$, $25\%$, and $50\%$. Consider the results in Fig.~\ref{fig:vq2d_ablation}. In both strategies, the search efficiency improves as we reduce $k$. The performance drops drastically for the 1st strategy where we subsample the search window, while it remains relatively stable for the second strategy where we preview a fraction of frames closest to the query. For example, we can achieve a search efficiency of \FEB{$48.0\%$} with only a \FEB{$6 - 16\%$ relative} drop in performance with $k=50\%$ in the 2nd strategy. However, the performance drops significantly if we reduce $k$ further. For example, we observe a reduction of \FEB{$38-60\%$} for $k=10\%$ with the 2nd strategy. This suggests that more intelligent methods that perform contextual search are needed to improve the search efficiency for VQ2D while maintaining good performance.

\begin{table*}[t]
\centering
\resizebox{1.0\textwidth}{!}{%
\begin{tabular}{@{}lc|cccccc|cccccc@{}}
\toprule
                   &              &                       \multicolumn{6}{c|}{Validation set}                     &                   \multicolumn{6}{c}{Test set}                            \\
Detector + Tracker &$\mathcal{P}$ &         Succ   &         tAP    &   tAP$_{25}$    &       stAP    &   stAP$_{25}$  &       rec\%    &        Succ    &         tAP     &      tAP$_{25}$ &        stAP    &    stAP$_{25}$ &         rec\%    \\ \midrule
Siam-RCNN + PF     & Simple       &     \FEB{32.4} &     \FEB{0.06} &  \FEB{0.14}     &  \FEB{0.02}   &     \FEB{0.06} &    \FEB{13.2}  &    \FEB{32.7}  &    \FEB{0.06}   &   \FEB{0.14}    &    \FEB{0.02}  &     \FEB{0.06} &       \FEB{12.9} \\
Siam-RCNN + KYS    & Simple       &     \FEB{33.0} &     \FEB{0.08} &  \FEB{0.15}     &  \FEB{0.03}   &     \FEB{0.08} &    \FEB{27.2}  &    \FEB{33.4}  &    \FEB{0.09}   &   \FEB{0.16}    &    \FEB{0.03}  &     \FEB{0.08} &       \FEB{26.9} \\
Siam-RCNN + KYS    & Residual     & \bd{\FEB{39.8}}& \bd{\FEB{0.12}}& \bd{\FEB{0.20}} &\bd{\FEB{0.04}}& \bd{\FEB{0.12}}& \bd{\FEB{32.2}}& \bd{\FEB{41.6}}& \bd{\FEB{0.12}} &\bd{\FEB{0.21}}  & \bd{\FEB{0.05}}& \bd{\FEB{0.13}}& \bd{\FEB{34.0}}  \\ \bottomrule
\end{tabular}
}
\caption{\small \textbf{Visual queries 2D localization results.} We compare the performance of various baselines on the VQ2D validation and test datasets. Column 1 indicates the detector and tracker. Column 2 indicates the projection architecture used in case of the Siam-RCNN model.}
\label{tab:vq2d_results}
\end{table*}

\subsubsection*{Visual queries 3D localization baseline}

\begin{figure}[t]
    \centering
    \includegraphics[width=0.49\textwidth]{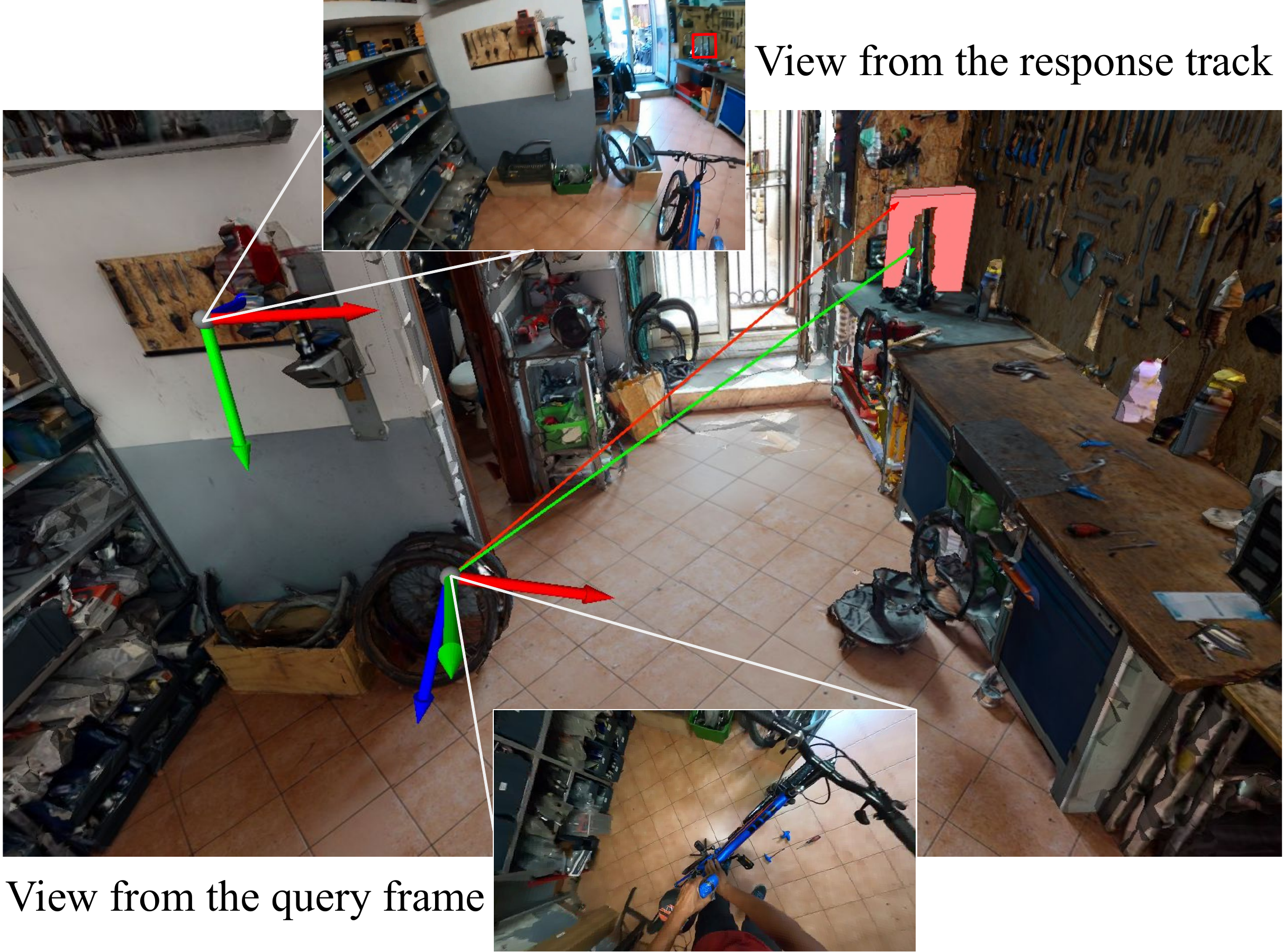}
    \caption{Visual queries 3D localization task demo. The top view is the view from the last frame of the response track with the target object annotated with a 2D red bounding box. The bottom view is the view from the query frame. The target object is annotated with a 3D red bounding box at the top right of the figure. The figure shows the ground-truth (green) and the predicted (red) 3D displacement vectors.}
    \label{fig:vq_3d_demo}
\end{figure}

Next we describe the baseline for the visual query with 3D localization task.
Recall the task definition: given a video, a query frame, and a visual crop of a target object, the goal is to output a 3D displacement vector from the camera center of the query frame to the center of the target object in 3D. The 3D position of the target object is defined at its most recent appearance in the video. Figure \ref{fig:vq_3d_demo} shows a sample of the task.

Our baseline strategy has three steps. We first estimate the camera poses of the video. Then we retrieve the most recent instance of the target object in the video. Lastly, we estimate the depth of the detected object and retrieve its 3D position from the query frame.

\paragraph{Camera pose estimation}
The camera poses are estimated using a keypoint matching strategy along with a Perspective-n-Point (PnP) resolution approach. At a high level our approach consists of the following four steps. First we estimate the camera intrinsic parameters using Structure-from-Motion (SfM). Secondly, we extract and match keypoints from each frame in the video to keypoints extracted from the Matterport3D panoramas. Then, using the matched keypoints we set up and solve a PnP problem for each frame in the video to estimate the corresponding camera pose. Lastly, we refine the poses using temporal constraints.

\textbf{Step 1: Camera intrinsics estimation}
We start by extracting a set of contiguous non-blurry frames from the video. In order to select non-blurry frames we compute the variance of the Laplacian on each image and select the ones with a value higher than a \FEB{100} threshold. We then select the largest contiguous set of non-blurry images. We cap the number of selected frames to \FEB{10} to limit the computational time of the SfM module. Once we have selected the images we run the automatic reconstruction module of COLMAP \cite{schonberger2016structure} to estimate the camera instrinsic parameters with a radial fisheye camera model.

\textbf{Step 2: Keypoint extraction and matching}
We use SuperGlue \cite{sarlin2020superglue} to extract and match keypoints. 
We first extract keypoints from the scan panoramas $\{k_{\{p,n\}}, p \in \mathcal{P}, n \in \mathcal{N}\}$ where $\mathcal{P}$ is the number of panoramas and $\mathcal{N}$ is the number of keypoints. The scan panoramas are generated using the Matterport SDK.\footnote{Matterport-SDK: {\scriptsize\url{https://matterport.github.io/showcase-sdk/sdk_intersection_inspector.html}}} We render RGB and depth images at each scan position and sweep over pitch values $ \in [-30,30]$ with a step size of 5 deg. and yaw values $\in [-180, 180]$ with a step size of 15 deg. We generate on average 7K images per scan. Note that while we are not releasing the panoramas because of data anonymization concerns, we are providing the precomputed keypoints.
Similarily, we extract keypoints from the video frames $\{k_{\{i,m\}}, i \in \mathcal{I}, m \in \mathcal{M}\}$ where $\mathcal{I}$ is the number of images in the video and $\mathcal{M}$ is the number of keypoints.
Once the keypoints are extracted we loop through each frame $i \in \mathcal{I}$ in the video and match the extracted frame keypoints $\{k_{\{i,m\}}, m \in \mathcal{M}\}$ to all the panoramas keypoints $\{k_{\{p,n\}}, p \in \mathcal{P}, n \in \mathcal{N}\}$.
We use the pretained models available\footnote{SuperGlue weights: {\scriptsize\url{https://github.com/magicleap/SuperGluePretrainedNetwork}}} of SuperPoint \cite{detone2018superpoint} for keypoints and descriptors extraction and SuperGlue \cite{sarlin2020superglue} for matching.

\textbf{Step 3: PnP resolution}
 We compute the camera pose for the video frames having at least 20 matched keypoints. We empirically find that a threshold of 20 provides a good tradeoff between the number of overall pose estimates and the quality of the estimations. The positions of the 3D keypoints are computed from a pinhole camera model of the Matterport camera using the rendered panorama depth, camera intrinsics, and camera pose. The positions of the 2D keypoints are directly extracted from the video frames pixels.
 We then use the OpenCV library to solve the PnP setup and estimate the camera pose from the matched pairs of 3D and 2D points and using the estimated camera intrinsic parameters.
Using this method we can estimate the camera pose of roughly $2\%$ of the total number of frames in the video. Next we incorporate temporal constraints to increase this number.

\textbf{Step 4: Temporal constraints and final pose estimation}
To increase the number of estimates we refine the pose estimation pipeline by incorporating temporal constraints in an iterative procedure. We start by extracting and matching 2D keypoints from localized frames to non-localized ones in the video. This step is similar to the above Step 2; we use the same SuperGlue \cite{sarlin2020superglue}. Using the matched keypoints and current estimated poses we triangulate new 3D keypoints for the non-localized images. 
We then solve a new PnP setup with the new keypoints. We apply this procedure iteratively until convergence.
After refinement we achieve a performance of $15\%$ of pose estimates of the total number of frames accross all video clips.

\begin{figure}[t]
    \centering
    \includegraphics[width=0.5\textwidth]{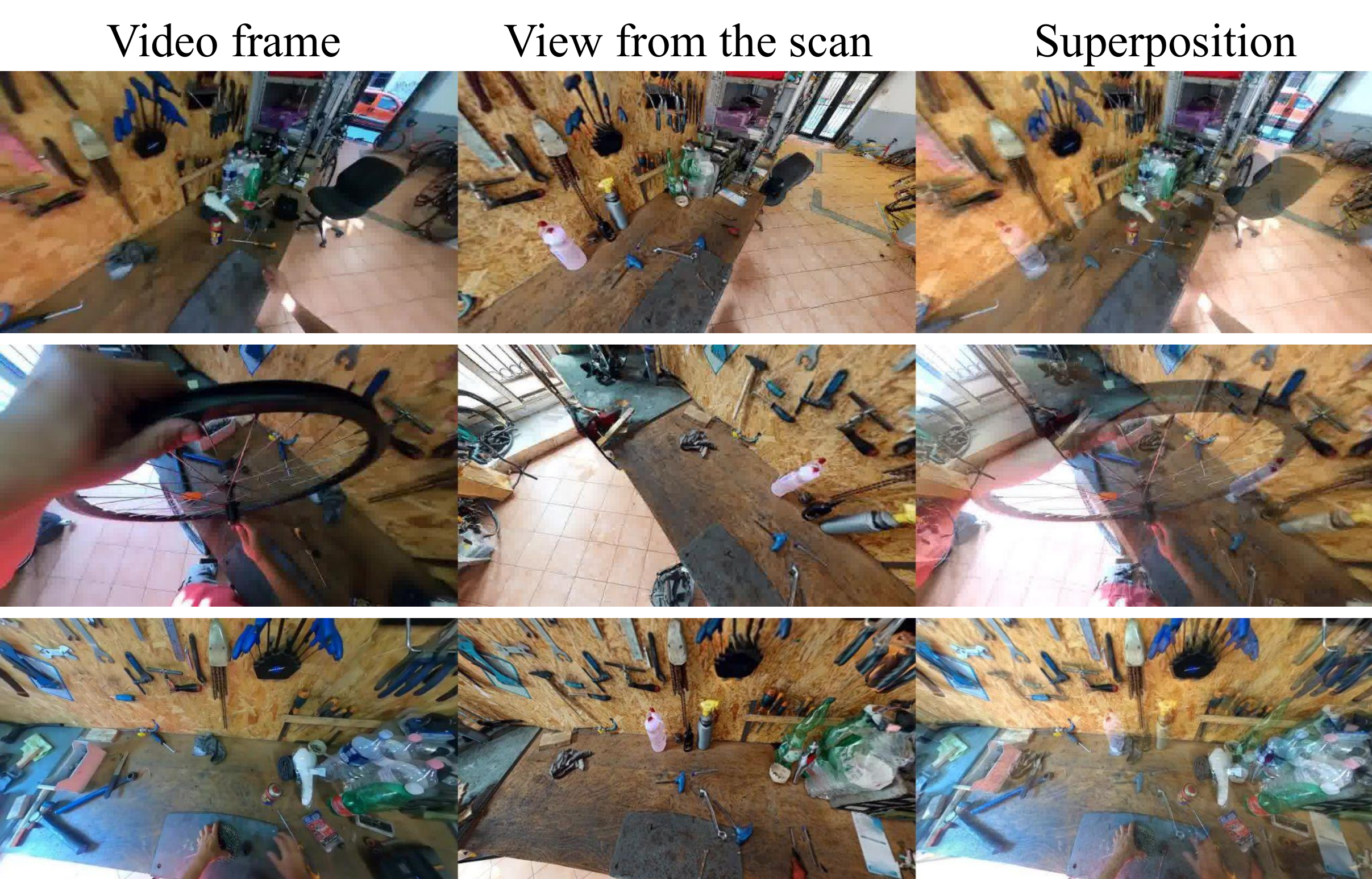}
    \caption{\textbf{Samples of camera pose estimation}. Left shows the frame from the egocentric video, middle has the view rendered from the estimated viewpoint in the scan and right is the superposition of both. We observe that even with big scene differences between the video and the scan (e.g., the wheel in the second row), the algorithm is able to accurately retrieve  the camera pose.}
    \label{fig:vq_3d_camera_pose}
\end{figure}

\paragraph{Camera pose estimation quality and sources of error}
We qualitatively evaluate the camera pose estimation pipeline by rendering the views in the 3D scans. Recall that the scans and videos have been recorded at different times and thus the scenes can contain large differences. Figure \ref{fig:vq_3d_camera_pose} shows camera poses estimates where left is the frame from the video, middle is the view from the scan, and right is the superposition. We see that even with large scene differences between the scan and video (e.g., the wheel in the middle example) the algorithm is capable of producing good pose estimates.

 The remaining unlocalized frames are due to abrupt motion (lost track) and when the view is too close-up to the scene (not enough keypoints matched). 

\paragraph{Target object retrieval}
We build our solution on top of the visual queries 2D localization baseline. The 2D localization baseline outputs a response track with 2D detections of the target object. Our baseline combines these 2D detections along with depth estimation and camera pose estimation to retrieve the 3D position of the object.

\begin{figure}[t]
    \centering
    \includegraphics[width=0.5\textwidth]{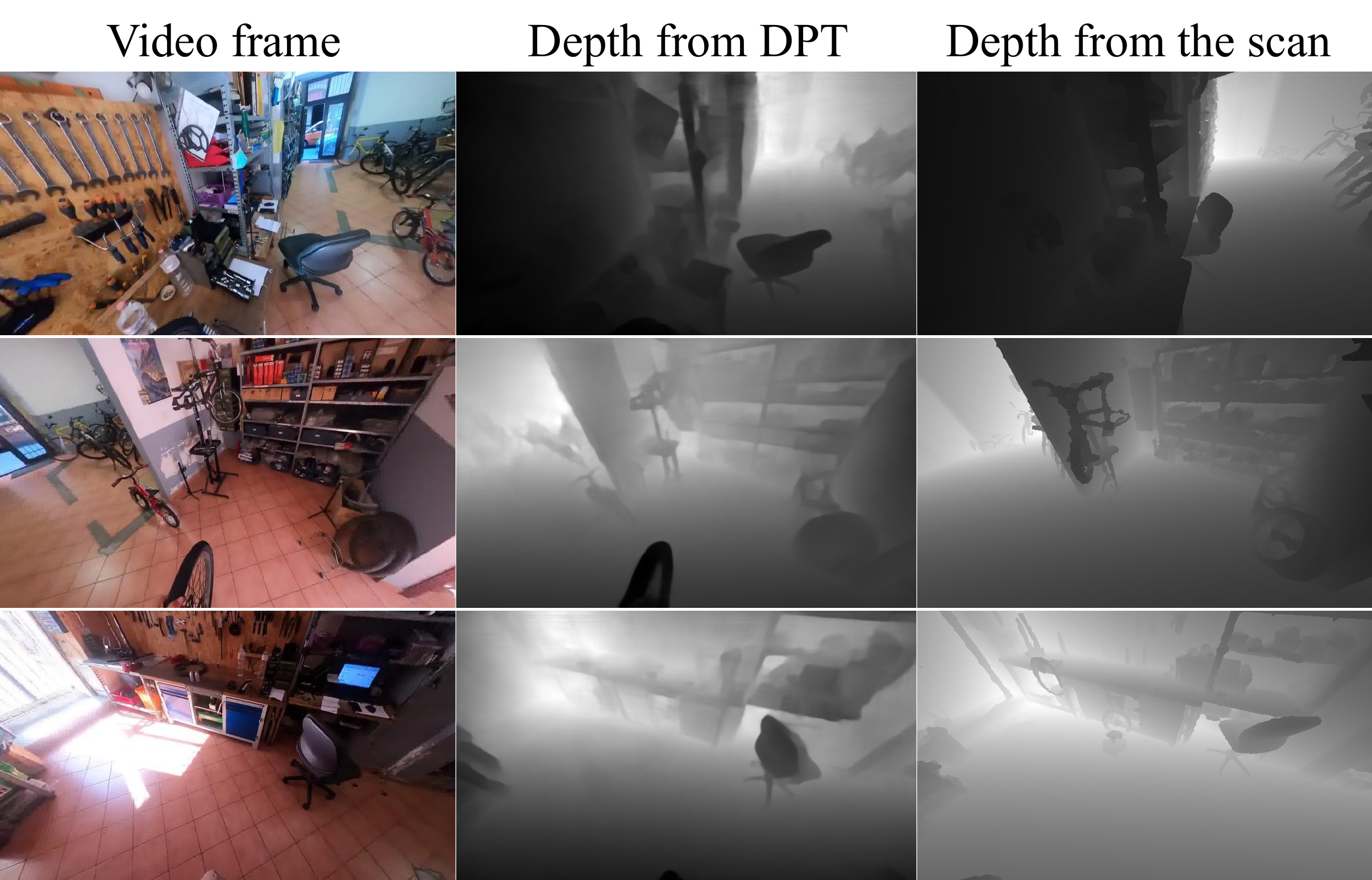}
    \caption{\textbf{Samples of depth estimation}. Left shows the frame from the egocentric video, middle has the estimated depth from DPT \cite{ranftl2021vision} and right has the depth from the scan rendered at the estimated viewpoint.}
    \label{fig:vq_3d_depth}
\end{figure}

\paragraph{Depth estimation}
We estimate the depth of the most recent frame of the response track for which we have a pose estimate. We use the DPT network \cite{ranftl2021vision} with pretrained weights on NYU\_v2 \cite{silberman2012indoor}. Figure \ref{fig:vq_3d_depth} shows depth estimation results where left is the frame from the video, middle is the estimated depth, and right is the depth from the scan rendered at the estimated viewpoint (not available to the baseline model). Note that due to scene differences between the video and the scan, the two depths frames will differ in some region of the image.
We then compute the depth value of the target centroid as the median of a square region centered at the 2D detection. 

\paragraph{3D displacement vector reconstruction}
Given the estimated depth $d$ of the object centroid $c$ in frame $f$ of the response track and the estimated camera instrisics $K$, we construct the 3D vector displacement $\hat{v}_f$ in the current frame $f$ coordinate system using a pinhole camera model:

\begin{gather}
\hat{v}_f =
\begin{bmatrix}
x\\
y\\
z
\end{bmatrix} 
=d K^{-1} c=
d K^{-1}
\begin{bmatrix}
u\\
v\\
1
\end{bmatrix}
\end{gather}
where $u,v$ are the pixel indices of the centroid $c$ in frame $f$. 
We then estimate the object centroid position $\hat{t}_s$ in the scan coordinate system:
\begin{gather}
    \hat{t}_s = {P^s_f} \hat{v}_f
\end{gather}
where $P^s_f$ is the camera pose for the frame $f$.
We further retrieve the displacement vector $\hat{v}_Q$ in the query frame $Q$ coordinate system:
\begin{gather}
    \hat{v}_Q = {P^s_Q}^{-1} \hat{t}_s
\end{gather}
where $P^s_Q$ is the camera pose of the query frame.

\paragraph{Experiments and results}
We compare the performance of multiple baselines along with ablation studies. We present the results in Table \ref{tab:vq3d_results}. Numbers are computed on the validation set (\FEB{164} queries) of the VQ3D task. We report the query ratio QwP, for which we have camera pose estimates for the response track and query frame. Additionally, we report the success rate Suc$c^{*}$ which is the success metric computed only for queries with associated pose estimates. 

Overall, we notice a low QwP ratio leading to a low success rate. These low metrics are due to a small number of camera pose estimates (15\% overall). Nonetheless, we observe that the best VQ2D baseline method combined with the pretrained DPT \cite{ranftl2021vision} depth estimator yields the best performances in terms of $L2$ and success. These numbers tell that there are opportunities for enhancement in designing better camera pose estimators. Additionally, we perform ablation studies using the ground-truth response tracks and different depth estimators (random, from the scan, using DPT). For the random experiment we uniformly sample a depth value between 0.1 and 10 meters. From the ablation experiments we note that rendering the depth from the scan at the estimated viewpoint increases the performances compared to using DPT (lines 2 and 3). This suggests that there is also room for improvement in designing better depth estimators.

\begin{table}[t]
\centering
\resizebox{0.5\textwidth}{!}{
\begin{tabular}{@{}cccccccccc@{}}
\toprule
RT           & depth  &   L2   &   angle &  Suc$c^{*}$\% & Succ\%   & QwP\% \\ \midrule
ground-truth & random &  \FEB{7.93}   &  \FEB{1.99} &  \FEB{0.00} & \FEB{0.00} & \FEB{1.83} \\
ground-truth &  scan  &  \FEB{2.92}   &  \FEB{1.10} & \FEB{76.47} & \FEB{1.22} & \FEB{1.83} \\
ground-truth  & DPT   &  \FEB{3.33}   &  \FEB{1.15} &  \FEB{76.47} & \FEB{1.22} & \FEB{1.83}   \\ \midrule
Siam-RCNN + PF         & DPT & \FEB{6.53}           & \FEB{1.64} & \FEB{25.00}          & \FEB{0.61} & \FEB{\textbf{0.61}} \\
Siam-RCNN + KYS (sim.) & DPT & \FEB{\textbf{5.78}}  & \FEB{\textbf{0.48}}          & \FEB{\textbf{36.36}} & \FEB{0.61} & \FEB{0.61} \\
Siam-RCNN + KYS (res.) & DPT & \FEB{5.98}           & \FEB{1.60}          & \FEB{30.77}          & \FEB{\textbf{1.22}} & \FEB{\textbf{1.83}} \\ \bottomrule
\end{tabular}
}
\caption{\textbf{Visual queries 3D localization results.} We compare the performance of various baselines on the val set of the VQ3D task. Column 1 indicates the VQ2D network used to predict the response track (RT). The last metric QwP measures the query ratio for which we have pose estimation for the response track and the query frame. The $L2$ metric is expressed in meters and angles are in radians. The first three rows are ablation studies using the ground-truth response tracks and with depth estimated randomly, using the scan and via the DPT \cite{ranftl2021vision} network.}
\label{tab:vq3d_results}
\end{table}

\begin{figure*}[t]
\begin{center}
\footnotesize
\includegraphics[width=0.9\textwidth]{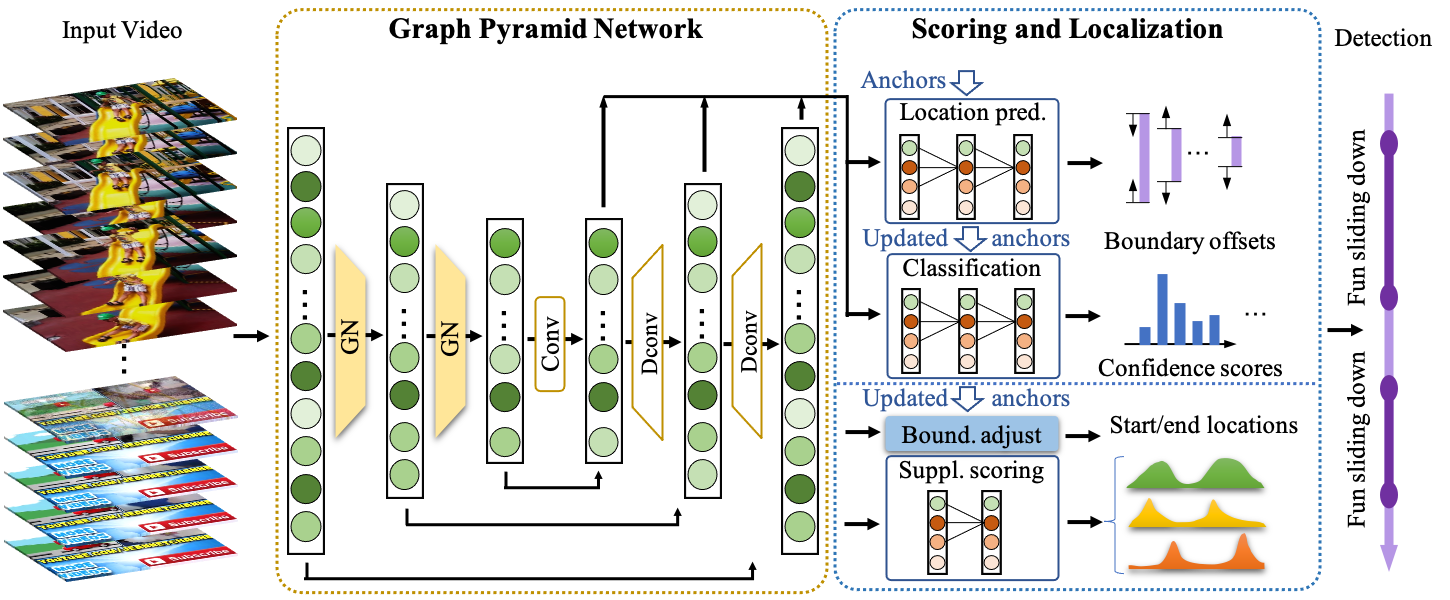} %

\end{center}
\caption{\small{\textbf{Baseline model architectures: {moment queries}.}  Its takes a video sequence and generates detected actions with start/end time, their categories, and confidence scores. It has two components: \textbf{graph pyramid network (GPN)}, and \textbf{scoring and localization (SoL)}. \textbf{GPN} is composed of multi-level encoder and decoder pyramids. The encoder aggregates features in different levels via a stack of  graph networks (GN) (yellow trapezoid area; the decoder restores the temporal resolution and generates  multi-level features for detection.  \textbf{SoL}  (blue dashed box) contains four modules, the top two predicting action scores and boundaries, the  bottom two producing supplementary  scores and adjusting boundaries. Figure is adapted from \cite{zhao2020video}.}}
\label{fig:moment_arch}
\end{figure*}
\subsubsection*{Natural language query baselines}
Since the natural language queries can be seen as a language-grounding problem in a video,
we adopt two prior methods in order to implement the baselines for this task.

\textbf{(a) 2D Temporal Adjacent Networks (2D-TAN)}\cite{2DTAN_2020_AAAI}:
We apply 2D-TAN  with a sliding window method to implement the natural language query baseline. %
The goal of 2D-TAN is to answer where the semantically corresponding video moment is, given a language query in an untrimmed video. The language query stems from one of the 13 template questions. %
The core idea of 2D-TAN is to consider adjacent moment candidates as the temporal context on a two-dimensional temporal map and retrieve the most relevant moment from the candidates. More concretely, 2D-TAN takes each moment candidate as one element in the 2D temporal map such that the adjacent moment candidates on the map can have much-overlapped content or share the same start or end time slot. It applies a convolutional neural network on the 2D map to predict the Intersection over Union of each moment candidate and the ground-truth moment. Please see \cite{2DTAN_2020_AAAI} for more details. 

Since the 2D-TAN enumerates all the possible combinations of start-end pairs, the $O(N^2)$ space complexity of the 2D map leads to a heavy model, especially when we require a precise moment boundary. To make 2D-TAN more appropriate to our problem, we further use a sliding window method on top of 2D-TAN. We break down the clip into a number of overlapping windows, where a window presents a small portion of the clip. The windows are taken as the input of the 2D-TAN model in both training and testing phases.

During the training of the 2D-TAN model, we use Ego4D's provided pre-extracted features for both the video clip and language query. The clip feature is from a SlowFast \cite{feichtenhofer2019slowfast} network pretrained on Kinetics 400 dataset, and the language feature is a based on the BERT model \cite{devlin-etal-2019-bert}. The window duration is 40s, and stride is 20s in the sliding window method. Notably, we only use windows that contain or are next to a ground-truth moment in training, but we use all the windows in testing. We keep all the other hyper-parameters in 2D-TAN the same as its default except for tIoU threshold and learning rate. We decreased the tIoU threshold from 0.5 to 0.3 to enable more positive samples during training and empirically set the learning rate to 0.001. 
We train the model for 100 epochs and report the test set performance on the best checkpoint on the validation set.
2D-TAN gives top-1 and top-5 recalls of \FEB{$5.80\%$} and \FEB{$13.90\%$} at IoU=0.3, respectively.
In addition, we also ablate the model to obtain performance by randomizing the video features 
($-$visual) and textual features ($-$text) for NLQ in Tab.~\ref{apptab:nlq_results}.

\begin{table}[t]
    \centering
    \begin{tabular}{c c c c c c}
        \toprule
        & \multirow{2}{*}{Baseline} & \multicolumn{2}{c}{IoU=$0.3$ (\%)} & \multicolumn{2}{c}{IoU=$0.5$ (\%)}\\
        & & r@1 & r@5 & r@1 & r@5\\
        \midrule
        \multirow{2}{*}{\rotatebox[origin=c]{90}{Val} $\begin{dcases} \\  \end{dcases}$} &
         2D-TAN \cite{2DTAN_2020_AAAI} & \FEB{5.04} & \FEB{\textbf{12.89}} & \FEB{2.02} & \FEB{5.88}\\
         & VSLNet \cite{zhang2020span} & \FEB{\textbf{5.45}} & \FEB{10.74} & \FEB{\textbf{3.12}} & \FEB{\textbf{6.63}}\\
         \midrule
         \multirow{6}{*}{\rotatebox[origin=c]{90}{Test} $\begin{dcases} \\ \\ \\ \\ \\ \end{dcases}$}
         & 2D-TAN \cite{2DTAN_2020_AAAI} & \FEB{\textbf{5.80}} & \FEB{\textbf{13.90}} & \FEB{2.34}& \FEB{5.96}\\
         & $\quad-$visual & \FEB{2.29} & \FEB{6.77} & \FEB{1.32} & \FEB{3.46}\\
         & $\quad-$text   & \FEB{3.46} & \FEB{10.13} & \FEB{1.78} & \FEB{4.38}\\
         & VSLNet \cite{zhang2020span} & \FEB{5.47} & \FEB{11.21} & \FEB{\textbf{2.80}} & \FEB{\textbf{6.57}}\\
         & $\quad-$visual & \FEB{1.80} & \FEB{5.44} & \FEB{0.90} & \FEB{2.45}\\
         & $\quad-$text  & \FEB{3.05} & \FEB{7.39} & \FEB{1.45} & \FEB{4.12}\\
         \bottomrule
    \end{tabular}
    \caption{Performance of the NLQ baselines on val and test splits.}
    \label{apptab:nlq_results}
\end{table}

\textbf{(b) Span-based Localization Network (VSLNet)}\cite{zhang2020span}:
Unlike traditional approaches in video natural language localization works, VSLNet treats the
input untrimmed video as a text passage, and uses a span-based approach to identify the relevant
sections semantically related to the given natural language query.
At its core, VSLNet first encodes the natural language query and video features using a common,
shared Transformer \cite{van2017transformation} network.
Next, it uses the encoded query to then attend to the relevant parts of the video clip
(akin to a text paragraph).
The attended sections are further refined using a query-guided highlighting (QGH) strategy by
extending the selection foreground of the video by a hyperparamter to capture more visual 
context. 
Please refer to \cite{zhang2020span} for more details on the motivation and architecture.

For our experiments, we maintain consistency with the other NLQ baselines and use pre-extracted
features for both the video clip (SlowFast network \cite{slowfast}) and natural language query 
(BERT \cite{devlin-etal-2019-bert}).
We use the implementation provided by the authors\footnote{\url{https://github.com/IsaacChanghau/VSLNet}} with the following changes:
(a) Set the video features size to $2304$ dimensions to accommodate the features extracted from
the SlowFast network,
(b) Replace the text encoder to a frozen, pretrained BERT \cite{devlin-etal-2019-bert} model,
(c) Set the internal dimension of the multimodal network to $128$, and project the pre-trained 
BERT features from $768$ to $128$.
We train the model for $200$ epochs and pick the model with the best performance on val split.
The corresponding test performance of this VSLNet model is reported in 
Tab. \ref{apptab:nlq_results}, along with visual and textual ablations.

\subsubsection*{Moments queries baseline}

We formulate a moment queries baseline as a temporal action detection method~\cite{lin2018bsn,xu2020g,zhao2020video}, plus simple post-processing.

The MQ task only expects predictions for the query categories, whereas the temporal action detection task returns the predictions for all categories. Therefore, we can first use a temporal action detection method to predict for all categories, and only output the results corresponding to the query categories.

To predict all categories, we adopt a recent method  VSGN~\cite{zhao2020video}, which was designed for temporal action detection in third-person videos. We use VSGN without the VSS component. Figure~\ref{fig:moment_arch} illustrates the architecture. It takes a video $\mathcal{V}$ as input, extracts features for each snippet in the video using a network such as SlowFast~\cite{slowfast}, and feeds these features into a graph pyramid network. The graph pyramid network contains a encoder and a decoder, where the encoder is comprised of multiple levels of graph convolutional networks, and the decoder is comprised of multiple levels of de-convolutional networks. It is an anchor-based method that pre-defines temporal segments for each feature level as prediction reference. It predicts the scores and refines the locations of the anchors in two stages. In the first stage, it uses a region proposal network (RPN) from the decoder to  predict  class labels and regress boundaries for each anchor; in the second stage, it applies a boundary adjustment module to refine the boundary offsets based on the updated anchors from the first stage.  It also has startness/endness predictions to provide auxiliary supervision and supplement scores for each predicted segment.  Its output predictions are formulated as $\Phi=  \left \{ \phi_m=\left ({t}_{m, s},{t}_{m, e},  {c}_m, {s}_m \right ) \right \}_{m=1}^{M}$,  where $m$ is the number of predictions, ${t}_{m, s}$ and ${t}_{m, e}$ are  start time and end time of the $m^{th}$ prediction respectively, ${c}_m$ is the predicted category, and ${s}_m$ is the confidence score. For more details, please refer to \cite{zhao2020video}.

Given a query category $c$, the retrieval results for the moment queries task are obtained as follows

\begin{equation}
    \Phi_c=  \left \{ \phi_m=\left ({t}_{m, s},{t}_{m, e},  {c}_m, {s}_m \right )  | c_m = c, 1 \leq m \leq M) \right \}.
\end{equation}

\paragraph{Implementation details}
For feature extraction, we use Ego4D's provided pre-extracted features using a SlowFast~\cite{slowfast} network pre-trained on Kinects400~\cite{kinetics} at 1.87 features per second. The feature dimension is 2304. 

Considering that the maximum clip length is 8 minutes, which has 897 features,  we make the input length of our network 928 frames to cover the longest video clip. We have 5 levels in the graph pyramid network, each with temporal length 232, 116, 58, 29, and 14 respectively. We pre-define two base anchors of sizes 4 and 12 for Level 1 and increase the sizes by 2 for each deeper layer. We train for 30 epochs with a batch size 32 and learning rate 0.0001. In inference, we only apply per-category NMS with a confidence threshold 0.0005.

\paragraph{Experiments and results}

We show our baseline performance in terms of mAP in Table~\ref{Tab:moment_episodic_mAP} and recall @ ${kx}$, tIoU=${m}$ in Table~\ref{Tab:moment_episodic_recall}.

\begin{figure}[h]
\centering
\includegraphics[width=0.49\textwidth]{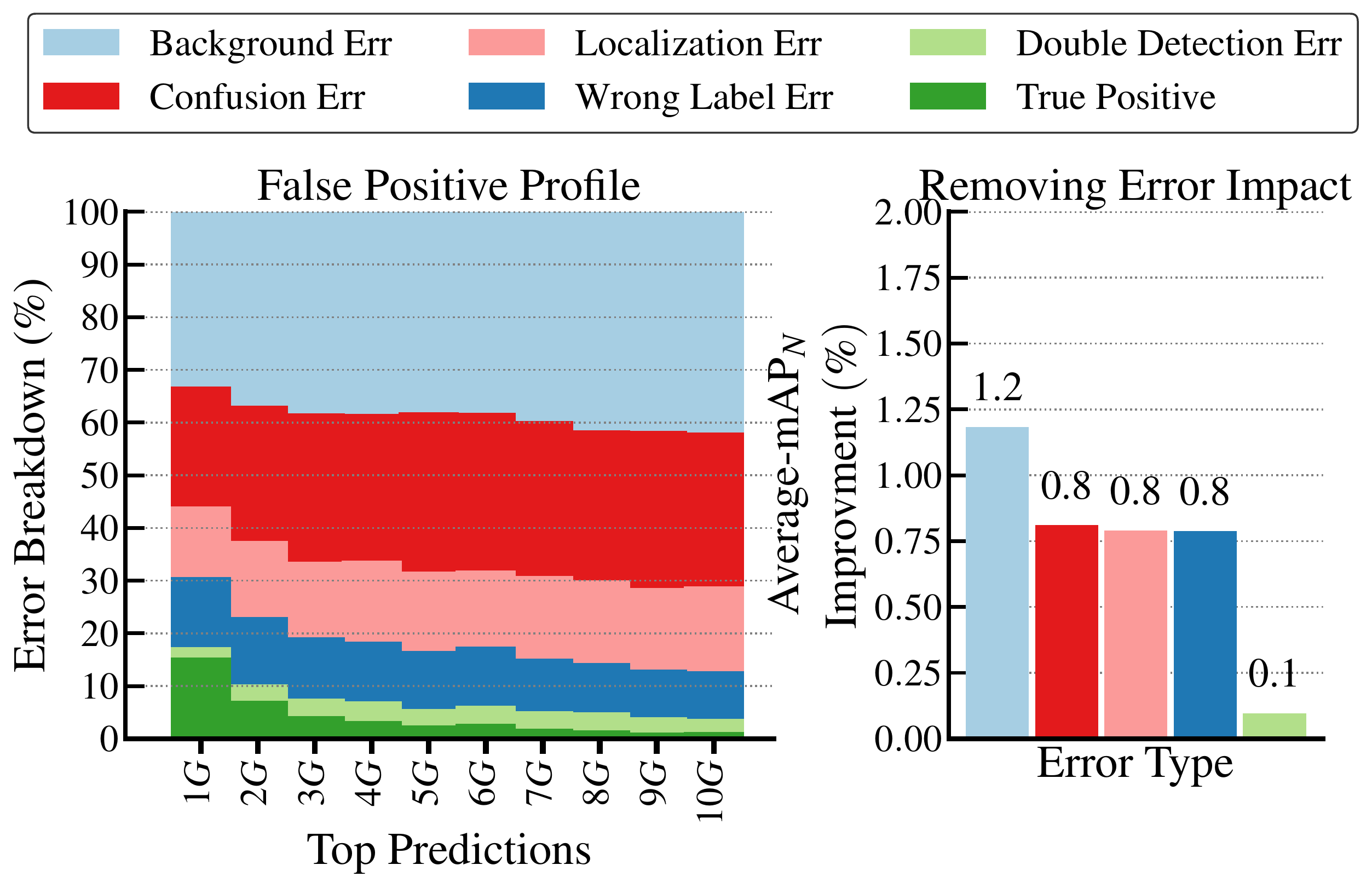}
\caption{\textbf{Moment queries results: false positive analysis}. The error types are determined by the tIoU between ground-truth and predicted moments, as well as the correctness of the predicted labels, according to \cite{detad}. Background error: $\textrm{tIoU} <1e^{-5}$; confusion error: $1e^{-5}<\textrm{tIoU}<\alpha$, label is wrong; wrong label error: $\textrm{tIoU}>=\alpha$, label is wrong; localization error: $1e^{-5}<\textrm{tIoU}<\alpha$, label is correct, where $\alpha$ refers to the tIoU thresholds \{0.1, 0.2, 0.3, 0.4, 0.5\}. `G' refers to the number of ground-truth instances. }
\label{appfig:moment_fp_analysis}
\vspace{-2mm}
\end{figure}

\begin{figure}[h]
\centering
\includegraphics[width=0.49\textwidth]{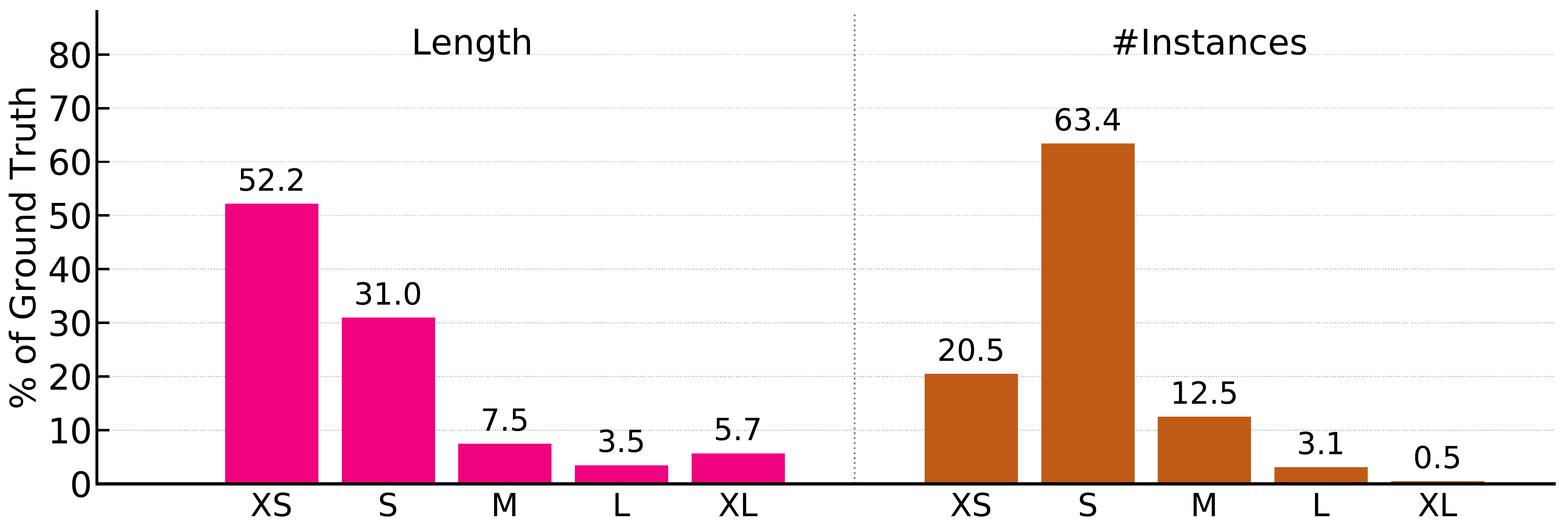}
\includegraphics[width=0.49\textwidth]{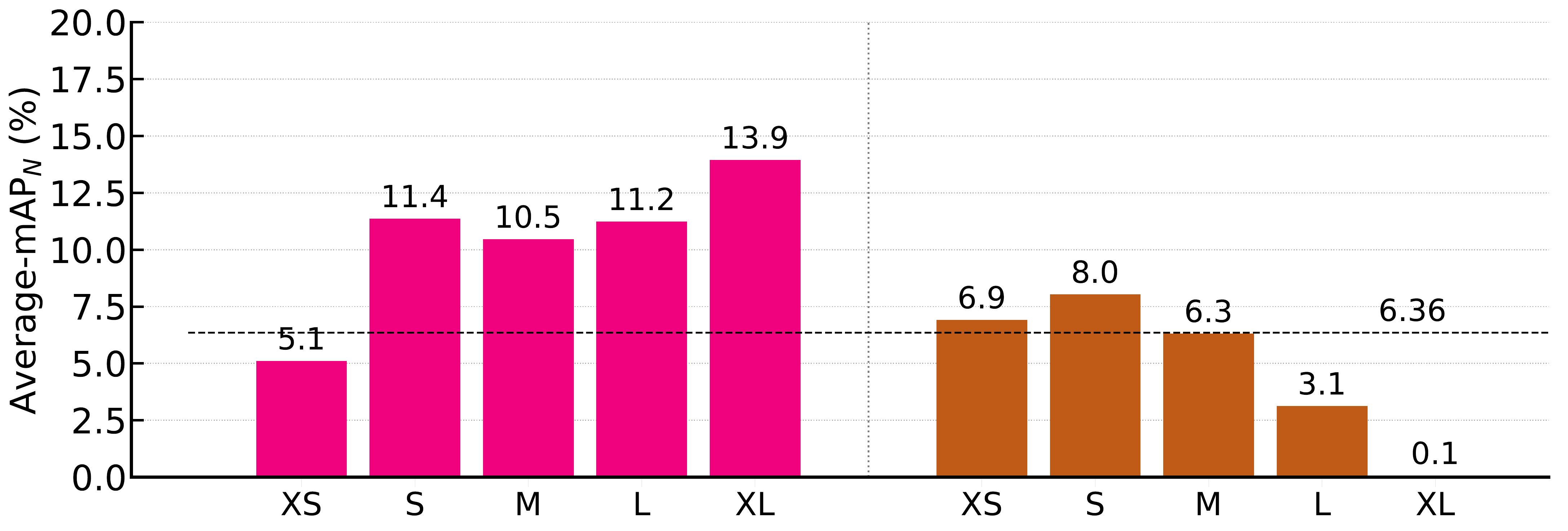}
\caption{\textbf{Moment queries results: sensitivity analysis.} Top: Distribution of instance per action characteristic: length; \# instances.  Bottom: average mAP$_{N}$ (\%)~\cite{detad} in each characteristic bucket. The `length' characteristic divides all moment instances 5 buckets based on the moments duration in seconds: XS (0, 10], S (10, 60], M (60, 180], L (180, 300], and XL (300, inf].  The `\# instances' characteristic divides all moment instances into 5 buckets based on the  number of instances belonging to the same category in one video clip: XS (0, 1], S (1, 3], M (3, 10], L (10, 20], and XL (20, inf]. }
\label{appfig:moment_sensitivity_analysis}
\vspace{-2mm}
\end{figure}

We provide further analysis on the average precision results using DETAD~\cite{detad}. In Fig~\ref{appfig:moment_fp_analysis}, we illustrate the proportion of each error type for the false positive predictions. It shows that both localization and classification are responsible for the false positive, improving either can increase the overall performance by a nontrivial amount. In Fig~\ref{appfig:moment_sensitivity_analysis}, we demonstrate the performance of different groups of moment instances based on moment duration and number of instances belonging to the same category per video clip. We notice that short moments tend to have low performance even though they are large in number. When there are 2-3 instances in one video, they are easiest to detect.

\begin{table}[t]
\centering
\caption{\textbf{Moment queries results} on the validation set and the test set, measured by mAP (\%) at different tIoU thresholds.  
}
\small
\setlength{\tabcolsep}{2.6 pt}
\begin{tabular}{l|ccccc|c}
\toprule
tIoU threshold     & 0.1 & 0.2 &  0.3 &  0.4 &  0.5 & Average \\ 
\hline

 {Validation set}  & \FEB{9.10}  & \FEB{7.16}  &\FEB{5.76}  &\FEB{4.62}   &\FEB{3.41}  & \FEB{6.03} \\
  {Test set}  &  \FEB{8.61} &  \FEB{6.52} & \FEB{5.43} & \FEB{4.30}  & \FEB{3.57} & \FEB{5.68}\\
\bottomrule
\end{tabular}
\label{Tab:moment_episodic_mAP}
\end{table}

\begin{table*}[t]
\centering
\caption{\textbf{Moment queries results on the validation set and the test set, measured by recall (R) @ $\textbf{kx}$, tIoU=$\textbf{m}$ (\%).}  
}
\small
\setlength{\tabcolsep}{2.5 pt}
\begin{tabular}{l | ccc|ccc|ccc}
\toprule
$m$  & \multicolumn{3}{c|}{$0.3$}  & \multicolumn{3}{c|}{$0.5$} & \multicolumn{3}{c}{$0.7$} \\
\hline
$k$ & $1$ & $3$ &  $5$ &  $1$ & $3$ &  $5$  & $1$ & $3$ &  $5$  \\ 
\hline

Validation Set &  \FEB{33.45} &  \FEB{51.26} & \FEB{58.43} & \FEB{25.16}  & \FEB{39.46} & \FEB{46.18} & \FEB{15.36} &  \FEB{22.67} & \FEB{25.81} \\
Test Set &  \FEB{33.56} &  \FEB{52.23} & \FEB{59.79} & \FEB{24.25}  & \FEB{39.22} & \FEB{46.22} & \FEB{14.83} & \FEB{23.15} & \FEB{26.28}  \\
\bottomrule
\end{tabular}
\label{Tab:moment_episodic_recall}
\end{table*}

\subsubsection*{Discussion}

Visual queries presents a novel and challenging task for object localization in egocentric videos. While our proposed baseline achieves a reasonable success rate of $42.9\%$, it only achieves a localization performance of $0.13$ tAP and $0.06$ stAP. Furthermore, the best performance is achieved with $0\%$ search efficiency, and na\"ive techniques to improve the search efficiency lead to drastic performance reductions. We hope that this task will spur future research into accurate and efficient techniques for object search. 

Natural language queries is a challenging multimodal task that has wide
applications in helping users search and retrieve relevant pieces of their
episodic memory, thanks to the flexibility of the queries.
The performance of the existing state-of-the-art video localization models 
highlights the needle-in-a-haystack nature of the task, due to shorter response 
windows of about $10s$ in a large video clip of $8$ minutes.
We hope that the NLQ dataset opens the door to future research that specializes
in identifying and retrieving a large diversity of language queries in longer 
egocentric video clips, moving a step closer to augmenting a user's episodic memory.

Moment queries in egocentric videos is a challenging task due to  the long-tailed distribution of categories and the large variation in moment duration. Our baseline achieves a reasonable result  according to the metric recall @$kx$, tIoU=$m$, which evaluates the performance of each query category and does not require correct classification of all categories. In contrast, its average mAP score of $5.96\%$ is low when all categories are evaluated. According to the false positive analysis in Fig~\ref{appfig:moment_sensitivity_analysis}, errors caused by wrong labels are significant. A more sophisticated classifier for all candidate moments can be explored in future work. In addition, as shown in Fig~\ref{appfig:moment_sensitivity_analysis}, the performance of short moments, which occupy a large proportion in the dataset, is not as good as that of long moments. Therefore, improving  short moments will significantly improve the overall performance.

\iftoggle{arxiv}{
\subsubsection*{Contributions statement}

Kristen Grauman led the Episodic Memory benchmark and paper writing, wrote annotation instructions,  contributed to data selection and taxonomy formation, and co-advised the VQ baseline development. Bernard Ghanem co-led the Episodic Memory benchmark, managed baseline development and evaluation for the MQ and NLQ tasks, and contributed to the annotation instructions, data selection, and taxonomy formation for the MQ and NLQ datasets. Jackson Hamburger contributed to the development of the annotation instructions and taxonomies of the NLQ, VQ, and MQ datasets along with the design of the early VQ baselines.

Santhosh Kumar Ramakrishnan led VQ data selection, annotation, analysis and auditing, contributed to the formulation and annotation instructions of VQ, data selection, and implemented the VQ baseline.
Vince Cartillier contributed to the VQ-3D formulation and annotation instructions, led VQ-3D data selection, annotation, analysis and auditing, and implemented the VQ-3D baseline. Dhruv Batra co-mentored Vince Cartillier on developing baselines and provided guidance on 3D scans using Matterport. Hyun Soo Park contributed to 3D reconstruction of egocentric videos with respect to 3D Matterport scans. Tien Do developed algorithms to reconstruct 3D egocentric camera poses with respect to 3D Matterport scans.

James Hillis provided background knowledge on human episodic memory function and contributed to early discussions on benchmark definition and annotation.
Satwik Kottur led the design of NLQ data selection and annotation instructions, contributed to the NLQ task formulation, coordinated NLQ data annotation and data analysis, implemented the VSLNet NLQ baseline, wrote part of the NLQ sections. Menmeng Xu designed and implemented the experiment pipeline for the NLQ task, implemented several NLQ methods, did NLQ result analysis and visualization, and wrote part of the NLQ sections.
Michael Wray contributed to early formulation of the benchmark tasks, definitions of NLQ queries and annotation instructions, provided input for dataset construction and evaluation metrics, and helped in the creation of the MQ taxonomy.  %

Chen Zhao designed and implemented the MQ baseline, proposed and implemented the new metric for MQ, wrote the MQ sections, did MQ result analysis and visualization, contributed to the formulation, data selection and annotation instructions of MQ. Tushar Nagarajan contributed to the MQ formulation and annotation instructions, developed the MQ label taxonomy, and led the data selection and annotation of the MQ dataset. Merey Ramazanova managed the datasets for the experiments of MQ and NLQ baselines, and assisted with the taxonomy formation for the MQ baseline. Antonino Furnari provided keypoint feature extraction from the Matterport3D panoramas for the VQ3D baseline.}
{}

\clearpage
\subsection{Hands and Objects Benchmark}
\label{appendix:hands-objects}

This section details the Hands and Objects benchmark including definitions, annotations, baseline models and results.

\subsubsection{Motivation}

\begin{figure}[t]
  \includegraphics[width=\linewidth]{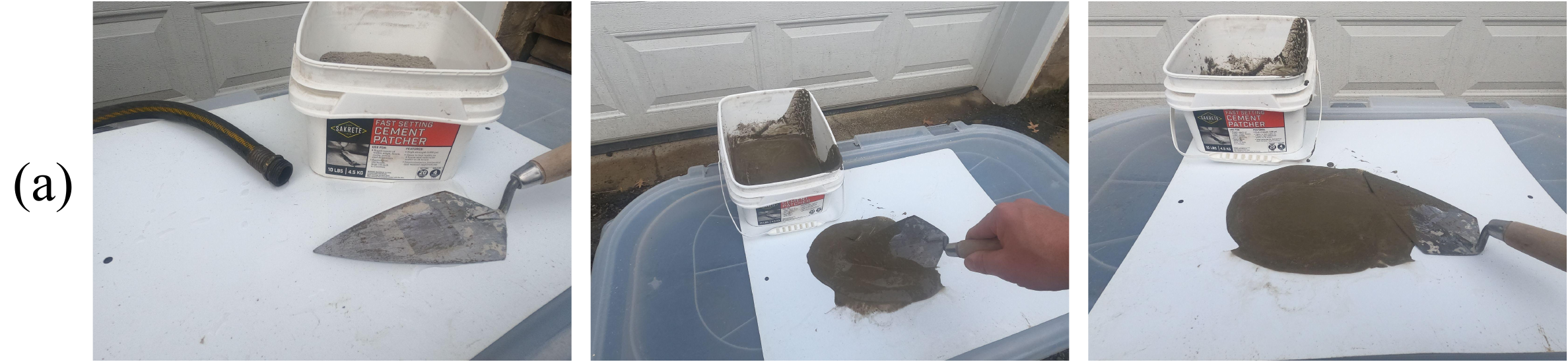}
  \includegraphics[width=\linewidth]{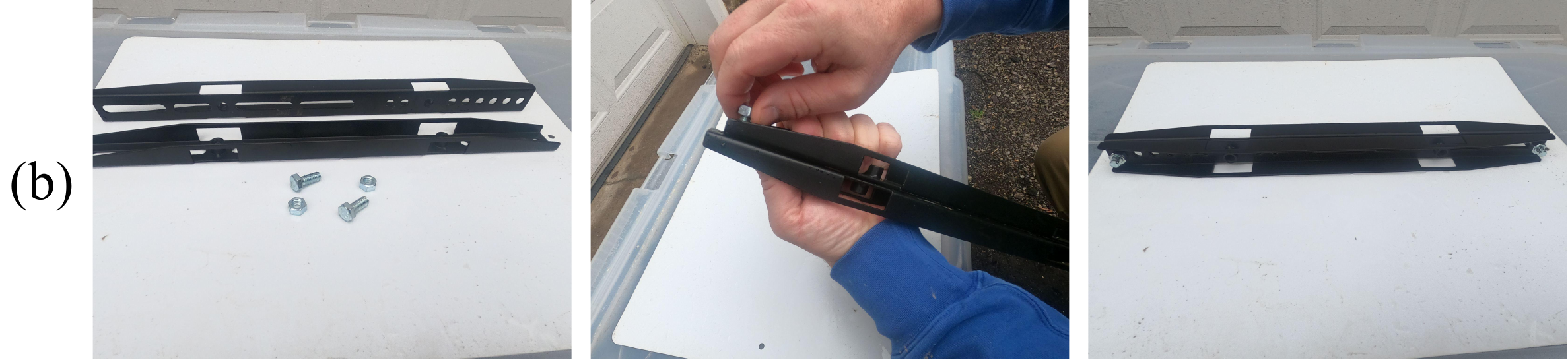} 
  \includegraphics[width=\linewidth]{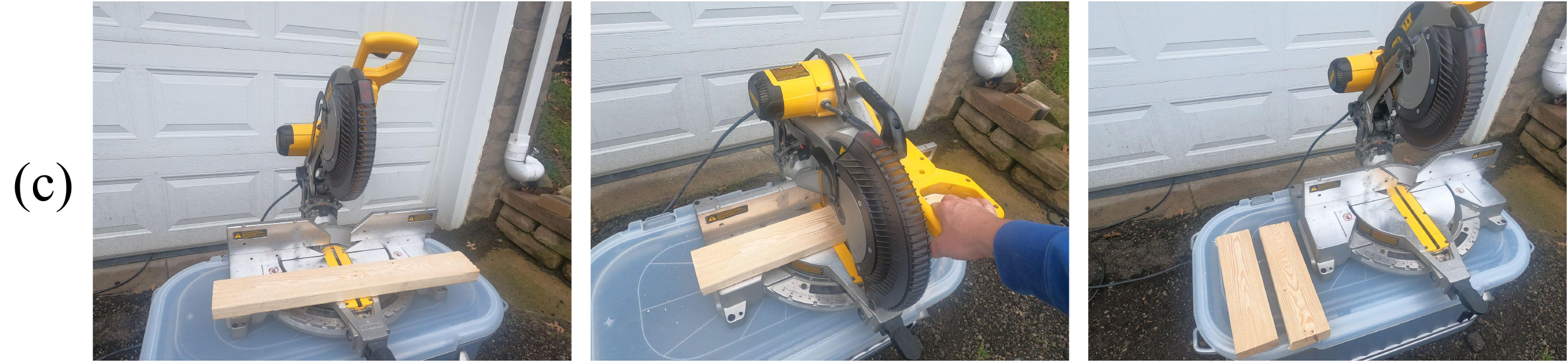}
  \caption{Examples of object state change. (a) State change through construction: attaching to two metal plates results in a new object.
  (b) State change through physical change: cutting a piece of wood results in two smaller pieces of wood.
  (c) State change through chemical reaction: combining two objects, water and cement powder, results in a new object, cement.}
\label{fig:state:change:example}
\end{figure}

In a video of a human operating and manipulating an object with their hands, there may exist an \emph{object state change}, \emph{i.e.}, the point where the state of the objects being operated changes, either temporarily or permanently in a way that cannot be easily reversed.
Examples of temporary state change include turning on a machine, while examples of permanent state changes include physical changes such as chopping a tomato into pieces and chemical changes such as mixing water and cement powder together to create a new composition of cement.
Some examples are illustrated in Figure \ref{fig:state:change:example}.

The concept of an object state change has been explored only in a limited manner in the video  literature \cite{fathi2013modeling, alayrac2017joint, damen2014} and the characterization of state changes has depended on many brittle vision-based component technologies, making it difficult to analyze state changes at scale. Fortunately, in the last decade we have seen tremendous advances in computer vision algorithms for understanding both objects and hands. As a result, we believe that now it is time to investigate the idea of characterizing state changes at scale and in depth.

Why is recognizing the impact of agents on objects and environments so critical? 
We believe that understanding, recognizing, and replicating object state changes are an essential aspect of creating artificial intelligence (AI) systems.
While current AI systems have the ability to replicate certain types of human actions such as assembling furniture \cite{knepper2013ikeabot} or cutting tomatoes \cite{sharma2019learning}, most systems do not  possess a general understanding of how the environment and the objects can be transformed as a result of interaction. Understanding the impact of interactions on objects and the environment is an important aspect of reasoning and can help AI systems perform more advanced tasks. For example, understanding the impact of interactions on the environment can help AI systems relate multiple ways to achieve the same change, discover efficient methods for achieving goal states, recognize the completion/incompletion of goals \cite{Heidarivincheh2018, Epstein2019}, recover from failure, and learn from mistakes.

In egocentric videos specifically, the object state changes offer rich and important information that are related to many other problems. For example, the object undergoing state change in an egocentric video can imply human-centric information such as human activity and intention. Moreover, the state change of an object shown provides cues about human-specific affordance and actionable information of an object or tool, which cannot be easily inferred from static images. Additionally, a joint understanding of human hands and the objects undergoing state change can benefit applications that require rich human demonstrations, such as robotic manipulation.

\textbf{Defining Object State Changes:} This benchmark focuses on identifying and localizing the state change of an object in an egocentric video. Specifically, a object state change can be represented by the three aspects in the video: temporal, spatial, and semantic.

\paragraph{Temporal:} An object state change can be represented by three distinct temporal points in the video.
\textit{(1) Point-of-no-return:}  The point-of-no-return (PNR) is the frame $I_{\text{pnr}}$ in a video that identifies the beginning of an object state change that cannot be easily reversed. 
\textit{(2) Pre-condition:} The pre-condition is defined as some frame $I_\text{pre}$ that marks a moment prior to the state-change in which the related objects were visible within the field of view of the camera. 
\textit{(3) Post-condition:} The post-condition is some frame $I_\text{post}$ at which the completion of the state change is visible after the point-of-no-return. 
These three frames mark the distinct temporal stages of the object state change: before and after the change, respectively. This proposal matches the Rubicon Boundaries proposed in~\cite{Moltisanti2017}.

\paragraph{Spatial:} An object state change can be represented by the bounding box of the object at the PNR, pre-condition and post-condition, along with any tools involved in performing the state change. Tools offer extended capabilities of the actor's hand, such as using an electric saw to cut a piece of wood in half. These bounding boxes represent the spatial dimensions of hands, tools and the objects undergoing the state change. 

\paragraph{Semantic:} We represent an object state change through the human action (verb), the object identity (noun) and the type of state change applied. The same state change can be performed on different objects using different tools. For example, cutting a piece of wood with electric saw and cutting a piece of paper with scissors are different interactions with different objects and different tools but they both result in the same object state change of \textit{being cut}.

\subsubsection{Related Work}

\paragraph{Object State Changes:} 
Existing approaches for modeling object states and/or their changes can be categorized into two research lines. The first deals with collections of images.
A representative dataset for this purpose is the MIT States dataset \cite{isola2015discovering}. By considering object states as object attributes (e.g. burnt, sliced), this line of work studies attribute-object composition, e.g. composition with context~\cite{composition:with:context}, modeling attributes as operators~\cite{attributes:as:operators}, and an architecture for compositional reasoning~\cite{task:driven:modular:networks:for:compositional:learning}. 

The second research line deals with video and views an action as a state transformation over time. 
One direction is the discovery of object states and/or manipulating actions, \eg in egocentric~\cite{fathi2013modeling,damen2014} and instructional videos~\cite{alayrac2017joint}. Fathi et al. \cite{fathi2013modeling} explore object state detection in video using a weakly supervised approach.
Another direction is the modeling of state transitions. 
Zhou et al.~\cite{zhou2016learning} study temporal transformations of a single object state in time-lapse videos.
Wang et al.~\cite{wang2016actions} propose to model state transformations in a high-level feature space with Siamese networks.
Doughty et al.~\cite{action:modifiers} leverage natural language and treat adverbs as modifiers for state transformations.
In terms of applications, Chang et al.~\cite{procedure:planning:in:instructional:videos} show state transformations can be utilized for procedure planning.

\paragraph{Human Hand Action Datasets:} Several video datasets have been proposed for human hand action recognition. The Yale human grasping dataset~\cite{yale:human:grasping:dataset} focuses on human grasping behavior and consists of 27.7 hours of annotated videos. The Something-Something dataset \cite{something:something} consists of 220,847 short videos annotated with 174 categories of general hand-object interactions.
The Jester dataset \cite{Jester} provides 148,092 short videos in 27 hand gesture types. 
Wang et al. \cite{wang:generative:model:human:object:interactions} construct a synthetic video dataset of human-object interaction through rendering hand and object CAD models. The recent Human Hands dataset~\cite{Shan20} annotates 100K single frames from web-based videos, focusing on hand interactions and the offset between the hand and the interacting object during interaction.

Several egocentric video datasets capture daily living activities by people~\cite{adl,utego,gtea,engagement,Damen2020RESCALING,charades-ego}.
In the Activities of Daily Living Dataset (ADL), subjects wear chest-mounted cameras and perform unscripted activities at home, with a total of 10 hours of video from 20 participants; the target task is activity recognition~\cite{adl}.  In the UT-Egocentric dataset (UT-Ego), subjects wear a head-mounted camera and perform long unscripted activities inside and outside of the home, with a total of 17 hours from 4 subjects (4-5 hours of continuous capture for each person); the target task is video summarization~\cite{utego}.  The UT Egocentric Engagement (UT EE) dataset consists of 14 hours of head-mounted camera video captured in public spaces like museums, malls, and grocery stores, and is annotated for moments of engagement by the camera wearer with the environment.  In the EGTEA+ dataset, 32 subjects wearing head-mounted cameras in a single kitchen environment capture 28 hours of video; the task is to recognize 44 meal preparation activities~\cite{gtea}.  
The EPIC-KITCHENS dataset consists of 100 hours of kitchen activities recorded in 45 unique environments, with a total of 89,977 different object interactions across 97 verb and 330 noun classes; the task is to recognize objects and activities and anticipate interactions in the next moment of video~\cite{Damen2020RESCALING}.  The Charades-Ego dataset consists of 34 hours of video from 71 participants, with both first- and third-person paired instances labeled for 156 actions~\cite{charades-ego}.

\subsubsection{Benchmark Definitions}

We now define the three tasks that comprise the Hands and Objects benchmark. The three tasks correspond to the three aspects of object state changes described above, namely, the temporal, spatial and semantic aspects of a state change.

\paragraph{(1) PNR Temporal Localization. }

The goal of Point-of-no-return (PNR) Temporal Localization is to predict $I_{\text{pnr}}$. One possible formulation is to view this problem as a per-frame classification problem, predicting the Point-of-no-return frame within a short video clip. The performance is evaluated only on the videos that contain object state change, and is measured by the absolute temporal error of $I_{\text{pnr}}$ prediction in seconds.

\FEB{The PNR was first discussed by P.~Gollwitzer in his well-cited handbook of behavior~\cite{gollwitzer}. Specifically, the book proposes the Rubicon Model of Action Phases, focusing on hand-object interaction. Action phases are delimited by three transition points: initiation of prior motion, PNR, and goal achievement. This was later experimentally assessed by our previous work~\cite{Moltisanti2017}, where PNR annotations
were acquired for three egocentric datasets, demonstrating
increased accuracy of annotations (see Fig.~10 in~\cite{Moltisanti2017}) and improved robustness in training models (see Sec.~5 in~\cite{Moltisanti2017}). Below, we find PNR closely aligns with the narration timestamps that we independently collected, suggesting PNR is a natural time point for human understanding (and thus narration) of the interaction.}

\paragraph{(2) State Change Object Detection. }

We define a State Change Object as the object that is manipulated by a person and  undergoes a change in its state. The goal of this task is to predict the 2D bounding boxes of the State Change Object in Point-of-no-return frame $I_{\text{pnr}}$ given three frames: Pre-condition $I_{\text{pre}}$, Point-of-no-return $I_{\text{pnr}}$, and Post-condition $I_{\text{post}}$. 
We expect that a good solution to this task would incorporate the visual information before and after state change to detect the State Change Object.
The detection performance is evaluated on the bounding boxes estimated in the Point-of-no-return frame $I_{\text{pnr}}$ and measured by Average Precision (AP).

\paragraph{(3) Object State Change Classification. }

The task of Object State Change Classification classifies a short video clip to a state change type. With $N$ object state change types defined, object state change classification is essentially an $(N+1)$-way classification problem, where the one additional category is ``without state change.'' Object State Change Classification is evaluated by classification accuracy.

\subsubsection{Data Selection}\label{app:data_anno}

Next we describe our data selection procedure and annotation pipeline, and we present the analysis of the data for the object state change benchmark. We begin by describing our procedure for selecting the subset of data to annotate for this benchmark.

\begin{figure*}\centering
\subfigure[\textbf{Scenarios}]{\label{fig:scen_dist}\includegraphics[width=0.65\linewidth]{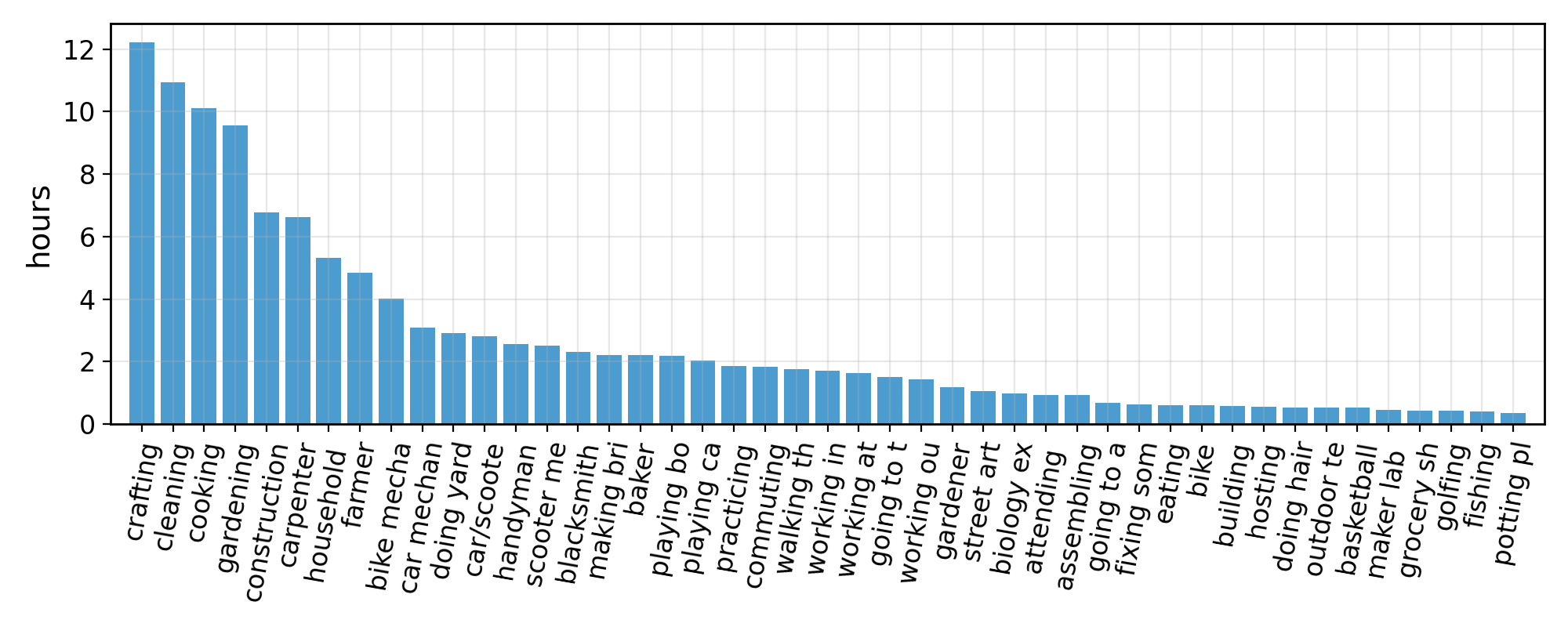}}
\subfigure[\textbf{Universities}]{\label{fig:uni_dist}\includegraphics[width=0.26\linewidth]{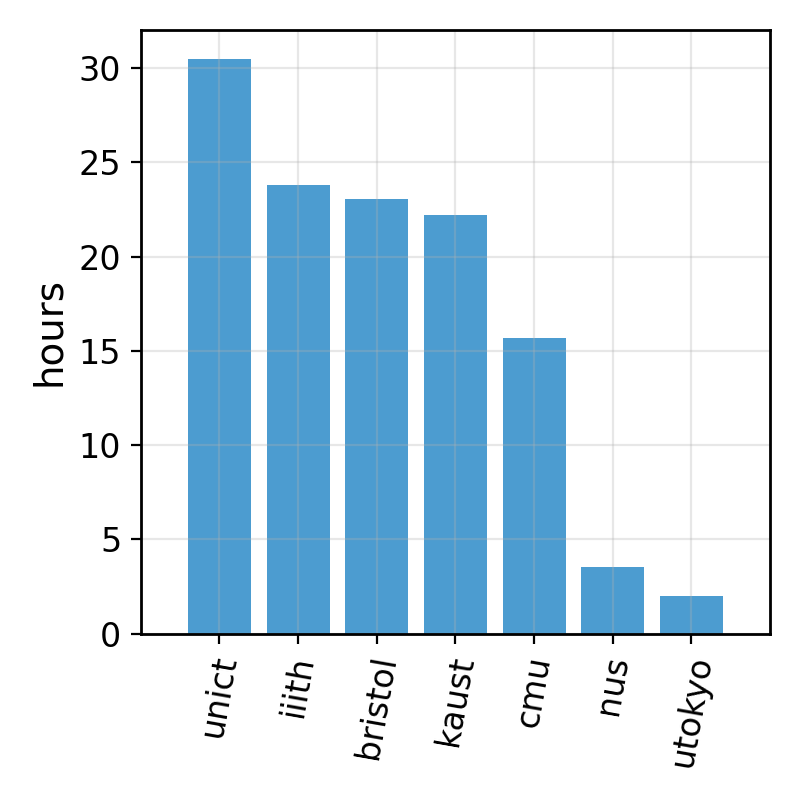}}\\[1mm] %
\caption{%
\textbf{Number of hours.} We show the distribution of the number of hours across scenarios\iftoggle{arxiv}{ (left) and universities (right)}.}
\label{fig:data_stats}\vspace{-2mm}
\end{figure*}

We start with a large pool of videos annotated with high-level scenario labels (\eg, gardening, cooking, landscaping, \etc) and narrations. We assess each scenario on the scale of 0 to 3 based on how likely it is to contain hand-object interactions (\eg, 0 for ``watching tv'', 3 for ``carpentery'', \etc). We then sample data to annotate following the resulting scenario distribution. Given a scenario and a target number of hours, we sample clips randomly in a hierarchical fashion: we first sample a participant, then a video, and finally a 5 minute clip from the video. If the video is shorter than 5 min we take the whole video. For each scenario, we balance the data across universities to maximize geographic diversity. 
The resulting scenario \iftoggle{arxiv}{and university distributions are}{distribution is} shown in Figure~\ref{fig:data_stats}. In total, our dataset has 120 hours representing 53 scenarios, 7 universities, and 406 participants.

\subsubsection{Data Annotation}

We annotate hand-object interactions corresponding to each narration within the selected 5 minute clips. We use the taxonomy from Section~\ref{sec:taxonomy} for semantic verb and noun labeling. The annotation pipeline consists of three sequential stages: critical frame labeling, pre-period labeling, and post-period labeling.

\ourparagraph{Critical frames.} Given a narration, we create an 8 second video snippet centered at the narration time point and present it to the annotators. We ask the annotators to first read the narration and select a corresponding verb from the taxonomy. The annotators can then play the video back and forth to select three critical frames in time: PNR, PRE, and POST. We ask the annotators to start with the PNR frame that identifies the beginning of the state change. This frame is less ambiguous and helps provide the context for the interaction. We then ask the annotators to label a frame prior to the state change (PRE) and a frame after the completion of the state change (POST). Note that the PRE and POST frames are not uniquely defined. We let the annotators pick any, as long as the relevant objects are \emph{fully} visible within the field of view of the camera.

\ourparagraph{Pre period.} Next, we label bounding boxes for the hands, tools, and objects, as well as the category names for the tools and objects. We do this in two steps. First we label the frames in the pre period, starting at PNR and going backward to the pre frame. The video frames are reversed and the annotators can play the video. We find that it is easier to start from the PNR frame since the hands and objects are clearly visible. To speed up hand box labeling, we initialize the hand boxes with a pre-trained object detector~\cite{Shan20} and ask the annotators to correct these.

\ourparagraph{Post period.} Finally, we ask the annotators to label spatial annotations and categories for the post frame. As before, we first present the annotators with the PNR frame. Note that in this case the PNR frame is already labeled which helps identify the hands and objects to label in the post frame.

\subsubsection{Data Analysis}

\begin{figure*}
\centering
\subfigure{\includegraphics[width=0.49\linewidth]{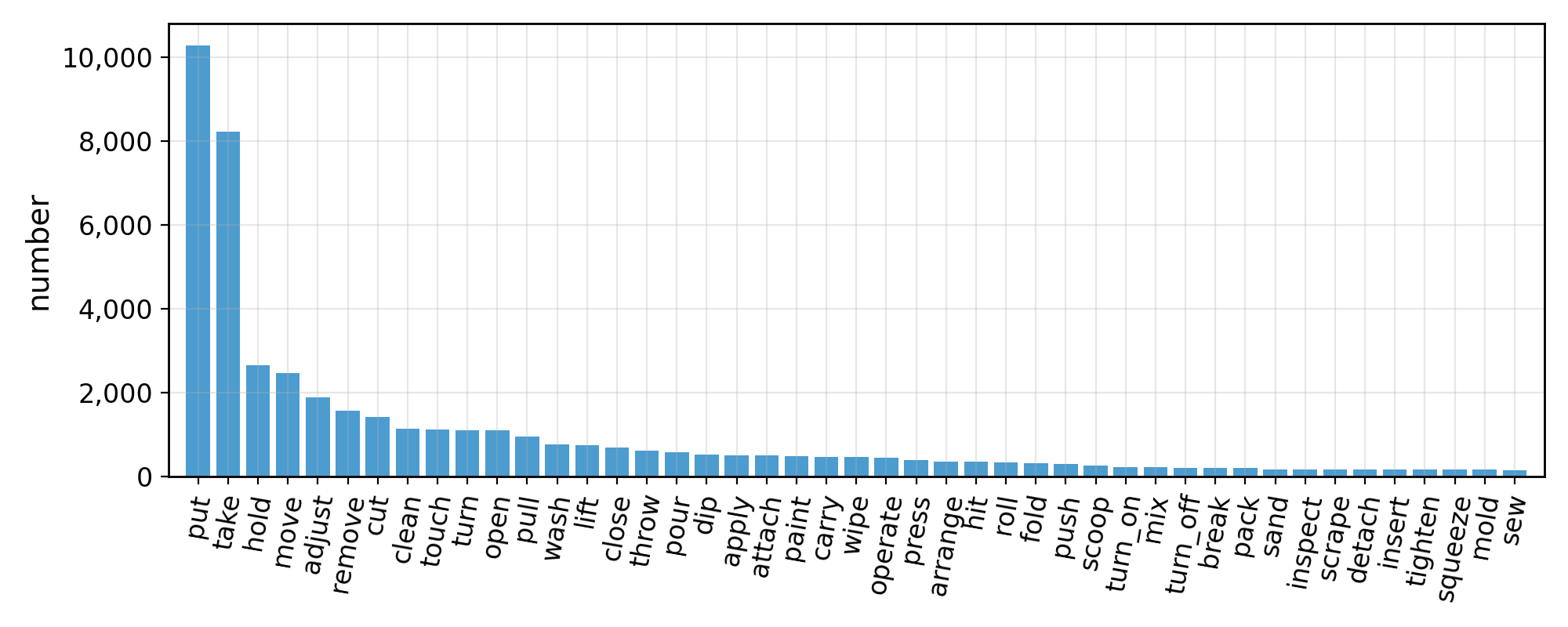}}
\hfill
\subfigure{\includegraphics[width=0.49\linewidth]{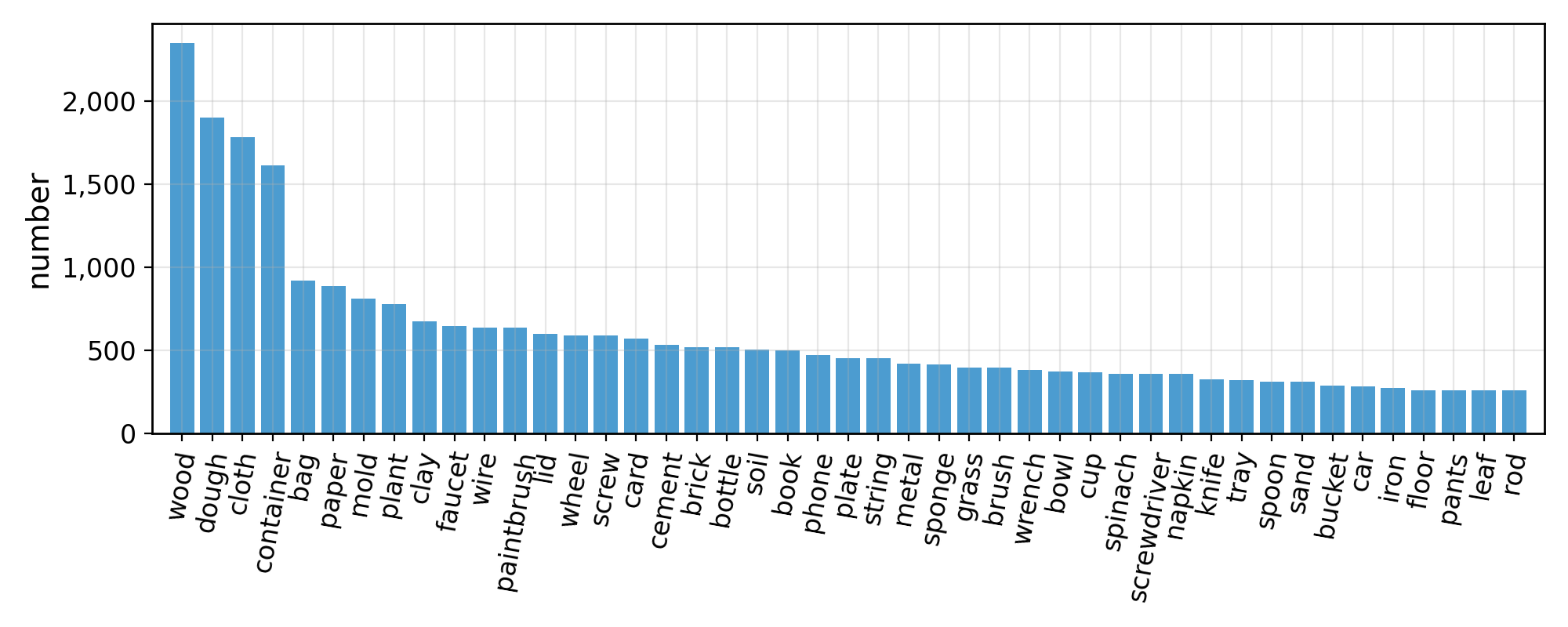}}\\[1mm]
\caption{%
\textbf{Labeled actions.} Distribution of verbs (left) and nouns (right) in annotated action instances. Top 45 verbs and nouns are shown for clarity. See Section~\ref{sec:taxonomy} for more details.
}
\label{fig:taxonomy}
\vspace{-3mm}
\end{figure*}

Finally, we present the analysis of our annotations.

\begin{figure}[t]\centering
\includegraphics[width=1.0\linewidth]{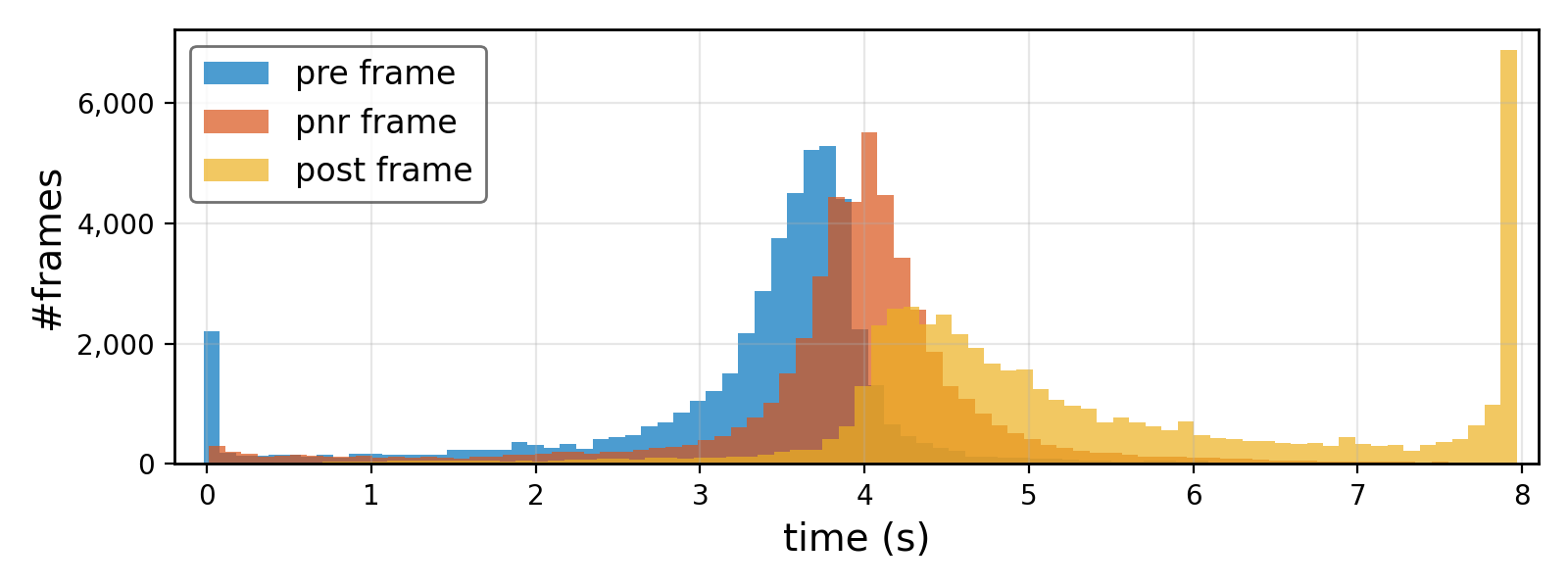}
\caption{\textbf{Critical frames.} Distribution of critical frame times. Shown relative to the 8s hand-object interaction snippet.}
\label{fig:critical_frame_stats}
\end{figure}

\ourparagraph{Critical frames.} In Figure~\ref{fig:critical_frame_stats} we show the temporal distribution of critical frames within the 8 second hand-object interaction snippets. First, we observe that the PNR frame distribution is centered around the middle of the 8 second snippet. Interestingly, this closely aligns with the narration point (4s mark). Next, we see that most of the pre and post frames come shortly before and after the PNR frame, respectively, highlighting the quick nature of these state changes, and thus the challenge in this benchmark. We also notice two additional modes for pre and post frames that come at the start and the end of the 8s interval, respectively. These correspond to long repetitive actions that start before or continue past the video snippet (\eg, knitting).

\begin{figure}[t]\centering
\includegraphics[width=1.0\linewidth]{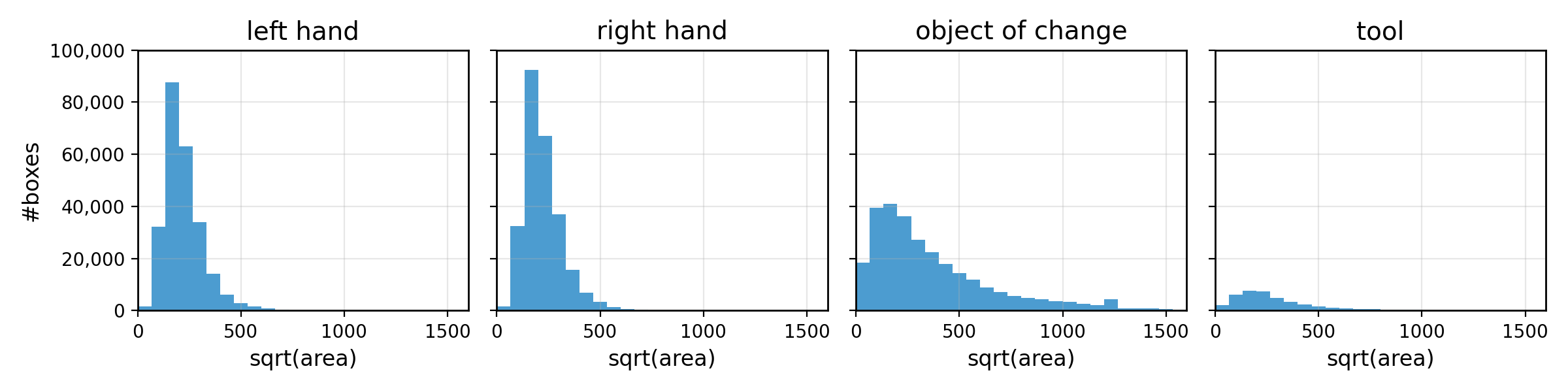}
\caption{\textbf{Hand and object sizes.} Distribution of bounding box sizes. Shown in terms of the square root of the box areas.}
\label{fig:box_size_stats}
\end{figure}
\begin{figure}[t]\centering
\includegraphics[width=1.0\linewidth]{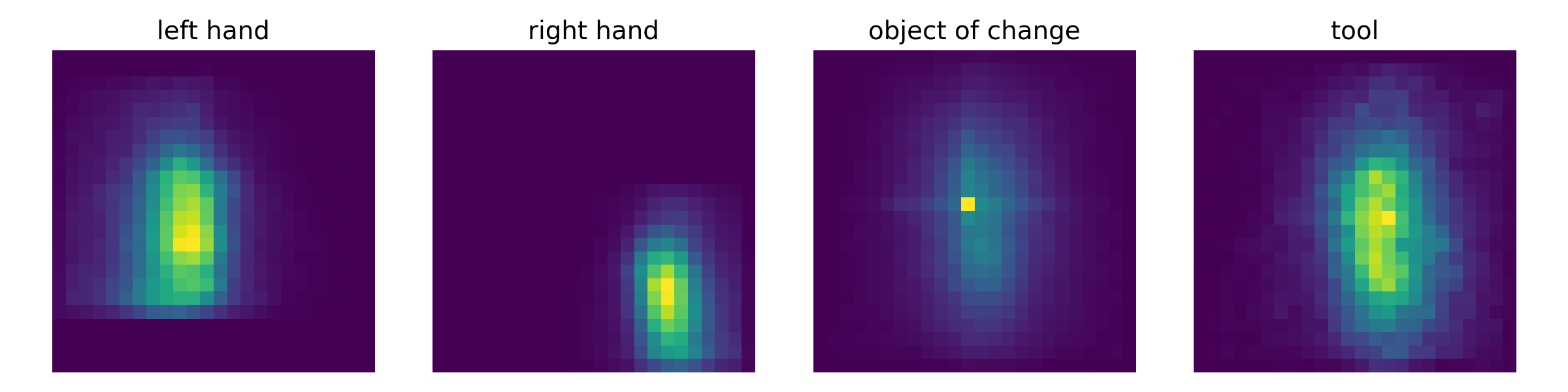}
\caption{\textbf{Hand and object locations.} Distribution of bounding box centers. Shown in normalized image coordinates.}
\label{fig:box_loc_stats}
\end{figure}

\ourparagraph{Hands and objects.} Our benchmark contains a large number of hands and objects annotated with bounding boxes. In total, we have $\sim$825K bounding boxes, including $\sim$245K for left hand, $\sim$260K for right hand, $\sim$280K for objects, and $\sim$40K for tools. In Figure~\ref{fig:box_size_stats} and Figure~\ref{fig:box_loc_stats}, we show the distributions of box sizes and locations, respectively. We observe that our data contains hands and objects at a variety of sizes and locations.

\ourparagraph{Actions.} One of the features of our benchmark is the diversity of interactions. We focus on low-level atomic actions rather than high-level actions. We show the distribution of verbs (Figure~\ref{fig:taxonomy}, left) and nouns (Figure~\ref{fig:taxonomy}, right). We see that we have a large number of verbs corresponding to common manipulation actions (\eg, put, take) and a natural long tail. The object distribution follows the same general trend. We note that our objects are common daily objects that are not typically present in object detection datasets (\eg, 442 out of our 478 object categories cover categories beyond the 80 COCO~\cite{lin2014microsoft} categories).

\subsubsection{Baselines: Object State Change Classification and PNR Temporal Localization}

We present the implementation of several baseline methods for the Object State Change Classification and PNR Temporal Localization tasks.
Among the implemented baseline models, in general there are one or two types of output network heads: a classification head for the video clip used for state change classification, and/or a per-frame classification head for temporal localization.
One can choose to train two models separately, or use the same backbone model but two network output heads and train the joint model with a multi-task loss function. 
The following baseline methods includes both types of model designs:

\textbf{I3D ResNet-50. }
We use I3D \cite{i3d} with ResNet-50 \cite{resnet} as backbone architecture of the model for both the Object State Change Classification and the PNR Temporal Localization tasks.
The ResNet backbone is followed by two network output heads: a state change classification head and a PNR temporal localization head.
The state change classification head is produced by global average pooling on the entire spatiotemporal feature tensor followed by a classification layer.
The PNR temporal localization head is produced by per-frame average pooling followed by a classification layer.
The overall training loss of the model is the combination of the loss of two heads which are both cross-entropy loss for classification.

\textbf{Boundary Matching Network (BMN). }
We use BMN \cite{bmn} as a baseline for the PNR Temporal Localization task.
BMN is a temporal segment detection method based on confidence prediction of dense temporal segment proposals.
We view the start of the video as the start of the temporal segment and Point-of-no-return $I_\textrm{pnr}$ as the end of the temporal segment, so we can convert the problem of localizing Point-of-no-return $I_\textrm{pnr}$ to the problem of detecting the temporal segment.
In our implementation, BMN uses ResNet as the backbone model. Furthermore, BMN is only used for the PNR temporal localization task.

\textbf{SlowFast + Perceiver. }
We implement a baseline model whose architecture consists of SlowFast \cite{slowfast} and Perceiver \cite{perceiver} for both object state change classification and PNR temporal localization.
SlowFast acts as the video deep feature extractor.
The features are then passed to a Perceiver model.
Similar to the previous BMN model, the SlowFast + Perciever model is only trained for temporal localization task.
The training loss of the model is the cross-entropy loss for per-frame classification.

\textbf{Bi-directional LSTM. }
We implement a Bi-directional LSTM model \cite{bi:lstm} for both the object state change classification and PNR temporal localization. We first pass individual frames to a ResNet model \cite{resnet} to extract deep features. The sequence of per-frame features is then passed to the Bi-directional LSTM as input, with the output sent to both the per-frame classification head and the whole-sequence classification head. The overall training loss of the model is the combination of the loss of two heads which are both cross-entropy loss for classification.

\begin{table}[t]
\centering
\small
\caption{Number of positive and negative video clips of object state change in train, validation and test splits. }
\label{tab:state:change:clip:stats}
\vspace{1ex}
\begin{tabular}{l|c|c|c}
\hline
Split & Positive & Negative & Total \\ \hline
Train & 20,041 & 21,044 & 41,085 \\ \hline
Val & 13,628 & 14,720 & 28,348 \\ \hline
Test & 13,561 & 14,870 & 28,431 \\ \hline
\end{tabular}
\end{table}

For the object state change classification tasks, in the current version we focus on the two-way classification problem of whether there is a object state change in the egocentric video.
In Table \ref{tab:state:change:clip:stats}, we illustrate the number of positive video clips that contains an object state change and the number of negative video clips that do not contain object state change in the train/val/test splits. In all three splits, the positive and negative clips are balanced in number.

Besides the above learnable baselines, for object state change classification, we also present the result of the naive baseline of always predicting the positive category as the prediction.
For the PNR temporal localization task, we additionally present the result of the naive baseline of always selecting the center frame of the trimmed video as the PNR frame, given the possible centre bias of the data.

The results for object state change classification task are illustrated in Table \ref{tab:state:change:classification}.
The naive baseline of always positive prediction yields state change classification accuracy of close to 50\%.
All the learnable baselines outperform the naive baseline and achieve accuracy of more than 60\% while Bi-directional LSTM baseline achieves the best performance. This shows that the learnable baselines can learn meaningful information about object state change, though there is clearly still space for improvement.
One challenge in this task is that there is very large variance in term of the types of object state changes and objects contained in the videos.

The results for the PNR temporal localization task are illustrated in Table \ref{tab:pnr:localization}.
The naive baseline of always predicting the center frame yields a temporal localization error of around 1{.}1 seconds.
Other learnable baselines can achieve better temporal localization error of around 0.85 seconds or less which shows the baseline models can learn meaningful information for temporal localization of object state change. Note that the SlowFast + Perceiver model achieves the best temporal localization performance of 0.425 seconds on validation set and 0.489 seconds on test set, which highlights the necessity of using attention-based mechanism to model the change of object state. One challenge for this task is that in some actions, \eg, cutting a piece of paper with scissors, the state change of an object does not necessarily cause significant change of visual appearance and therefore it is difficult to localize the PNR.

\begin{table}[t]
\centering
\small
\caption{Results of State Change Classification accuracy (\%). }
\label{tab:state:change:classification}
\vspace{1ex}
\begin{tabular}{l|c|c}
\hline
Baseline &  Val & Test \\ \hline
Always Positive &  48.1 &	47.7 \\ \hline
Bi-directional LSTM \cite{bi:lstm} & 65.3 & 63.8  \\ \hline
I3D ResNet-50 \cite{i3d} & 68.7 & 67.6 \\ \hline
\end{tabular}
\end{table}

\begin{table}[t]
\centering
\small
\caption{Results of Point-of-no-return temporal localization error (seconds). }
\label{tab:pnr:localization}
\vspace{1ex}
\begin{tabular}{l|c|c}
\hline
Baseline &  Val & Test \\ \hline
Always Center Frame &  1.032	& 1.056  \\ \hline
BMN \cite{bmn} & 0.780 & 0.805   \\ \hline
I3D ResNet-50 \cite{i3d} & 0.739 & 0.755 \\
\hline
Bi-directional LSTM \cite{bi:lstm} & 0.790 & 0.759  \\ \hline
SlowFast \cite{slowfast} + Perceiver \cite{perceiver} & 0.804 & 0.828 \\ \hline
\end{tabular}
\end{table}

\subsubsection{Baselines: State Change Object Detection}

While we expect that new methods developed for the tasks of state change object detection will utilize all three input frames (pre, PNR, post), in this initial stage of the benchmark, we only evaluate single-frame detection baselines, where only the PNR frame $I_\text{pnr}$ is used as input. We limited our input as many methods for object detection are primarily designed to work with a single image.

We present the implementation of several baseline methods for the state change object detection task. In general, the baseline models for the task can be categorized into two types: (1) directly detecting the bounding box of the state change object including Faster-RCNN \cite{faster:rcnn}, CenterNet \cite{centernet:zhou}, and DETR \cite{detr}, and (2) detecting hand bounding boxes first then predict state change object bounding boxes given the hands such as the 100DOH model \cite{100doh}. Specifically, the baseline methods are the following:

\textbf{Faster-RCNN \cite{faster:rcnn} } is a two-stage anchor-based 2D object detector on a single RGB image. In its classification head, the state change object is the only positive category. We train Faster-RCNN on our benchmark and use it to directly detect the bounding boxes of state change objects in PNR frames.

\textbf{CenterNet \cite{centernet:zhou} } is another object detection method on a single RGB image. It estimates object keypoints to find object center points and regresses all other object properties, such as size, 3D location, and orientation.
We train CenterNet to directly detect the bounding boxes of state change objects.

\textbf{DETR \cite{detr}} is an object detection model on a single RGB image based on Transformer \cite{transformer}. It views object detection as a direct set prediction problem and uses a transformer encoder-decoder architecture to produce a set of object predictions including bounding box information as well as other information such as category. We train DETR to directly detect the bounding boxes of state change objects.

\textbf{100DOH Model \cite{100doh} } first detects the bounding boxes of the human hand and objects as well as the relational vectors that links from each hand bounding box center to an object bounding box center. The final prediction of the objects are decided as the object predictions that satisfies the both the predictions of hand and relational vectors. 
We used the 100DOH model pre-trained on 100DOH dataset \cite{100doh} to first detect hand bounding boxes and then predict state change object bounding boxes given the hands.

\begin{table}[t]
\centering
\small
\caption{Number of State Change Object and hand bounding boxes in train, validation and test splits. }
\label{tab:state:change:box:stats}
\vspace{1ex}
\begin{tabular}{l|c|c}
\hline
Split & State Change Object & Hand \\ \hline
Train & 19,347 & 33,254 \\ \hline
Val & 12,912 & 22,098 \\ \hline
Test & 13,118 & 22,576 \\ \hline
\end{tabular}
\end{table}

\begin{table}[t]
\centering
\small
\setlength{\tabcolsep}{3.5pt}
\caption{Results of single-frame State Change Object Detection. The performance is measured in Average Precision (AP). }
\label{tab:active:object:det:single}
\vspace{1ex}
\begin{tabular}{l|c|c|c|c}
\hline
Baseline & Backbone &  AP & AP50 & AP75 \\ \hline
Faster-RCNN \cite{faster:rcnn}
 &  ResNet-101 \cite{resnet} &  13.4 & 25.6 & 12.5 \\ \hline
DETR \cite{detr} &  ResNet-50 \cite{resnet} &  15.5 & 32.8 & 13.0  \\ \hline
CenterNet \cite{centernet:zhou} & DLA-34 \cite{dla} & 6.4 & 11.7 & 6.1
 \\ \hline
100DOH Model \cite{100doh}&ResNet-101 \cite{resnet} & 10.7  & 20.6 & 10.1 \\ \hline
\end{tabular}
\end{table}

We show the number of state change objects and hand bounding boxes contained in our dataset in Table \ref{tab:state:change:box:stats}.
The results of single-frame State Change Object Detection are illustrated in Table \ref{tab:active:object:det:single}. All baselines struggle in detecting the State Change Objects with only one frame as input as an AP of 8-14\%. There are several challenges in this task. First, the bounding box sizes of state change objects have large variance. For example, the size of state change objects can be as large as half of image in the action of ``painting the wall'' and as small as a few pixels in the action of ``igniting the match.'' Second, when only using one frame as input, the detection models did not consider the change of object appearance across different frames. As future work, we hope the researchers will investigate using models that take multiple frames as input and perhaps develop frameworks that incorporate tracking or association.

\subsubsection{Discussion}

This novel benchmark explores three aspects of objects undergoing state changes as a result of hand manipulation: the \emph{when} (i.e. temporal localization of state change), \emph{where}~(i.e., spatial localization of objects that undergo change) and \emph{what}~(i.e., semantic notion of action and object transformation). As a first step, we have explored these independently using readily available localization and classification methods. However, approaches that aim to tackle this challenge should focus on jointly understanding the manipulation with its spatio-temporal impact on objects as these are transformed. For example, knowing an object is being split should offer a strong prior to the PNR localisation and detect two or more bounding boxes after the point-of-no-return. Such methods that tackle the dependencies between the tasks are yet to be developed. We hope this benchmark will spur innovative approaches that bridge the gap between action perception and the impact of actions on objects and environments.

\iftoggle{arxiv}{
\subsubsection{Contributions statement}

Kris Kitani helped formulate and write the object state change benchmark, designed the annotations and tasks for the HO benchmark. Dima Damen helped with the formulation and writing of the object state change benchmark, designed the annotations for the Hands and Objects (HO), and Forecasting benchmarks.   Ilija Radosavovic coordinated  HO data annotation, annotation analysis, and contributed to the definition and writing of the HO benchmarks.  Rohit Girdhar helped coordinate the HO data annotation and annotation analysis. 
Abrham Gebreselasie adapted the SlowFast+Perceiver model for PNR temporal localization.   Qichen Fu implemented all of the state change object detection baselines. Raghava Modhugu implemented the BMN baseline for PNR temporal localization.  Kristen Grauman contributed to the formulation and writing of object state change benchmark. Siddhant Bansal helped with the processing of HO data, development of HO data loader for PNR temporal localization and implemented the I3D ResNet-50 baselines. Xingyu Liu was the lead coordinator and mentor of the HO benchmark baseline implementations, and also contributed to the definition and writing of HO benchmarks. Xuhua Huang developed of the initial SlowFast+Perceiver model. Yifei Huang implemented the Bi-directional LSTM baseline for the PNR temporal localization and state change classification.
}{}
\clearpage

\subsection{Audio-Visual Diarization Benchmark}\label{sec:appendix-av}

This section details the Audio-Visual Diarization (AVD) benchmark task definitions, annotations, baseline models, and results.
\srt{As noted in Appendix~\ref{sec:deid-appendix}, 
the AVD benchmark uses only video %
where informed consent for capturing identities is explicitly collected from all participants in the scene, including faces and voice.}

\subsubsection{Motivation} \label{sec:avd-expanded-intro}

Egocentric human perception is driven by inferring useful information from all the primary senses. 
While visuals captured by the eyes are one of the main {\it information channels}, sounds as captured by the ears are equally relevant. 
In particular, for understanding humans' interaction with the environment from the first-person perspective, 
detecting, localizing, tracking (both in 3D space and time) and understanding sounds by combining the necessary acoustic information 
with visual signals becomes even more critical. 
Several psychophysical studies have proven that humans are remarkably good at 
locating where a sound came from in 3D space with respect to their head position \cite{micheyl2008evaluation}. 
Sensitivity of humans to moving sounds in horizontal and vertical planes is also well documented \cite{perrott1990minimum, kolarik2016auditory}.

For a long time, the computer vision community has studyied the problem of precise localization of objects and people, 
robustly tracking and segmenting them using images. 
In this effort, we aim to bring audio (human speech in particular) into the mix. 
Truly audio-visual systems not only enable richer capture and analysis of the environment (and a user's interaction with it), 
but they also help build technologies for visually or acoustically impaired users (e.g., hearing aids, augmented reality). 

The \textbf{goal of this benchmark} is to help advance the state of the art in audio-visual understanding from the egocentric viewpoint.  
Specifically, from a conversational perspective, the benchmark aims to understand \textit{who is talking when, and about what}. 
From a visual perspective, we are also interested in \textit{where the speaker is located}.
Given an egocentric video, the proposed tasks require extracting the spatial location of the speakers, 
their voice activity across the length of the video, and the content of their speech.  

Egocentric data presents several unique attributes to this problem. 
Firstly, sound sources may be visible within all, some, or none of the visual frames, 
depending on their movement within the scene and the movement of the camera wearer. 
Secondly, although the camera wearer is never visible (due the head mounted camera device) they are clearly audible 
and in fact often amplified compared to the other conversation participants due to the closeness to the microphone that captures the video.
Third, natural dynamics in the scene (camera wearer walking, running, rapid changes in head movement etc.) add 
significant blur and distortion to the visual stream---some such noise is structured and relevant for understanding the context and semantic content in the scene.

\subsubsection{Related Audio Visual Learning Work} \label{sec:related-avlearning-work}

There is a recent resurgence of work on audio-visual analysis within and beyond the computer vision community. 
These works tackle various aspects of audio-visual understanding, including source localization, 
cross-modal feature learning, audio spatialization, and audio source separation, as we briefly review next.

On audio-visual detection and tracking, recent works on multimodal learning explore ways to localize sounds in a given video frame~\cite{arandjelovic2017objects,Senocak_2019_PAMI,tian2018audio} and infer spatialized sound from video~\cite{gao2019visualsound,morgadoNIPS18}. 
Capturing and processing multi-channel audio is being studied in audio and microphone array signal processing communities, 
specifically from a user's perspective to understand a given scene \cite{nikunen2014direction, irie2019seeing}. 
Building upon these, it is reasonable to expect that human-centric audio 
has information content that can directly improve visual object categorization and recognition. 
Indeed, this is observed in some recent work where audio disambiguates certain visually ambiguous actions 
\cite{kazakos2019epic,audiovisual-slowfast}. 
For actions and activity, %
audio events can also be directly used to %
perform summarization \cite{arabaci2018multi}.  
In particular, capturing ego-driven actions and activity and separating them from general background actions 
and activity in the scene is critical. 

Alternatively, visual information has been used to disambiguate certain audio tasks like speech transcription. 
Specifically, audio-visual speech recognition has received a lot of attention in the last decade 
with multiple studies suggesting that automatic speech recognition (ASR) could benefit from visuals of the scene, 
or other non-acoustic information \cite{iwano2007audio, afouras2018deep}. 
As shown in here, it is reasonable to expect that lip reading from a first person point of view would also benefit ASR systems. 

In addition, audio-visual cross-modal learning may provide insight and solutions to one of the 
oldest problems in egocentric human communication ecology, referred to as cocktail party problem (CPP).
The essence of CPP is ``How do we recognize what one person is saying when others are speaking at the same time?'' 
Human listeners must  perceptually integrate the simultaneous sounds originating from one person's voice (e.g., harmonics and speech formants) 
and segregate these from the concurrent sounds of other talkers. 
In such situations, humans leverage visual information such as from lip movements to better understand, 
while their auditory system helps with focusing on a particular speaker characteristic while ignoring other speech/noise. 
Recent work on audio-visual diarization~\cite{avdiar} and multimodal source separation from video 
show that CPP and its variations can benefit from visual signals 
\cite{owens2018audio,afouras2018conversation,ephrat2018looking,gao2018objectSounds,zhao2018sound,gao2019coseparation,visual-voice}.

Furthermore, humans are pretty good in understanding the context of a conversation even when words are incomprehensible. 
They are able to fill in the missing details using their context knowledge.
This can be extended to sound sources that are non-humans as well. 
For a more detailed account of CPP please refer to \cite{bee2008cocktail}. 
Fully addressing CPP requires not only identifying and separating the different sound sources in the scene, 
but also understanding the auditory {\it attention} of the camera wearer---in other words, 
which sound source is the user attending to at the moment, or which one may the user want to attend to in the near future.

\subsubsection{Related Datasets and Benchmarks} \label{sec:avd-related-data-bench}

{\bf EPIC-Kitchens:} \cite{Damen2018EPICKITCHENS, damen2020epic}
EPIC-Kitchens is among the most widely known ego-centric dataset with first-person view events and annotations. 
The dataset comprises of multi-faceted, audio-visual, non-scripted recordings in native environments, 
i.e. the wearers' homes, capturing all daily activities in the kitchen over multiple days. 
The dataset is 100 hours, 20M frames, 90K actions in 700 variable-length videos, capturing
long-term unscripted activities in 45 environments using head-mounted cameras.
Annotations are collected using a Pause-and-Talk narration interface.
The dataset is widely used in action recognition, action detection, action anticipation, cross-modal retrieval, 
as well as unsupervised domain adaptation for action recognition. 

{\bf VoxCeleb:} \cite{Nagrani17, Chung18b}
VoxCeleb 1 and 2 comprise recordings of more than $6$K speakers spanning a wide range of different ethnicities, accents, professions, and ages.
The data is non-egocentric and is annotated for active speaker face bounding boxes, face tracks, and anonymous person IDs. 
VoxCeleb 2 in particular is defined for boosting research in speaker recognition, and it contains over a million utterances. 
Videos included in the dataset are shot in a large number of challenging visual and auditory environments. 
These include interviews from red carpets, outdoor stadiums and quiet indoor studios, speeches given to large audiences, 
excerpts from professionally shot multimedia, and even crude videos shot on hand-held devices. 
Audio segments present in the dataset are degraded with background chatter, laughter, overlapping speech and varying room acoustics.

{\bf VoxConverse:} \cite{chung2020spot}
VoxConverse is a related audio-visual diarization dataset consisting of over $50$ hours 
of multi-speaker clips of human speech, extracted from YouTube videos. 
Similar to VoxCeleb, this data is also non-egocentric. 
This dataset was proposed to boost research in speaker diarization for audio-visual inputs. 
A bulk of the data instances are from political debates and news anchors so as to capture 
conversational scenarios with overlapping and interrupting speech. 

{\bf AVA:} \cite{chaudhuri2018ava, roth2019ava}
The AVA spoken activity datasets are AVA speech and AVA active speaker. 
AVA speech is a densely annotated audio-based speech activity collection of AVA 1.0 third-person videos, 
and explicitly labels $3$ background noise conditions, resulting in approximately $46,000$ labeled segments spanning $45$ hours of data.
AVA active speaker associates speaking activity with a visible face, 
resulting in $3.65$ million frames labeled across approximately $39,000$ face tracks. 

{\bf AVDIAR:} \cite{gebru2017a}
The closest egocentric dataset for audio-visual diarization is AVDIAR.
It consists of $23$ staged sequences, with each sequence duration ranging from ten seconds to three minutes 
(a total of $27$ minutes of video). 
Each sequence comprises of 1-4 speakers some standing and some walking around in the visual FOV and having a conversation. 
The capture is done via a head mounted capture on a dummy head.

{\bf EASYCOM:} \cite{donley2021easycom}
EASYCOM is a recent dataset open sourced for the purpose of boosting egocentric audio-visual learning research 
with a focus on multi-channel data and CPP. 
The dataset corresponds to $5$ hours of conversational content with $3-5$ participants in a closed room setting. 
The content involves playing games, ordering food from a menu, and a general discussion on a prespecified list of topics. 
During the recording of the conversations, restaurant-like noise was played on loudspeakers in the room to mimic a real restaurant scene. 
The EASYCOM capture device use glasses with $6$ mics attached to the frame. 
Although rich in terms of multi-channel egocentric acoustic content, 
the setup is constrained in terms of realism, the data is not in the wild, and most importantly the dataset is small. 

\textbf{Existing audio-visual datasets vs. Ego4D:}
Of these existing datasets, EPIC-Kitchens, AVDIAR and EASYCOM are egocentric. 
However, EPIC-Kitchens focuses on solitary activity by the camera wearer, 
and neither the video nor annotations accommodate audio-visual conversation tasks requiring multiple people.
Although EASYCOM contains audio-visual conversation, it is a small dataset containing partly scripted conversations that are not in-the-wild. 
The participants in the sessions also do not move around. 
AVDIAR does include some participants who move around, but the camera wearer is a dummy head and, similar to 
EASYCOM, the data is not in-the-wild (sessions all are done in the same environment/scene). 
Ego4D accounts for all these aspects. 
Lastly, in contrast to VoxCeleb, VoxConverse and AVA, 
Ego4D offers first-person video and its conversation videos take place in casual daily-life environments with multiple speakers.

\subsubsection{Tasks: Definition and Annotations} \label{sec:avd-extend-tasks}

Here we detail the task definitions, the corresponding annotations, and the evaluation metrics. 
We propose a suite of tasks for the Audio-Visual Diarization (AVD) benchmark. 
These tasks are abbreviated as: 
\textit{Localization \& Tracking}, \textit{Active Speaker Detection}, \textit{Diarization} and \textit{Transcription}. 
These tasks jointly capture who is talking when, to whom, and about what in a given egocentric conversational scene. 
Observe that these tasks are implicitly tied to each other; each subsequent task is driven in some form by a previous task 
(as further clarified in the task descriptions below).\footnote{ 
Note that although speech transcription and source localization are distinct from audio-only speaker diarization---
all of which are well defined research paradigms in mainstream audio, speech and vision community---
we cumulatively refer to all these together under the umbrella of audio-visual diarization for Ego4D.} 

\paragraph{Task 1: Localization \& Tracking: 
Where is the person in the visual field of view?}

This first task in AVD captures the spatial position of all the probable speakers in the scene, 
from the point of view of the camera wearer. 
The goal of the task is to compute bounding boxes for them. 
Unlike classical face detection benchmarks, this task is challenging in the sense that
the dynamics of the camera wearer's head (coming from natural conversations) 
leads to significant movement in a speaker's apparent spatial location. 

{\it Annotations:}
For each speaker present in the $5$ min clip a bounding box is provided. 
Each frame of the video is annotated for the task. %
We first utilized a face detection and tracking model to estimate these bounding boxes, 
and then a team of human annotators validated and corrected these machine-generated boxes to improve annotation quality. 
A bounding box is considered a valid human annotation if it captures $80\%$ of the speaker's face; we peform a quality check steup to ensure this. 
Sideways looking faces are also annotated. 
Note that speakers who are very far from the camera wearer (oftentimes several meters away in the scene) 
and who do not come into conversational contact with the wearer are not annotated. 

{\it Evaluation:}
Recall that the goal of the task is to localize {\it as well as} track the speakers in the scene. 
Hence the evaluation metrics proposed account for the accuracy of trajectory of detected bounding boxes.
We follow the standard multiple object tracking (MOT) metrics to quantify the speaker tracking results. 
There are many different MOT metrics, in which we are most interested in
the MOTA in the CLEARMOT metrics \cite{CLEARMOT}, and IDF1, IDP, IDR in the Identity metrics \cite{IdentityMOT}. 
MOTA, the multiple obtect tracking accuracy, is a combined metric of false alarms, false positives and identity switches. 
MOTA is based on matching the tracking results with the ground truth 
at frame level, while the IDP (ID precision), IDR (ID Recall) and IDF1 (ID F1 score) are 
based on the tracking result to ground truth matching at the trajectory level. 
ID metrics give a tracker's performance on maintaining correct identification for each target.

\paragraph{Task 2: Active Speaker Detection: 
Who is speaking? }

The next task in AVD is to detect the active speaker in the scene. 
This task is in principle similar to active speaker detection---where the goal is to detect which of the visible people in the scene are speaking at a given time \cite{roth2019ava}. 
It builds on top of the previous localization and tracking task 
to recognize each of the speakers whose face bounding boxes are detected.
Hence, this task does not take into account speakers who are not visible in the camera's FOV. 
Note that active speaker detection is also an important aspect of speaker diarization (which is the next task in the benchmark). 

{\it Annotations:}
We provide an anonymous speaker label (e.g., speaker $1$, $2$ etc.) for each speaker visible in the clip. 
The camera wearer is assigned the label $C$. 
This is done by utilizing the face bounding box tracks annotations and labeling each track one at a time. 
Hence, each face track gets assigned one unique label, and multiple tracks within a single clip may share the same label 
(corresponding to the same speaker). 
However, the labels are clip-specific, i.e., a speaker who may be present across multiple clips 
does not get assigned a shared unique label across the clips. 
Again, speakers who are never in the visual FoV are not assigned a label. 

{\it Evaluation:}
We use the object detection mAP to quantify the active speaker detection result.
This is \srt{a} frame-wise metric. 
In a video frame, if the \srt{intersection over union} (IoU) between a detected face bounding box and the ground truth face
bounding box exceeds a predefined threshold, i.e. 0.5, we have a positive face detection. 
Each detection has an associated class to indicate whether it corresponds to an active speaker. 
Active speaker detection methods give a confidence score of the active speaker class for each detected face bounding box \cite{tao2021someone}. 

\srt{{\it Camera Wearer's Voice Activity Detection}:}
Note that the camera wearer's face is never visible in the camera's field of view, 
and so they do not have any face tracks associated with them.
However, in many cases, they are the dominant speakers. %
This is mainly because they are driving the interactions in many cases, %
and since their mouths are the closest to the microphones, 
their voice is in general amplified in the audio stream compared to other speakers. 
We propose to also consider them as active speakers and detect their voice. 
We use the object classification mAP to quantify the result of the camera wearer's voice activity detection.

\paragraph{Task 3: Diarization:  
Who spoke when?}

\srt{This next task further expands on the temporal aspect of active speaker detection (from the previous task).}
Given the set of speakers and their spatial localization in the visual field of view, 
this task aims to capture the voice activity of the speakers. 
It is identical to speaker diarization, a well studied research problem in the speech and audio domains \cite{anguera2012speaker, park2021review}
and answers the question, ``who spoke when". 
While speech from speakers that overlap with each other is one of the biggest issues to solve in this task, 
the egocentric perspective adds more complexity in terms of head motions and other dynamics associated with natural conversations. 
Note that the outputs of active speaker detection (the earlier task in the benchmark) also drive this task. 

{\it Annotations:}
For every active speaker label (where the annotations are from the previous Active Speaker Detection task), 
a human annotator marks the start and end time of that person speaking.
We account for overlapping speech segments where multiple speakers talk over each other, 
but we ignore speech not relevant to the conversation such as background speech from a TV or speech further away from the camera wearer. 
Note that speech segments from the camera wearer are also annotated.
The annotators rely both on the audio and the visual stream for creating these labels. 

{\it Evaluation:}
Diarization error rate (DER) is the {\it de facto} evaluation metric for speaker diarization \cite{anguera2006robust}, 
and it is well studied in the audio and speech processing community. 
DER measures the fraction of total time (in a given clip) that is not attributed correctly to a speaker or to non-speech. 
It is defined as follows:
\begin{equation}
    \textrm{DER (\%)} = (E_{miss} + E_{fa} + E_{spk}) \times 100,
\end{equation}
where $E_{miss}$ denotes the fraction of time that has been predicted to be non-speech 
while that segment is attributed to a speaker in the reference. 
$E_{fa}$ denotes the fraction of time that has been predicted to be associated with a speaker, 
but is actually labelled as non-speech in the reference, and $E_{spk}$ denotes the fraction of time where speech is associated with the wrong speaker. 
\srt{All errors are computed as a fraction of the total amount of speech.}

\paragraph{Task 4: Transcription: 
What did the speaker say?}

The final task of AVD is to transcribe the speech of each speaker, i.e., performing ASR. 
Similar to the diarization task, some of the challenges associated with the transcription task include overlapping speech and environmental noise.
In addition, the camera wearer's head movement results in a significant change of the audio volume of the speech recorded from others. 

{\it Annotations:}
Since the clips contain multiple speakers with overlapping speech segments and with different volumes, the final transcriptions are obtained in multiple passes. In the first pass, initial human annotations based on voice segments are merged with automatic annotations for regions with low volume. In a subsequent pass, human annotators had to correct and assign segments of transcriptions to the corresponding voice activity segments per speaker while also annotating overlapping speech. Note that annotators had both the audio and video available for annotation and, besides spoken words,  the occurrence of other artifacts such as unintelligible speech or incomplete words have also been annotated. 
The final transcription annotations for a clip consist of a sequence of segments labeled with begin time, end time, transcript and speaker ID within the clip. In evaluations, we applied ASR to these segments individually and computed the performance over all of these segments. Please note that the time segments associated with the transcripts are not the same as the ones used in diarization because we separately annotated the overlapping regions here to reduce transcription errors and account for speakers talking in low volume. 
This allows us to also distinguish voice activity from speech activity. In addition, the use of time-segmented transcriptions is also slightly different from standard ASR datasets in speech community which mainly have text and no timestamps.

{\it Evaluation:}
We utilize the Word Error Rate (WER), a standard ASR metric, for evaluating this task \cite{klakow2002testing}. 
First, the minimum edit or Levenshtein distance is computed between 
the reference and hypothesized transcription. 
WER then measures the ratio of the number of word substitutions ($S$), deletions ($D$) and insertions ($I$), 
i.e. the total number of edits necessary to convert the hypothesized transcription into the reference relative to the total number of words ($N_w$) in the reference:
\begin{align}
     \textrm{WER (\%)} = \frac{S + D + I}{N_w}\times 100.
\end{align}

\subsubsection{Data Statistics} \label{sec:avd-datastat}

From across the \numhours hours of video in Ego4D, approximately $764$ hours of data contains conversational content, 
and are directly relevant for the AVD and Social benchmarks. 
Please refer to Section \ref{appendix:social-collection} for a complete description of
the experimental design and scenarios used in these sessions. 
From this set, a randomly chosen subset of $572$ clips (each $5$ minutes long) are annotated for this first version release. 
Of these $572$ clips, $389$ clips are marked for training, $50$ clips for validation, and the remainder is the testing set.

\begin{figure*}[!tbh]
    \centering
      \includegraphics[width=0.31\linewidth]{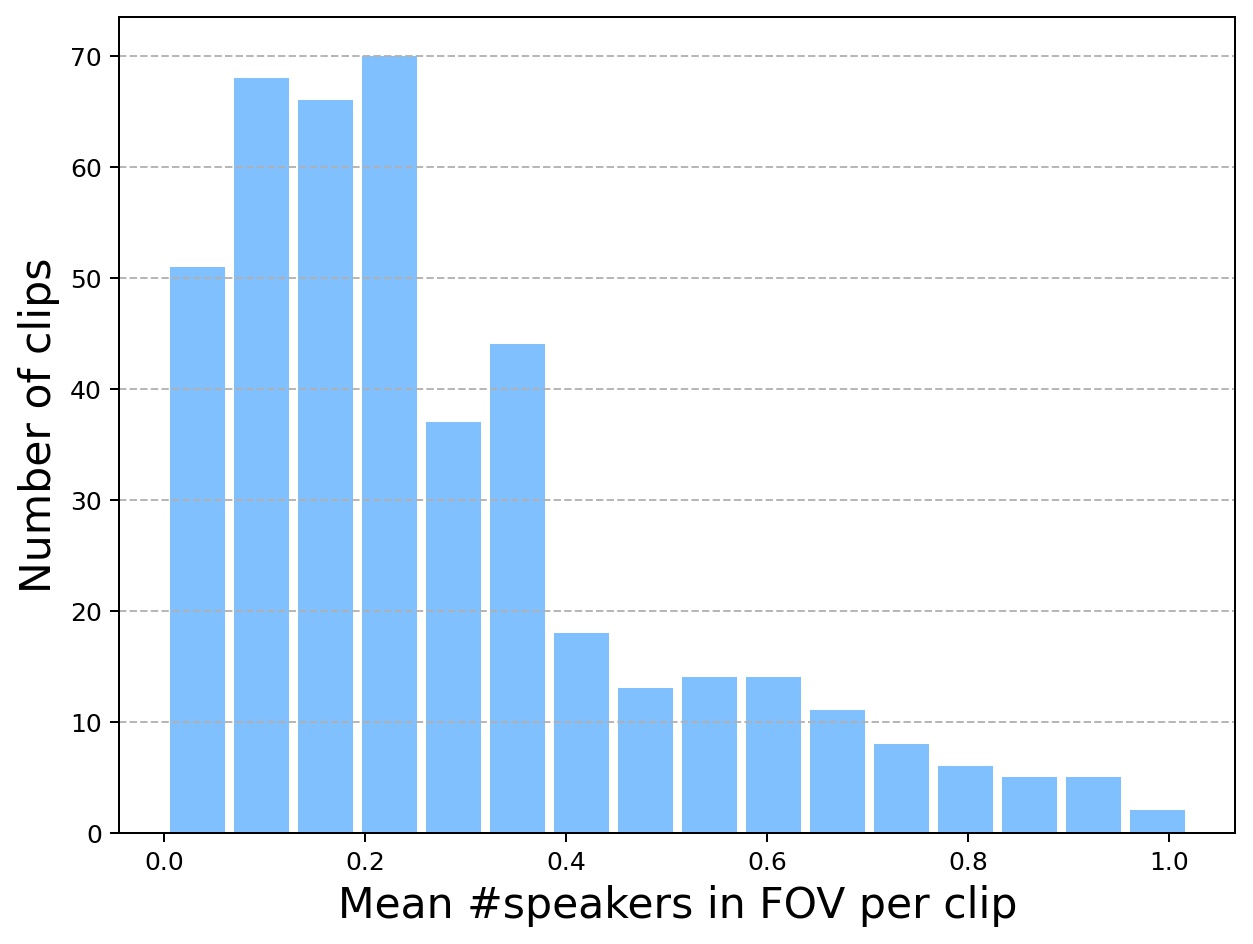}
      \includegraphics[width=0.31\linewidth]{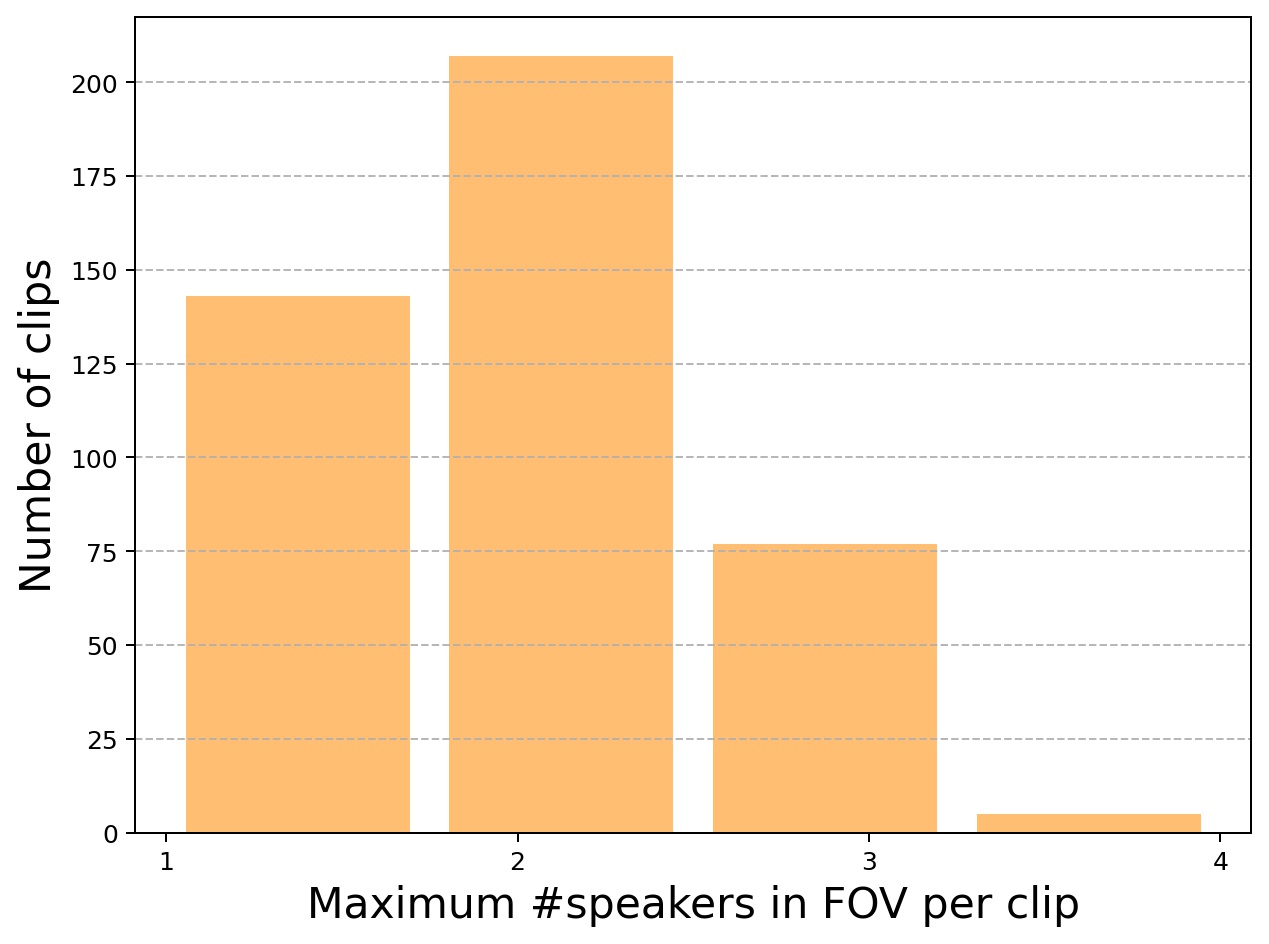}
      \includegraphics[width=0.31\linewidth]{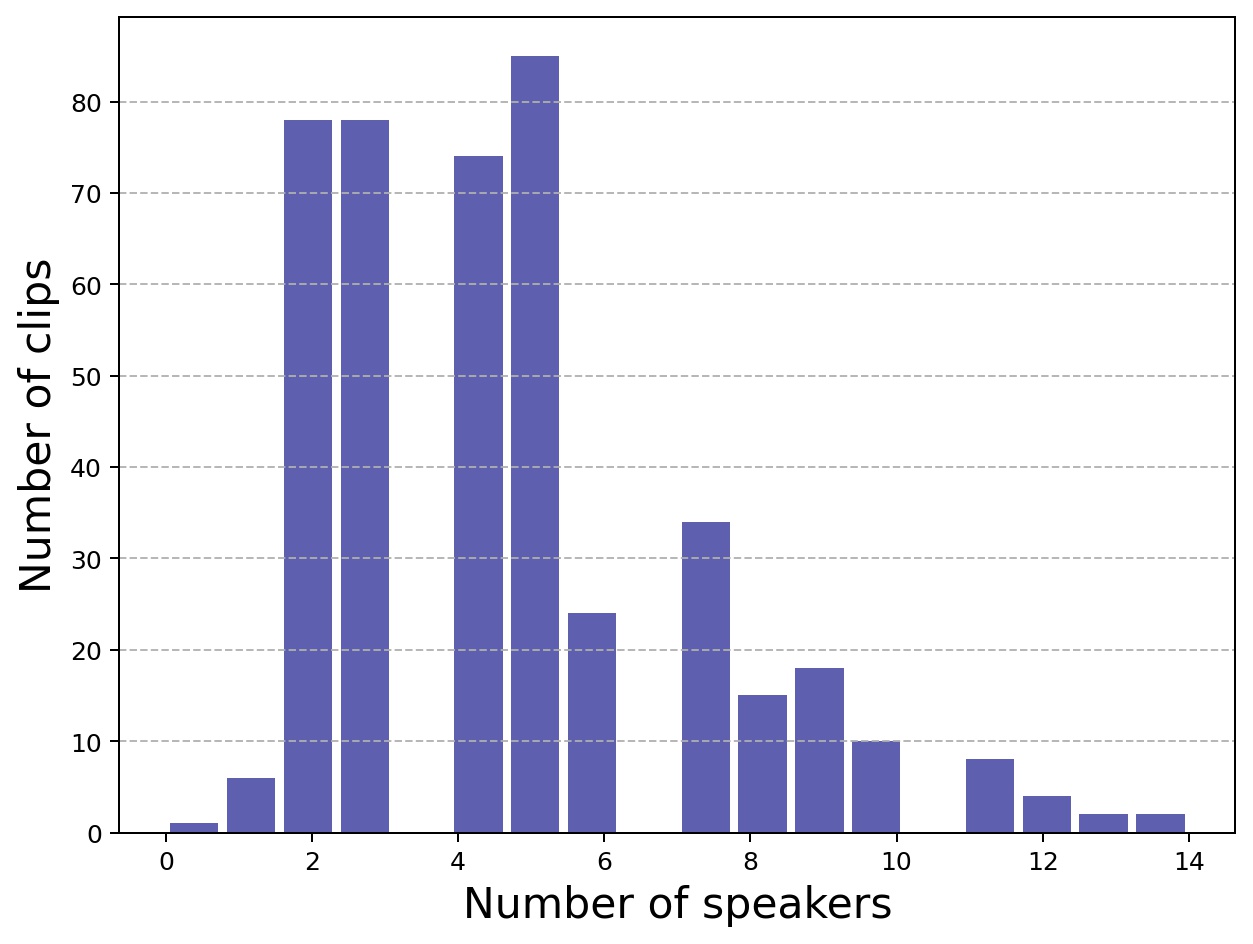}
  \caption{\label{fig:AVD-stat} AV Diarization data statistics. Mean and maximum number speakers in FOV, and number speakers per clip.}
\end{figure*}

Table \ref{tab:AVD-stat} and Figure \ref{fig:AVD-stat}  summarize  statistics about the speaker content from across these clips. 
Observe the long tails of mean and maximum number of speakers in the dataset.  
We note that in the first version of the data release, due to the fact that the total number of clips is relatively small, 
the test and/or validation batches may be biased in terms of changes in speakers' accents, 
changes in vocabulary usage (since the participants are from different cultural backgrounds from across the world), 
and in general changes in nature of interactiveness between speakers in a scene.  
There is marginal distributional shift among the 
training, testing and validation splits. 
This is mainly because of the smaller number of annotations in this version of AVD for Ego4D. We expect these distributional shifts to be less significant in future releases and as more data will be annotated.

\begin{table}[!h]
    \centering
    \begin{tabular}{|l|r|}
		\toprule
		Statistic (Avg.) & Value \\
    	\hline
		Speakers per clip & 4.71  \\
		Speakers per frame &  0.74 \\
		Speaking time in clip & 219.81 sec \\
		Speaking time per person in clip & 43.29 sec \\
		Camera wearer speaking time & 77.64 sec \\
		\bottomrule
	\end{tabular}
\caption{\label{tab:AVD-stat} AVD Data Statistics.}
\end{table}

\begin{figure*}[!tbh]
      \includegraphics[width=\textwidth]{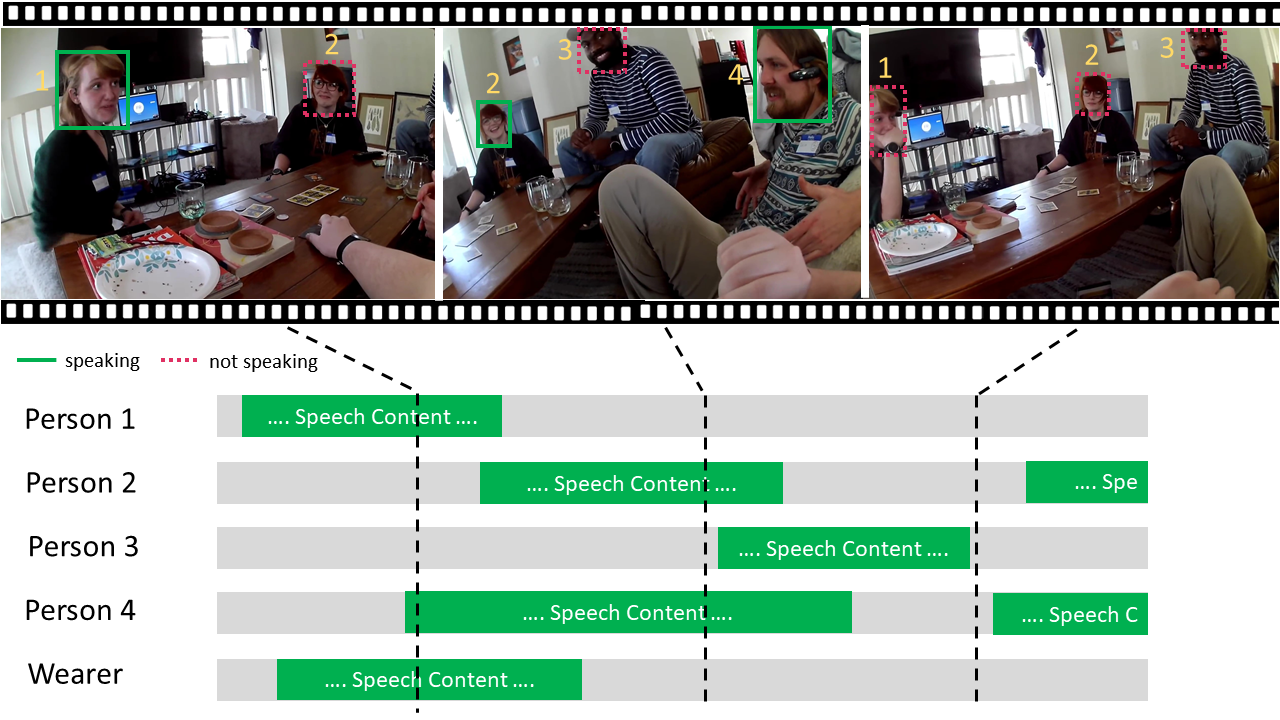}
  \caption{\label{fig:AVD-summary} AV Diarization benchmark annotations summary. The four tasks are annotated in a sequential fashion, starting with localization and tracking of speakers, active speaker detection labels, diarization time stamps, and finally transcriptions. 
  The figure shows the face detections (highlighted by bounding boxes), 
  speaker detection (shown by the anonymous person IDs 1, 2, etc.), active speaker (highlighted in green) 
  and voice activity (shown below in green highlighted time segments). 
  \srt{Speakers in the visual FOV who are not talking are highlighted in dotted red boxes.} 
  The clips used for AVD (and Social Interaction) have consent from participants to leave their faces unblurred.}
\end{figure*}

\begin{figure*}[!tbh]
      \includegraphics[width=\textwidth]{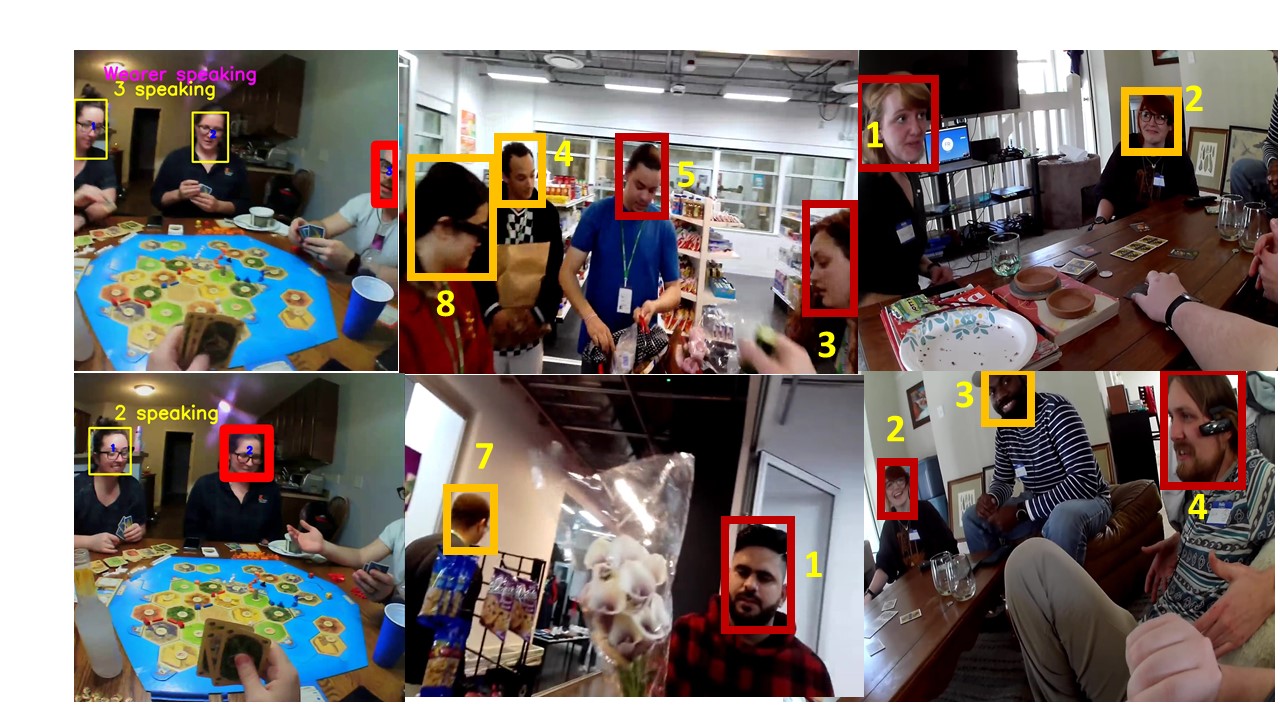}
  \caption{\label{fig:AVD-summary} Example annotations showing the face detections (highlighted by bounding boxes), 
  speaker detection (shown by the anonymous person IDs 1, 2, etc.), active speaker (highlighted in red) 
  and voice activity (shown below in blue highlighted time segments).  
  \srt{As illustrated here, the data for AVD includes people walking around and talking, sitting and playing games etc.}
  The clips used for AVD have consent from participants to leave their faces unblurred.}
\end{figure*}

\subsubsection{Baseline Modeling Framework} \label{par:avd_baseline}

Recall that the $4$-part tasks in this benchmark are tied to each other, 
in the sense that representations learned from one task may be relevant for the others. 
To that end, we propose a baseline learning framework that addresses each task in a sequential fashion. 
The framework includes the following steps:
\begin{itemize}[noitemsep]
    \item We first detect people's heads and do short term tracking in the video. 
    The short term tracker follows each detected head by expanding a set of trajectories based on their positions, 
    sizes and the appearance of the person. 
    The trajectories may end when occlusion happens or when the tracked person goes out of the field of view. 
    New trajectories can also be added to the trajectory set.
    
    \item The short term tracker's trajectory for each person is often fragmented into  multiple parts. 
    Hence, we then optimize the grouping of the tracklets in step one so that 
    the trajectories of each person can be linked together. 
    We formulate the problem as a constrained combinatorial optimization problem. 
    Integer programming can be used to solve the problem directly but it has exponential complexity. 
    For efficiency, we develop a greedy approach which is much faster and still gives strong results.
    
    \item We then classify each person/head in each video frame as an active speaker or not. 
    Based on the classification result and the corresponding detected long-term trajectories, 
    we further associate the audio/speech to each person in the video.
    We use this preliminary list of audio \srt{feature} embeddings to further extract and match un-associated audio segments to speaker labels. 
        
    \item We then \srt{use two methods to detect the camera wearer's voice activity}. 
    \srt{The first method uses} high energy audio segment in the clip 
        (under the assumption that their voice has natural amplification compared to the remaining speakers).
      \srt{The second method is 
        a deep classifier that predicts whether the wearer is speaking.} 
    \item \srt{Lastly, we applied ASR to the speech regions based on the ground truth segmentation and evaluated the WER across all segments. 
    Evaluating the system by using another segmentation method is challenging especially in the case of overlapping speech segments. 
    Jointly modeling time segments and transcriptions will be a challenging problem (as we discuss in Section \ref{sec:avd-discussion}).}

\end{itemize}

We describe further details about each of these steps below, 
and Tables \ref{tab:AVD-tracking}--\ref{tab:AVD-transcription} summarize the resulting performance metrics for the tasks.

\paragraph{Audio Only Models for Speaker Diarization} %

The problem of speaker diarization from audio has been studied to a considerable extent in the 
field of speech processing \cite{anguera2012speaker, park2021review}. 
For the audio-only baseline system, the VBx diarization approach has been utilized \cite{landini2022bayesian} 
for having shown superior results on different types of datasets such as CALLHOME \cite{NISTSRE2000evalplan} (telephone conversations), 
AMI \cite{carletta2005ami} (meetings) and DIHARD II \cite{DIHARD19} (myriad of domains ranging from audiobooks to YouTube videos). 
This method requires speech activity regions and these were obtained using the ASpIRE model based on a 
time delay neural network (TDNN) with statistics pooling, available %
with the Kaldi speech recognition toolkit \cite{Povey_ASRU2011}. 
We refer to this as kVAD (the Kaldi VAD model). 
Although this kVAD has been trained 
on slightly different data (telephone conversations), 
and thus does not provide the best possible results, it has been chosen for the baseline system  because of its general availability. 

The speech activity regions are uniformly segmented to obtain shorter segments and speaker embeddings 
(so-called x-vectors \cite{snyder2018x}) are extracted one per subsegment. 
The x-vectors are obtained with a ResNet101 extractor \cite{he2016deep} trained to produce speaker-discriminative embeddings. 
The input to the network are log Mel-filter bank features every 10~ms, and given a segment of speech, it computes a single 256 dimensional vector that represents the whole segment. 
The information of the whole segment is aggregated with a statistical pooling layer which computes the mean and standard deviation of activations over the time domain. A linear transformation is then used to reduce the dimensionality to 256. 
The training data consisted of VoxCeleb1 \cite{Nagrani17}, VoxCeleb2 \cite{Chung18b} and CN-CELEB \cite{fan2020cn} together, totalling 2877 hours of speech from 8178 speakers.

The x-vectors are initially clustered to a few dozens of classes using agglomerative hierarchical clustering. This initial clustering is fed as initialization to a Bayesian hidden Markov model which estimates altogether the number of speakers in the recording as well as the assignment of x-vectors to the states. 
Each state in the model corresponds to one speaker and the probability of observing a particular x-vector in a particular state 
can be interpreted as the corresponding speaker producing the corresponding segment of speech. 
\srt{The most relevant hyperparameters of the model were fine-tuned to obtain the best DER performance on the Ego4D validation set.}
The VBx implementation published by Brno University of Technology is publicly available %
as well as the training recipe published by Phonexia Research. %

\paragraph{Short-term People Tracking} 

The goal here is to track people's faces. 
However, our method can also be used to track the whole body of each person. 
The short-term tracker maintains a set of trajectories. 
The trajectories include the attributes such as the person-ID, the frames tracked, a life counter, the appearance
features and the positions of the tracked bounding boxes.  Throughout, we use the term ``person-ID" to refer to an anonmyous tag for a person in the video (person 1, person 2, etc.); no actual identities are available in the data, and the benchmark does not aim to perform any person identification.
There are two kinds of trajectories. 
If a trajectory's tracked frames are less than a threshold, e.g. 5, it is in probation and is not counted as
a real trajectory even though we maintain all the information for them. 
When a trajectory's tracked frames are greater than the threshold,
it becomes a real trajectory. Each trajectory also has a life span.
The life of a new trajectory starts from a fixed value. 
The life of a trajectory is restored to a fixed maximum value, such as 10, 
if the trajectory is matched to a candidate person head bounding boxes. 
Otherwise, the trajectory goes into a maintenance mode and its life decreases by 1 each time it fails to find a match. 
If the life of a trajectory goes to 0, it is removed from the trajectory set.

The key component of the short-term tracker is matching trajectories
to the candidate head bounding boxes in each frame. This can be formulated
as the following optimization problem:

\begin{align}
	&\min\sum_{(i,j)}c_{i,j}x_{i,j} \\
	\mbox{s.t.  } & x_{i,j} \mbox{ forms a max-matching},\nonumber \\ 
	     & x_{i,j}=0, \mbox{ if } (i,j)\in E,\nonumber\\
	& x_{i,j}=0,1, \nonumber
\end{align}	
where $x_{i,j}$ is 1 if trajectory $i$ matches candidate head box $j$ and 0 otherwise. 
$E$ is a set in which the pairs of trajectory and candidate cannot match each other, 
examples include cases such as the candidate is too far away, the size is too different or the appearance does not match. 
$c_{i,j}$ is the cost of matching trajectory $i$ and candidate head detection $j$. 
This cost of matching, $c_{i,j}$, is computed as a linear combination of the normalized bounding box distances 
and the difference of the appearance features. 
The normalized bounding box distance is defined
as the ratio of the Euclidean distance between the two corners of
the last bounding box in the trajectory and the detected head bounding box in the image to the size of the detected bounding box. 
Each trajectory also maintains a feature vector to characterize the most recent appearance of the tracked person. 
This feature vector is obtained from a \srt{feature} embedding network trained on a large person head dataset.

This optimization problem can be solved efficiently using the Hungarian algorithm or the primal dual algorithm. 
Due to the imperfect features, the optimization may have an identity switching problem if two targets cross paths. 
To solve the problem, we enforce the longer trajectories to have higher priority to match. 
We use a two-step matching scheme. 
We first match all the trajectories that are longer than a specific threshold chosen empirically. 
Once done, we then match the shorter trajectories. 
This scheme naturally gives higher priority to longer trajectories, thereby reducing mismatches among them. 
This is more robust than a single stage matching where all trajectories are handled together. 

In our implementation, the person detector is a Yolo-V3 detector~\cite{redmon2018yolov3} which
detects the head and person bounding box simultaneously. The detector
is trained on images from the Google OpenImage dataset~\cite{kuznetsova2020open} and
a fisheye image dataset~\cite{fu2019datasets}. We use the detected head bounding boxes for
people tracking. The person head appearance's feature is extracted
using the person embedding network, which is trained on the VoxCeleb2
dataset using the triplet loss. The network has the structure of a ResNet-18.

\paragraph{Long-term Tracking by Trajectory Matching} %

The short term tracker generates fragmented person trajectories. If
a person is occluded or goes out of the field of view and reappears,
it will receive a new ID. The fragmented trajectories are referred to
as tracklets. We need to group the tracklets throughout the whole
video to generate the final trajectories for each person. The grouping
problem can be formulated as follows:

\begin{align}
	& \min\sum_{m,n}D_{m,n}y_{m,n} \\
	\mbox{s.t.  } & y_{m,n}=y_{n,m},\forall m,n, \nonumber \\
	& y_{m,k}+y_{k,n} \le 1+y_{m,n},\forall m,n, \nonumber \\
        & y_{m,n}=0, \mbox{if m and n overlap in time or } D_{m,n}>g, \nonumber \\
        & y_{m,n} \mbox{ is binary }, \nonumber
\end{align}	
where $y_{m,n}=1$ if tracklet $m$ and $n$ can be grouped together and otherwise $y_{m,n}=0$. 
$D_{m,n}$ is the appearance distance between the trackelet $m$ and $n$ and $g$ is a threshold. 
Here $D_{m,n}=\min_{\{i\in T_{m},j\in T_{n}\}}||f_{i}-f_{j}||^{2}$, 
where $T_{i}$ is the \srt{set of} person head boxes in tracklet $i$ and $f_{i}$ is the corresponding feature embedding. 
The constraints require the grouping to be reflective: if \srt{tracklet} $m$ and $n$ can be grouped together so can $n$ and $m$,
transitive: if $m$ and $k$ can be grouped together and so can $k$ and $n$, then $m$ and $n$ can be grouped together. 
Two tracklets cannot be grouped together if they have time overlap or their distance is greater than a threshold $g$. 
The optimization can be solved using integer programming. However, this method has exponential complexity. 
We propose a fast greedy algorithm to solve the problem.

The greedy algorithm starts by treating each initial tracklet as a
trajectory and progressively groups two trajectories with the closest
$D$ until no trajectories can be grouped together. 
Since the distance between two trajectories can be computed by finding the minimum of 
all the ``element'' tracklet pair distances, 
the merging procedure is efficient if we pre-compute and cache the element pair distance.
This greedy approach gives strong results while maintaining low complexity. 

\begin{algorithm*}
\caption{Greedy Tracklet Grouping}
\begin{algorithmic}
      \State Initialize sets $P$=\{$S_{1},$$S_{2}$, ..., $S_{N}$\}, where
      $S_{i}=\{T_{i}\}$, $T_{i}$ is the tracklet $i$ and $N$ is the
      number of tracklets.
      
      \For {(m, n), m=1..N and n=1..N}
          \State compute $D(m,n)$
      \EndFor
      
      \While{True}
      
          \For { ($S_{m},S_{n})$, $S_{m}\in P$ and $S_{n}\in P$,
      and ($S_{m},S_{n})$ do not have time conflict} 
                \State compute $F(S_{m},S_{n})=\min_{T_{a}\in S_{n},T_{b}\in S_{m}}D(a,b)$
          \EndFor
      
          \State $(m^{*},n^{*})$ = argmin($F(S_{m},S_{n}))$
      
          \If{$(m^{*},n^{*})$ is empty or $F(S_{m^{*}},S_{n^{*}})>g$}
              break
	  \EndIf    
      
          \State $S_{m^{*}}=S_{m^{*}} \cup S_{n^{*}}$ and P.pop($S_{n^{*}}$)
      \EndWhile	  
      
      \State P includes the grounded trajectories
\end{algorithmic}
\end{algorithm*}

The algorithm reduces to the minimum spanning tree method if there is conflict
between each pair of trajectories. However, if there are time-conflicting
tracklets, there is no guarantee the greedy algorithm gives the
globally optimal solution. \cc{We illustrate the method 
through a simple example: Assume there are trackelets \{$T_1$,$T_2$,$T_3$,$T_4$\}, 
$T_1$ and $T_2$ have time conflict, and
$T_3$ and $T_4$ have time conflict. D($T_1$,$T_3$) = 10, D($T_2$,$T_4$) = 1, D($T_1$,$T_4$) = 3 and D($T_2$,$T_3$) = 4.
We assume $g=20$.
Using the proposed greedy method, the solution P is \{\{$T_2$,$T_4$\},\{$T_1$,$T_3$\}\}
whose overall cost is 11. However, the optimal solution is \{\{$T_1$,$T_4$\},\{$T_2$,$T_3$\}\}
whose overall cost is 7. Even though the greedy method does not guarantee
the global optimal solution, empirically we observe that the proposed method give
strong results. In fact, if the person embedding is accurate, these
corner cases would probably never occur and the greedy result would approach the 
globally optimal solution.}

Table \ref{tab:AVD-tracking} summarizes the tracking metrics MOTA, MOTP, IDF1, IDR, and IDP on the \srt{validation and} test sets.
\begin{table}[t]
    \centering
    \begin{tabular}{|l|r|r|}
		\toprule
		Metric & Valid & Test \\
		\hline
		MOTA &  74.52 & 71.94 \\
		MOTP &  79.07 & 79.17  \\
		IDF1 &  84.92 & 80.07 \\
		IDR  &  80.40 & 73.52 \\
		IDP  &  89.97 & 87.90 \\
		\bottomrule
	\end{tabular}
\caption{\label{tab:AVD-tracking} Localization and tracking baseline metrics \srt{on the 
validation and the test sets respectively}. }
\end{table}

\paragraph{Active Speaker Detection:} %

We use two approaches for active speaker detection. One approach is based on mouth region classification, 
and the second method is a transformer based audio-visual method for active speaker detection \cite{tao2021someone}.

\noindent \textbf{RegionCls:}
Our first approach is based on the classification of mouth regions.
It first computes the 3D head orientation using a regression
network. In our implementation, the $z$ direction is into the image; if the head 3D
orientation $z$ coordinate on the unit sphere is greater than 0.3, we
assume the face is away from the camera. If the face is facing away from the camera, we ignore the
image and the active speaker detection result is set to null.  For faces looking at the camera,
our method first regresses the facial key points using the image within
the person's head bounding box. We use the mouth key points to crop out the mouth image. 
The cropped mouth image is then sent to a classification network to classify whether the speaker is talking or not. 

Note that we also explored using multiple images, wherein
we stack a short sequence of cropped mouth images in a time interval
for active speaker classification. Our experiments show the multiple
mouth images input do not significantly improve the result. This is
probably due to the fast movement of the camera and sometimes difficult
angles of the face. This causes inaccurate cropped mouth regions.

\begin{figure*}[!tbh]
    \centering
      \includegraphics[width=0.75\linewidth]{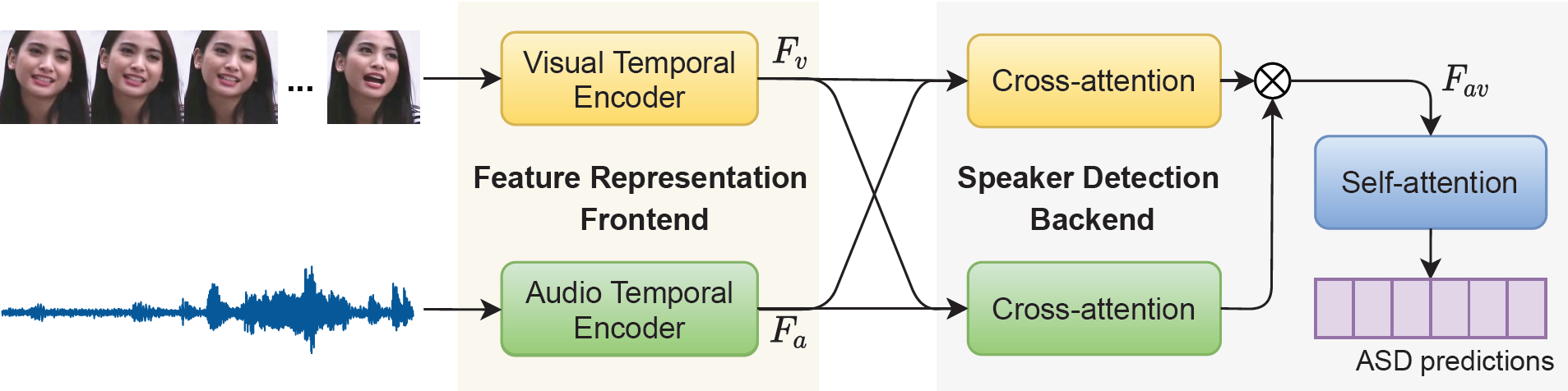}
  \caption{\label{fig:talknet-arch} TalkNet: An audio-visual temporal network for detecting and tracking the active speaker in a video \cite{tao2021someone}.  Figure is from~\cite{tao2021someone}.}
\end{figure*}

\noindent \textbf{TalkNet:} \cite{tao2021someone}
TalkNet is an end-to-end pipeline that takes the cropped face video
and corresponding audio as input, and decides if the person is speaking in each video frame. 
It consists of a feature representation frontend and a speaker detection backend classifier, as illustrated in Figure \ref{fig:talknet-arch}. 
The frontend contains an audio temporal encoder and a video temporal encoder. 
They encode the frame-based input audio and video signals into the time sequence of audio and video embeddings, representing temporal context. 
The backend classifier consists of an inter-modality cross-attention mechanism to dynamically align audio and visual content, 
and a self-attention mechanism to observe speaking activities from the temporal context at the utterance level.

Tables \ref{tab:AVD-activedetection1}\srt{, \ref{tab:AVD-activedetection2}, \ref{tab:AVD-activedetection1:val} and \ref{tab:AVD-activedetection2:val}} summarize the resulting performance. 
For each of the two proposed baseline models, 
we report performance summaries with pretraining based on AVA \srt{ and also models
trained using only videos from the Ego4D training dataset.}
Note that the video-only approach can be combined with any voice activity detection to remove false alarms. Here we use such an algorithm from  \cite{SileroVAD}, and we refer to this as sVAD
This can greatly improve the active speaker detection results. 
The max-filtering has a window size of 11. 
TalkNet also has a built-in smoothness filtering to post-process the raw classification result.
\begin{table}[!h]
    \centering
    \begin{tabular}{|l|r|}
		\toprule
		Model & mAP@0.5 \\
		\hline
		RegCls w/o smoothing & 29.68 \\
		RegCls + max-filtering & 31.95 \\
		RegCls + max-filtering + sVAD & 33.72   \\
		TalkNet & 34.75   \\
		TalkNet + sVAD & 34.56 \\
    	Always Speak & 24.46  \\		
		\bottomrule
	\end{tabular}
\caption{\label{tab:AVD-activedetection1} Active speaker detection baseline metrics \srt{on the test set} with pre-training using AVA.
 In Always Speak, all the detected faces are classified as active speakers.}
\end{table}

\begin{table}[!h]
    \centering
    \begin{tabular}{|l|r|}
		\toprule
		Model & mAP@0.5\\
		\hline
		RegCls w/o smoothing & 29.65 \\
		RegCls + max-filtering & 32.77 \\
		RegCls + max-filtering + sVAD & 34.35   \\
		TalkNet & 50.90   \\
		TalkNet + sVAD & 49.66 \\ 
    	\bottomrule
	\end{tabular}
\caption{\label{tab:AVD-activedetection2} \srt{Active speaker detection baseline metrics
on the test set using training videos in the Ego4D dataset.}}
\end{table}

\begin{table}[!h]
    \centering
    \begin{tabular}{|l|r|}
		\toprule
		Model & mAP@0.5 \\
		\hline
		RegCls w/o smoothing & 22.09 \\
		RegCls + max-filtering & 22.88 \\
		RegCls + max-filtering + sVAD & 25.53   \\
		TalkNet & 34.36   \\
		TalkNet + sVAD & 34.65 \\
    	Always Speak & 20.94  \\		
		\bottomrule
	\end{tabular}
\caption{\label{tab:AVD-activedetection1:val} \srt{Active speaker detection baseline metrics on the validation set with models trained on AVA dataset. In Always Speak, all the detected faces are classified as active speakers.}
}
\end{table}

\begin{table}[!h]
    \centering
    \begin{tabular}{|l|r|}
		\toprule
		Model & mAP@0.5\\
		\hline
		RegCls w/o smoothing & 20.33 \\
		RegCls + max-filtering & 21.93 \\
		RegCls + max-filtering + sVAD & 24.60   \\
		TalkNet & 51.04  \\
		TalkNet + sVAD & 50.58 \\
    	\bottomrule
	\end{tabular}
\caption{\label{tab:AVD-activedetection2:val} \srt{Active speaker detection baseline metrics on the validation set using training videos in the Ego4D dataset.}}
\end{table}

\paragraph{Matching Speakers Outside FoV:}
Based on the tracked heads and the active speaker detection results,
we can associate the audio to the visible people in the scene. However, this
is still not complete because there are cases in which the speaker is outside of the visual field of view. To solve this problem, we first create an audio-signature for each visible person in the video. 

We extract one second of audio centered at each video frame time instant. 
If the audio corresponds to a speaking head in the image,
we compute the audio embedding of the one second audio and insert the feature into the audio signature library of the person. 
The audio embeddings can be obtained from any speech representation learning methods. 
\srt{We explored several models including a modified
ResNet18 which takes audio spectrogram logarithm magnitude in one-second windows as the input and trained on the VoxCeleb2 dataset using triplet loss, 
and a version of wav2vec 2.0 \cite{baevski2020wav2vec}---a self-supervised approach to speech representation learning.}

We parse the video and find instants when a particular person is not
in the video frame and match the audio embedding to the person's audio signature library. 
We find the minimum distance of this audio embedding to all the signature audio embeddings in the library. 
If the distance is less than a predefined threshold, we classify the person as speaking and otherwise not.  
 Note that the audio embedding is used only within the same 5 minute video clip and never across video clips. 
 Person IDs are always anonymous tags (person 1, 2, etc.).

We use this method to detect all the background audio of the people
of interest when they are not visible. This method assumes that the
active speaker is perfect. In reality, active speaker gives noisy
results. This would cause other people's voice feature to be included
in a person's signature library and affect the final audio classification
result. 

\paragraph{Tracking Camera Wearer's Audio:} %

The camera wearer is a special participant because their face is invisible in
the egocentric videos. The active speaker detection method thus cannot be used to associate the wearer with their voice. 
We use two methods to detect the camera wearer's voice.

{\bf Method I:} The first method uses energy filtering followed by audio matching.
This method does not need ground truth labeling of the camera wearer's voice activities.
Since the microphone of the camera is usually closer to the wearer's mouth than other subjects in the scene, 
the amplitude of the wearer's voice often has higher energy than other participant's voices. 
We use this heuristic to extract candidates of the wearer's voice by choosing portions of audio with energy higher
than certain threshold. Since different recordings have different levels of loudness,
we normalize the audio using the maximum energy and then choose the possible
wearer's voice using a fixed percentage of the maximum energy. 
This threshold percentage is set to be as high as possible to avoid false alarms. 
Once the candidate audio is selected, we use the same audio matching method described in the previous
section to find all the audio that belongs to the camera wearer.
This simple method works reasonably well as summarized in Table \ref{tab:AVD-camera-activity}. 
The approach fails when the wearer never talks or talks in a very low voice, 
and in general the baseline works better for near range microphones than long range microphones. 

{\bf Method II:} In the second method, we directly classify the audio at each time instant to two categories: 
wearer's voice or not wearer's voice. The logarithm magnitude of the spectrogram at 40ms window is the input. 
The network is a modified ResNet. 
The network is trained on the Ego4d AV training dataset using a standard cross-entropy loss.

We use classification mAP to quantify the wearer audio activity detection result.
We report the average mAP on \srt{ both the test videos and validation videos in Table~\ref{tab:AVD-camera-activity}}.
\begin{table}[!h]
    \centering
    \begin{tabular}{|l|r|r|}
		\toprule
		Model & \srt{Valid} & Test\\
		\hline
		Method I & 43.95 & 50.61  \\
		Method II & 72.00 & 74.29 \\
    	Always Speak & 21.30 & 26.09 \\		
		\bottomrule
	\end{tabular}
\caption{\label{tab:AVD-camera-activity} Camera wearer activity detection baseline metrics (mAP) \srt{ on the validation and test sets respectively.}  
Always Speak assigns that the wearer speaking in each video frame.}
\end{table}

\paragraph{Speaker Diarization} 
Tables \ref{tab:AVD-diarization} \srt{, \ref{tab:AVD-diarization:val}} and \ref{tab:AVD-audio-diarization} summarize 
the speaker diarization DER metrics for the baseline models proposed in the earlier sections. 
We report the results with training only on Ego4d data as well as on with training on existing diarization datasets. 
Note that the audio-only DER is aggregated while the audio-visual DER is averaged. 
\srt{Also note the impact of the VAD on the diarization performance with the audio-only baseline. It should be noted that a model more tailored to Ego4D-like data could be used to obtain better performance. Nevertheless, this aspect still poses challenges on the AVD benchmark.}

\begin{table}[!h]
    \centering
    \begin{tabular}{|l|c|c|r|}
		\toprule
		Model & trained & sVAD & DER [\%] \\ 
		      & on Ego4D &  &  \\ 
		\hline
		RegionCls & no & no & 84.79 \\
		RegionCls & no & yes & 83.88   \\
        TalkNet & no & no & 86.68 \\
        TalkNet & no & yes & 85.85 \\
		RegionCls & yes, only & no & 80.52 \\
		RegionCls & yes, only & yes & 80.17   \\
        TalkNet & yes, only & no & 73.14 \\
        TalkNet & yes, only & yes & 73.32 \\        
        Always Speak & - & - & $>$100 \\
        Never Talk & - & - & 100 \\
		\bottomrule
	\end{tabular}
\caption{\label{tab:AVD-diarization} Diarization Baseline Metrics showing DER on \srt{the test set}.
In Always Speak, all the detected people are labeled as "speaking" in each video frame.
In Never Talk, all the detected people are labeled as "not speaking" in each video frame.}
\end{table}

\begin{table}[!h]
    \centering
    \begin{tabular}{|l|c|c|r|}
		\toprule
		Model & trained & sVAD & DER [\%] \\ 
		      & on Ego4D &  &  \\ 
		\hline
		RegionCls & no & no & 98.82 \\
		RegionCls & no & yes & 90.98   \\
        TalkNet & no & no & 99.73 \\
        TalkNet & no & yes & 92.14 \\
		RegionCls & yes, only & no & 81.66 \\
		RegionCls & yes, only & yes & 79.97   \\
        TalkNet & yes, only & no & 80.58 \\
        TalkNet & yes, only & yes & 79.30 \\        
        Always Speak & - & - & $>$100 \\
        Never Talk & - & - & 100 \\
		\bottomrule
	\end{tabular}
\caption{\label{tab:AVD-diarization:val} Diarization baseline metrics showing DER \srt{on the validation set}.
In Always Speak, all the detected people are labeled as "speaking" in each video frame.
In Never Talk, all the detected people are labeled as "not speaking" in each video frame.}
\end{table}
\begin{table}[!h]
    \centering
    \begin{tabular}{|l|r|r|}
		\toprule
		Type of VAD & Valid & Test \\ 
		\midrule
		kVAD & 67.24 & 65.28 \\ 
        Ref. Activity & 36.56 & 39.99 \\  
		\bottomrule
	\end{tabular}
\caption{\label{tab:AVD-audio-diarization} Diarization performance with audio-only models for \srt{validation and test sets} using kVAD and reference (ground truth) voice activity annotations.}
\end{table}

\paragraph{Transcription} 

To obtain baseline transcriptions, we used the pre-trained Gigaspeech model provided in the ESPNet model zoo \cite{espnet-model-zoo}. 
This model is trained on the Gigaspeech dataset \cite{chen2021gigaspeech} which contains 10000 hours of speech. 
Input features to the model are logmel features augmented using the SpecAugment method \cite{park2019specaugment} 
and normalized by global mean-variance normalization. 
The encoder of the acoustic model is based on macaron-style conformer \cite{gulati2020conformer} with 12 blocks and 
8 attention heads and the decoder is based on a 6-layer transformer \cite{vaswani2017attention} with 8 attention heads. 
In both the encoder and decoder, linear layers have 2048 units and the encoder output is 512 dimensional. 
The decoder output has 5000 sentencepiece \cite{kudo2018sentencepiece} units. 
The model is trained using a joint CTC and attention objective \cite{kim2017joint}. 
For decoding, no language model is used. For decoding, we used CTC weight of 0.3 and beam size 20 
which we did not fine-tune on the Ego4D dataset. 
The pre-trained model obtained from \cite{espnet-model-zoo} cannot support 5-min videos, hence, 
we used oracle segment information from the transcription annotations to segment the data and we decoded each segment separately. The final WER is obtained by counting the total number of errors over the whole validation or test set.

\srt{In Table \ref{tab:AVD-transcription}, we summarize the WER results depending on the VAD segmentation method on both validation and test sets. To compute the final WER, we 1) removed punctuation from both the reference and the ASR hypothesis, 2) allowed soft-match on contractions such as (I will vs. I'll) using the English global mapping file from Kaldi repository \cite{kaldi-glm}, and 3) used the NIST sclite tool \cite{nist-sclite}.
As we can see from Table \ref{tab:AVD-transcription},
on both the test and validation sets, the WERs are quite high.  This shows that the dataset is challenging for an off-the-shelf ASR model because of overlapping speech, noise, different volume levels for different speakers, occasional foreign word usage, etc.} 
\begin{table}[!h]
    \centering
    \begin{tabular}{|l|r|r|}
		\toprule
		Speech Segments & Valid & Test  \\
		\midrule
		Ground Truth & 64.8 & 59.2 \\
		\bottomrule
	\end{tabular}
\caption{\label{tab:AVD-transcription} \srt{ASR transcription WERs (\%) on the validation and test data using the reference speech segmentation.}}
\end{table}

\subsubsection{Discussion} \label{sec:avd-discussion}

Although AV diarization presents a task suite composed of reasonably well understood tasks from the vision, 
speech and audio communities, our baseline results clearly suggest that efficient speaker localization, 
tracking, diarization and transcription is a rather complex problem in the egocentric perspective and with in-the-wild data.
This is specifically evident from the performance of the \srt{joint audio and video driven} diarization and transcription baselines (with DER of $>80\%$ and WER of $>60\%$). 
\srt{Overlapping speech makes both these tasks particularly difficult to annotate as well as evaluate any proposed models. 
Performing some audio-visual source separation prior to these tasks may improve the efficacy, nevertheless sensitivity to changes and difference in speech amplitudes of overlapping speakers would still be challenging to address.}

Novel cross-modal learning approaches that jointly model audio and visual modalities \srt{while accounting for such attributes (overlapping speakers, interruptions, noise in the wild etc.)} are needed to further improve these performances. 
The baseline framework we utilized here also does not account for efficient information sharing across the four tasks in the benchmark. 
Specifically, the relationship between robust localization and tracking with multi-speaker diarization is not studied, 
and this is also not well understood in the literature. We expect this to be a challenging problem.  

We also observed that subjective attributes in conversations, like speaker accents, changes in vocabulary usage based on cultural differences etc., 
influence both the content of the speech and the clarity with which it can be captured in human annotations. 
The camera wearer's head motion adds significant blur to speakers' faces. 
To account for such aspects we performed quality checks on human annotations, 
and we expect novel unsupervised and self-supervised learning will help further address such subjective attributes. 

In future versions, we expect to increase the scope of the task suite (i.e., proposing new tasks and annotations), 
thereby opening new avenues for both core machine learning in first person perspective, and also for robust multi-modal representation learning.  
We could also investigate research directions focused on spatial audio by creating 3D environments coupled with SoundSpaces~\cite{soundspaces}. 
This enables new research and tasks in audio-visual sound source localization, audio-visual direction-of-arrival estimation and related immersive reality applications. 
We note that a small fraction of our dataset does comprise of binaural audio captured using in-ear microphones and an audio recorder (Tascam, Appendix~\ref{sec:collection}).

\iftoggle{arxiv}{
\subsubsection{Contributions statement}

Vamsi Krishna Ithapu co-led the audio-visual diarization benchmark workstream, the corresponding tasks definition, 
data selection methodology, data annotation tooling and guidelines and writing.
Christian Fuegen co-lead the audio-visual benchmark workstream, the diarization and transcription tasks definition, the corresponding annotation guidelines and paper writing.
Hao Jiang worked on data annotation tooling, tasks definition for localization and tracking, active speaker detection 
and diarization; also worked on building the baseline models for these tasks and writing. 
Federico Landini and Jachym Kolar worked on baseline models for audio-only voice activity detection and diarization, and writing. 
Leda Sari worked on transcription task definition, corresponding annotation guidelines and baseline modeling. 
Eric Zhongcong Xu worked on data selection methodology and the baseline modeling of active speaker detection.  
Ruijie Tao and Mike Zheng Shou worked on the modeling of active speaker detection.
Hanbyul Joo worked on data annotation tooling and data selection methodology. 
Christoph Feichtenhofer worked on the task definition and metrics.
Anurag Kumar worked on active speaker detection and diarization tasks definition, and on audio embeddings modeling for these tasks. 
Morrie Doulaty worked on baseline models for voice activity detection and diarization and data analysis of annotations.
Lorenzo Torresani worked on the tasks definition and annotation guidelines.
Kristen Grauman contributed to the benchmark formulation and writing.}
{}
\clearpage
\subsection{Social Interaction Benchmark}
\label{appendix:social-tasks}

This section details the Social Interaction benchmark task definitions, annotations, baseline models, and results. We also provide details on the video data collection process for multi-person capture with participants who consented to have their faces unblurred and conversation recorded (Appendix~\ref{appendix:social-collection}).
\srt{As noted in Appendix~\ref{sec:deid-appendix}, 
the social benchmark videos were screened to remove any information (e.g. last names or social media accounts) that could directly identify participants. However, participants' faces and voices are present as per our informed consent.}

\subsubsection{Formal Task Definition}
\label{appendix:formal}

LAM and TTM are defined  as  follows: (1) LAM: $y = f(\mathbf{I}, \mathbf{B})$; (2) TTM: $y = f(\mathbf{I}, \mathbf{A}, \mathbf{B})$ where $\mathbf{I}=\{I_t \}_{-T_1}^{T_2}$,  $\mathbf{A}=\{A_t \}_{-T_1}^{T_2}$, and $\mathbf{B}=\{B_t\}_{-T_1}^{T_2}$ are time-synchronized past sequences of video, audio, and bounding boxes, respectively, where $T_1$ and $T_2$ are the length of the past and future time horizon, respectively, and $t=0$ is the center frame. The bounding box indicates the target person to classify. $y$ is a binary classification label defined by:
\begin{align}
    y=\left\{\begin{array}{ll}1 &{\rm if~target~looks/talks~at~camera~wearer} \\ 0 &{\rm otherwise.}\end{array}\right.
\end{align}

\begin{figure}[h!]
  \centering  
      \subfigure[Annotation tool]{\label{Fig:ex_annotation}\includegraphics[width=\linewidth]{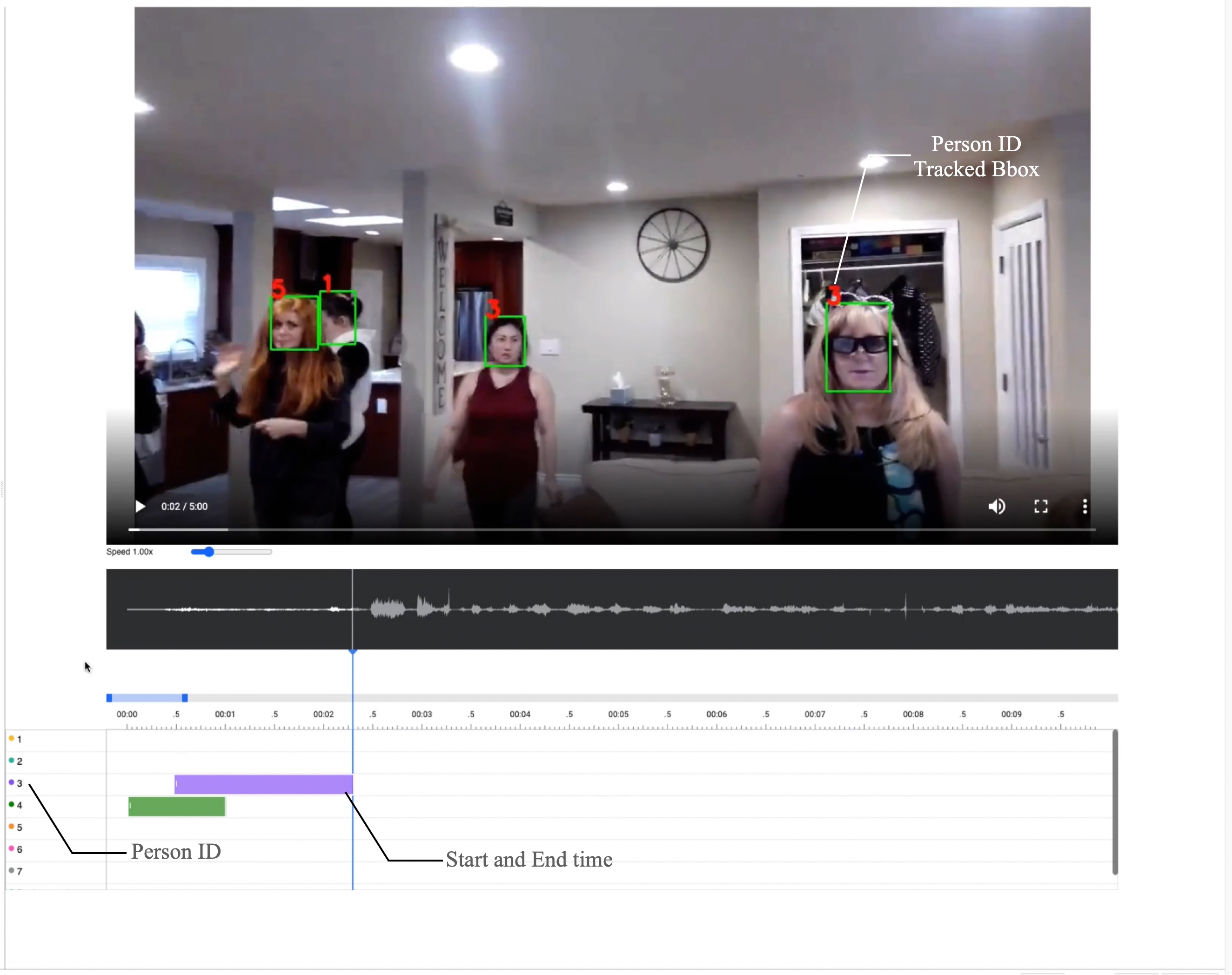}}\\
      \subfigure[Visualization of annotations.   ]{\label{Fig:ex_vis}\includegraphics[width=\linewidth]{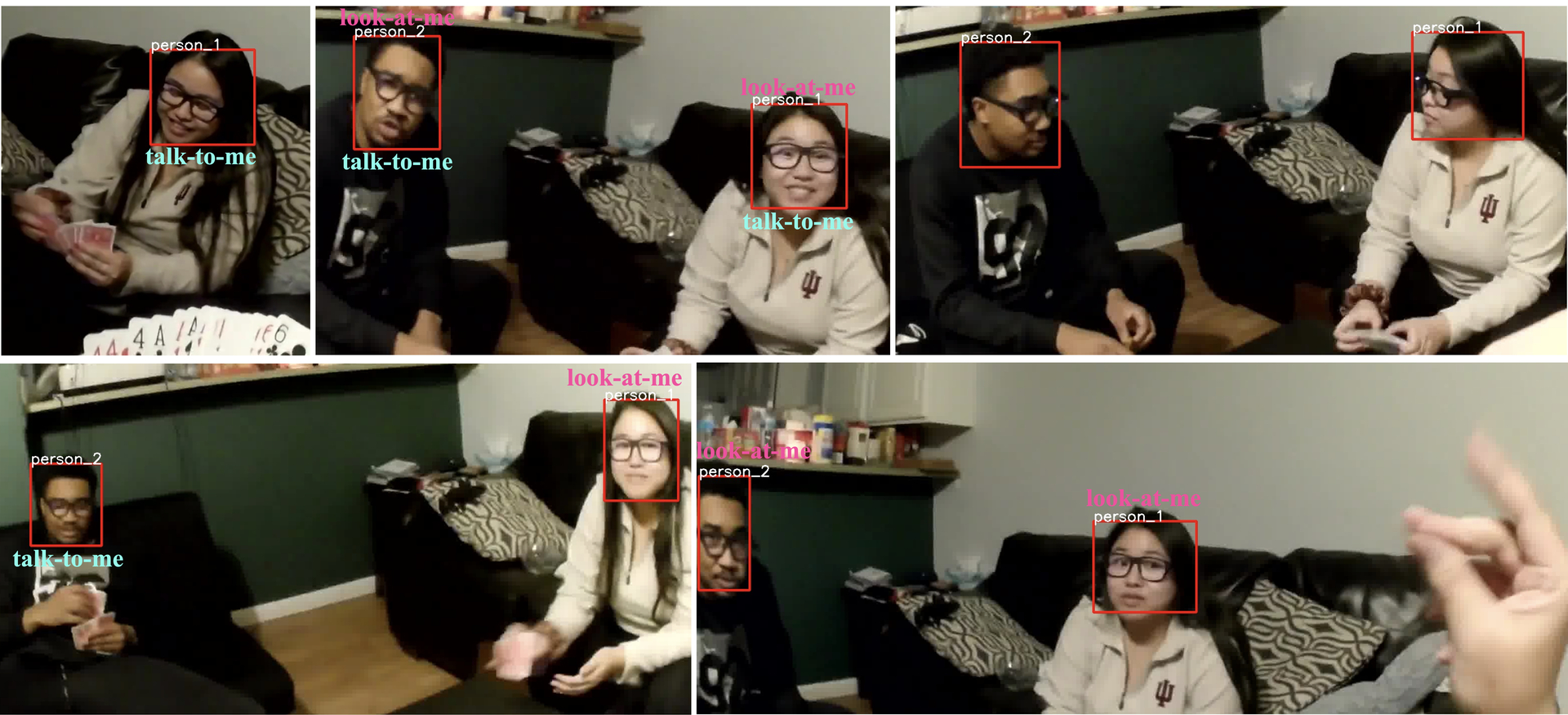}} 
  \caption{(Top) The GUI of the annotation tool;  (Bottom) Visualization of example annotations. Note that LAM (denoted by magenta text) and TTM (denoted by cyan text) may not necessarily occur together as shown in these examples.} 
  \label{fig:task-viz}
\end{figure}

The LAM and TTM tasks are defined as a frame-level prediction $y$, which stands in contrast to audio analysis tasks where labels are often assigned at the level of audio frames or segments. A desired model must be able to make a consolidated decision based on the video and audio cues over the time course of an utterance. For example, if the speaker turns their head to the side momentarily while speaking to the camera-wearer, then a frame where the speaker is looking away would have $y_{\rm LAM} = 0$ while $y_{\rm TTM}=1$. Figure~\ref{fig:task-viz} gives some frame level visualization of annotations that illustrate the task definitions.

\subsubsection{Annotation Statistics}
\label{appendix:social-stats}
The social task annotations, LAM and TTM, build on the same video clips used in the AV diarization tasks and described in Appendix~\ref{sec:avd-datastat}. Fig~\ref{fig:social-annotation-stat} summarizes the statistics of LAM and TTM annotations across these clips. We compute the percentage of the frames with LAM or TTM annotations in each clip and show the histograms in Fig~\ref{fig:social-annotation-stat} (a) and (b), respectively. In many clips, these events happen rarely (10 \% or lower), and the frames with LAM annotations are less frequent than TTM cases. We also list the duration of each LAM or TTM annotation (the duration between start and end time) in Fig~\ref{fig:social-annotation-stat} (c) and (d), in order to illustrate the significant variations in length. The most frequent case is short-duration LAM or TTM behaviors, lasting 1 or 2 seconds. The data was organized as follows for baseline model training in Section~\ref{appendix:social-baselines}: 389 clips were held out for training, comprising 32.4 hours in total. An additional 50 clips (4.2 hours) and 133 clips (11.1 hours) were held out to form the validation and testing sets, respectively.

\begin{figure}[t]
    \subfigure[\% of LAM per clips]{
        \centering
        \includegraphics[width=0.47\linewidth]{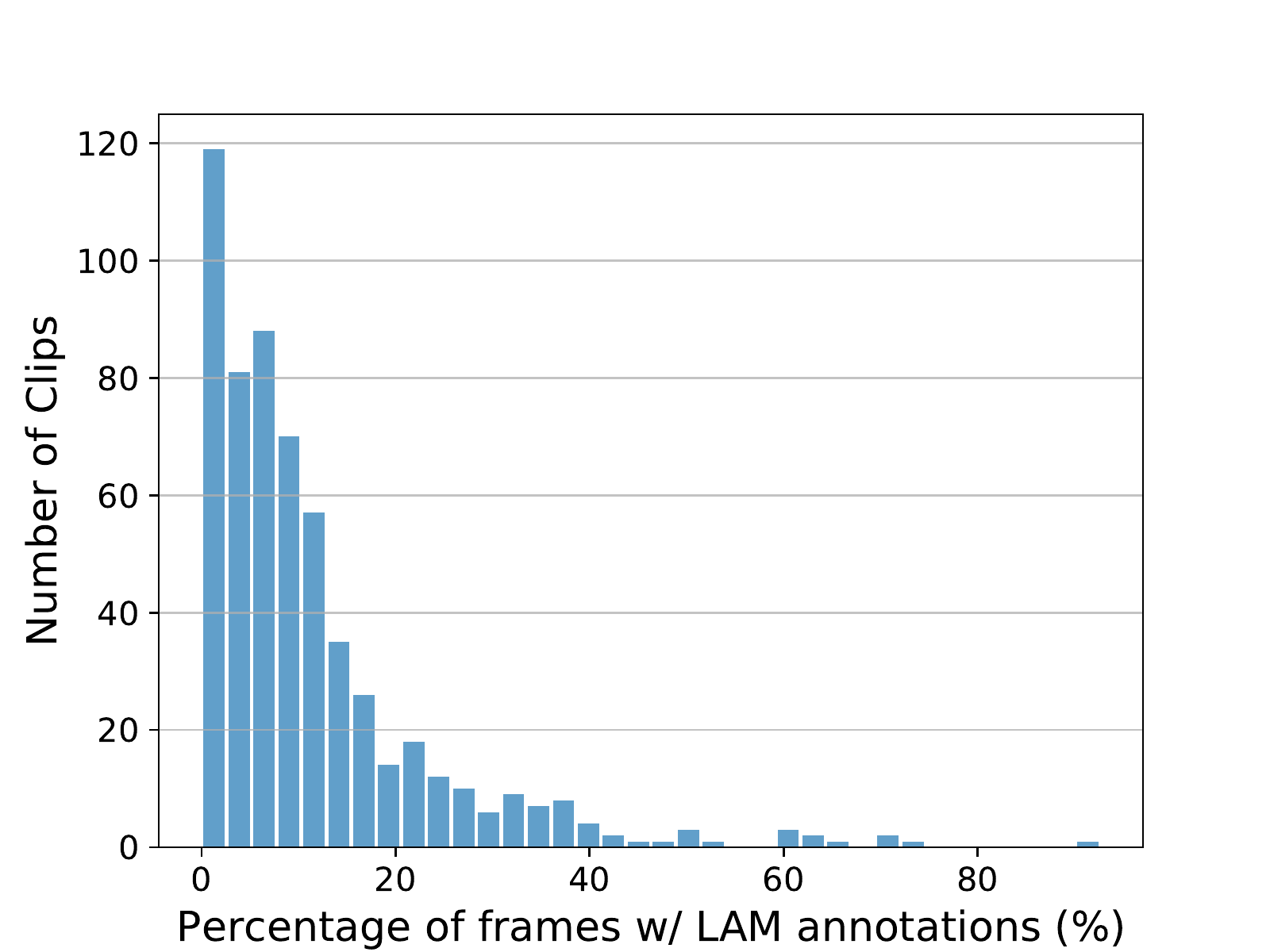}
    }
    \subfigure[\% of TTM per clips]{
        \centering
        \includegraphics[width=0.47\linewidth]{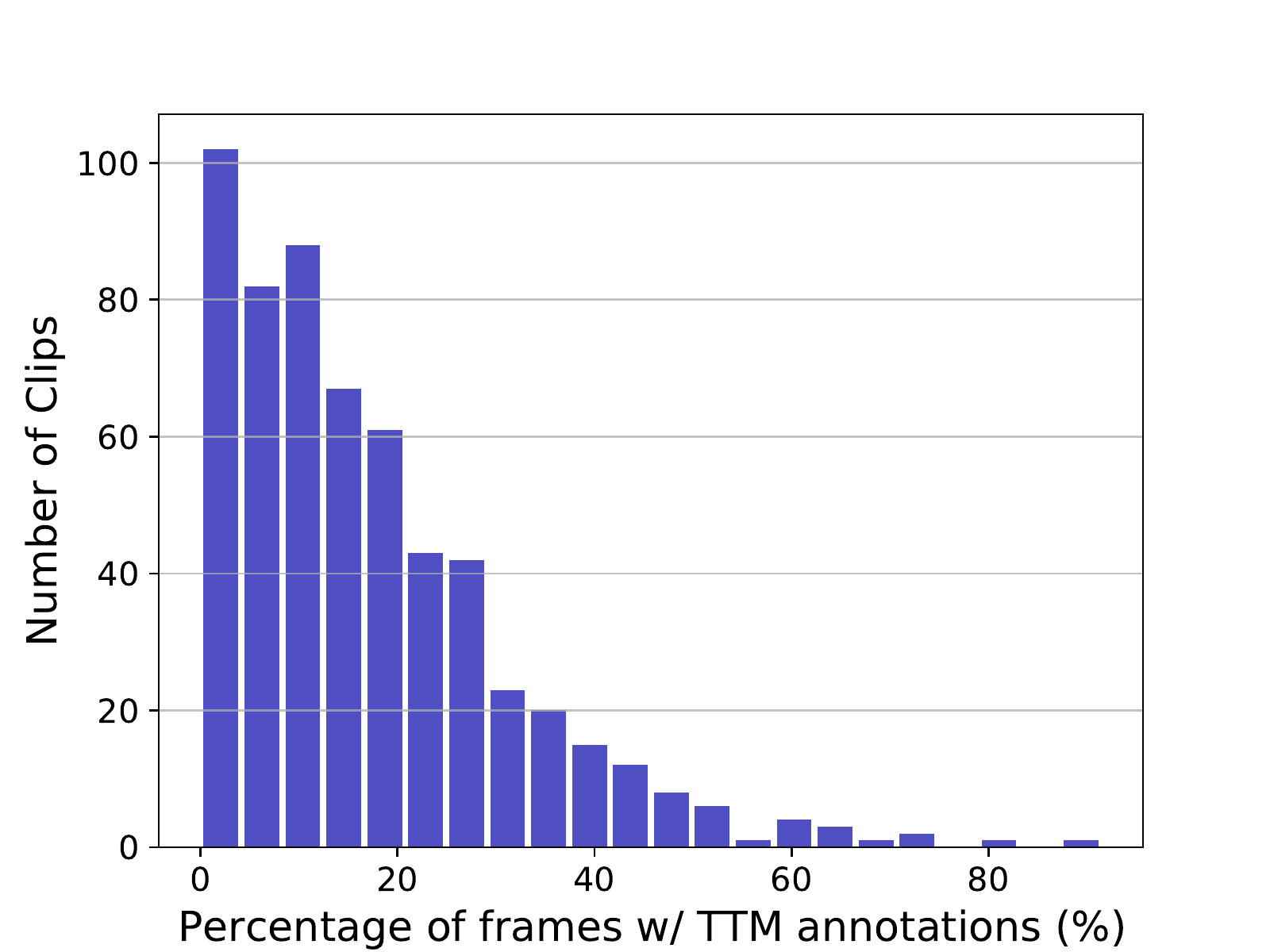}
    }\\
    \subfigure[Duration of LAM]{
        \centering
        \includegraphics[width=0.47\linewidth]{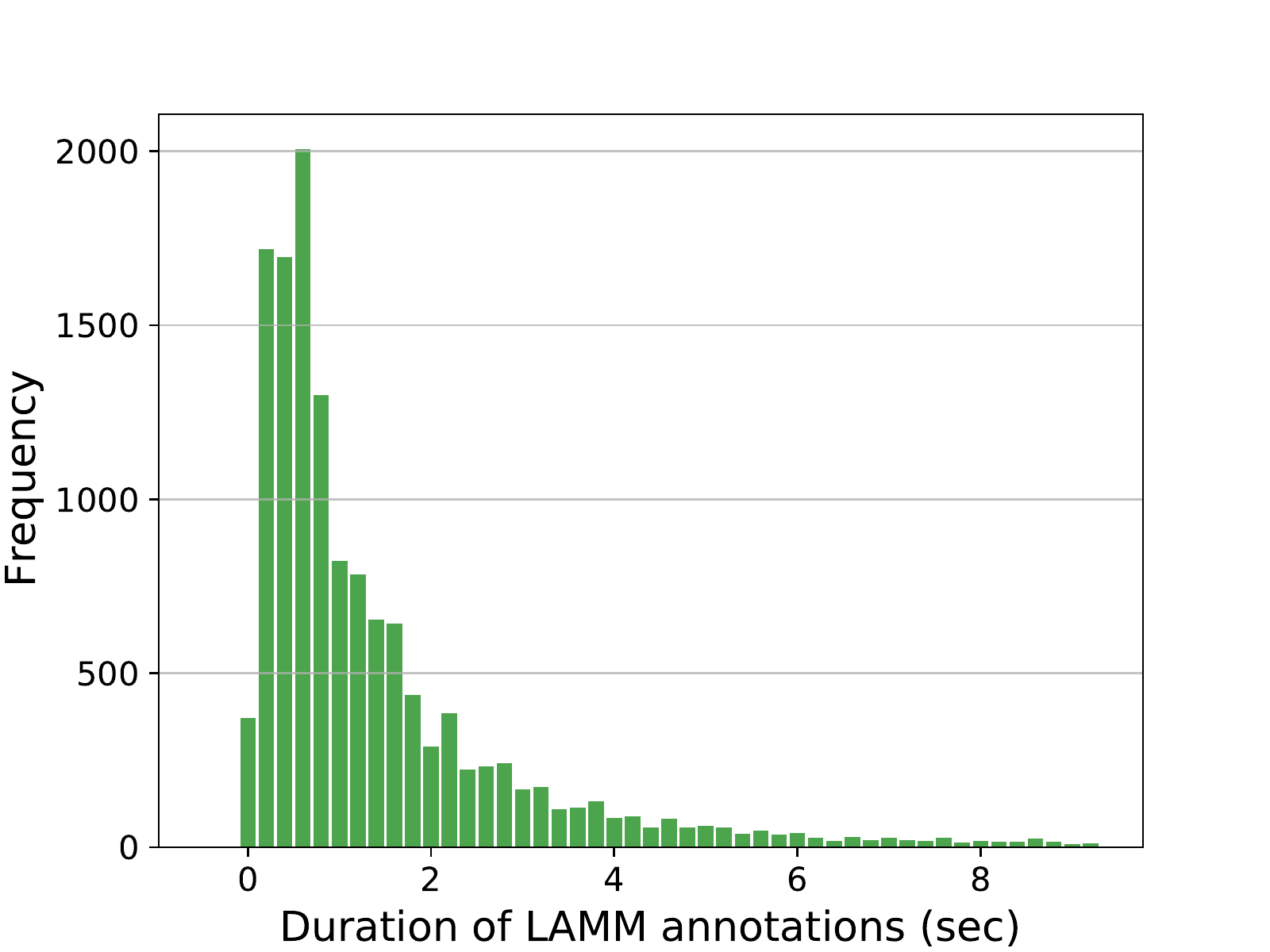}
    }
    \subfigure[Duration of TTM]{
        \centering
        \includegraphics[width=0.47\linewidth]{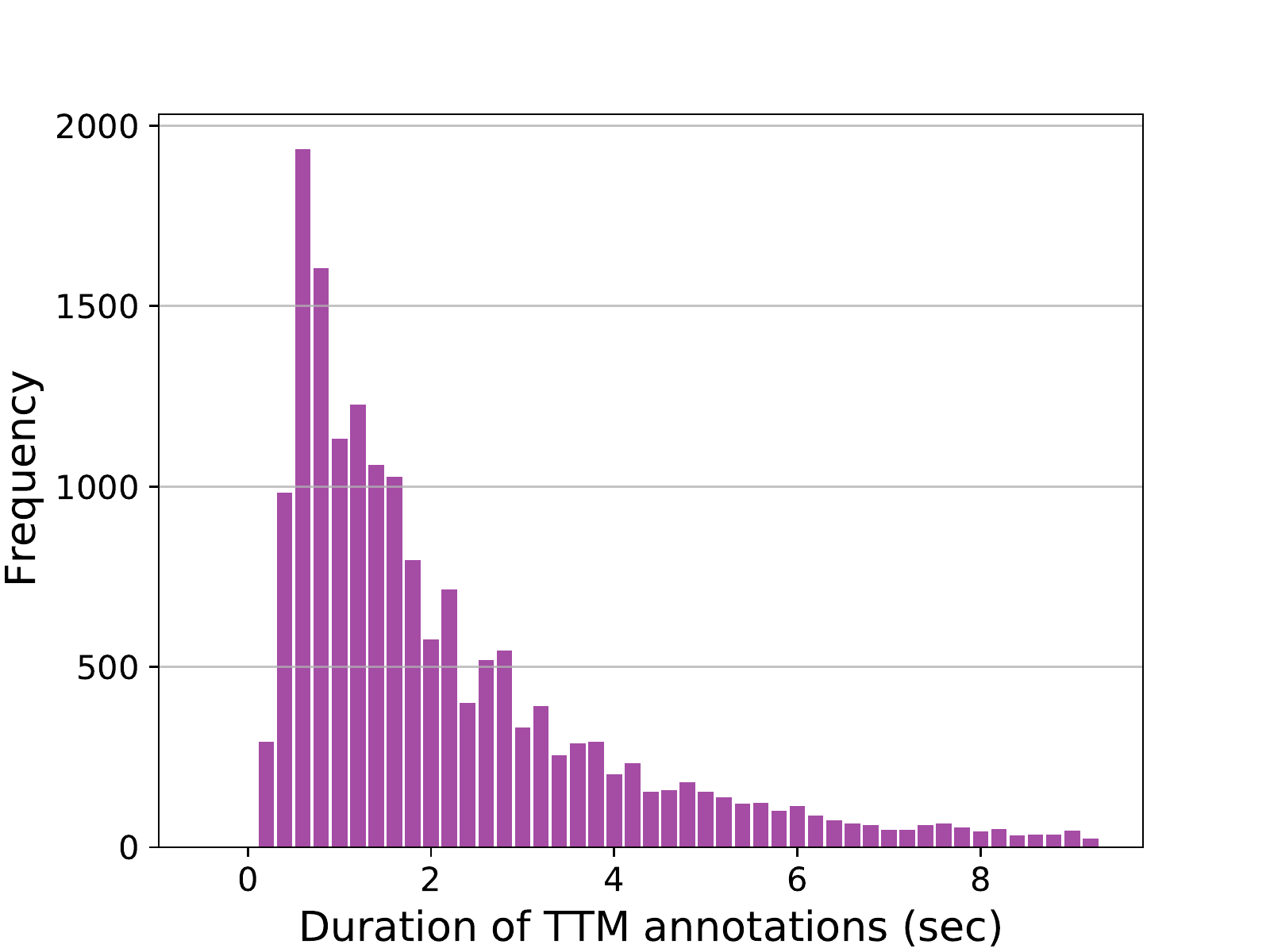}
    }
  \caption{\label{fig:social-annotation-stat} Social task annotation statistics. (a) Histogram showing the number of clips vs. the percentage of frames with look-at-me annotations; (b) Histogram showing the number of clips vs. the percentage of frames with talk-to-me annotations in each clip; (c) Histogram showing the duration of look-at-me annotations; (d) Histogram showing the duration of talk-to-me annotations.}
\end{figure}  

\subsubsection{Social Baseline Models and Results}
\label{appendix:social-baselines}

\begin{figure}[t]
    \subfigure[LAM]{
        \begin{minipage}[t]{\linewidth}
        \centering
        \includegraphics[width=0.97\linewidth]{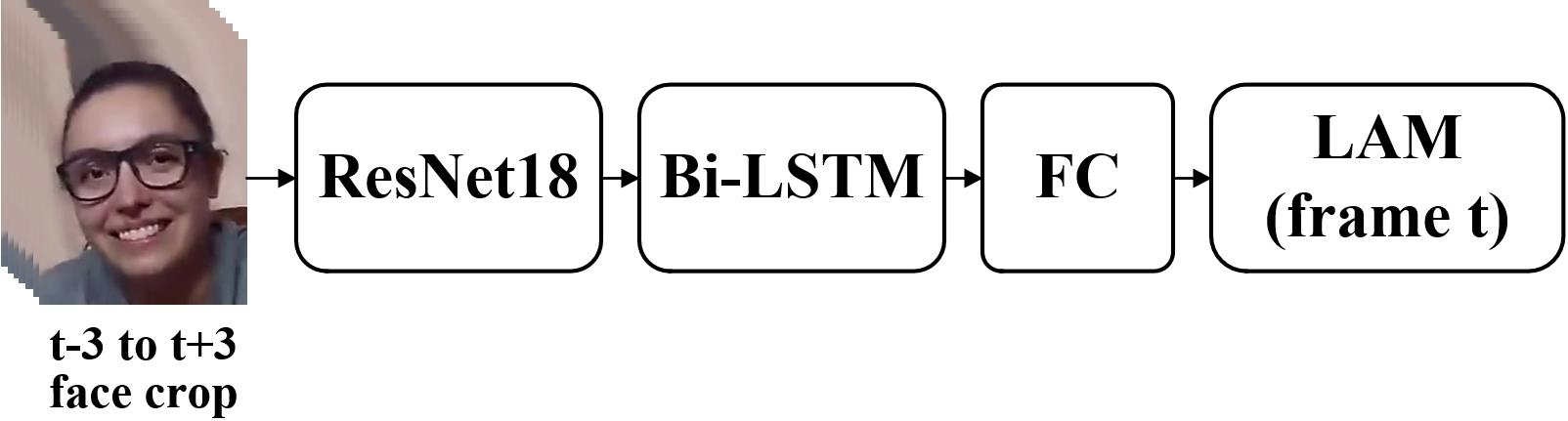}
        \label{baseline_arch:lam}
        \end{minipage}
    }
        \subfigure[TTM]{
        \begin{minipage}[t]{\linewidth}
        \centering
        \includegraphics[width=\linewidth]{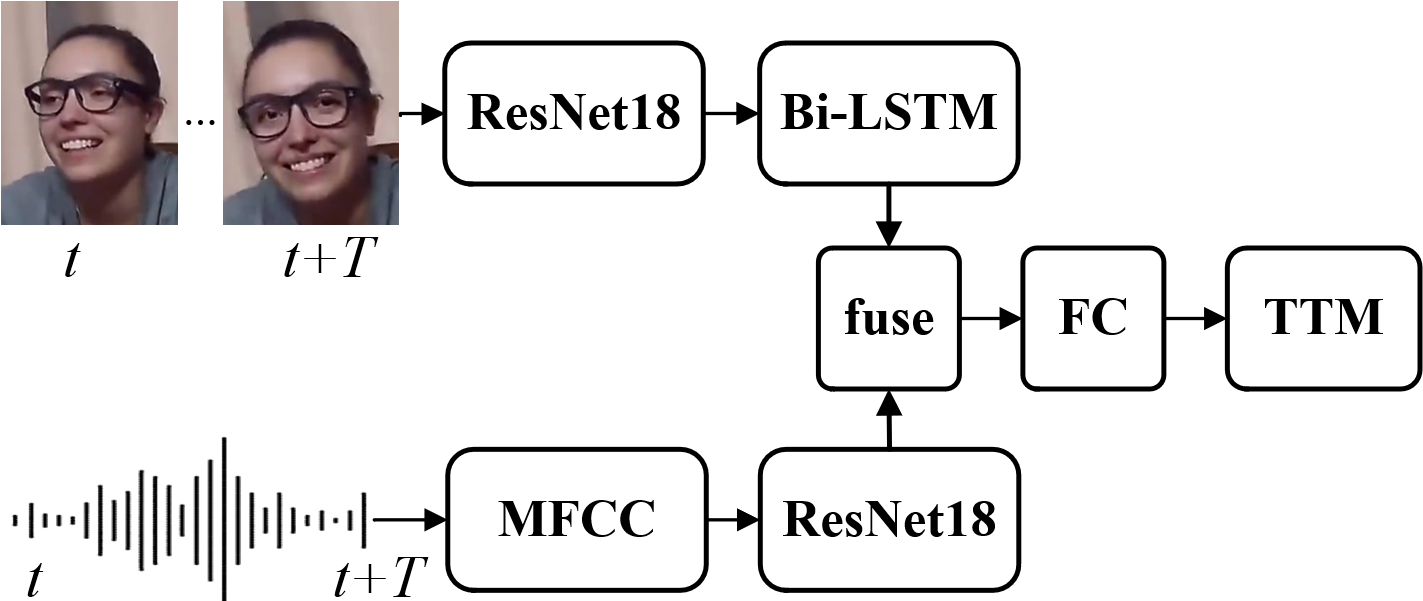}
        \label{baseline_arch:ttm}
        \end{minipage}
    }
    \centering
    \caption{\textbf{Baseline model architectures.} (a) LAM model uses a ResNet-18 as a backbone to extract the feature of each frame. A Bidirectional-LSTM then takes the sequence and encode the features into one embedding. We pass the embedding to FC layer that predicts the LAM result. (b) TTM model has two encoders. The video encoder is the same as LAM. The audio encoder extracts the MFCC frequency map of the audio segment and the feature is fed into a ResNet-18 network. The visual and audio embeddings are concatenated and passed through the FC layer to predict the target of this utterance.}
    \label{fig:baseline_arch}
\end{figure}

\paragraph{LAM} Our baseline model for LAM is a video-based model using ResNet-18 and Bidirectional LSTM. %
Our model uses the cropped face regions in video as input in order to focus on cues about the head pose and social attention visible in the face.
The architecture of our baseline is similar to the Gaze360 \cite{Kellnhofer2019}. As illustrated in Fig~\ref{baseline_arch:lam}, we input 7 consecutive frames ($T_1=3$ and $T_2=3$) from one face tracklet, and each image is resized to 224$\times$224. Each frame is then processed by the ResNet-18 backbone independently to generate 256 dimensional face features. The feature sequence is encoded by a Bidirectional LSTM, which has two recurrent layers with dimensionality 256. The output is fed into a classification head to predict the binary LAM result for the center frame at the $t$-th timestamp. The LAM task has a class imbalance issue, and we use weighted cross-entropy loss. Since the architecture is similar to Gaze360, we have two options for the initialization: first, initializing the backbone from a pretrained Gaze360 model; second, initializing the model randomly and training from scratch on Ego4D. During training, we sample center frames with a stride of 3. The network is optimized by Adam with a learning rate of $5\times 10^{-4}$. 

The results are shown in Table \ref{tab:lam_results}. Our baseline model achieves an mAP of 66.07\% on the test split when initialized randomly, and the performance is higher at 78.07\% when initialized from Gaze360. These findings highlight the close relationship between the LAM task and gaze estimation. The random guess model achieves about 8\% accuracy because the negative samples account for 92\% of the test split and the model always predicts looking at me.

\paragraph{TTM} The baseline model for TTM digests multi-modal inputs: each audio segment is paired with an associated face crop. Since the audio segments vary substantially in duration, we break the long utterances into short segments whose maximum duration is limited to 1.5s. If the segment is shorter than 0.15s, we skip it in the training stage. The associated faces are also resized to 224$\times$224, and the video encoder is the same as LAM. However, sometimes the speakers leave the field of view or become invisible due to the rapid motion. In this case, we pad the face sequences with blank images. The MFCC feature is extracted every 10ms with a 25ms window length. The feature is then fed into the audio backbone, a ResNet-18 designed for audio tasks \cite{chung2020in}. Following the encoders, we concatenate the audio and visual embeddings and pass them to the final classification head to get the TTM result for the visible faces associated with the segment. To train the model in parallel, we first sort the short segments based on the length and group the segments into a batch if they have the same duration. The batch size is restricted by the GPU memory; we use a batch size of 400. The model is also optimized using Adam with a learning rate of $5\times 10^{-4}$. 

Table \ref{tab:ttm_results} summarizes the TTM results. TTM is more challenging than LAM. We can see that our baseline model only increases the mAP by 9.77\% on the test split in comparison to the random guess model.

\subsubsection{Discussion}

While the benchmark tasks of detecting when attention and speaking behaviors are directed towards the first-person are closely related to existing analysis tasks, it is clear from the baseline performance that there is substantial room for improvement, with mAP of 78.07 for LAM and 55.06 for TTM. 

The TTM task is particularly challenging because it requires  analysis of the audio content to understand the target audience of an utterance, as well as the fusion of audio and video cues. The most complete solution to this problem will require an understanding of the semantics of the utterance in the context of an evolving conversational interaction. Future work on this task might involve more sophisticated language modeling and possibly hierarchical analysis approaches that allow the integration of cues at multiple levels, e.g. at the dialog level to understand who is participating in a conversational exchange, at the utterance level to access semantics, and at the audio level to exploit prosodic and other cues. 

The LAM task presents additional challenges such as the need to deal with motion blur and fast head movements, and may also benefit from a more explicit modeling of head movement and the patterns of gaze behavior that arise in conversational interaction.

\begin{table}[t]
\centering
\begin{tabular}{l|ll|lll}
\multicolumn{1}{l|}{} &\multicolumn{2}{c|}{val} &\multicolumn{2}{c}{test}\\    \cline{2-5}
\multicolumn{1}{l|}{} & Acc   & mAP & Acc   & mAP \\ \hline
\multicolumn{1}{l|}{Random Guess}  &8.57 &51.19    &7.98      & 50.96  \\
\multicolumn{1}{l|}{Baseline (Gaze360)}&91.78 &79.90 & 87.97 & 78.07        \\
\multicolumn{1}{l|}{Baseline (Random)}  &86.45 &72.11    & 75.38 & 66.07       \\
\hline
\end{tabular}
\caption{\textbf{Results of LAM.} The baseline model was initialized from Gaze360~\cite{Kellnhofer2019} (2nd row) and at random (3rd row).}
\label{tab:lam_results}
\end{table}

\begin{table}[t]
\centering
\begin{tabular}{l|ll|lll}
\multicolumn{1}{l|}{} &\multicolumn{2}{c|}{val} &\multicolumn{2}{c}{test}\\  \cline{2-5}
\multicolumn{1}{l|}{} & Acc   & mAP & Acc   & mAP \\ \hline
\multicolumn{1}{l|}{Random Guess}  &32.44 &53.82    &47.41      & 50.16  \\
\multicolumn{1}{l|}{Baseline}   &64.31 &56.50   & 49.75 & 55.06       \\
\hline
\end{tabular}
\caption{\textbf{Results of TTM.} The baseline model is initialized randomly.}
\label{tab:ttm_results}
\end{table}

\subsubsection{Social Dataset Collection}
\label{appendix:social-collection}

The Ego4D Social data collection process was designed to achieve: 1) naturalistic interactions, 2) multi-modal capture, and 3) diverse participants and environments. %
Participants consisted of friends and family groups and data was captured in residences and local neighborhoods, ensuring naturalistic interactions. Capture hardware varied across sites but included wearable cameras, wearable eye trackers\iftoggle{arxiv}{ (at Georgia Tech and Indiana University),}{,} binaural recording systems, and smart watches\iftoggle{arxiv}{ (at Georgia Tech).}{.} Protocols included highly-structured settings, where participants were asked to play games over a period of a few hours in a residence, and unstructured settings where participants captured social interactions in daily life over a period a week or more. Sample social interaction contexts included playing board and card games, preparing meals, and going on walks.  The bulk of the data collection took place during the COVID-19 pandemic, and the resulting study protocols were designed to safeguard participants against additional risk.

 The social data consists of data collected at five sites\iftoggle{arxiv}{: Atlanta, Bloomington, Redmond, Twin Cities, and Singapore.}{.} %
In total, 764 hours of video and audio were collected for the social benchmark task. A detailed summary of the data collection practices at each site can be found in Appendix~\ref{sec:collection}.

\subsubsection{Derived Tasks for Future Social Benchmarks}

The core tasks of LAM and TTM define a starting point for analyzing multi-modal egocentric data and inferring social interactions. We now describe two groups of potential future tasks, attention tasks and speaking tasks, that could be supported via the existing annotations in Ego4D Social and the gaze data collected from eye trackers.

\paragraph{Egocentric Attention Prediction (EAP)} Prior work~\cite{Li2013,Li2021} has demonstrated the feasibility of predicting where the camera-wearer is looking (i.e. their egocentric attention) using only egocentric video captured from a head-worn camera. This work leveraged the context of hand-eye coordination tasks, which require gaze to be coordinated with hand movements and objects. A subset of the Ego4D Social data includes gaze measurements produced by wearable eye trackers by Indiana University and Georgia Tech participants (e.g., Pupil Invisible), which will greatly expand the size of data for hand-eye coordination in the wild.

\paragraph{Social Gaze Prediction (SGP)} The LAM task addresses the special case of social gaze: a person looks at the camera-wearer. It is possible to generalize the task by predicting the social gaze target for each of the visible faces in an egocentric video, i.e., $y^p \in \{0,1,\ldots,M\}$, where $M$ is the total number of participants in a group social interaction, and $p \in \{0,1,\ldots,M\}$. $p$ is the index for social members. The case $y^p=q$ means that target $p$ was looking at participant $q$. The case $y^p=0$ captures alternative gaze targets, including non-social gaze targets (e.g. looking at an object), looking at people who are not wearing an egocentric camera (with the result that ground truth annotations are not available), and looking at unknown targets not captured in any of the egocentric videos. Let $\hat{y}^{q,p}$ denote the LAM label for target person $p$ visible in frame of egocentric video $\mathbf{I}_q$ captured by participant $q$. Then the SGP label is given by $y^p = \argmax_q\{\hat{y}^{q,p}\}$. The Ego4D Social data includes synchronized videos from multiple social members, which will allow us to expand the annotation by matching the person ID with the camera-wearers. 
Note that since the video recorders are not genlocked, the identification of corresponding frames will only be approximate. However, since gaze behaviors persist over multiple frames we do not believe this will be an issue.

A key issue in defining the task is the determination of the participant set. For a 2D version of SGP, termed SCG-2D, the participant set is defined by participants who are visible in frame $t$. This is a social version of the video-based gaze follow task~\cite{Chong2020attended}, where the goal is to predict whether each target participant is looking at any of the other participants who are visible in the frame. A more challenging 3D version of the task, SCG-3D, uses all of the participants who are present in the social scene at the time of frame $t$. This task requires the ability to predict which participant the target person $p$ is looking at in the case where that participant is not visible in frame $t$. This can in principle be accomplished by maintaining a birds-eye view layout map of the social scene, that captures the approximate spatial relationships between the participants. Such a layout map could be used in conjunction with an approach like Gaze360~\cite{Kellnhofer2019} to solve the SCG-3D task. Note that this task could potentially benefit from taking recorded binaural audio as an additional input, as the ability to localize sound sources could provide additional cues for determining the locations of gaze targets which are not visible in the video.

\paragraph{Utterance Target Prediction (UTP)} The TTM task can be generalized to the full set of participants in the same way that LAM can be extended to SGP. The input space is the same as TTM and the output space is similar to SGP, where $y^p=q$ means that participant $p$ is talking to participant $q$, and $y^p=0$ denotes the cases where the participant is not talking to anyone, or is talking to someone who is not wearing an egocentric camera (and therefore ground truth cannot be determined). In contrast to SGP, UTP requires the identification of all of the target recipients of an utterance. In fact, our TTM annotation already supports this task, as it differentiates the case where the utterance is directed to multiple participants including the camera wearer. This additional label is ignored in the design of the simpler TTM task.

\paragraph{Transcript-based Variants} For all of the previously-defined social tasks it is possible to define a variant which utilizes a transcript of the audio file as an additional input modality. For example, the TTM-T task is the variant of TTM with the prediction defined as $y^p=f(\mathbf{I}, \mathbf{A}, \mathbf{T}, \mathbf{B})$, where $\mathbf{T}$ the transcript (time-stamped sequence of words) obtained from $\mathbf{A}$. This can potentially simplify the use of dialog cues to identify the intended targets for utterances and social gaze. 

\iftoggle{arxiv}{
\subsubsection{Contributions statement}

James M. Rehg co-led the social benchmark effort and paper writing. Hanbyul Joo co-led the social benchmark effort and data annotation. Mike Zheng Shou co-led the social benchmark effort and problem formulation and modeling experiments. David Crandall led data collection at the Bloomington site and contributed to the social benchmark formulation and paper writing. Vamsi Ithapu contributed to the social benchmark formulation and data annotation. Hyun Soo Park led data collection at the Twin Cities site and contributed to the social benchmark formulation and paper writing. 

Hao Jiang contributed to model development and data annotation. Yunyi Zhu contributed to model implementation and experiments. Eric Zhongcong Xu contributed to the social benchmark data preparation and the model implementation and experiments, and contributed to all data collection related tasks for the Singapore site. Ruijie Tao contributed to data collection for the Singapore site. Fiona Ryan led the data collection effort for the Atlanta site, including protocol design, multimodal sensor deployment and synchronization, and de-identification. Miao Liu contributed to data collection and analysis for the Atlanta site. Audrey Southerland contributed to the protocol design, IRB authoring and submission, participant recruiting, and data ingestion for the Atlanta site. Jayant Sharma contributed to participant recruiting, data collection, IRB submission, analysis, and data ingestion for the Twin Cities site. Yuchen Wang contributed to the protocol design, participant recruiting, and data collection for the Bloomington site. Weslie Khoo developed the multi-camera synchronization and de-identification pipelines at the Bloomington site.

\paragraph{Acknowledgements} The social benchmark team would like to acknowledge the following additional contributions from individuals at each site: \textbf{Atlanta}: Jeffrey Valdez (recruitment and data collection), Gabriella Stripling, Ruth Stolovitz, and Andrea Sucre-Pardo (recruitment and dataset de-identification). 
\textbf{Twin Cities}: Reese Kneeland, Angad Cheema, Silong Tan, Anjali Oleksy, Zhiteng Cao, Diana Begelman (data collection and annotation) \textbf{Facebook:} Samuel Clapp and Peter Dodds (binaural audio recording and multimodal synchronization). \textbf{Bloomington:} Zunaeed Salahuddin, Zehua Zhang, Ziwei Zhao. 

}{}

\clearpage
\subsection{Forecasting Benchmark}
\label{appendix:forecasting}

This section details the Forecasting benchmark task definitions, annotations, baseline models, and results.

\subsubsection{Formal tasks definitions}
\label{appendix:forecasting_tasks}

As noted in the main paper, there are four forecasting tasks: future locomotion movement prediction, future hand prediction, short-term object interaction anticipation, and long-term action anticipation.

\subsubsection*{Future Locomotion Movements Prediction}

This task aims to predict the future locomotion of a user given a sequence of past images. We formulate the problem as:
\begin{align}
    \mathcal{X} = \begin{bmatrix}\mathbf{x}_{t+1} &\cdots & \mathbf{x}_{t+F}\end{bmatrix}^\mathsf{T} = f(\mathbf{x}_{t-T},\cdots,\mathbf{x}_{t-1}; \mathcal{I}), \label{Eq:locomotion}
\end{align}
where $\mathcal{X}$ is the future trajectory, $T$ and $F$ are the past and future time horizons, respectively, $\mathbf{x}_{t}$ is the point on the trajectory at time~$t$, and $\mathcal{I}$ is the egocentric image at time~$t$. With an assumption that the person walks over a major plane (e.g., ground plane), we represent the trajectory in a 2D plane, i.e., $\mathbf{x}_t \in \mathds{R}^2$.

The essence of the locomotion task is to design a function~$f$ to predict a set of plausible $K$ future trajectories $\{\mathcal{X}^k\}_{k}$ given the \FEB{current image}. %
Since there exists a number of plausible future trajectories with different topology, e.g., trajectories that bifurcate at an Y-junction, we predict $K$ future trajectories.

\subsubsection*{Future Hand Prediction}

\begin{figure*}
    \centering
    \includegraphics[width=\linewidth]{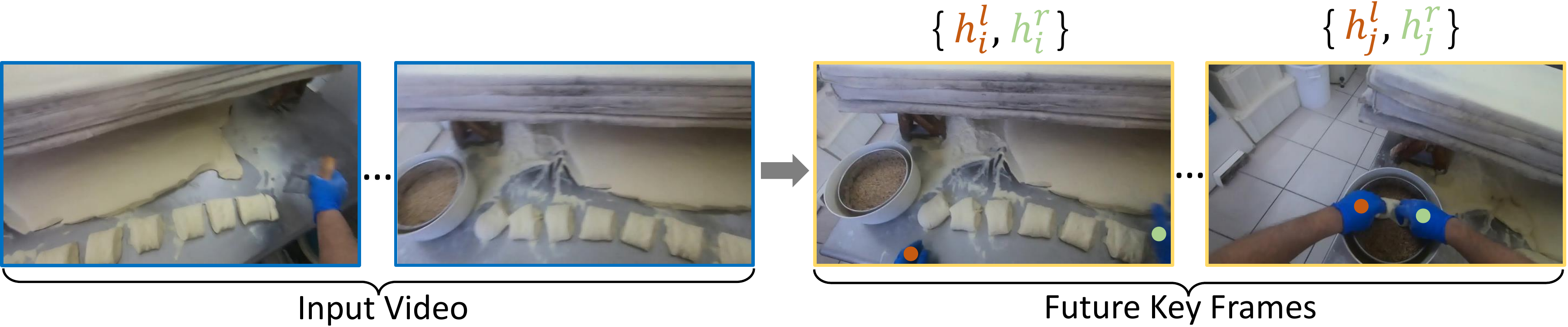}
    \caption{Example of future hand prediction.}
    \label{fig:hand_ant_setup}
\end{figure*}

In addition to future locomotion movements prediction, we consider another challenging task of predicting future hand positions of key-frames (see visual illustration in Fig.~\ref{fig:hand_ant_setup}). Specifically, we denote the contact frame\footnote{The contact frame is defined as the first frame in which the user touches the object, hence the moment in which the object becomes active.} as $x_c$, pre-condition frame\footnote{As defined in Section \ref{appendix:hands-objects}, the pre-condition frame marks a moment prior to the
state-change of an object.} as $x_p$, and the three frames preceding the pre-condition frame by $0.5s$, $1s$ and $1.5s$ as $x_{p_1}$, $x_{p_2}$, $x_{p_3}$, respectively. Formally, given an input egocentric video $1.5s$ before the pre-condition time step (denoted as $x=\{x_{{p_3}-t_o-1}, ..., x_{{p_3}-1}\}$, with $t_o$ referred as observation time), this task seeks to predict the positions of both hands $(h_i^l, h_i^r)$ in the future key frames, where $i\in \{c,p,{p_1},{p_2},{p_3}\}$.

\subsubsection*{Short-Term Object Interaction Anticipation}

This task aims to predict the next human-object interaction happening after a given timestamp. Given an input video, the goal is to anticipate:
\begin{itemize}
    \item {The spatial positions of the active objects, among those which are in the scene (e.g., bounding boxes around the objects). We consider the next active object to be the next object which will be touched by the user (either with their hands or with a tool) to initiate an interaction;}
    \item {The category of each of the detected next active objects (e.g., ``knife'', ``tomato'');}
    \item {How each active object will be used, i.e., what action will be performed on the active objects (e.g., ``take'', ``cut'');}
    \item {When the interaction with each object will begin (e.g., ``in 1 second'', ``in 0.25 seconds''). This is the time to the first frame in which the user touches the active object (time to contact). This prediction can be useful in scenarios which involve human-machine collaboration. For instance, an assistive system could give an alert if a short time to action is predicted for a potentially dangerous object to touch.}
\end{itemize}

\begin{figure*}
    \centering
    \includegraphics[width=\linewidth]{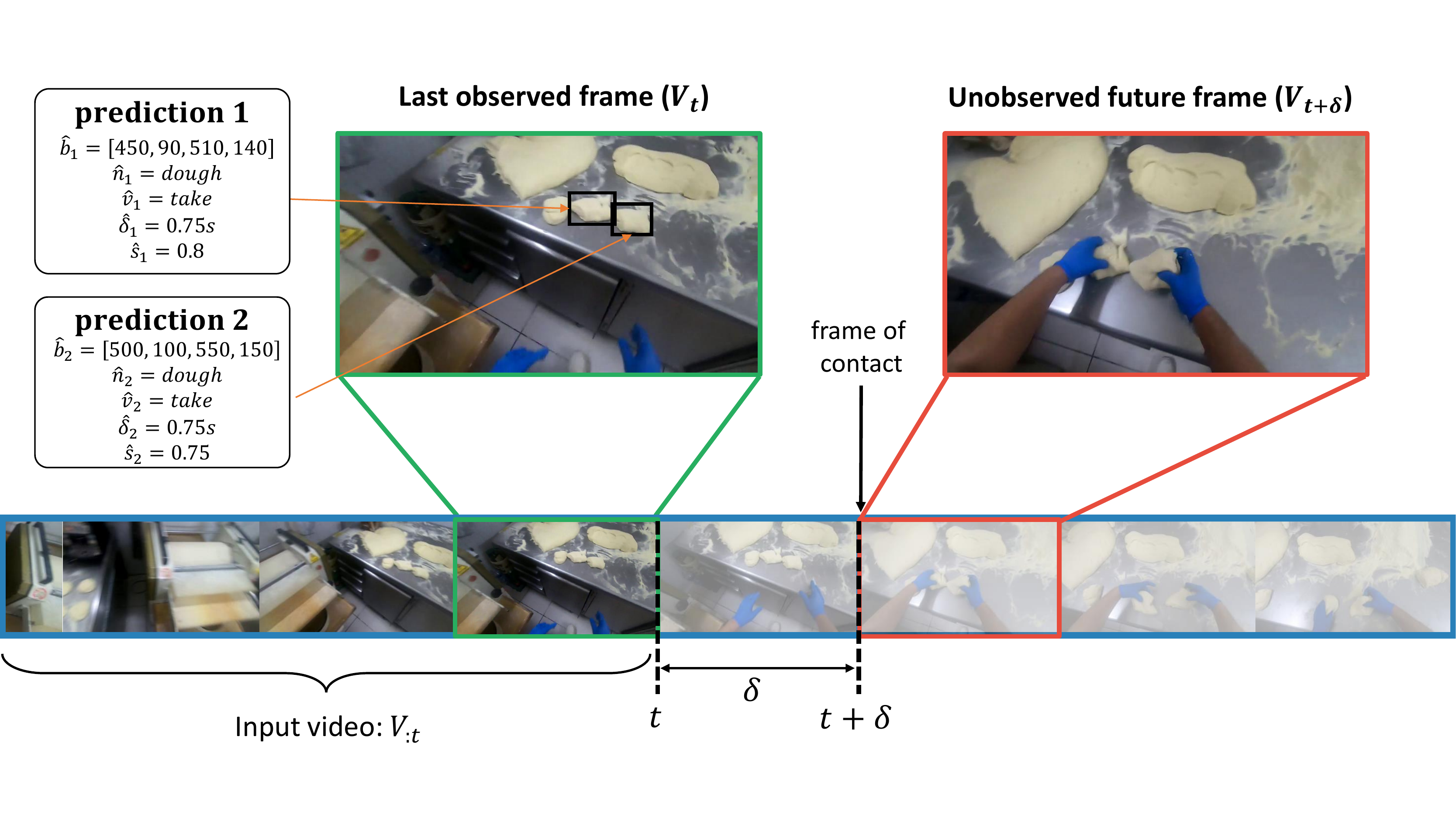}
    \caption{Example of short-term object interaction anticipation.}
    \label{fig:object_interaction_anticipation}
\end{figure*}
In this task, models are required to make predictions \textit{at a specific timestamp}, rather than densely throughout the video. Figure~\ref{fig:object_interaction_anticipation} illustrates the set-up. {The model is allowed to process the video up to frame $t$, at which point it must anticipate the next active objects, and how they will take part in an interaction in $\delta$ seconds, where $\delta$ is unknown. The model can make zero or more predictions. Each prediction indicates the next active object in terms of noun class~($\hat n$) and bounding box ($\hat b$), a verb indicating the future action~($\hat v$), as well as the time to contact ($\hat \delta$), which estimates how many seconds in the future the interaction with the object will begin. Each prediction also comprises a confidence score ($\hat s$) used for evaluation.}

Specifically, let $V$ be an untrimmed video.  
We will denote with $V_t$ the frame of $V$ occurring at time-step $t$ and with $V_{:t}$ the video segment starting at the beginning of $V$ (timestamp $0$) and ending at timestamp $t$.
Given a timestamp $t$, denoted as ``stopping time'', the short-term object interaction anticipation task requires that a model is able to exploit the observed video $V_{:t}$ to predict {$N$ tuples (where $N$ is arbitrary)}:
\begin{equation}
\label{eq:prediction_tuple}
{\{(\hat b_{i}, \hat n_{i}, \hat v_{i}, \hat \delta_{i}, \hat s_{i})\}_{i=1}^N}
\end{equation}
where:
\begin{itemize}
    \item {$\hat b_{i} \in \mathbb{R}^4$ is a bounding box indicating the position of the predicted next active object;}
    \item {$\hat n_{i} \in \mathcal{N}$ is a \textit{noun} indicating the class of the next active object, where $\mathcal{N}$ is the set of possible nouns. 
    \item {$\hat v_{i} \in \mathcal{V}$ is a \textit{verb} indicating the action which will be performed, where $\mathcal{V}$ is the set of possible verbs;}}
    \item {$\hat \delta_{i} \in R^+$ is the \textit{time to contact}, a positive number which estimates how many second into the future the interaction with the object will begin;}
    \item {$\hat s_{i} \in [0,1]$ is a confidence score associated to the prediction. Objects with a large confidence value are deemed to be likely next-active.}
\end{itemize}

The model is {allowed} to perform $N$ predictions for each observed example {(with $N$ arbitrary) both to} account for the {presence of multiple next-active-objects and to handle the} multi-modality of future predictions.
Each of the $N$ predictions is intended as a plausible {future object interaction.}
Figure~\ref{fig:object_interaction_anticipation} illustrates the proposed task. Given a video $V_{:t}$, a method should be able to {detect} the next active objects ({e.g., two instances of} ``dough''), {predict} the action which will be performed with that object (e.g.,``take''), {and} the time to contact (e.g., $0.75s$).

\subsubsection*{Long-Term Action Anticipation}
\begin{figure*}[t]
    \centering
    \includegraphics[width=\linewidth]{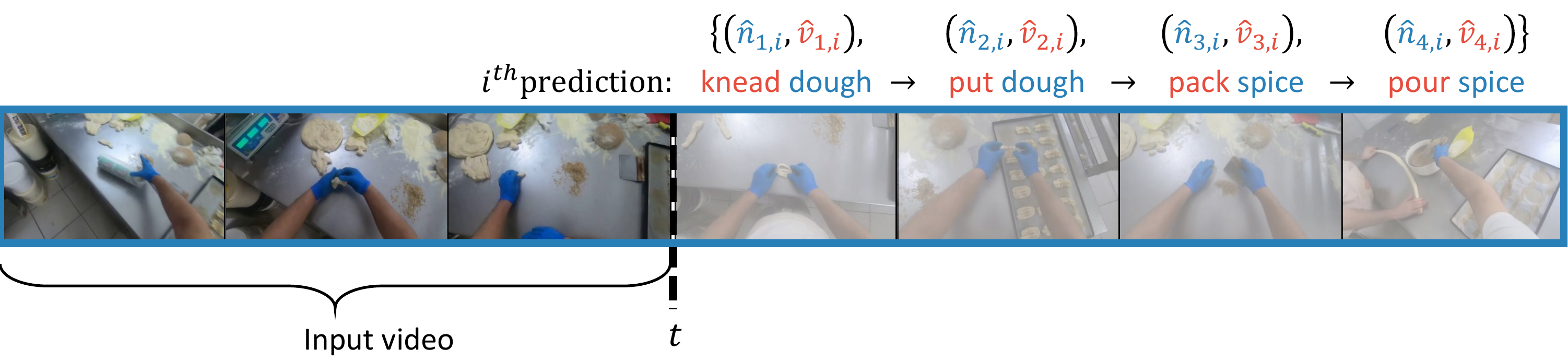}
    \caption{\textbf{Example of long-term action anticipation.} After observing a video up to a particular timestep $t$, a method should be able to predict the sequence of actions that will likely occur, in the correct order (e.g., first ``take dough'', next ``put dough'' etc.) 
    }
    \label{fig:long_term_anticipation}
\end{figure*}

Long-term action anticipation aims to predict further into the future. Rather than predict the next action at a given timestamp, models will be required to predict the sequence of $Z$ future actions which the camera-wearer is likely to perform. This is important for long-horizon planning where a sequence of actions is required to be performed in a specific order to achieve a goal. Critically, these actions occur over long time horizons, may be of variable length and do not occur uniformly across time (e.g., an action every 5s). Thus, the task is defined at a more abstract level --- models are required to predict sequences of action classes (verb and noun), rather than time to action or to next active objects bounding boxes in the current frame.

More formally, given an untrimmed video $V$ and a \emph{stopping time} $t$ as described above, the long-term action anticipation model must observe $V_{:t}$ and predict $N$ sets of sequences of $Z$ plausible future actions:
\begin{equation}
\{\{(\hat n_{z,i}, \hat v_{z,i})\}_{z=1}^{Z}\}_{i=1}^{N}
\end{equation}
where:
\begin{itemize}
    \item $\hat n_{z,i} \in \mathcal{N}$ is the predicted noun and $\hat v_{z,i} \in \mathcal{V}$ is the predicted verb of the $z$-th future action.
    \item $\{(\hat n_{z,i}, \hat v_{z,i})\}_{z=1}^{Z}$ represents the sequence of future actions sorted by the predicted order in which they will appear in the video.
\end{itemize}

Like the short-term object interaction anticipation task, the model is allowed to generate $N$ sets of predictions to account for the multi-modal nature of future prediction. Figure~\ref{fig:long_term_anticipation} illustrates the proposed task.

\subsubsection{Data Selection}
\label{appendix:forecasting_data}

\subsubsection*{Future Locomotion Movements Prediction}
Egocentric videos for locomotion and hand-object interaction are nearly mutually exclusive. Among these videos, we skim through each video to manually identify video clips (beginning and end frames) that satisfy the following selection criteria. (1) Locomotion, by definition, involves diverse activities associated with walking. The clip should include substantial translational movement. (2) Each video clip must be longer than 10 seconds for past trajectory observation and future prediction. (3) The videos must observe surrounding scenes. This differs from the videos for hand-object interaction where the camera is deliberately tilted down to focus on the hand manipulation. We consider videos from glass-mounted cameras of which field of view approximately aligns with the first person. (4) 3D reconstruction and ground plane need to be accurate. After running structure from motion, we ensure 3D reconstruction from the videos achieves reasonable quality by checking 2D reprojection of the point cloud and ground plane. Given a set of these video clips, we choose frames for training/testing data for every second. 

\subsubsection*{Remaining Tasks}

For the remaining tasks we first manually ranked the scenarios based on their applicability to the forecasting tasks. For instance, scenarios like carpentery were high priority for forecasting whereas walking in the park was low-priority. We scored all scenarios from 1-3 based on this priority. We impose constraints on the minimum number of hours and participants to sub-select scenarios that have sufficient data for training (each participant should have contributed at least 15 minutes; and there should be at least 20 minutes of videos for that scenario).
Next, we chunk our videos into 5 minute clips, and use the following algorithm to select clips to be labeled. 
To ensure geographical diversity, we distribute the total hours over universities and randomly select clips from each to fill the hours allocated to that university. If there are universities that contributed less, then their hours are distributed across the other universities. To select the clips given a university and the hours allocated; we would first sample a participant, then sample a video for that participant, and sample 1 clip from that video. For certain repetitive scenarios (like brick making), we reject this clip if we already have selected at least 2 clips from the same video. We repeat the process until the required number of hours are selected.

\subsubsection{Data Annotation}
\label{appendix:forecasting_data_annotations}
\subsubsection*{Future Locomotion Movements Prediction}

\begin{figure*}[t]
  \centering  
      \subfigure[Geometry]{\label{Fig:notation}\includegraphics[height=0.18\textheight]{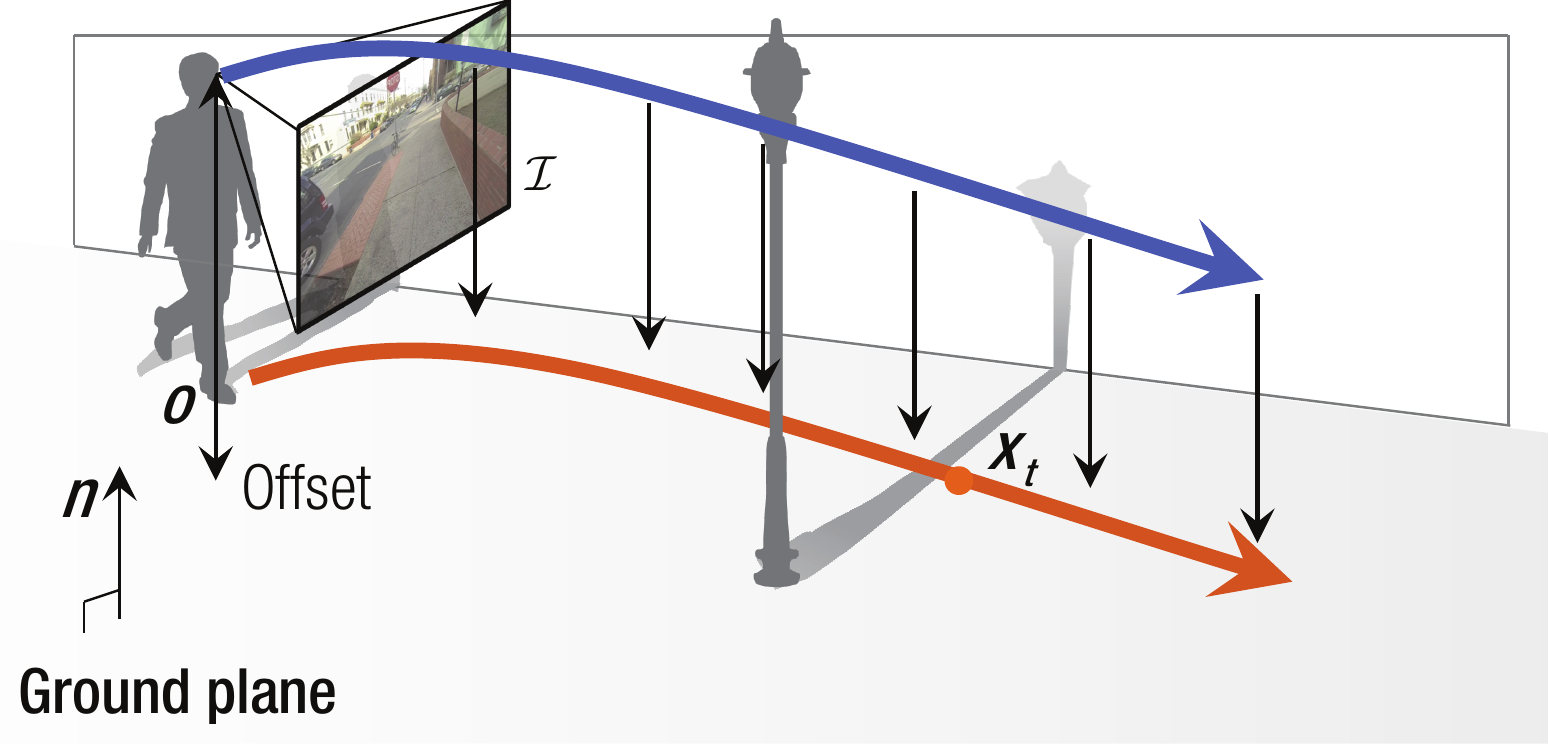}}~~~~~
      \subfigure[Future trajectory prediction]{\label{Fig:depth}\includegraphics[height=0.18\textheight]{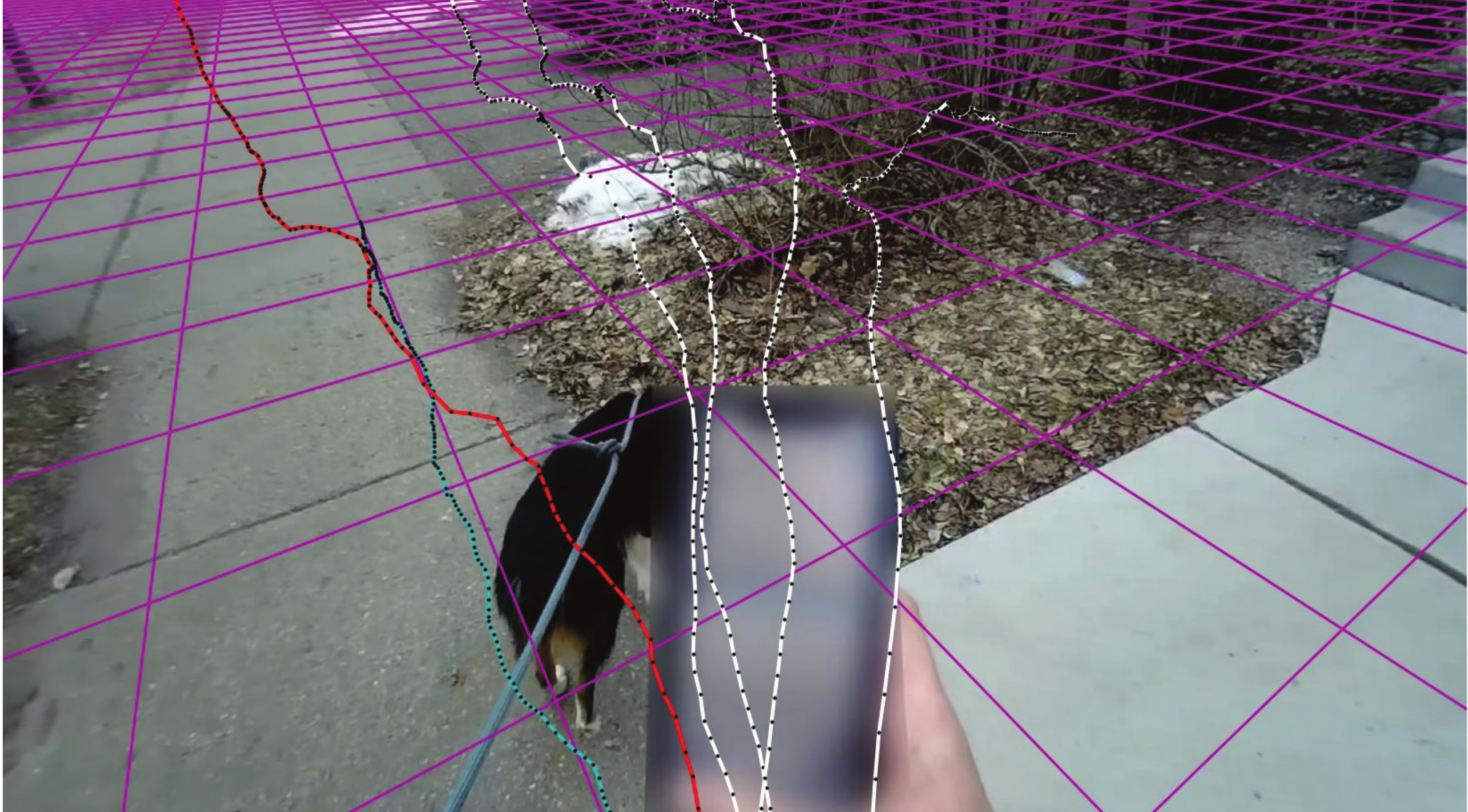}} 
  \caption{(a) We represent the future trajectory of a person using the ground plane. Given the 3D reconstruction of the camera trajectory, we project it into the estimated ground plane to form the future trajectory. (b) The ground truth future trajectory (blue) and the predicted trajectories (red and white) are shown in the egocentric image with the ground plane coordinate (magenta grid). We predict top 5 trajectories where the top prediction is marked in red.} 
  \label{Fig:PTP}
\end{figure*}

We generate the ground truth of future trajectories using 3D reconstruction of the camera trajectories. Given a sequence of egocentric images, we reconstruct the 3D egomotion and scene geometry using a standard structure from motion pipeline with a few modification to handle a large number of images. With the 3D scene point cloud, we estimate the ground plane using RANSAC with the ground plane normal prior. The 3D reconstructed camera trajectory is projected onto the ground plane to form the 2D future trajectory as shown in Figure~\ref{Fig:PTP}.

\srt{Our image dataset includes locomotion in outdoor, indoor, and mixed scenes. We split the image data into training/validation/testing sets with approximately 70\%/15\%/15\%, respectively. The ratio across scenes does not exactly match because the split is performed based on the (anonymous) participant ID. The summary of the data split can be found in Table~\ref{tab:locomotion_data}.}

\begin{table}[t]
    \centering
    \footnotesize
    \begin{tabular}{l|cccc}
    \hline
    Data & Outdoor & Indoor & Mixed & Total \\
    \hline
    Train & 34.1k & 0.41k & 16.7k & 51.3k\\
    Val & 7.5k & 0.23k& 6k & 13.9k\\
    Test & 7.4k & 0.18k & 3k & 10.6k\\
    \hline
    \end{tabular}
    \caption{We split the image data for locomotion prediction based on scenes that including outdoor, indoor, and mixed.}
    \label{tab:locomotion_data}
\end{table}

\subsubsection*{Future Hands Movements Prediction}

For the the future hand position and trajectory prediction, the annotation will be performed by labeling bounding boxes around hands in the frame in which
the user touches the active objects as well as in frames preceding each object interactions. Hands bounding boxes will be associated to a label useful to distinguish among left and right hands. Therefore, given an object interaction, we will annotate key frames preceding the beginning of the interaction. Specifically, $t_c$ and $t_p$ denote the time step of contact frame and pre-condition frame, and $t_{p_1}$, $t_{p_2}$, $t_{p_3}$, denote time steps $0.5s$, $1s$ and $1.5s$ before the pre-condition time step. Therefore, for each interaction there will be 5 key frames labeled with bounding boxes of hands, including the contact frame. We use the bouding box center as the ground truth of hands positions.

\subsubsection*{Short-Term Object Interaction Anticipation} \label{sec:annotation_object_interaction_short}
\begin{figure}
    \centering
    \includegraphics[width=\linewidth]{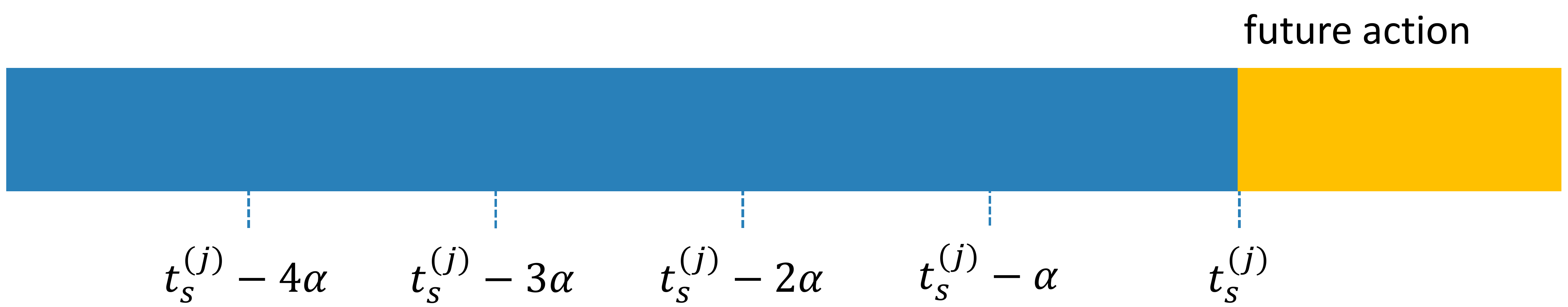}
    \caption{An example of how frames are sampled to be labeled with next active object annotations. For a given action $i$, we sample $m$ frames at regular intervals $\alpha$. If we set $m=4$ and $\alpha=0.5$, we label the frame of contact as well as $4$ frames along a segment of $2s$ preceding the beginning of the action at a framerate of $2fps$.}
    \label{fig:next_active_object_sampling}
\end{figure}

\begin{figure*}
    \centering
    \includegraphics[width=\linewidth]{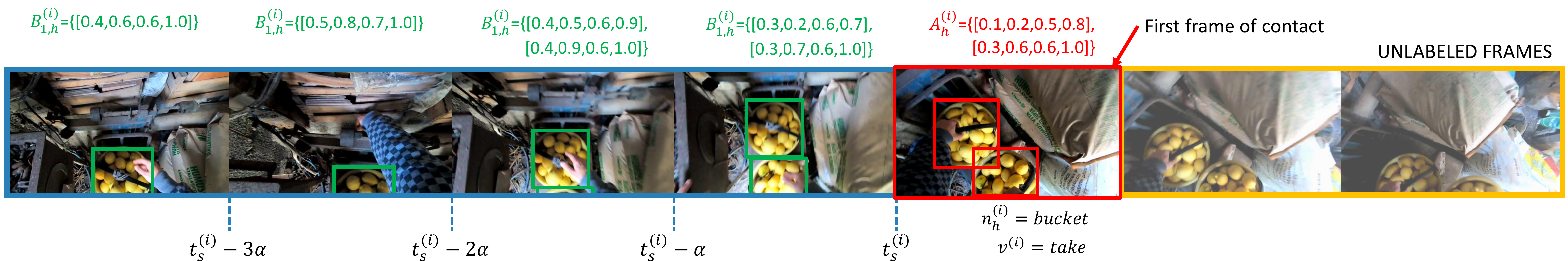}
    \caption{Example of annotations for the short-term object interaction anticipation task.}
    \label{fig:object_interaction_anticipation_labeling}
\end{figure*}

Each video $V$ of the dataset is labeled with a set of short term object interaction anticipation annotations $\mathcal{S}_V=\{S_V^{(j)}\}_j$ indicating the occurrence of object interactions in the video.
Each annotation
\begin{equation}
S_V^{(j)}=(t_s^{(j)}, \{n^{(j)}_h\}_h, v^{(j)}, \{A^{(j)}_h\}_h,
\{B^{(j)}_h\}_h)    
\end{equation} 
includes: 
\begin{itemize}
    \item $t_s^{(j)}$: the timestamp indicating the beginning of the interaction with the active objects. This is the first frame in which the user touches at least one of the active objects; 
    \item $\{n^{(j)}_h\}_h$: the set of categories of the $h$ interacted objects;
    \item $v^{(j)}$: the class of the action involving the active objects;
    \item $\{A^{(j)}_h\}_h$: the bounding box annotations for the active objects. The cardinality of $\{A^{(j)}_h\}_h$ is equal to the cardinality of $\{n_h^{(j)}\}$, i.e., $|\{A^{(j)}_h\}_h|=|\{n_h^{(j)}\}|$. The $h^{th}$ set $\{A^{(j)}_h\}_h$ contains bounding box annotations for the active objects of category $n_h$ at timestamp $t_s^{(j)}$;
    \item $\{B_h^{(j)}\}_h$: the bounding box annotations for the next active objects. The cardinality $\{B_h^{(j)}\}_h$ is equal to the cardinality of $\{A^{(j)}_h\}_h$, i.e., $|\{B_h^{(j)}\}_h| = |\{A^{(j)}_h\}_h|$. The $j^{th}$ set $B_h^{(j)}$ contains the bounding box annotations of next active objects of class $n_h$. 
    In particular, $B_h^{(j)}$ contains annotations for the same object instances annotated in $A^{(j)}_h$, tracked in frames preceding $t_s^{(j)}$. 
    Specifically, $B^{(j)}_h = \{B_{l,h}^{(j)} | l=1,...,m\}$, where $B_{l,h}^{(j)}$ is the set of bounding box annotations of next active object of class $n_h$ annotated at timestamp $t_s-l \alpha$. 
    Here $m$ indicates the number of frames preceding the beginning of the interaction in which objects are annotated, whereas $\alpha$ is the temporal distance between the sampled frames. 
    For instance, setting {$\alpha=0.5s$} and {$m=4$}, we will label the frame in which the object is interacted as well as {$4$ frames in a $2s$ segment preceding the interaction.}
    \figurename~\ref{fig:next_active_object_sampling} shows an example of how frames are sampled with the considered scheme.
\end{itemize}
Figure~\ref{fig:object_interaction_anticipation_labeling} reports a sample clip with the discussed annotations. The timestamp $t_s$ is selected as the first one in which the user touches the active objects. The frames following this timestamp are not labeled. Active object bounding boxes are labeled at timestamp $t_s$, whereas next active object bounding boxes are labeled in frames preceding $t_s$.

\begin{figure*}
    \centering
    \includegraphics[width=\linewidth]{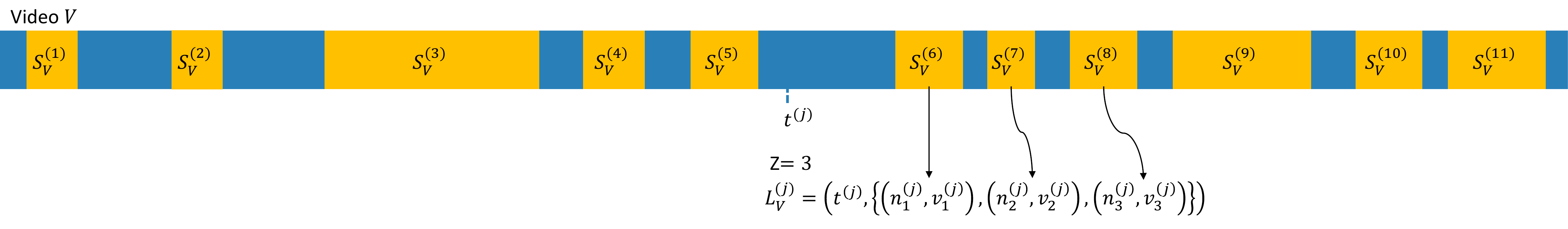}
    \caption{An example of a long-term annotation $L^{(V_{:t})}$ for an untrimmed video $V$ at timestamp $t$ can be obtained from short-term annotations $S_V^{(i)}$. In the example, $Z=3$, hence the long term annotation is obtained by considering the first three actions beginning after timestamp $t$. }
    \label{fig:next_active_object_sampling_long_term}
\end{figure*}

\subsubsection*{Long-Term Action Anticipation}

Each video $V$ is labeled with a set of long-term action annotations $\{L_V^{(j)}\}_j$, corresponding to a \emph{stopping time} until which the video can be observed, and a sequence of $Z$ future action labels defined as follows:
\begin{equation}
    L_{V}^{(j)}= ( t^{(j)}, \{(n_z^{(j)}, v_z^{(j)})\}_{z=1}^Z )
\end{equation}
where: 
\begin{itemize}
    \item $t^{(j)}$: the timestamp until which the video can be observed (i.e., $V_{:t^{(j)}}$) before making predictions of future actions;
    \item $n_z^{(j)}$: the noun category of the primary interacted object in the $z$-th future action;
    \item $v_z^{(j)}$: the verb describing how the objects will be interacted with in the $z$-th future action .
\end{itemize}
For each video, $t^{(j)}$ are selected from the \emph{last} timestamp of each annotated object interaction. It is worth noting that once short-term annotations $S_V^{(i)}$ are available (see Section \ref{sec:annotation_object_interaction_short}) and a value for $Z$ has been chosen, the long-term annotations $L_V^{(j)}$ can be easily obtained by sampling the first $Z$ actions annotated in video $V$ beginning after timestamp $t^{(j)}$. More formally, the future action labels for $L_V^{(j)}$ are obtained as: 

{\small
\begin{align}
\{(n^{(i_z)}_0, v^{(i_z)})\ | (t_s^{(i_z)}, \{n^{(i_z)}_h\}_h, v^{(i_z)},  \{A^{(i_z)}_h\}_h, \{B^{(i_z)}_h\}_h) \in \mathcal{S}_V \land \nonumber\\
t_s^{(i_z)} \geq t^{(j)} \land \nonumber\\
t_s^{(i_1)} \leq \ldots \leq t_s^{(i_Z)} \land \nonumber \\
 \nexists S_V^{(j)} \in \mathcal{S}_V | t^{(j)} \notin \{i_1,\ldots, i_Z\}, t \leq t_s^{(j)} < t_s^{(i_Z)}
\}_{z=1}^Z  \nonumber
\end{align}}
where $n^{(i_z)}_0$ refers to the primary interacted object from the set of interacted objects $\{n^{(i_z)}_h\}_h$. Figure~\ref{fig:next_active_object_sampling_long_term} illustrates an example of how long-term annotations are obtained form short-term annotations.

\subsubsection*{Annotation analysis}\label{sec:annot_analysis}

\noindent
\textit{Dataset statistics} 
As discussed earlier, one of our primary objectives when selecting the data to annotate was to maximize the diversity in terms of activities and geographic locations. Our dataset includes scenarios spanning a wide range of everyday activities (\eg, gardening, cleaning, fishing, \etc). In addition to diversity across scenarios, there is also geographic diversity within scenarios. For example, cooking may look very different in Italy, India, Saudi Arabia, or Japan. In Figure~\ref{fig:data_stats}, we show the resulting scenario and university distributions. Overall, our benchmark consists of \FEB{120 hours of annotated video coming from 53 scenarios, 7 universities, and 406 participants.}\\

\noindent
\textit{Temporal structure of activities} 
Human activity is goal-driven and structured over time, with certain action sequences being favored over others. We measure this temporal structure using Normalized Pointwise Mutual Information (NPMI)~\cite{church1990word} over pairs of actions following prior work~\cite{gu2018ava}. NPMI is a measure of how likely actions follow each other. In our dataset, typical patterns include ``pull grass $\rightarrow$ throw grass (0.87)'', ``hold spinach $\rightarrow$ cut spinach (0.83)'', ``turn-on faucet $\rightarrow$ turn-off faucet (0.68)'', ``take cloth $\rightarrow$ fold cloth (0.49)'' etc. Several actions also occur in sequence with high NPMI scores due to the repetitive nature of the activity. For example, ``flip page $\rightarrow$ flip page (0.83)'' while reading, or ``cut carrot $\rightarrow$ cut carrot (0.82)'' while cooking. Finally, we see common action sequences involving multiple objects like ``fill tire $\rightarrow$ close valve (0.89)'', or ``lift vacuum-cleaner $\rightarrow$ clean staircase (0.87)''. This structure is valuable and can inform long-term action anticipation models. 

\textit{Dataset split} To facilitate future research and comparisons, we construct training, validation, and test splits containing 40$\%$, 30$\%$, and 30$\%$ of the data, respectively. We note, however, that we do not release the ground truth annotations for the test set. Following common practice, evaluation on the test set will be supported through the public evaluation server and leader board. We assign data to splits randomly at the level of 5 minute clips. This ensures that all interactions within a 5 minute clip were labeled by an annotator and provides enough temporal context for long-term video tasks, like long-term action anticipation.\\

\subsubsection{Evaluation measures}
\label{appendix:forecasting_measures}

\subsubsection*{Future Locomotion and Hands Movements Prediction}

\noindent
\textit{Future Locomotion} We measure the accuracy of the prediction using two metrics. (1) K best mean trajectory error (K-MTE): we measure K best trajectory error:
\begin{align}
    {\rm K-MTE} = \underset{\{\mathcal{X}_k\}_{k=1}^K}{\operatorname{argmin}}~\frac{1}{\sum_t v_t} \sum_{t} v_t\|\mathbf{x}_t - \widehat{\mathbf{x}}_t\|,
\end{align}
$\mathbf{x}_t\in \mathds{R}^2$ is the predicted location at time $t$, $\widehat{\mathbf{x}}_t$ is the ground truth location, and $v_t$ is the visibility. The visibility indicates the availability of the ground truth trajectory, i.e., due to severe egocentric videos, the ground truth trajectories may include missing data. $v_t=0$ indicates missing data at time $t$. (2) Probability of correct trajectory (PCT): we measure the success rate of the correct trajectory retrieval:
\begin{align}
    {\rm PCT}{\epsilon} = \frac{1}{K}\delta \left(\frac{1}{\sum_t v_t} \sum_{t} v_t\|\mathbf{x}_t - \widehat{\mathbf{x}}_t\| < \epsilon\right),
\end{align}
where $\delta(\cdot)$ is one if the statement is true and zero otherwise. $\epsilon$ is the trajectory error tolerance, i.e., if the trajectory error is smaller than the error tolerance, it is considered as a correct trajectory prediction. ${\rm PCT}{\epsilon}$ measures how many trajectories among $K$ retrieved trajectories are close to the ground truth trajectory. \\

\noindent
\textit{Future Hand Movement} 
As for the future hands movements prediction, we only consider the key frame prediction, and therefore adopt Mean Key Frame Displacement Error Contact (M.Disp.) Key Frame Displacement Error as evaluation metrics (C.Disp.): 

\begin{itemize}
    \item {
Mean Key Frame Displacement Error (M.Disp.): 
\begin{equation}
\small
D_{m} =  \dfrac{1}{n} \sum_{i\in H_t} \ \Vert h_i - \hat{h}_i\Vert
\label{eq:meandisp}
\end{equation}
\noindent $H_t$ refers to the set of visible hand positions of key frames, and $n$ is the length of set $H_t$. $h_i$ denotes the predicted hand position in the image coordinate, while $\hat{h}^i$ denotes the ground truth hand positions.
    }
    
    \item{
    Contact Key Frame Displacement Error (C.Disp.): 
\begin{equation}
\small
D_{c} =  \Vert h_c -\hat{h}_c\Vert
\label{eq:contactdisp}
\end{equation}
$h_c$ refers to the hand positions at Contact frame.
    }
\end{itemize}
Note that all reports are reported on downsampled video frames with height of 256 and original aspect ratio.

\subsubsection*{Short-Term Object Interaction Anticipation} \label{sec:short_term_evaluation}
Methods will be evaluated at the timestamps in which next-active objects have been annotated, i.e.,
\begin{eqnarray}
{\Big\{t | t }&=&{ t_s - l\cdot \alpha\ \nonumber} \\
&& {\forall t_s \in \{t_s^{(j)} | \exists \overline h : B_{\overline h}^{(j)} \neq \emptyset\}_j} \nonumber \\
&&{\forall l \in \{1,...,m\}\Big\}}
\end{eqnarray}
where ${\{t_s^{(j)} | \exists \overline h : B_{\overline h}^{(j)} \neq \emptyset\}_j}$ {is the set of all timestamps indicating the beginning of an interaction, for which at least one next active object has been annotated, }
 and $\alpha$ and $m$ are defined in Appendix~\ref{appendix:forecasting_data_annotations}.

{Since detecting next active objects is a major part of the task, we base our evaluation measures on mean Average Precision (mAP), as defined in the Pascal VOC challenge~\cite{everingham2010pascal}. As in standard mAP, we first match each of the detected next active objects to ground truth annotations. A predicted and a ground truth bounding boxes are a possible match if their Intersection Over Union (IOU) value exceeds $0.5$ and if some matching criteria are met. We will define matching criteria later. Predictions are matched to ground truth annotations belonging to the same evaluated example in a greedy fashion, prioritizing predictions with higher confidence scores and choosing matches corresponding to larger IOU values. A ground truth annotation can be matched at most with one predicted box. All matched predictions are counted as true positives, whereas all unmatched predictions are counted as false positives. Performance on the whole test set is summarized using the mean of the Average Precision values obtained for each class. 

To account for the multi-modal nature of future predictions (i.e., more than one next active object can be likely), we ``discount'' the number of false positives obtained in a given example by the number of available ground truth annotations in that example multiplied by $K-1$, where $K$ is a parameter of the evaluation measure. Specifically, if an example contains two ground truth annotation, we ignore the $(K-1)*2$ false positives with the highest scores. This effectively implements a ``Top-K mean Average Precision'' criterion which does not penalize methods for predicting up to $K-1$ possibly likely next active objects which are not annotated. Given a generic prediction $(\hat b_i, \hat n_i, \hat v_i, \hat \delta_i \hat s_i)$ and a generic ground truth annotation $(b_j, n_j, v_j, \delta_j)$, we define the following variants of this Top-K evaluation measure considering different matching criteria:}
\begin{itemize}
    \item Noun Top-K mAP: prediction $i$ and annotation $j$ are a possible match if the following conditions are satisfied:
    \begin{itemize}
        \item[*] $IOU(\hat b_i, b_j)>0.5$;
        \item[*] $\hat n_i = n_j$;
    \end{itemize}
    \item Noun + Verb Top-K mAP: prediction $i$ and annotation $j$ are a possible match if the following conditions are satisfied:
    \begin{itemize}
        \item[*] $IOU(\hat b_i, b_j)>0.5$;
        \item[*] $\hat n_i = n_j$;
        \item[*] $\hat v_i = v_j$.
    \end{itemize}
    \item Noun + TTC Top-K mAP: prediction $i$ and annotation $j$ are a possible match if the following conditions are satisfied:
    \begin{itemize}
        \item[*] $IOU(\hat b_i, b_j)>0.5$;
        \item[*] $\hat n_i = n_j$;
        \item[*] $|\hat \delta_i - \delta_j|<T_{\delta}$.
    \end{itemize}
    \item Overall Top-K mAP: prediction $i$ and annotation $j$ are a possible match if the following conditions are satisfied:
    \begin{itemize}
        \item[*] $IOU(\hat b_i, b_j)>0.5$;
        \item[*] $\hat n_i = n_j$;
        \item[*] $\hat v_i = v_j$;
        \item[*] $|\hat \delta_i - \delta_j|<T_{\delta}$.
    \end{itemize}

    Where $T_{\delta}$ is a tolerance threshold, parameter of the evaluation measure.
    
\end{itemize}
{The goal of the different measures is to assess the ability of the model to predict next object interactions at different levels of granularity. We use $K=5$ and $T_{\delta}=0.25$.}

\subsubsection*{Long-Term Action Anticipation}

\label{sec:long_term_evaluation}

Methods will be evaluated at the set of timestamps specified by the end of each annotated object interaction in a video $V$. Let $L_V^{(j)}=\{(n_z^{(j)}, v_z^{(j)})\}_{z=1}^Z$ be the ground truth annotation related to video $V$ at time-stamp $t^{(j)}$ and let $\{\{(\hat n_{z,k}^{(j)}, \hat v_{z,k}^{(j)})\}_{z=1}^Z\}_{k=1}^K$ be the $K$ predicted sequences of $Z$ actions.
We will consider single noun/verb/action predictions correct following the definitions discussed in Section~\ref{sec:short_term_evaluation}. The $K$ predicted sequences will hence be evaluated using the edit distance metric as follows.

For a given $k$, this is obtained by evaluating the edit distance between a predicted sequence and the ground truth sequence of future actions. The edit distance \begin{eqnarray}
\Delta_{E}(\{(\hat n_{z,k}^{(j)}, \hat v_{z,k}^{(j)})\}_{z=1}^Z, \{(n_z^{(j)}, v_z^{(j)})\}_{z=1}^Z) \nonumber
\end{eqnarray} 
is computed as the Damerau-Levenshtein distance~\cite{levenshtein1966binary,damerau1964technique} over sequences of predictions of verbs, nouns and actions.  The goal of this measure is to assess performance in a way which is robust to some error in the predicted order of future actions. 
A predicted verb/noun is considered ``correct'' if it matches the ground truth verb label at a specific time-step.  
The allowed operations to compute the edit distance are insertions, deletions, substitutions and transpositions of any two predicted actions.
Following the ``best of many'' criterion, the $K$ predictions are evaluated considering the smallest edit distance between the ground truth and any of the $K$ predictions: 
\begin{eqnarray}
&\Delta_{E}(\{\{(\hat n_{z,k}^{(j)}, \hat v_{z,k}^{(j)})\}_{z=1}^Z\}_{k=1}^K,
\{(n_z^{(j)}, v_z^{(j)})\}_{z=1}^Z) =\nonumber \\
& \min\limits_{k=1..K} \Delta_{E}(\{(\hat n_{z,k}^{(j)}, \hat v_{z,k}^{(j)})\}_{z=1}^Z, \{(n_z^{(j)},  v_z^{(j)})\}_{z=1}^Z)\nonumber
\end{eqnarray}

Note that we consider edit distance over simple accuracy based measures. Treating predictions for each future time-step independently and calculating accuracy does not account for the sequential nature of the prediction task where the order of predictions is important. We evaluate each metric independently for verbs, nouns and actions (verb and noun together). We report edit distance at $Z=20$ (ED@20) and use $K = 5$ in our experiments. We select $Z = 20$ as baselines begin to predict actions at random for higher values of $Z$.

\subsubsection{Baseline definitions and implementation details}
\label{appendix:forecasting_baselines}

\subsubsection*{Future Locomotion Movements Prediction}

We make use of the method by Park et al.~\cite{park-cvpr2016} for a baseline algorithm. The method models the trajectory prediction function in Equation~(\ref{Eq:locomotion}) using KNN classification with CNN image encoding, i.e., 
\begin{align}
    \{\mathcal{X}\} = KNN\left(\{\phi(\mathcal{I}_i)\}, \phi(\mathcal{I})\right)
\end{align}
where $KNN(A,B)$ finds the K nearest neighbor of $B$ given the set $A$, and $\phi(\mathcal{I})\in \mathds{R}^n$ is a function that extracts the image feature of $\mathcal{I}$. We use the AlexNet image feature extractor for $\phi$.

Notably, the baseline algorithm leverages a polar coordinate system to represent the trajectory, i.e., $\mathbf{X}_{j}^{2D}=\left[\begin{array}{ccccc}r_{j}&\theta_{j}\end{array}\right]^\mathsf{T}$ is a 2D trajectory on the ground plane where $r_i$ and $\theta_i$ are the polar coordinates of the trajectory represented in the egocentric coordinate system, i.e., distance (radial) and direction (angle) with respect to the person's feet location as shown in Figure~\ref{Fig:PTP}:
\begin{align}
    \mathbf{X}_{j}^{2D} = \mathtt{cart2polar} (\mathbf{r}_1^\mathsf{T}\mathbf{X}_j, \mathbf{r}_2^\mathsf{T}\mathbf{X}_j)
\end{align}
where $\mathbf{r}_1$ and $\mathbf{r}_2$ are the two spanning vectors of the ground plane that are aligned with the rotation matrix $\mathbf{R}_t$. $\mathbf{r}_1$ is the facing direction and $\mathbf{r}_2$ is lateral direction. Both are perpendicular to the ground plane normal $\mathbf{n}$ as shown in Figure~\ref{Fig:PTP}. $\mathtt{cart2polar}$ is a coordinate transform from cartesian to polar coordinates.

\subsubsection*{Future Hands Movements Prediction }

\noindent
\textit{Baseline Description}
The proposed future hand movement prediction task can be factorized as a regression problem. To address this task, we adopt a baseline that utilizes the I3D network as the backbone to extract the spatial-temporal video representations of the input video sequence, and then use a linear mapping function as the regressor to predict the future keyframe hand positions. 
We adopt the smoother l1 loss as the objective function:
\begin{equation}
  L_{h} =
    \begin{cases}
      0.5*w*(h-\hat{h})^2/\beta, & \text{if} \: \vert h-\hat{h} \vert < \beta\\
      w*(\vert h-\hat{h} \vert -0.5*\beta), & \text{otherwise} \\
    \end{cases}    
\label{eq:handant_loss}
\end{equation}
where $h\in R^{20}$ is a vector that represents the x,y coordinates of both left and right hands in the aforementioned five future key frames. If the hand is not observed in the keyframe, we pad 0 into the $\hat{h}$, and adopt a binary mask $w$ to prevent the gradients propagation of these unobserved instances.\\

\noindent
\textit{Training Details}  
We adopt the I3D model as the backbone network and a regression header, composed of two linear operations, to predict the hand positions in the future key frames. For our experiments, we set observation time $T_o$ as $2$s.  For training, we applied several data augmentation techniques, including random flipping, rotation, cropping and color jittering to avoid overfitting. Our baseline model was trained with a batch size of $64$ for $25$ epochs using a cosine learning rate decay with a initial learning rate of $0.0375$. We set $\beta$ to 5 in the weighted smoothed L1 loss as introduced in Eq.~\ref{eq:handant_loss}.

\subsubsection*{Short-Term Object Interaction Anticipation }

\begin{figure}
    \centering
    \includegraphics[width=\linewidth]{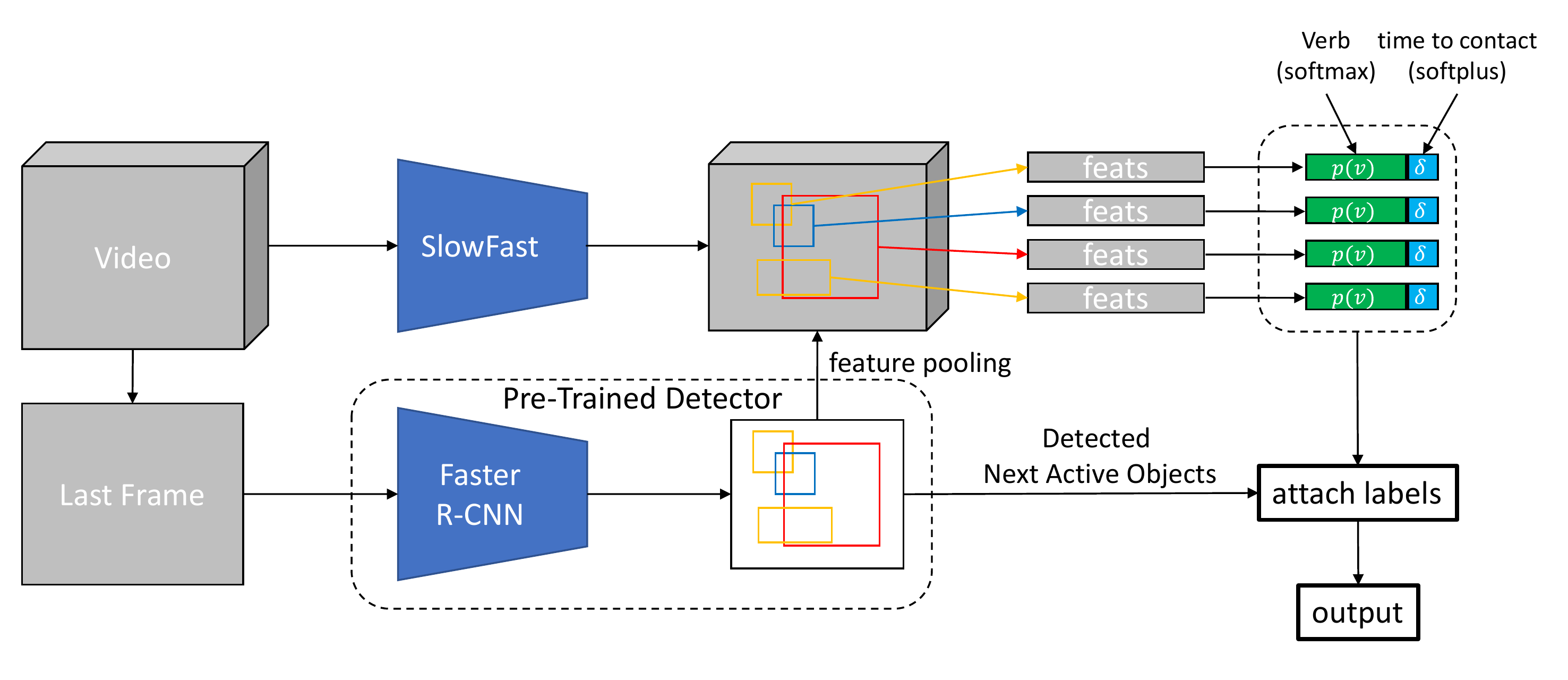}
    \caption{Short-Term object interaction anticipation baseline.}
    \label{fig:short_term_baseline}
\end{figure}
\noindent

\textit{Data and annotations used for the experiments}
We performed our experiments on a subset of the data and annotations to obtain verb and noun taxonomies consistent with the Short-Term Object-Interaction Anticipation task. We started by considering all annotated actions for which a contact frame has been specified by the annotators. Note that these constitute about $30\%$ of the whole set of annotated actions and that the notion of a contact frame is fundamental to our task.
We then gathered all annotated frames and referenced them to their respective contact frames, computing the time to action targets.
We discarded all those annotations which comprised a verb or a noun class marked by the annotator as ``null''.
We further discarded annotations related to nouns which had been labeled inconsistently and non-object classes such as ``wall'' or ``wallpaper''.
We similarly removed all annotations related to the verb ``talk'' which do not involve interactions with objects.

To avoid having an over-specific noun taxonomy, we clustered selected noun classes into homogeneous groups. For instance the nouns ``okra'', ``apple'', ``celery'' and ``avocado'' have all been grouped under the ``vegetable\_fruit'' class.
We also grouped verbs which have similar semantic when anticipated. For instance, the verbs ``take'', ``carry'', ``lift'', ``pull'' and ``remove'' have all been grouped in the ``take'' cluster. Note that while these actions may be visually different, they all have similar effects on objects, which makes them indistinguishable when anticipated.
We further removed all annotations related to nouns appearing less than $50$ times in the test set (we follow the common split defined for this benchmark). 
We choose to retain only nouns appearing at least $50$ times in the test set to allow for a reliable evaluation through the mAP measure.

\begin{figure*}
    \centering
    \includegraphics[width=\linewidth]{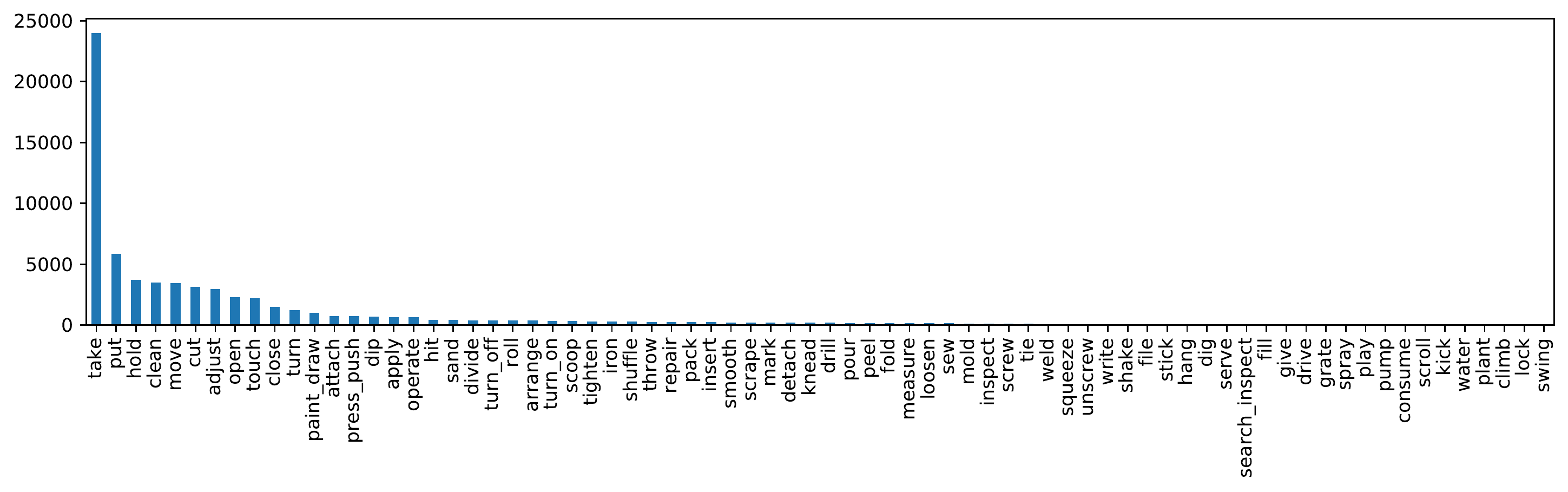}
    \caption{Verb distribution in the Short-Term Object-Interaction Anticipation data. }
    \label{fig:sta_verb_distribution}
\end{figure*}

\begin{figure*}
    \centering
    \includegraphics[width=\linewidth]{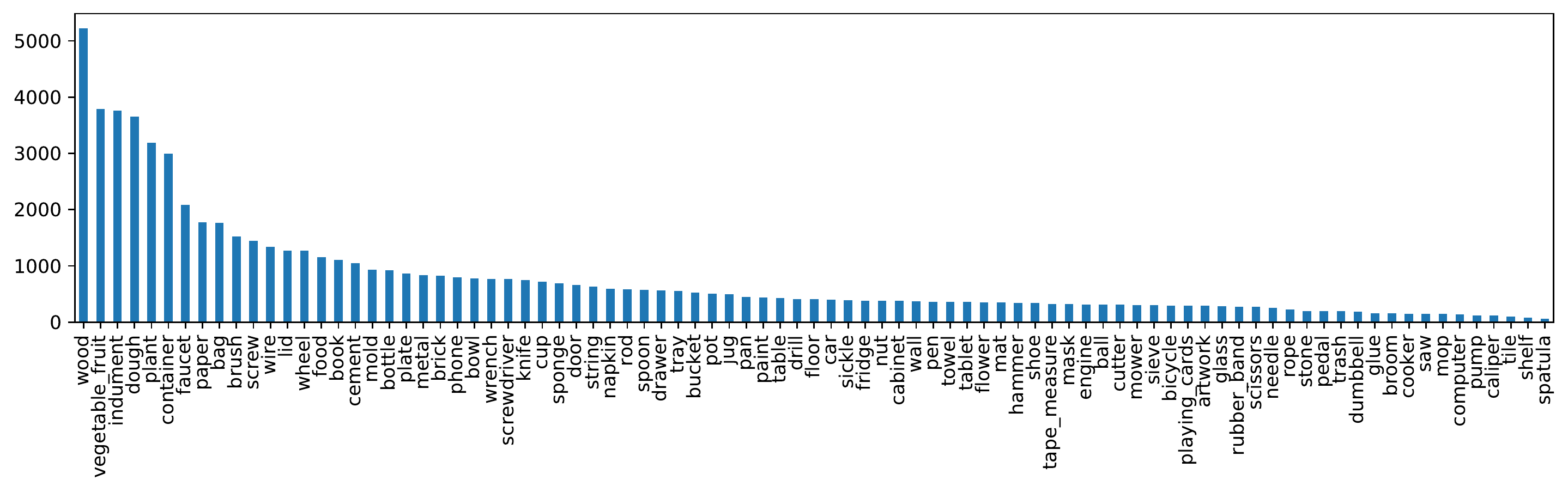}
    \caption{Noun distribution in the Short-Term Object-Interaction Anticipation data.
    }
    \label{fig:sta_noun_distribution}
\end{figure*}

The final set of data includes \FEB{$64,798$} annotated examples in total with \FEB{$87$} nouns and \FEB{$74$} verbs. Our taxonomy is adapted from the one presented in Figure~\ref{fig:taxonomy}. Figure~\ref{fig:sta_verb_distribution} and Figure~\ref{fig:sta_noun_distribution} report the distributions of verb and noun annotations in the selected data. Among the \FEB{$64,798$} annotations, \FEB{$27,801$} are in the training set, \FEB{$17,217$} are in the validation set, and \FEB{$19,780$} are in the test set. \\

\textit{Baseline Description}
Figure~\ref{fig:short_term_baseline} illustrates the proposed baseline for short-term object interaction anticipation.
The baseline includes two main components. 
A Faster R-CNN object detector~\cite{girshick2015fast} is used to detect next active objects in the last frame of the input video clip processed at full resolution.
A SlowFast 3D CNN~\cite{feichtenhofer2019slowfast} is hence used to predict a verb label and a time to action for each predicted object. 
This is done by obtained a fixed-length representation of each object through ROI pooling~\cite{girshick2015fast}. 
Two linear layers are hence used to predict a probability distribution over verbs and a positive quantity for time to contact prediction respectively.
Verb probability distributions are obtained using a softmax layer, whereas a softplus activation is used for time to contact prediction to make sure that the prediction is a positive number.
The final output of the model is obtained by attaching the predicted verb and time to contact to each detected next active object.
The noun label and confidence scores are copied from the output of the Faster R-CNN component.\\

\noindent
\textit{Training Details} 
We first train the Faster R-CNN component on all frames with annotated next active objects. 
We use the Faster RCNN detector based on ResNet50 using the ``3x'' training schedule provided with the Detectron2 library\footnote{https://github.com/facebookresearch/detectron2}.
After this stage, the weights of the Faster R-CNN component are not updated anymore.
We hence train a SlowFast model based on ResNet50. We follow the configuration provided in the PySlowFast library\footnote{https://github.com/facebookresearch/SlowFast} to tackle the AVA detection task (``SLOWFAST\_32x2\_R50\_SHORT.yaml''). The SlowFast model takes as input video clips of 32 frames sampled with a temporal stride of $1$ frame. 
During training, we match each detected object to the ground truth instance with largest Intersection Over Union (IOU), provided that it is larger than $0.5$.
We hence attach the verb and time to contact labels of the ground truth boxes to the matched ones.
We then train the model applying the following loss only to boxes which have been matched to ground truth instances:
\begin{equation}
\mathcal{L}=\mathcal{L}_v + \lambda \mathcal{L}_{ttc}
\end{equation}
where $\mathcal{L}_v$ is the cross entropy loss for verb prediction, $\mathcal{L}_{ttc}$ is the smooth L1 loss~\cite{girshick2015fast} applied to time to contact prediction, and we set $\lambda=10$ to control the contributions of the two losses.
To regulate the number of frames processed by the slow branch, we set $\alpha=8$.
We train the model on $4$ NVIDIA V100 GPUs with a batch size of $64$ for $50$ epochs using a cosine learning rate policy with a base learning rate of $0.001$. We validate the model at the end of each epoch and consider the weights which achieved the best overall top-5 $mAP$ on the validation.

\subsubsection*{Long-Term Action Anticipation } \label{sec:long_term_baseline}

\noindent
\textit{Baseline Description} The goal of the baseline model is to take as input a trimmed video of arbitrary length, and predict $N$ different plausible sequences of future actions. The baseline models thus consist of three components: (1) the encoder backbone for obtaining clip level features, (2) the aggregation module for combining the obtained features from different clips, and (3) the decoder network for decoding the plausible sequences of future actions. For encoder backbones, we consider state of the art video recognition networks from both convolutional model, namely, \mbox{SlowFast}~\cite{feichtenhofer2019slowfast} and the newly proposed video transformer models, namely, MViT~\cite{fan2021multiscale}. For aggregation module, we experiment with simple concatenation operators that concatenates the obtained clip features from multiple input clips as well as transformer based self-attention modules. For the decoder networks we consider the following options: 
\begin{itemize}[leftmargin=*]
\itemsep0em 
    \item No Change: A simple recognition baseline that assumes no future change in the current action and simply predicts the \textit{currently} observed action as a duplicated static future sequence for $Z$ steps.
    \item MultiHead: This model trains $Z$ independent heads in parallel, one for each future time step. The final sequence is simply the conjoined predicted actions of each head. 
\end{itemize}

\begin{figure}
    \centering
    \includegraphics[width=\linewidth]{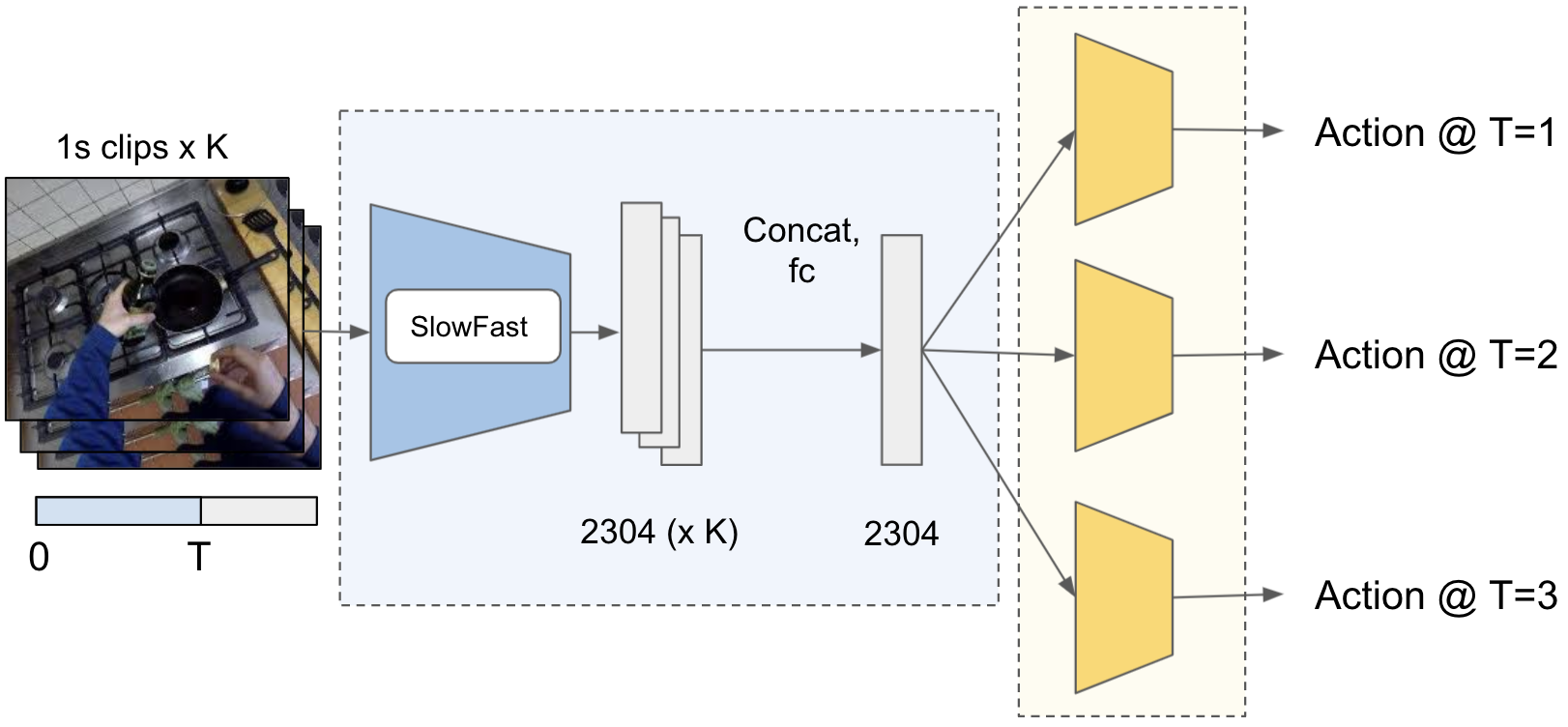}
    \caption{\textbf{Long-Term Action Anticipation baseline}. A baseline model with a SlowFast backbone, and $Z = 3$ is shown here. Blue box: clip encoder network. Yellow box: multiple classifier heads, one for each future action. See Sec.~\ref{sec:long_term_baseline} for more details.}
    \label{fig:long_term_baseline}
\end{figure}

Finally, to generate $N$ plausible future sequences for constructing multimodal baselines, we simply sample the predicted future action distribution $N$ times.   
The framework for a particular instantiation of the MultiHead baseline is illustrated in Figure~\ref{fig:long_term_baseline}.\\

\noindent
\textit{Training Details} 
For each video, we sample multiple input clips to process with our backbone network. A single clip length for both the backbones, SlowFast and MViT, comprises of $16$ frames sampled $4$ frames apart. Each clip is processed independently by the same encoder weights and combined with the aggregation module. The aggregated feature is decoded with the decoder module where the output behavior changes during training and testing. In training, the decoder predicts the next action probability distributions for each future step. We calculate the sum of losses for each prediction as our total loss:

\begin{equation}
\mathcal{L}_{lta} = \sum_{z=1}^Z \mathcal{L}_{v}((p^{n}_z, p^{v}_z), (n_z, v_z))
\end{equation}
where $\mathcal{L}_v$ is cross entropy loss, $p^{*}_z$ refers to the predicted probability distribution over verbs and nouns, and $(n_z, v_z)$ refer to the ground truth future action labels.

During testing, we sample action class labels $(\hat{n}_z, \hat{v}_z)$ from the predicted distribution independently for each future step. We repeat this sampling procedure $N$ times to generate multiple cancidate sets of predictions for evaluation described in Section~\ref{sec:long_term_evaluation}. 

We use the taxonomy presented in Figure~\ref{fig:taxonomy} for our experiments. We finetune a Kinetics-400~\cite{kay2017kinetics} pretrained encoder backbones on Ego4D action recognition and use this model for all baselines to extract the clip level features. The aggregation module and decoder networks are trained from random initialization directly on the forecasting task. The encoder weights are kept unchanged during the decoder network training. We set $Z=20$ for long horizon future evaluation and $K = 5$ as the number of plausible future sequences predicted by the model. For all baselines, we sample $2$ input clips to capture past context unless otherwise specified. We train the model on 8 NVIDIA V100 GPUs with a batch size of 64 for 30 epochs and a base learning rate of 0.0001.

\subsubsection{Results}
\label{appendix:forecasting_results}

\subsubsection*{Future Locomotion Movements Prediction}

\srt{
We evaluate the KNN based baseline algorithm by measuring mean trajectory error (K-MTE) and probability of correct trajectory (PCT) given an error tolerance. The trajectory length ranges from 7 to 15 seconds (70-150 points in a trajectory given 10 FPS). Our baseline achieves mean error \FEB{8.81}m for $1-{\rm MTE}$ and \FEB{0.39} for ${\rm PCT}_{\epsilon=3m}$. The result is summarized in Table~\ref{tab:locomotion_results} and \ref{tab:locomotion_results1}.}

\begin{table}[t]
    \centering
    \footnotesize
    \begin{tabular}{ll|cc}
    \hline
    Set & Metric & Mean & Median \\
    \hline
    Val  & 5-MTE& \FEB{5.11m} & \FEB{2.53m}\\
    Val  & 3-MTE& \FEB{6.19m} & \FEB{2.99m}\\
    Val  & 1-MTE& \FEB{8.81m} & \FEB{4.63m}\\
        \hline
    Test  & 5-MTE& \FEB{4.84m} & \FEB{2.69m}\\
    Test  & 3-MTE& \FEB{5.54m} & \FEB{3.24m}\\
    Test  & 1-MTE& \FEB{7.66m} & \FEB{4.73m}\\
    \hline
    \end{tabular}
    \caption{Results of the locomotion prediction task. We report mean/median for 7-15 second predictions. We use $K=1,3,5$.}
    \label{tab:locomotion_results}
\end{table}

\begin{table}[t]
    \centering
    \footnotesize
    \begin{tabular}{l|cccccc}
    \hline
    Set & $\epsilon=1m$ & $\epsilon=2m$& $\epsilon=3m$& $\epsilon=4m$& $\epsilon=5m$& $\epsilon=6m$ \\
    \hline
    Val  & \FEB{0.14} & \FEB{0.29} & \FEB{0.39} & \FEB{0.46} & \FEB{0.51} & \FEB{0.54}\\
    Test  & \FEB{0.16} & \FEB{0.31} & \FEB{0.40} & \FEB{0.47} & \FEB{0.53} & \FEB{0.58}\\
        \hline
    \end{tabular}
    \caption{Results of the locomotion prediction task. We report the probability of correct trajectory (PCT) as varying the error threshold $\epsilon$.}
    \label{tab:locomotion_results1}
\end{table}

\subsubsection*{Future Hands Movements Prediction }
\begin{figure*}
    \centering
    \includegraphics[width=\linewidth]{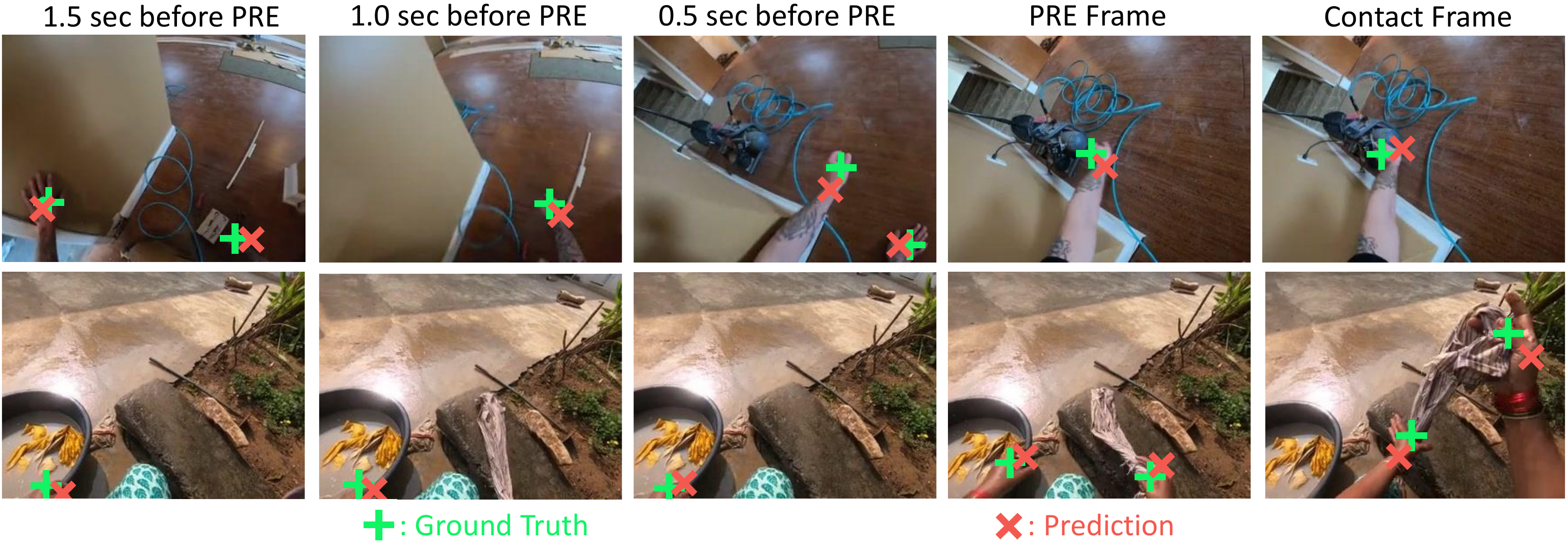}
    \caption{Qualitative examples of future hands movements prediction using the proposed baseline. The ground truth hands positions are plotted as green crosses, while the predicted hands positions are plotted as red crosses.}
    \label{fig:handant_vis}
\end{figure*}

For future hands movements prediction task, we report mean displacement error (M.Disp.) and contact frame displacement error (C.Disp.) on both validation  and test sets  in Table~\ref{table:handant_results}. Our baseline model achieves M.Disp. of (\FEB{$52.98/53.68$})  and C.Disp. of (\FEB{$56.37/56.17$}) for left/right hand position prediction on the test set. It is worth noting that predicting hand positions on contact frame is more challenging than on other key frames. This is because, by the definition of contact frame and pre-condition frame, the anticipation temporal footprint of contact frame is larger than other key frames. We further provide qualitative results of our baseline method in Fig.~\ref{fig:handant_vis}. Notably, the model can make reasonable predictions on future hand positions. However, the model is more likely to fail when there is drastic embodied motions.

\begin{table}[t]
    \centering
    \footnotesize
    \begin{tabular}{cc|cc|cc}
     \hline
  \multirow{2}{*}{Set} &\multirow{2}{*}{Method}  &\multicolumn{2}{c|}{Left Hand} &\multicolumn{2}{c}{Right Hand} \\ \cline{3-6}
    & & M.Disp.$\downarrow$   & C.Disp.$\downarrow$        &  M.Disp.$\downarrow$   & C.Disp.$\downarrow$        \\ \hline 
    Val &I3D+Reg   & \FEB{54.11} & \FEB{57.29} &\FEB{54.73} &\FEB{57.94} \\\hline 
    Test &I3D+Reg   & \FEB{52.98} & \FEB{56.37} &\FEB{53.68} &\FEB{56.17} \\   \hline
    \end{tabular}
    
    \caption{Results of future hand movement prediction task. Note that the left and right hands movements are evaluated separately. $\downarrow$ indicates lower is better}
    \label{table:handant_results}
\end{table}

\subsubsection*{Short-Term Object Interaction Anticipation }
\begin{figure*}[t]
    \centering
    \includegraphics[width=0.49\textwidth]{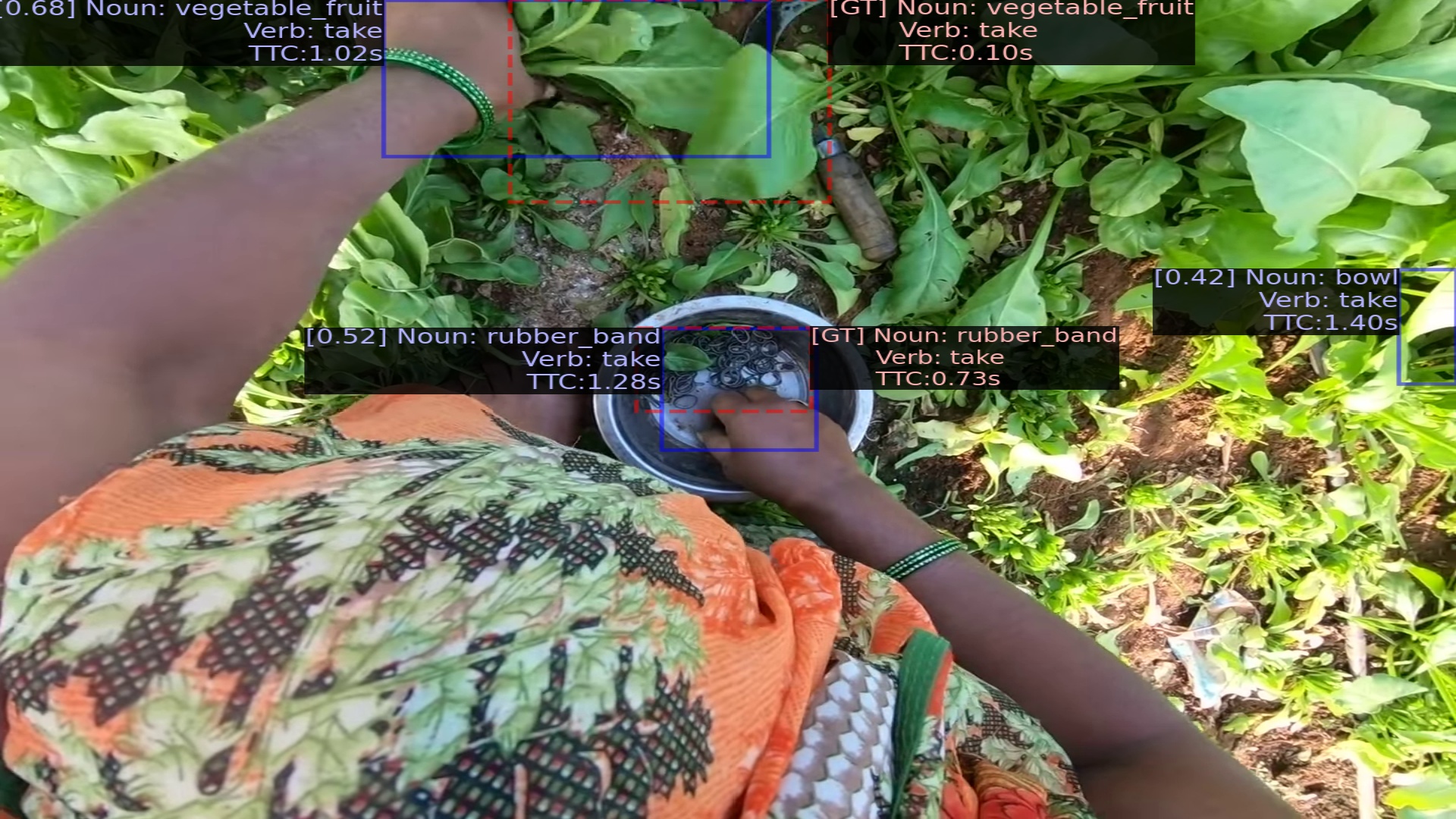}
    \includegraphics[width=0.49\textwidth]{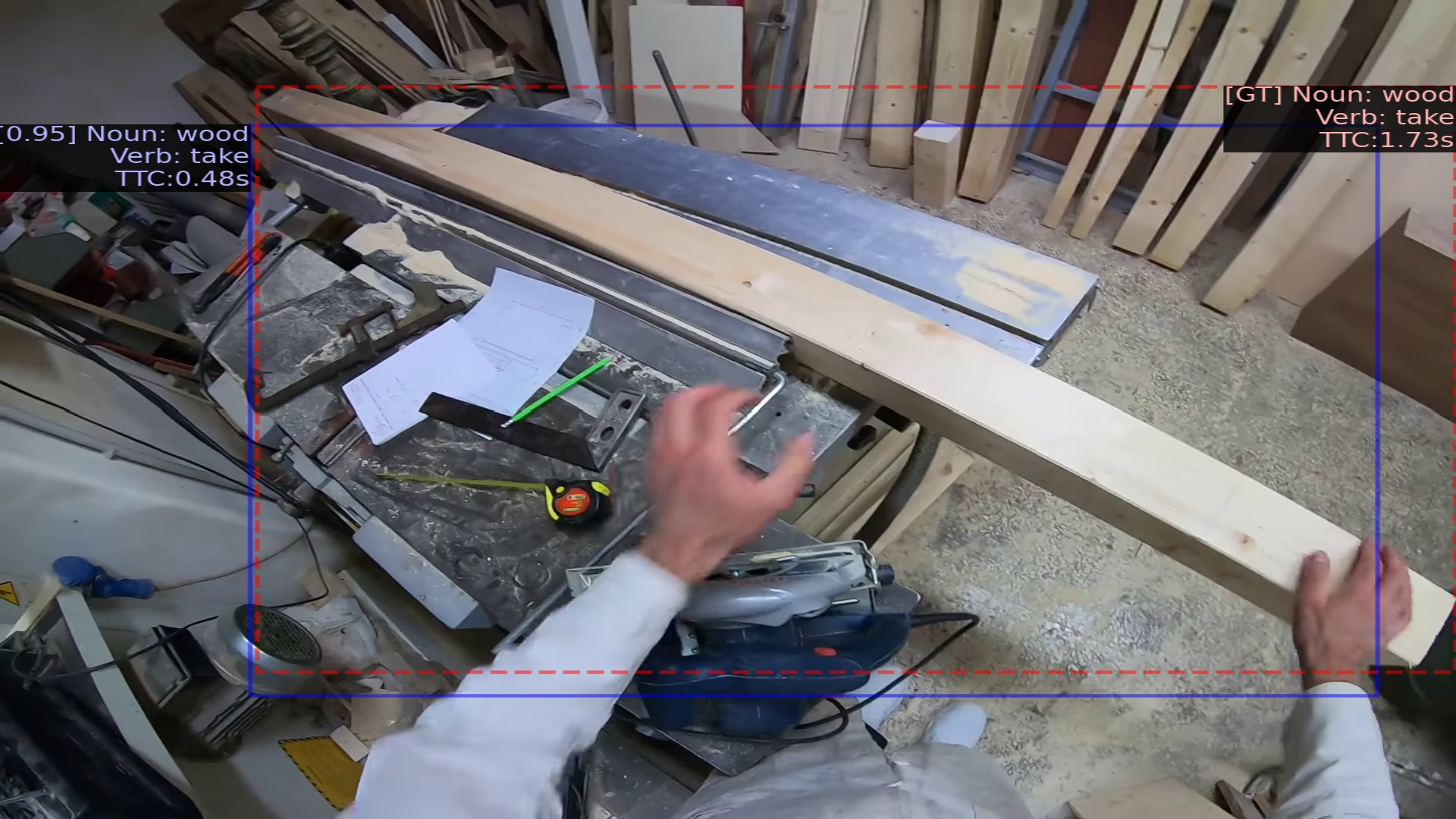}
    \includegraphics[width=0.49\textwidth]{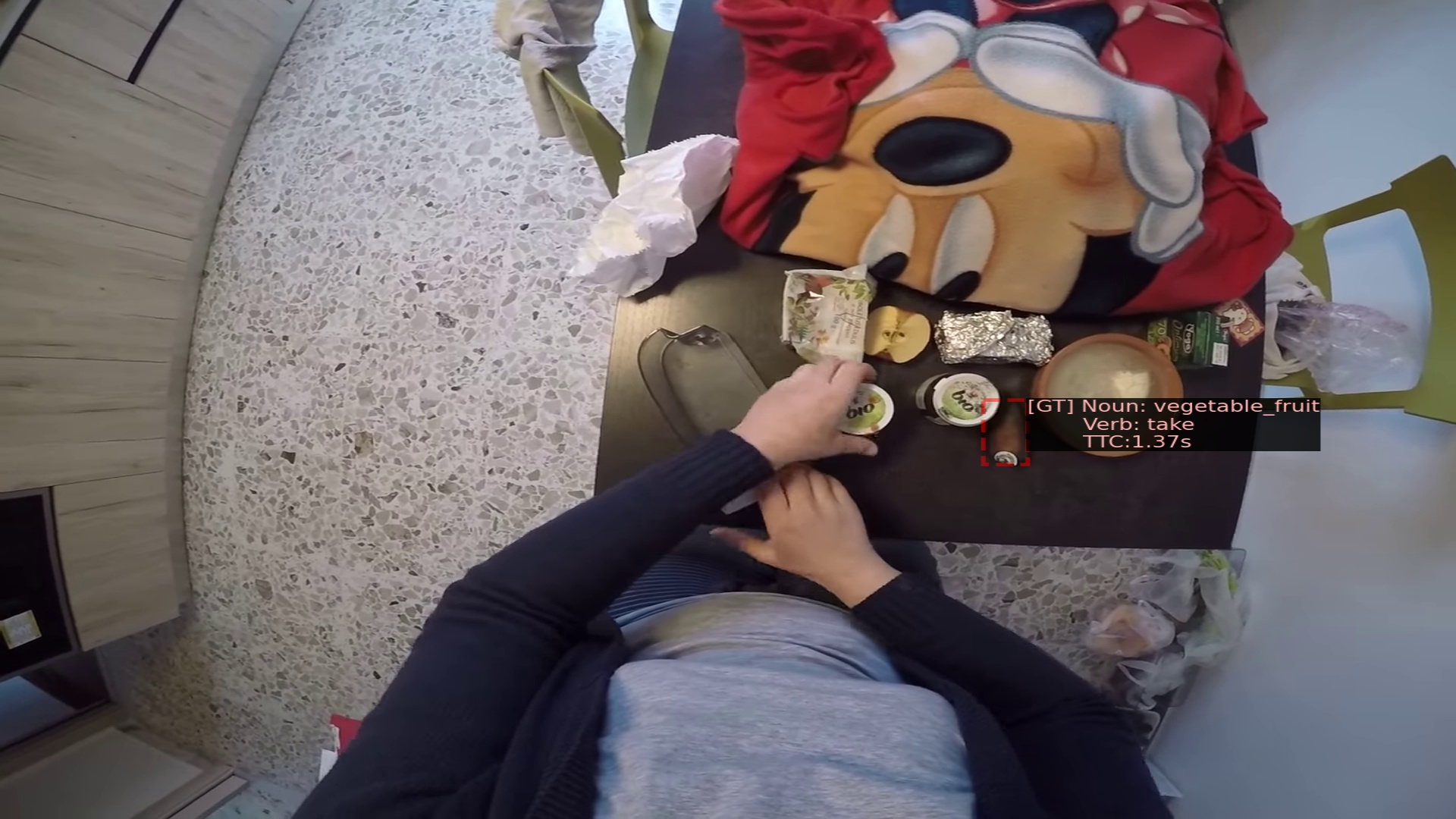}
    \includegraphics[width=0.49\textwidth]{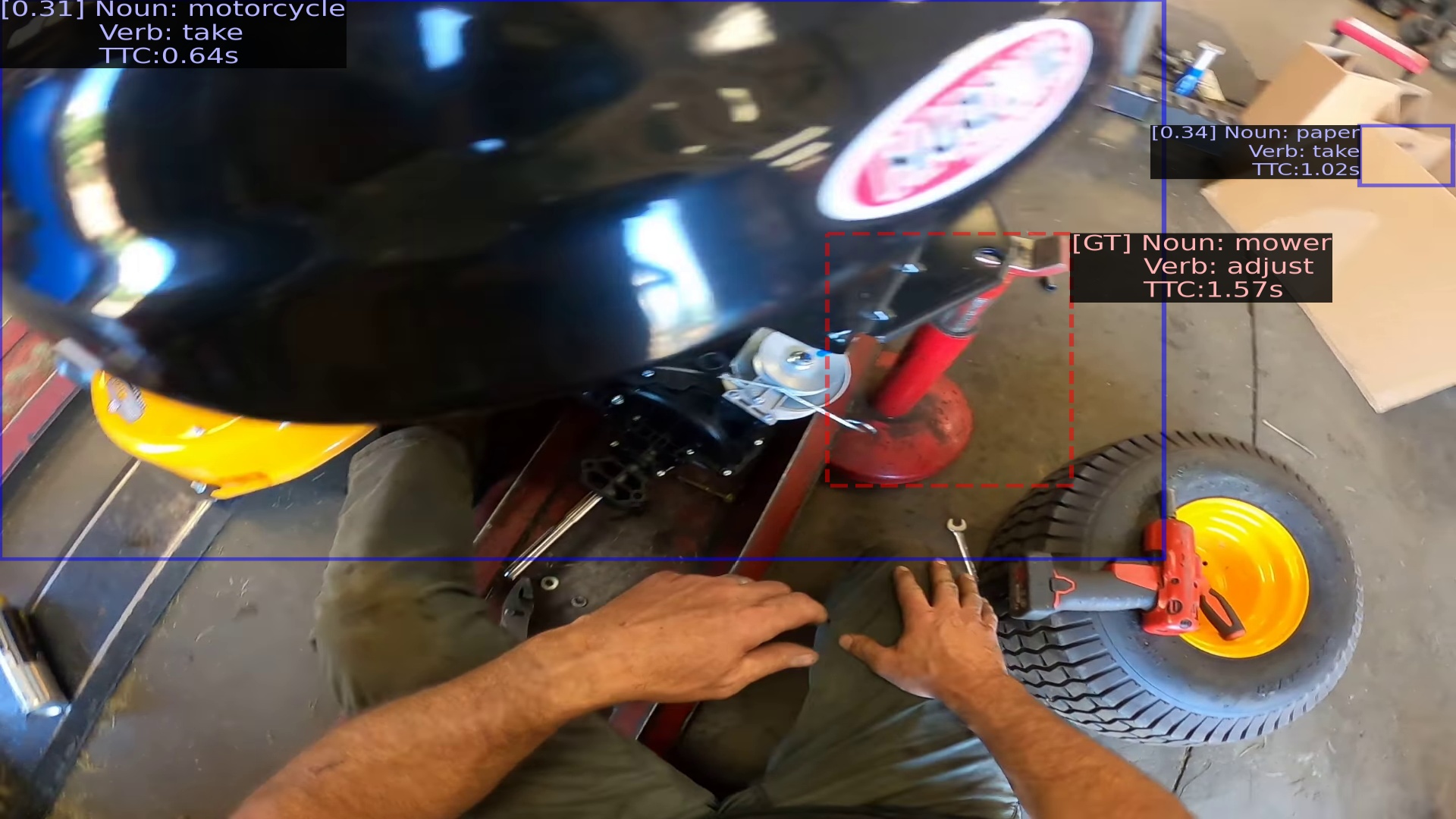}
    \caption{Qualitative examples of short-term object interaction anticipation using the proposed baseline. The numbers in brackets represent the confidence scores associated to the predictions. The ground truth next-active object is highlighted using a dashed red line, whereas model predictions are reported in blue solid lines. }
    \label{fig:short_term_qualitative}
\end{figure*}

Table~\ref{tab:short_term_results} reports the results for the short-term object interaction anticipation task on both the validation and test sets.
We compare the proposed baseline based on Faster RCNN and SlowFast (``FRCNN+SF'' in the table) with a simpler baseline which uses Faster RCNN to detect object and predict their classes, but draws verb and TTC predictions randomly from the training set distribution (``FRCNN+Rnd.'' in the table).
Results are reported in Top-5 mAP\% according to the different matching criteria discussed in Appendix~\ref{appendix:forecasting_measures}.  
As can be noted, the proposed baseline outperforms random prediction by big margins when verbs and TTCs are predicted on both the validation and test sets.
This suggests that, despite being simple, the baseline can leverage the observed video to anticipate future object interactions.
Figure~\ref{fig:short_term_qualitative} reports some qualitative examples of the baseline. The model is sometimes able to detect the next active objects and predict suitable verbs and TTCs, but performance tends to be limited especially in complex scenarios.

\begin{table}[t]
    \centering
    \footnotesize
    \begin{tabular}{ll|cccc}
    \hline
    Set & Method & Noun & Noun+Verb & Noun+TTC & Overall \\
    \hline
        Val & FRCNN+Rnd. & \FEB{17.55} & \FEB{1.56} & \FEB{3.21}& \FEB{0.34}\\
        Val & FRCNN+SF & \FEB{17.55} & \FEB{5.19} & \FEB{5.37}  & \FEB{2.07} \\
    \hline
        Test & FRCNN+Rnd. & \FEB{20.45} & \FEB{2.22} & \FEB{3.86} & \FEB{0.44} \\
        Test & FRCNN+SF & \FEB{20.45} & \FEB{6.78} & \FEB{6.17} & \FEB{2.45} \\
         \hline
    \end{tabular}
    \caption{Results of the short-term object interaction anticipation task. See text for discussion.}
    \label{tab:short_term_results}
\end{table}
\subsubsection*{Long-Term Action Anticipation }
Table~\ref{tab:long_term_results} shows our results  on both the validation and test sets. The \emph{No Change} baseline simply predicts the current action as the next $Z$ actions, and performs poorly at predicting future actions. Explicitly training multiple heads improves performance on verbs, nouns and actions. Changing the backbone architecture from SlowFast to MViT greatly improves verb forecasting prediction performance, but deteriorates noun forecasting performance, highlighting the trade-off between the two despite similar action classification performance on Kinetics. Finally, including larger video context information in the form of multiple input clips by using the transformer based aggregator module results in the best performance. 

\begin{figure*}
    \centering
    \includegraphics[width=\linewidth]{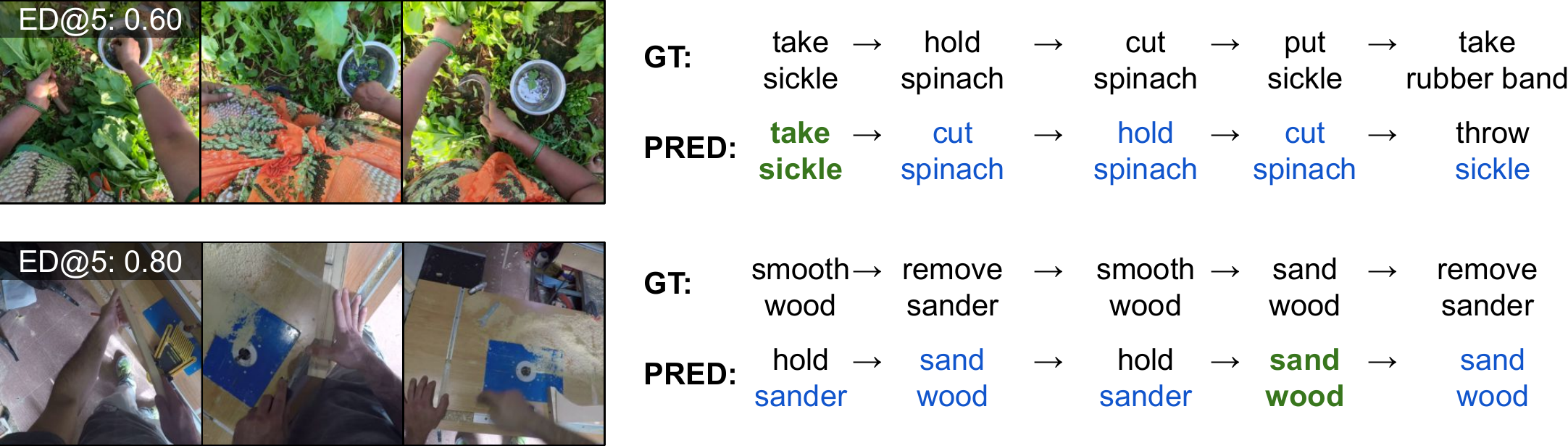}
    \caption{Long term action anticipation - qualitative results. Actions in green represent correct predictions (correct action, at the correct position). Actions in blue represent incorrect ordering of valid actions. Our edit-distance metric accounts for both cases. }
    \label{fig:long_term_qual}
\end{figure*}

Figure~\ref{fig:long_term_qual} shows some qualitative results of our method. In each row, the ground truth future actions are shown along with the predictions from our model (for 5 time-steps). Correct predictions are highlighted in green, while valid actions that are incorrectly ordered (or paritally correct) are highlighted in blue. Note that though not perfectly aligned, incorrectly ordered sequences are given partial credit via the edit-distance metric.

\begin{table}[t]
\centering
\footnotesize
\begin{tabular}{lll|lll}

                              \multicolumn{2}{c}{Val set} & & \multicolumn{3}{c}{ED@(Z=20)}  \\ \hline
\multicolumn{1}{l|}{Backbone} & \multicolumn{1}{l|}{Aggregator}  & Decoder   & Verb   & Noun    & Action \\ \hline
\multicolumn{1}{l|}{SlowFast} & \multicolumn{1}{l|}{Concat}      & No Change & 0.766  & 0.830   & 0.960       \\
\multicolumn{1}{l|}{SlowFast} & \multicolumn{1}{l|}{Concat}      & MultiHead & 0.747  & 0.808   & 0.952       \\
\multicolumn{1}{l|}{MViT}     & \multicolumn{1}{l|}{Concat}      & MultiHead & \textbf{0.707}  & 0.901   & 0.972   \\ 
\multicolumn{1}{l|}{SlowFast} & \multicolumn{1}{l|}{Transformer} & MultiHead & 0.745  & \textbf{0.779}   & \textbf{0.941}       \\ \hline
\\

                                                            \multicolumn{2}{c}{Test set} & & \multicolumn{3}{c}{ED@(Z=20)}  \\ \hline
\multicolumn{1}{l|}{Backbone} & \multicolumn{1}{l|}{Aggregator}  & Decoder   & Verb   & Noun    & Action \\ \hline
\multicolumn{1}{l|}{SlowFast} & \multicolumn{1}{l|}{Concat}      & No Change & 0.761  & 0.810   & 0.959       \\
\multicolumn{1}{l|}{SlowFast} & \multicolumn{1}{l|}{Concat}      & MultiHead & 0.743  & 0.791   & 0.948       \\
\multicolumn{1}{l|}{MViT}     & \multicolumn{1}{l|}{Concat}      & MultiHead & \textbf{0.697}  & 0.904   & 0.969   \\ 
\multicolumn{1}{l|}{SlowFast} & \multicolumn{1}{l|}{Transformer} & MultiHead & 0.739  & \textbf{0.780}   & \textbf{0.943}       \\ \hline

\end{tabular}
\caption{Results of the long-term action anticipation task. Lower is better. See text for discussion.
}
\label{tab:long_term_results}
\end{table}

\subsubsection{Discussion}

\subsubsection*{Data Annotation}
Annotating the videos for forecasting tasks posed a number of interesting challenges. First, we found the diversity of the data led to a large and diverse taxonomy, which some annotators found hard to navigate. Hence, we found a number of annotators used the "OTHER" option, which we eventually manually mapped to the taxonomy where possible. In future annotations, we plan to ask annotators to always pick the closest taxonomy item even if writing in a free-form OTHER label, to encourage them to stick to the taxonomy as much as possible. 
Second, we noticed annotators struggled with defining bounding boxes over ``stuff'' categories. For example, when labeling ``cutting grass'', it was often challenging to draw a box that covers the full extent of the object of change (\ie ``grass''). Finally, it was sometimes challenging to define what the object of change was, when using large tools. For example, if using a lawn mower to clear grass, does one consider the mower as the tool and hence the grass as the object of change, or the levers and buttons inside the mower as the object of change. We chose to rely on the narrators to define which interaction to label (\ie pushing the lever/button vs cutting grass), and asked the annotators to label tools and objects accordingly.

\subsubsection*{Future Locomotion Movements Prediction}
The baseline quantitative results on the locomotion prediction task imply that the visual cues, e.g., side walk, obstacles, and road, in egocentric images are highly indicative of future movement. However, the baseline method that encodes the visual semantics of an image with a global feature is not detailed enough to model complex walking movement, e.g., avoiding pedestrians. This opens an opportunity for challenge participants to incorporate a fine-grained visual representation.  

\subsubsection*{Future Hands Movements Prediction }
Our baseline model for future hands movements prediction suffers from the drastic head movements in egocentric video and the stochastic nature of future forecasting. We speculate that explicitly modeling the head movements and next-active objects may complement the video representations for predicting future hands movements.

\subsubsection*{Short-Term Object Interaction Anticipation }
The short-term object interaction anticipation results highlight that the proposed task is challenging, with the baseline achieving an overall Top-5 $mAP$ of \FEB{$2.07\%$} on the validation set and \FEB{$2.45\%$} on the test set.
The key challenges are likely due to the uncertain nature of future predictions as well as to the inability of the object detector to correctly detect next active objects and ignore the others.
Nevertheless, the proposed baseline, even if simple, allows to greatly improve over a combination of an object detector and a random prediction of verbs and time to contact quantities. This suggests that methods can learn to analyze the input video in order to make reasonable predictions about the future.

\subsubsection*{Long-Term Action Anticipation }
We discuss several important aspects of the long-term action forecasting problem through our experiments and ablation studies. All ablations are run with SlowFast backbone networks, and models are trained for 30 epochs.
\\

\noindent
\textit{How important is Ego4D action recognition pre-training?}
Table~\ref{tab:lt_results_kinetics} shows the performance of our models when pretrained \emph{only} on Kinetics-400 action recognition (as opposed to further fine-tuning on Ego4D action recognition). All models benefit greatly from training on Ego4D data in two ways. First, there is a large domain gap between Kinetics and Ego4D both in terms of visuals (third-person vs. egocentric viewpoint) and the diversity of activities they contain, which pre-training helps account for. Second, action recognition models benefit from biases in the label structure of future actions as seen from the performance of the \emph{No Change} baseline in Table~\ref{tab:long_term_results}.\\

\begin{table}[t]
\centering
\resizebox{\columnwidth}{!}{
\begin{tabular}{lll|lll}

                   \multicolumn{3}{c|}{Val Set}&       \multicolumn{3}{c}{ED@(Z=20)}

                   \\ \hline
\multicolumn{1}{l|}{Init}   & \multicolumn{1}{l|}{Backbone} & \multicolumn{1}{l|}{Aggregator}  & Verb& Noun & Action \\ \hline

\multicolumn{1}{l|}{K400}   & \multicolumn{1}{l|}{SlowFast} & \multicolumn{1}{l|}{Concat}      & 0.752  & 0.820   & 0.958       \\
\multicolumn{1}{l|}{+Ego4D} & \multicolumn{1}{l|}{SlowFast} & \multicolumn{1}{l|}{Concat}      & \textbf{0.747}  & \textbf{0.808}   & \textbf{0.952}       \\ \hline
\multicolumn{1}{l|}{K400}   & \multicolumn{1}{l|}{SlowFast} & \multicolumn{1}{l|}{Transformer} & 0.746  & 0.809   & 0.953       \\
\multicolumn{1}{l|}{+Ego4D} & \multicolumn{1}{l|}{SlowFast} & \multicolumn{1}{l|}{Transformer} & \textbf{0.745}  & \textbf{0.779}   & \textbf{0.941}       \\ \hline

\end{tabular}
}
\caption{Long term anticipation - varying pretraining data. MultiHead decoder used for all models. Ego4D action recognition pretraining greatly improves downstream forecasting performance. 
}
\label{tab:lt_results_kinetics}
\end{table}

\begin{table}[t]
\centering
\resizebox{\columnwidth}{!}{
\begin{tabular}{lll|lll}

                      \multicolumn{3}{c|}{Val Set}&      \multicolumn{3}{c}{ED@(Z=20)}  
                      
                      \\ \hline
\multicolumn{1}{l|}{\# clips}   & \multicolumn{1}{l|}{Backbone} & \multicolumn{1}{l|}{Aggregator}  & Verb& Noun & Action \\ \hline
\multicolumn{1}{l|}{2}   & \multicolumn{1}{l|}{SlowFast} & \multicolumn{1}{l|}{Transformer}      & \textbf{0.743}  & 0.790   & 0.946       \\
\multicolumn{1}{l|}{4}   & \multicolumn{1}{l|}{SlowFast} & \multicolumn{1}{l|}{Transformer} & 0.744  & 0.796   & 0.947       \\ 
\multicolumn{1}{l|}{8} & \multicolumn{1}{l|}{SlowFast} & \multicolumn{1}{l|}{Transformer}      & 0.745  & \textbf{0.779}   & \textbf{0.941}       \\ \hline

\end{tabular}
}
\caption{Long term anticipation - varying number of input clips. MultiHead decoder used for all models. Performance increases with more input context. 
}
\label{tab:lt_input_clips}
\end{table}

\noindent
\textit{How important is past context for transformer based models?}
Our transformer aggregation modules aggregate information across a larger temporal history controlled by the number of input clips to the model. Table~\ref{tab:lt_input_clips} shows the sensitivity of these models to the amount of past context video that it has access to. Overall, performance increases as more context information is provided to the model, however this increase comes at the cost of memory consumption --- 8 is the maximum number of clips that can be fit in GPU memory. \\

\begin{figure}
    \centering
    \includegraphics[width=\columnwidth]{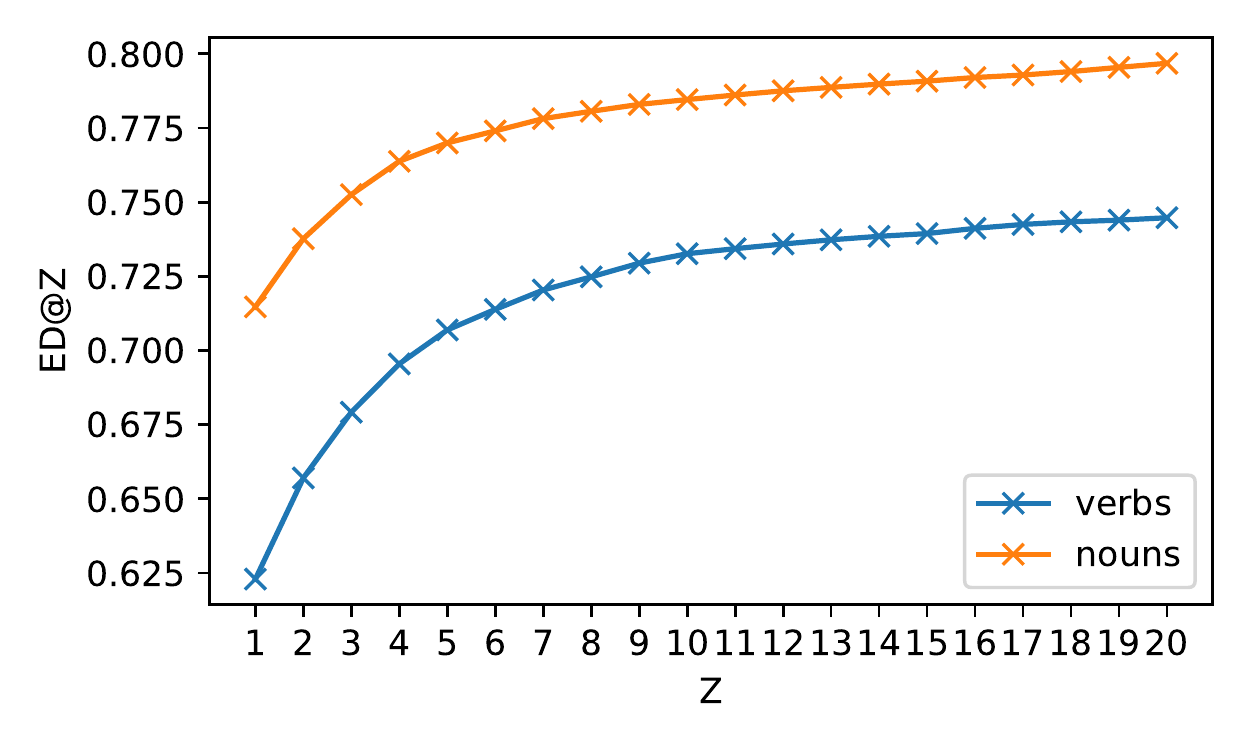}
    \caption{Performance vs.~number of future actions $Z$. Predicting further into the future is naturally more difficult. Models begin to predict close to random actions for very high values of $Z$. }
    \label{fig:long_term_perf_vs_z}
\end{figure}

\noindent
\textit{How far into the future can models predict?}
As mentioned in Section~\ref{sec:long_term_evaluation} we report results for predictions at $Z=20$ as baselines begin to predict actions at random for higher values of $Z$. Figure~\ref{fig:long_term_perf_vs_z} shows the plot of edit distance vs. $Z$ for our baseline models. As expected, it is far easier to anticipate actions that occur immediately next, which gets more difficult as $Z$ increases, and steadily plateaus.\\

\noindent
\textit{How to generate multiple candidate predictions?}
As mentioned in Section~\ref{sec:long_term_evaluation} we evaluate the best of $K=5$ predictions to arrive at our final results. To generate the $K$ predictions, we sample each classifier head independently, however there are several methods to improve this including heuristic search algorithms (like beam search). Ideally, the multi-modal nature of future prediction should be accounted for in the model design itself. Moreover, decoder models that take into account the sequential nature during inference should be considered. These include transformer based decoders that are popular in recent language models (e.g., BERT, GPT) This is an important future direction of research.

\iftoggle{arxiv}{
\subsubsection{Contributions statement}

\textit{Giovanni Maria Farinella} led the Forecasting Benchmark working on the definition of the proposed tasks, on the collection, and writing the paper.
\\\textit{Rohit Girdhar} co-led the Forecasting Benchmark working on the definition of the proposed tasks, on the collection, and writing the paper.
\\\textit{Antonino Furnari} contributed to the definition of the proposed benchmark tasks and in particular to the Short-Term Object Interaction Anticipation task and has been key driver of implementation, collection, annotation development throughout the project, and writing the paper.
\\\textit{Ilija Radosavovic} worked on the definition of tasks and has been key driver of implementation, collection, annotation development throughout the project, and writing  the paper.
\\\textit{Tushar Nagarajan} contributed to the definition of the proposed benchmark tasks and in particular to the Long-Term Action Anticipation task and has been key driver of implementation, collection, annotation development throughout the project, and writing the paper. 
\\\textit{Tullie Murrell} worked on baseline implementation of the Long-Term Action Anticipation task. 
\\\textit{Karttikeya Mangalam} worked on baseline implementation, experiments and writing the Long-Term Action Anticipation task. 
\\\textit{Christoph Feichtenhofer} oversaw the development of the task, baselines and implementation of the Long-Term Action Anticipation task.
\\\textit{Miao Liu}  worked on the definition of Future Hands Movement Prediction task and has been key driver of implementation, collection, annotation development throughout the project, and writing  the paper. 
\\\textit{Wenqi Jia} worked on baseline implementation of the Future Hands Movement Prediction task. 
\\\textit{Zachary Chavis} worked on the Locomotion Forecasting task and has been key driver of implementation, collection, and annotation development throughout the project.
\\\textit{Hyun Soo Park} worked on the definition of Locomotion Forecasting tasks, collection, annotation, and writing the paper. }
{}
\clearpage
\subsection{Societal Impact}\label{sec:appendix-societal}

\srt{Our contribution can positively impact video understanding.  It offers the research community a large-scale resource captured with rigorous privacy and ethics standards (detailed in Appendix~\ref{sec:collection} and~\ref{sec:deid-appendix}) together with a diversity of subjects, and the benchmarks will promote reproducible technical advances.  More broadly, egocentric perception has the potential to positively impact society in many application domains, including assistive technology, education, fitness, entertainment and gaming, eldercare, robotics, and augmented reality.  

Nonetheless, future research in this area must guard against the potential negative societal impact if technology for egocentric vision were misused.  

First, there are risks surrounding privacy. %
As we begin to see a proliferation of wearable cameras in public spaces, producers of these wearable devices will need to develop and implement protocols for notice and consent regarding the collection of data in public spaces, as well as user controls for how such data may be used, stored, and shared with any third parties.
Similarly, models that may be used to transcribe speech or perform other tasks related to footage should include robust user controls such as the ability to remove or obscure personal data or sensitive content.

Note that for all our audio-visual and social benchmarking work, the data used has full consent from the participants in the video, i.e., to use their unblurred faces and audio of their conversation.  To date, the research community has lacked any large-scale data resource with which to study these kinds of problems; Ego4D will help the community to consider new solutions while leveraging real-world, diverse data that respects the privacy protocols of different countries.  Furthermore, the Ego4D data is available only for users who sign a license that enumerates the allowable uses of the data, which is intended to hinder potential negative applications. 

Second, there is a risk that our large-scale collection could inspire future collection efforts without the same level of care or attention to the privacy and ethical concerns as were taken in Ego4D.  To mitigate this risk, we have aimed to be comprehensive in our descriptions of all parts of our procedures, and we will include our best practices recommendations when publicly disseminating the results of the project.

Finally, despite our best efforts as discussed in the main paper, there are still some imbalances in the dataset.   For example, the data from Rwanda is relatively small, and though 74 cities represents a leap in coverage, they do not capture all possible demographics.  We acknowledge that no matter how far one goes, full global coverage of daily life activity is elusive.    Still, we can mitigate this risk by continuing to grow global collaborations with researchers and participants in underrepresented areas.}

\clearpage

{\small
\bibliographystyle{ieee_fullname}
\bibliography{refs,gaze,social,av-refs,forecasting,ho,em}

\begin{thebibliography}{100}\itemsep=-1pt

\bibitem{espnet-model-zoo}
Github repository of the {ESPNet} model zoo.
\newblock \url{https://github.com/espnet/espnet_model_zoo}.
\newblock We used the \url{Shinji
  Watanabe/gigaspeech_asr_train_asr_raw_en_bpe5000_valid.acc.ave} model.

\bibitem{kaldi-glm}
Kaldi {English GLM} file.
\newblock
  \url{https://github.com/kaldi-asr/kaldi/blob/master/egs/ami/s5/local/english.glm}.

\bibitem{NISTSRE2000evalplan}
{NIST SRE 2000 Evaluation Plan}.
\newblock
  \url{https://www.nist.gov/sites/default/files/documents/2017/09/26/spk-2000-plan-v1.0.htm_.pdf}.

\bibitem{abu2018will}
Yazan Abu~Farha, Alexander Richard, and Juergen Gall.
\newblock When will you do what?-anticipating temporal occurrences of
  activities.
\newblock In {\em Computer Vision and Pattern Recognition}, pages 5343--5352,
  2018.

\bibitem{afouras2018deep}
Triantafyllos Afouras, Joon~Son Chung, Andrew Senior, Oriol Vinyals, and Andrew
  Zisserman.
\newblock Deep audio-visual speech recognition.
\newblock {\em IEEE transactions on pattern analysis and machine intelligence},
  2018.

\bibitem{afouras2018conversation}
Triantafyllos Afouras, Joon~Son Chung, and Andrew Zisserman.
\newblock The conversation: Deep audio-visual speech enhancement.
\newblock In {\em Interspeech}, 2018.

\bibitem{Afouras2020}
Triantafyllos Afouras, Andrew Owens, Joon~Son Chung, and Andrew Zisserman.
\newblock {Self-supervised Learning of Audio-Visual Objects from Video}.
\newblock In {\em Proceedings of the European Conference on Computer Vision
  (ECCV 20)}, volume 12363 LNCS, pages 208--224, 2020.

\bibitem{alayrac2017joint}
Jean-Baptiste Alayrac, Josef Sivic, Ivan Laptev, and Simon Lacoste-Julien.
\newblock Joint discovery of object states and manipulation actions.
\newblock {\em ICCV}, 2017.

\bibitem{detad}
Humam Alwassel, Fabian Caba~Heilbron, Victor Escorcia, and Bernard Ghanem.
\newblock Diagnosing error in temporal action detectors.
\newblock In {\em Proceedings of the European Conference on Computer Vision
  (ECCV)}, 2018.

\bibitem{anguera2012speaker}
Xavier Anguera, Simon Bozonnet, Nicholas Evans, Corinne Fredouille, Gerald
  Friedland, and Oriol Vinyals.
\newblock Speaker diarization: A review of recent research.
\newblock {\em IEEE Transactions on audio, speech, and language processing},
  20(2):356--370, 2012.

\bibitem{anguera2006robust}
Xavier Anguera~Mir{\'o}.
\newblock {\em Robust speaker diarization for meetings}.
\newblock Universitat Polit{\`e}cnica de Catalunya, 2006.

\bibitem{vqa}
Stanislaw Antol, Aishwarya Agrawal, Jiasen Lu, Margaret Mitchell, Dhruv Batra,
  C.~Lawrence Zitnick, and Devi Parikh.
\newblock {VQA}: {V}isual {Q}uestion {A}nswering.
\newblock In {\em International Conference on Computer Vision (ICCV)}, 2015.

\bibitem{arabaci2018multi}
Mehmet~Ali Arabac{\i}, Fatih {\"O}zkan, Elif Surer, Peter Jan{\v{c}}ovi{\v{c}},
  and Alptekin Temizel.
\newblock Multi-modal egocentric activity recognition using audio-visual
  features.
\newblock {\em arXiv preprint arXiv:1807.00612}, 2018.

\bibitem{arandjelovic2017objects}
Relja Arandjelovi{\'c} and Andrew Zisserman.
\newblock Objects that sound.
\newblock In {\em ECCV}, 2018.

\bibitem{baevski2020wav2vec}
Alexei Baevski, Henry Zhou, Abdelrahman Mohamed, and Michael Auli.
\newblock wav2vec 2.0: A framework for self-supervised learning of speech
  representations.
\newblock {\em arXiv preprint arXiv:2006.11477}, 2020.

\bibitem{Bambach_2015_ICCV}
Sven Bambach, Stefan Lee, David~J. Crandall, and Chen Yu.
\newblock Lending a hand: Detecting hands and recognizing activities in complex
  egocentric interactions.
\newblock In {\em The IEEE International Conference on Computer Vision (ICCV)},
  December 2015.

\bibitem{bee2008cocktail}
Mark~A Bee and Christophe Micheyl.
\newblock The cocktail party problem: what is it? how can it be solved? and why
  should animal behaviorists study it?
\newblock {\em Journal of comparative psychology}, 122(3):235, 2008.

\bibitem{IdentityMOT}
Keni Bernardin, Alexander Elbs, and Rainer Stiefelhagen.
\newblock Multiple object tracking performance metrics and evaluation in a
  smart room environment.
\newblock In {\em Sixth IEEE International Workshop on Visual Surveillance, in
  conjunction with ECCV}, volume~90. Citeseer, 2006.

\bibitem{CLEARMOT}
Keni Bernardin and Rainer Stiefelhagen.
\newblock Evaluating multiple object tracking performance: the clear mot
  metrics.
\newblock {\em EURASIP Journal on Image and Video Processing}, 2008:1--10,
  2008.

\bibitem{bertasius2016first}
Gedas Bertasius, Hyun~Soo Park, Stella~X. Yu, and Jianbo Shi.
\newblock First-person action-object detection with egonet.
\newblock In {\em Proceedings of Robotics: Science and Systems}, July 2017.

\bibitem{Beyan2018}
Cigdem Beyan, Francesca Capozzi, Cristina Becchio, and Vittorio Murino.
\newblock {Prediction of the leadership style of an emergent leader using audio
  and visual nonverbal features}.
\newblock {\em IEEE Transactions on Multimedia}, 20(2):441--456, 2018.

\bibitem{bhat2020kys}
Goutam Bhat, Martin Danelljan, Luc Van~Gool, and Radu Timofte.
\newblock Know {{Your Surroundings}}: {{Exploiting Scene Information}} for
  {{Object Tracking}}.
\newblock {\em arXiv:2003.11014 [cs]}, May 2020.

\bibitem{brachmann2014learning}
Eric Brachmann, Alexander Krull, Frank Michel, Stefan Gumhold, Jamie Shotton,
  and Carsten Rother.
\newblock Learning 6d object pose estimation using 3d object coordinates.
\newblock In {\em European conference on computer vision}, pages 536--551.
  Springer, 2014.

\bibitem{gpt3}
Tom~B. Brown, Benjamin Mann, Nick Ryder, Melanie Subbiah, Jared Kaplan,
  Prafulla Dhariwal, Arvind Neelakantan, Pranav Shyam, Girish Sastry, Amanda
  Askell, Sandhini Agarwal, Ariel Herbert-Voss, Gretchen Krueger, Tom Henighan,
  Rewon Child, Aditya Ramesh, Daniel~M. Ziegler, Jeffrey Wu, Clemens Winter,
  Christopher Hesse, Mark Chen, Eric Sigler, Mateusz Litwin, Scott Gray,
  Benjamin Chess, Jack Clark, Christopher Berner, Sam McCandlish, Alec Radford,
  Ilya Sutskever, and Dario Amodei.
\newblock Language models are few-shot learners, 2020.

\bibitem{yale:human:grasping:dataset}
Ian~M Bullock, Thomas Feix, and Aaron~M Dollar.
\newblock The yale human grasping dataset: Grasp, object, and task data in
  household and machine shop environments.
\newblock {\em IJRR}, 2015.

\bibitem{cai2016understanding}
Minjie Cai, Kris~M Kitani, and Yoichi Sato.
\newblock Understanding hand-object manipulation with grasp types and object
  attributes.
\newblock In {\em RSS}, 2016.

\bibitem{detr}
Nicolas Carion, Francisco Massa, Gabriel Synnaeve, Nicolas Usunier, Alexander
  Kirillov, and Sergey Zagoruyko.
\newblock End-to-end object detection with transformers.
\newblock In {\em European Conference on Computer Vision}, pages 213--229.
  Springer, 2020.

\bibitem{carletta2005ami}
Jean Carletta, Simone Ashby, Sebastien Bourban, Mike Flynn, Mael Guillemot,
  Thomas Hain, Jaroslav Kadlec, Vasilis Karaiskos, Wessel Kraaij, Melissa
  Kronenthal, et~al.
\newblock {The AMI meeting corpus: A pre-announcement}.
\newblock In {\em International workshop on machine learning for multimodal
  interaction}, pages 28--39. Springer, 2006.

\bibitem{i3d}
Joao Carreira and Andrew Zisserman.
\newblock Quo vadis, action recognition? a new model and the kinetics dataset.
\newblock In {\em proceedings of the IEEE Conference on Computer Vision and
  Pattern Recognition}, pages 6299--6308, 2017.

\bibitem{procedure:planning:in:instructional:videos}
Chien-Yi Chang, De-An Huang, Danfei Xu, Ehsan Adeli, Li Fei-Fei, and
  Juan~Carlos Niebles.
\newblock Procedure planning in instructional videos.
\newblock {\em arXiv preprint arXiv:1907.01172}, 2019.

\bibitem{chaudhuri2018ava}
Sourish Chaudhuri, Joseph Roth, Daniel~PW Ellis, Andrew Gallagher, Liat Kaver,
  Radhika Marvin, Caroline Pantofaru, Nathan Reale, Loretta~Guarino Reid, Kevin
  Wilson, et~al.
\newblock Ava-speech: A densely labeled dataset of speech activity in movies.
\newblock {\em arXiv preprint arXiv:1808.00606}, 2018.

\bibitem{soundspaces}
C. Chen, U. Jain, C. Schissler, S.~V.~Amengual Gari, Z. Al-Halah, V. Ithapu, P.
  Robinson, and K. Grauman.
\newblock Soundspaces: Audio-visual navigation in 3d environments.
\newblock In {\em ECCV}, 2020.

\bibitem{chen2019audio}
Changan Chen, Unnat Jain, Carl Schissler, Sebastia Vicenc~Amengual Gari, Ziad
  Al-Halah, Vamsi~Krishna Ithapu, Philip Robinson, and Kristen Grauman.
\newblock Audio-visual embodied navigation.
\newblock {\em environment}, 97:103, 2019.

\bibitem{chen2021gigaspeech}
Guoguo Chen, Shuzhou Chai, Guanbo Wang, Jiayu Du, Wei-Qiang Zhang, Chao Weng,
  Dan Su, Daniel Povey, Jan Trmal, Junbo Zhang, et~al.
\newblock Gigaspeech: An evolving, multi-domain asr corpus with 10,000 hours of
  transcribed audio.
\newblock {\em arXiv preprint arXiv:2106.06909}, 2021.

\bibitem{chen2015microsoft}
Xinlei Chen, Hao Fang, Tsung-Yi Lin, Ramakrishna Vedantam, Saurabh Gupta, Piotr
  Doll{\'a}r, and C~Lawrence Zitnick.
\newblock Microsoft coco captions: Data collection and evaluation server.
\newblock {\em arXiv preprint arXiv:1504.00325}, 2015.

\bibitem{Chong2020eyecontact}
Eunji Chong, Elysha Clark-Whitney, Audrey Southerland, Elizabeth Stubbs, Chanel
  Miller, Eliana~L Ajodan, Melanie~R Silverman, Catherine Lord, Agata Rozga,
  Rebecca~M Jones, and James~M Rehg.
\newblock {Detection of eye contact with deep neural networks is as accurate as
  human experts}.
\newblock {\em Nature Communications}, 11(1):6386, dec 2020.

\bibitem{Chong2020attended}
Eunji Chong, Yongxin Wang, Nataniel Ruiz, and James~M. Rehg.
\newblock {Detecting Attended Visual Targets in Video}.
\newblock In {\em Proceedings of the IEEE Conference on Computer Vision and
  Pattern Recognition (CVPR 20)}, pages 5395--5405, Seattle, WA, 2020.

\bibitem{chung2020in}
Joon~Son Chung, Jaesung Huh, Seongkyu Mun, Minjae Lee, Hee~Soo Heo, Soyeon
  Choe, Chiheon Ham, Sunghwan Jung, Bong-Jin Lee, and Icksang Han.
\newblock In defence of metric learning for speaker recognition.
\newblock In {\em Interspeech}, 2020.

\bibitem{chung2020spot}
Joon~Son Chung, Jaesung Huh, Arsha Nagrani, Triantafyllos Afouras, and Andrew
  Zisserman.
\newblock Spot the conversation: speaker diarisation in the wild.
\newblock {\em arXiv preprint arXiv:2007.01216}, 2020.

\bibitem{Chung18b}
J.~S. Chung, A. Nagrani, and A. Zisserman.
\newblock {VoxCeleb2: Deep Speaker Recognition}.
\newblock In {\em INTERSPEECH}, 2018.

\bibitem{church1990word}
Kenneth Church and Patrick Hanks.
\newblock Word association norms, mutual information, and lexicography.
\newblock {\em Computational linguistics}, 16(1):22--29, 1990.

\bibitem{damen2020epic}
Dima Damen, Hazel Doughty, Giovanni Farinella, Sanja Fidler, Antonino Furnari,
  Evangelos Kazakos, Davide Moltisanti, Jonathan Munro, Toby Perrett, Will
  Price, et~al.
\newblock The epic-kitchens dataset: Collection, challenges and baselines.
\newblock {\em IEEE Transactions on Pattern Analysis \& Machine Intelligence},
  (01):1--1, 2020.

\bibitem{Damen2020RESCALING}
Dima Damen, Hazel Doughty, Giovanni~Maria Farinella, , Antonino Furnari, Jian
  Ma, Evangelos Kazakos, Davide Moltisanti, Jonathan Munro, Toby Perrett, Will
  Price, and Michael Wray.
\newblock Rescaling egocentric vision.
\newblock {\em IJCV}, 2021.

\bibitem{Damen2018EPICKITCHENS}
Dima Damen, Hazel Doughty, Giovanni~Maria Farinella, Sanja Fidler, Antonino
  Furnari, Evangelos Kazakos, Davide Moltisanti, Jonathan Munro, Toby Perrett,
  Will Price, and Michael Wray.
\newblock Scaling egocentric vision: The epic-kitchens dataset.
\newblock In {\em European Conference on Computer Vision (ECCV)}, 2018.

\bibitem{damen2014}
Dima Damen, Teesid Leelasawassuk, Osian Haines, Andrew Calway, and Walterio
  Mayol-Cuevas.
\newblock {Y}ou-{D}o, {I}-{L}earn: {D}iscovering task relevant objects and
  their modes of interaction from multi-user egocentric video.
\newblock In {\em BMVC}, 2014.

\bibitem{damen2016you}
Dima Damen, Teesid Leelasawassuk, and Walterio Mayol-Cuevas.
\newblock You-do, i-learn: Egocentric unsupervised discovery of objects and
  their modes of interaction towards video-based guidance.
\newblock {\em CVIU}, 2016.

\bibitem{damerau1964technique}
Fred~J Damerau.
\newblock A technique for computer detection and correction of spelling errors.
\newblock {\em Communications of the ACM}, 1964.

\bibitem{del2016summarization}
Ana~Garcia Del~Molino, Cheston Tan, Joo-Hwee Lim, and Ah-Hwee Tan.
\newblock Summarization of egocentric videos: A comprehensive survey.
\newblock {\em IEEE Transactions on Human-Machine Systems}, 47(1), 2016.

\bibitem{deng2009imagenet}
Jia Deng, Wei Dong, Richard Socher, Li-Jia Li, Kai Li, and Li Fei-Fei.
\newblock Image{N}et: A large-scale hierarchical image database.
\newblock In {\em CVPR}, 2009.

\bibitem{detone2018superpoint}
Daniel DeTone, Tomasz Malisiewicz, and Andrew Rabinovich.
\newblock Superpoint: Self-supervised interest point detection and description.
\newblock In {\em CVPR Workshop}, 2018.

\bibitem{bert}
Jacob Devlin, Ming-Wei Chang, Kenton Lee, and Kristina Toutanova.
\newblock Bert: Pre-training of deep bidirectional transformers for language
  understanding.
\newblock {\em arXiv:1810.04805}, 2018.

\bibitem{devlin-etal-2019-bert}
Jacob Devlin, Ming-Wei Chang, Kenton Lee, and Kristina Toutanova.
\newblock {BERT}: Pre-training of deep bidirectional transformers for language
  understanding.
\newblock In {\em Proceedings of the 2019 Conference of the North {A}merican
  Chapter of the Association for Computational Linguistics: Human Language
  Technologies, Volume 1 (Long and Short Papers)}, pages 4171--4186,
  Minneapolis, Minnesota, June 2019. Association for Computational Linguistics.

\bibitem{donley2021easycom}
Jacob Donley, Vladimir Tourbabin, Jung-Suk Lee, Mark Broyles, Hao Jiang, Jie
  Shen, Maja Pantic, Vamsi~Krishna Ithapu, and Ravish Mehra.
\newblock Easycom: An augmented reality dataset to support algorithms for easy
  communication in noisy environments.
\newblock {\em arXiv preprint arXiv:2107.04174}, 2021.

\bibitem{DoostiBoosting}
Bardia Doosti, Ching{-}Hui Chen, Raviteja Vemulapalli, Xuhui Jia, Yukun Zhu,
  and Bradley Green.
\newblock Boosting image-based mutual gaze detection using pseudo 3d gaze.
\newblock In {\em Thirty-Fifth {AAAI} Conference on Artificial Intelligence},
  pages 1273--1281, 2021.

\bibitem{action:modifiers}
Hazel Doughty, Ivan Laptev, Walterio Mayol-Cuevas, and Dima Damen.
\newblock Action modifiers: Learning from adverbs in instructional videos.
\newblock {\em arXiv preprint arXiv:1912.06617}, 2019.

\bibitem{TREK150}
Matteo Dunnhofer, Antonino Furnari, Giovanni~Maria Farinella, and Christian
  Micheloni.
\newblock Is first person vision challenging for object tracking?
\newblock In {\em IEEE/CVF International Conference on Computer Vision
  Workshops (ICCVW) - Visual Object Tracking Challenge}, 2021.

\bibitem{ephrat2018looking}
Ariel Ephrat, Inbar Mosseri, Oran Lang, Tali Dekel, Kevin Wilson, Avinatan
  Hassidim, William~T Freeman, and Michael Rubinstein.
\newblock Looking to listen at the cocktail party: A speaker-independent
  audio-visual model for speech separation.
\newblock In {\em SIGGRAPH}, 2018.

\bibitem{Epstein2019}
Dave Epstein, Boyuan Chen, and Carl Vondrick.
\newblock Oops! predicting unintentional action in video.
\newblock In {\em Arxiv}, 2019.

\bibitem{DIHARD19}
N.~Ryant et. al.
\newblock {The Second DIHARD Diarization Challenge: Dataset, task, and
  baselines.}
\newblock In {\em Proceedings of Interspeech}, 2019.

\bibitem{everingham2010pascal}
Mark Everingham, Luc Van~Gool, Christopher~KI Williams, John Winn, and Andrew
  Zisserman.
\newblock The pascal visual object classes (voc) challenge.
\newblock {\em International journal of computer vision}, 88(2):303--338, 2010.

\bibitem{caba2015activitynet}
Bernard~Ghanem Fabian Caba~Heilbron, Victor~Escorcia and Juan~Carlos Niebles.
\newblock Activitynet: A large-scale video benchmark for human activity
  understanding.
\newblock In {\em Proceedings of the IEEE Conference on Computer Vision and
  Pattern Recognition}, pages 961--970, 2015.

\bibitem{fan2019lasot}
Heng Fan, Haibin Ling, Liting Lin, Fan Yang, Peng Chu, Ge Deng, Sijia Yu, Hexin
  Bai, Yong Xu, and Chunyuan Liao.
\newblock {{LaSOT}}: {{A High}}-{{Quality Benchmark}} for {{Large}}-{{Scale
  Single Object Tracking}}.
\newblock In {\em 2019 {{IEEE}}/{{CVF Conference}} on {{Computer Vision}} and
  {{Pattern Recognition}} ({{CVPR}})}, pages 5369--5378, {Long Beach, CA, USA},
  June 2019. {IEEE}.

\bibitem{fan2021multiscale}
Haoqi Fan, Bo Xiong, Karttikeya Mangalam, Yanghao Li, Zhicheng Yan, Jitendra
  Malik, and Christoph Feichtenhofer.
\newblock Multiscale vision transformers.
\newblock {\em arXiv preprint arXiv:2104.11227}, 2021.

\bibitem{fan2020cn}
Yue Fan, JW Kang, LT Li, KC Li, HL Chen, ST Cheng, PY Zhang, ZY Zhou, YQ Cai,
  and Dong Wang.
\newblock {CN-CELEB: a challenging Chinese speaker recognition dataset}.
\newblock In {\em ICASSP 2020-2020 IEEE International Conference on Acoustics,
  Speech and Signal Processing (ICASSP)}, pages 7604--7608. IEEE, 2020.

\bibitem{Fang2021}
Yi Fang, Jiapeng Tang, Wang Shen, Wei Shen, Xiao Gu, Li Song, and Guangtao
  Zhai.
\newblock {Dual Attention Guided Gaze Target Detection in the Wild}.
\newblock In {\em Proceedings of the IEEE Conference on Computer Vision and
  Pattern Recognition (CVPR 21)}, 2021.

\bibitem{disney-social}
Alireza Fathi, Jessica~K. Hodgins, and James~M. Rehg.
\newblock Social interactions: A first-person perspective.
\newblock In {\em CVPR}, 2012.

\bibitem{Fathi2012social}
A. Fathi, J.~K. Hodgins, and J.~M. Rehg.
\newblock {Social interactions: A first-person perspective}.
\newblock In {\em Proceedings of the IEEE Conference on Computer Vision and
  Pattern Recognition (CVPR 12)}, pages 1226--1233. IEEE, jun 2012.

\bibitem{fathi-state-change}
A. Fathi and J. Rehg.
\newblock Modeling actions through state changes.
\newblock In {\em CVPR}, 2013.

\bibitem{fathi2013modeling}
Alireza Fathi and James~M Rehg.
\newblock Modeling actions through state changes.
\newblock In {\em CVPR}, 2013.

\bibitem{slowfast}
Christoph Feichtenhofer, Haoqi Fan, Jitendra Malik, and Kaiming He.
\newblock Slowfast networks for video recognition.
\newblock In {\em ICCV}, 2019.

\bibitem{feichtenhofer2019slowfast}
Christoph Feichtenhofer, Haoqi Fan, Jitendra Malik, and Kaiming He.
\newblock Slowfast networks for video recognition.
\newblock In {\em Proceedings of the IEEE/CVF international conference on
  computer vision}, pages 6202--6211, 2019.

\bibitem{nist-sclite}
Jonathan Fiscus.
\newblock {NIST} sclite sscoring toolkit.
\newblock \url{https://github.com/usnistgov/SCTK}.

\bibitem{fu2019datasets}
Jianglin Fu, Ivan~V Baji{\'c}, and Rodney~G Vaughan.
\newblock Datasets for face and object detection in fisheye images.
\newblock {\em Data in brief}, 27:104752, 2019.

\bibitem{furnari2017next}
Antonino Furnari, Sebastiano Battiato, Kristen Grauman, and Giovanni~Maria
  Farinella.
\newblock Next-active-object prediction from egocentric videos.
\newblock {\em Journal of Visual Communication and Image Representation},
  49:401--411, 2017.

\bibitem{furnari2020rolling}
Antonino Furnari and Giovanni Farinella.
\newblock Rolling-unrolling lstms for action anticipation from first-person
  video.
\newblock {\em IEEE Transactions on Pattern Analysis and Machine Intelligence},
  2020.

\bibitem{furnari2019rulstm}
Antonino Furnari and Giovanni~Maria Farinella.
\newblock What would you expect? anticipating egocentric actions with
  rolling-unrolling lstms and modality attention.
\newblock In {\em International Conference on Computer Vision}, 2019.

\bibitem{gao2017red}
Jiyang Gao, Zhenheng Yang, and Ram Nevatia.
\newblock Red: Reinforced encoder-decoder networks for action anticipation.
\newblock {\em BMVC}, 2017.

\bibitem{gao-2018-separation}
R. Gao, R. Feris, and K. Grauman.
\newblock Learning to separate object sounds by watching unlabeled video.
\newblock In {\em ECCV}, 2018.

\bibitem{gao2018objectSounds}
Ruohan Gao, Rogerio Feris, and Kristen Grauman.
\newblock Learning to separate object sounds by watching unlabeled video.
\newblock In {\em ECCV}, 2018.

\bibitem{gao2019visualsound}
Ruohan Gao and Kristen Grauman.
\newblock 2.5d visual sound.
\newblock In {\em CVPR}, 2019.

\bibitem{gao2019coseparation}
Ruohan Gao and Kristen Grauman.
\newblock Co-separating sounds of visual objects.
\newblock In {\em ICCV}, 2019.

\bibitem{visual-voice}
R. Gao and K. Grauman.
\newblock Visual{V}oice: Audio-visual speech separation with cross-modal
  consistency.
\newblock In {\em CVPR}, 2021.

\bibitem{avdiar}
I. Gebru, S. Ba, X. Li, and R. Horaud.
\newblock Audio-visual speaker diarization based on spatiotemporal bayesian
  fusion.
\newblock {\em PAMI}, 2018.

\bibitem{gebru2017a}
Israel~D. Gebru, Sil{\`e}ye Ba, Xiaofei Li, and Radu Horaud.
\newblock Audio-visual speaker diarization based on spatiotemporal bayesian
  fusion.
\newblock {\em IEEE Transactions on Pattern Analysis and Machine Intelligence},
  39, 2017.

\bibitem{georgakis2016multiview}
Georgios Georgakis, Md~Alimoor Reza, Arsalan Mousavian, Phi-Hung Le, and Jana
  Ko{\v{s}}eck{\'a}.
\newblock Multiview rgb-d dataset for object instance detection.
\newblock In {\em 2016 Fourth International Conference on 3D Vision (3DV)},
  pages 426--434. IEEE, 2016.

\bibitem{avt}
Rohit Girdhar and Kristen Grauman.
\newblock Anticipative video transformer.
\newblock In {\em ICCV}, 2021.

\bibitem{girshick2015fast}
Ross Girshick.
\newblock Fast r-cnn.
\newblock In {\em Proceedings of the IEEE international conference on computer
  vision}, pages 1440--1448, 2015.

\bibitem{gkioxari2015actiontubes}
Georgia Gkioxari and Jitendra Malik.
\newblock Finding action tubes.
\newblock In {\em 2015 {{IEEE Conference}} on {{Computer Vision}} and {{Pattern
  Recognition}} ({{CVPR}})}, pages 759--768, {Boston, MA, USA}, June 2015.
  {IEEE}.

\bibitem{gollwitzer}
P. Gollwitzer.
\newblock {\em Action phases and mind-sets, Handbook of motivation and
  cognition: Foundations of social behavior}.
\newblock 1990.

\bibitem{something:something}
Raghav Goyal, Samira~Ebrahimi Kahou, Vincent Michalski, Joanna Materzynska,
  Susanne Westphal, Heuna Kim, Valentin Haenel, Ingo Fruend, Peter Yianilos,
  Moritz Mueller-Freitag, et~al.
\newblock The" something something" video database for learning and evaluating
  visual common sense.
\newblock In {\em ICCV}, 2017.

\bibitem{bi:lstm}
Alex Graves, Santiago Fern{\'a}ndez, and J{\"u}rgen Schmidhuber.
\newblock Bidirectional lstm networks for improved phoneme classification and
  recognition.
\newblock In {\em International conference on artificial neural networks},
  pages 799--804. Springer, 2005.

\bibitem{gu2018ava}
Chunhui Gu, Chen Sun, David~A Ross, Carl Vondrick, Caroline Pantofaru, Yeqing
  Li, Sudheendra Vijayanarasimhan, George Toderici, Susanna Ricco, Rahul
  Sukthankar, et~al.
\newblock Ava: A video dataset of spatio-temporally localized atomic visual
  actions.
\newblock In {\em Proceedings of the IEEE Conference on Computer Vision and
  Pattern Recognition}, pages 6047--6056, 2018.

\bibitem{gulati2020conformer}
Anmol Gulati, James Qin, Chung-Cheng Chiu, Niki Parmar, Yu Zhang, Jiahui Yu,
  Wei Han, Shibo Wang, Zhengdong Zhang, Yonghui Wu, et~al.
\newblock Conformer: Convolution-augmented transformer for speech recognition.
\newblock {\em arXiv preprint arXiv:2005.08100}, 2020.

\bibitem{he2018maskrcnn}
Kaiming He, Georgia Gkioxari, Piotr Doll{\'a}r, and Ross Girshick.
\newblock Mask {{R}}-{{CNN}}.
\newblock {\em arXiv:1703.06870 [cs]}, Jan. 2018.

\bibitem{resnet}
Kaiming He, Xiangyu Zhang, Shaoqing Ren, and Jian Sun.
\newblock Deep residual learning for image recognition.
\newblock In {\em CVPR}, 2016.

\bibitem{he2016deep}
Kaiming He, Xiangyu Zhang, Shaoqing Ren, and Jian Sun.
\newblock Deep residual learning for image recognition.
\newblock In {\em CVPR}, 2016.

\bibitem{Heidarivincheh2018}
Farnoosh Heidarivincheh, Majid Mirmehdi, and Dima Damen.
\newblock Detecting the moment of completion: Temporal models for localising
  action completion.
\newblock In {\em BMVC}, 2018.

\bibitem{spacy}
Matthew Honnibal, Ines Montani, Sofie Van~Landeghem, and Adriane Boyd.
\newblock {spaCy: Industrial-strength Natural Language Processing in Python},
  2020.

\bibitem{huang2021got10k}
Lianghua Huang, Xin Zhao, and Kaiqi Huang.
\newblock {{GOT}}-10k: {{A Large High}}-{{Diversity Benchmark}} for {{Generic
  Object Tracking}} in the {{Wild}}.
\newblock {\em IEEE Transactions on Pattern Analysis and Machine Intelligence},
  43(5):1562--1577, May 2021.

\bibitem{hussein2019timeception}
Noureldien Hussein, Efstratios Gavves, and Arnold~WM Smeulders.
\newblock Timeception for complex action recognition.
\newblock In {\em CVPR}, 2019.

\bibitem{irie2019seeing}
Go Irie, Mirela Ostrek, Haochen Wang, Hirokazu Kameoka, Akisato Kimura,
  Takahito Kawanishi, and Kunio Kashino.
\newblock Seeing through sounds: Predicting visual semantic segmentation
  results from multichannel audio signals.
\newblock In {\em ICASSP 2019-2019 IEEE International Conference on Acoustics,
  Speech and Signal Processing (ICASSP)}, pages 3961--3964. IEEE, 2019.

\bibitem{StatesAndTransformations}
Phillip Isola, Joseph~J. Lim, and Edward~H. Adelson.
\newblock Discovering states and transformations in image collections.
\newblock In {\em CVPR}, 2015.

\bibitem{isola2015discovering}
Phillip Isola, Joseph~J Lim, and Edward~H Adelson.
\newblock Discovering states and transformations in image collections.
\newblock In {\em CVPR}, 2015.

\bibitem{iwano2007audio}
Koji Iwano, Tomoaki Yoshinaga, Satoshi Tamura, and Sadaoki Furui.
\newblock Audio-visual speech recognition using lip information extracted from
  side-face images.
\newblock {\em EURASIP Journal on Audio, Speech, and Music Processing},
  2007:1--9, 2007.

\bibitem{perceiver}
Andrew Jaegle, Felix Gimeno, Andrew Brock, Andrew Zisserman, Oriol Vinyals, and
  Joao Carreira.
\newblock Perceiver: General perception with iterative attention.
\newblock {\em arXiv preprint arXiv:2103.03206}, 2021.

\bibitem{lemma}
Baoxiong Jia, Yixin Chen, Siyuan Huang, Yixin Zhu, and Song-Chun Zhu.
\newblock A multi-view dataset for learning multi-agent multi-task activities.
\newblock In {\em ECCV}, 2020.

\bibitem{jiang2017seeing}
Hao Jiang and Kristen Grauman.
\newblock Seeing invisible poses: Estimating 3d body pose from egocentric
  video.
\newblock In {\em CVPR}, 2017.

\bibitem{kinetics}
Will Kay, Joao Carreira, Karen Simonyan, Brian Zhang, Chloe Hillier, Sudheendra
  Vijayanarasimhan, Fabio Viola, Tim Green, Trevor Back, Paul Natsev, et~al.
\newblock The kinetics human action video dataset.
\newblock {\em arXiv preprint arXiv:1705.06950}, 2017.

\bibitem{kay2017kinetics}
Will Kay, Joao Carreira, Karen Simonyan, Brian Zhang, Chloe Hillier, Sudheendra
  Vijayanarasimhan, Fabio Viola, Tim Green, Trevor Back, Paul Natsev, et~al.
\newblock The kinetics human action video dataset.
\newblock {\em arXiv preprint arXiv:1705.06950}, 2017.

\bibitem{kazakos2019epic}
Evangelos Kazakos, Arsha Nagrani, Andrew Zisserman, and Dima Damen.
\newblock Epic-fusion: Audio-visual temporal binding for egocentric action
  recognition.
\newblock In {\em Proceedings of the IEEE International Conference on Computer
  Vision}, pages 5492--5501, 2019.

\bibitem{Kellnhofer2019}
Petr Kellnhofer, Simon Stent, Wojciech Matusik, and Antonio Torralba.
\newblock {Gaze360: Physically Unconstrained Gaze Estimation in the Wild}.
\newblock In {\em Proceedings of the IEEE International Conference on Computer
  Vision (ICCV 19)}, 2019.

\bibitem{kim2017joint}
Suyoun Kim, Takaaki Hori, and Shinji Watanabe.
\newblock Joint ctc-attention based end-to-end speech recognition using
  multi-task learning.
\newblock In {\em 2017 IEEE international conference on acoustics, speech and
  signal processing (ICASSP)}, pages 4835--4839. IEEE, 2017.

\bibitem{kitani:2012}
Kris~M. Kitani, Brian Ziebart, James~D. Bagnell, and Martial Hebert.
\newblock Activity forecasting.
\newblock In {\em ECCV}, 2012.

\bibitem{klakow2002testing}
Dietrich Klakow and Jochen Peters.
\newblock Testing the correlation of word error rate and perplexity.
\newblock {\em Speech Communication}, 38(1-2):19--28, 2002.

\bibitem{Knapp2014}
Mark~L. Knapp, Judith~A. Hall, and Terrence~G. Horgan.
\newblock {\em {Nonverbal Communication in Human Interaction}}.
\newblock Wadsworth Cengage Learning, 8th edition, 2014.

\bibitem{knepper2013ikeabot}
Ross~A Knepper, Todd Layton, John Romanishin, and Daniela Rus.
\newblock Ikeabot: An autonomous multi-robot coordinated furniture assembly
  system.
\newblock In {\em 2013 IEEE International conference on robotics and
  automation}, pages 855--862. IEEE, 2013.

\bibitem{kolarik2016auditory}
Andrew~J Kolarik, Brian~CJ Moore, Pavel Zahorik, Silvia Cirstea, and Shahina
  Pardhan.
\newblock Auditory distance perception in humans: a review of cues,
  development, neuronal bases, and effects of sensory loss.
\newblock {\em Attention, Perception, \& Psychophysics}, 78(2):373--395, 2016.

\bibitem{koppula2016anticipating}
Hema~S. Koppula and Ashutosh Saxena.
\newblock Anticipating human activities using object affordances for reactive
  robotic response.
\newblock {\em Pattern Analysis and Machine Intelligence}, 38(1):14--29, 2016.

\bibitem{krishna2017dense}
Ranjay Krishna, Kenji Hata, Frederic Ren, Li Fei-Fei, and Juan~Carlos Niebles.
\newblock Dense-captioning events in videos.
\newblock In {\em International Conference on Computer Vision (ICCV)}, 2017.

\bibitem{krishna-wacv2016}
Alexei A.~Efros Krishna Kumar~Singh, Kayvon~Fatahalian.
\newblock Krishnacam: Using a longitudinal, single-person, egocentric dataset
  for scene understanding tasks.
\newblock In {\em IEEE Winter Conference on Applications of Computer Vision
  (WACV)}, 2016.

\bibitem{kristan2020vot}
Matej Kristan, Ales Leonardis, Jiri Matas, Michael Felsberg, Roman Pflugfelder,
  Joni-Kristian Kamarainen, Luka~{\v C}ehovin Zajc, Martin Danelljan, Alan
  Lukezic, Ondrej Drbohlav, Linbo He, Yushan Zhang, Song Yan, Jinyu Yang,
  Gustavo Fernandez, and et {al.}
\newblock The eighth visual object tracking {{VOT2020}} challenge results,
  2020.

\bibitem{kudo2018sentencepiece}
Taku Kudo and John Richardson.
\newblock Sentencepiece: A simple and language independent subword tokenizer
  and detokenizer for neural text processing.
\newblock {\em arXiv preprint arXiv:1808.06226}, 2018.

\bibitem{kuznetsova2020open}
Alina Kuznetsova, Hassan Rom, Neil Alldrin, Jasper Uijlings, Ivan Krasin, Jordi
  Pont-Tuset, Shahab Kamali, Stefan Popov, Matteo Malloci, Alexander
  Kolesnikov, et~al.
\newblock The open images dataset v4.
\newblock {\em International Journal of Computer Vision}, 128(7):1956--1981,
  2020.

\bibitem{cmu-kitchens}
F.~De la Torre, J. Hodgins, J. Montano, S. Valcarcel, R. Forcada, and J. Macey.
\newblock Guide to the carnegie mellon university multimodal activity
  (cmu-mmac) database.
\newblock In {\em Tech. report CMU-RI-TR-08-22, Robotics Institute, Carnegie
  Mellon University}, 2009.

\bibitem{lacheze2009audio}
Loic Lacheze, Yan Guo, Ryad Benosman, Bruno Gas, and Charlie Couverture.
\newblock Audio/video fusion for objects recognition.
\newblock In {\em 2009 IEEE/RSJ International Conference on Intelligent Robots
  and Systems}, pages 652--657. IEEE, 2009.

\bibitem{lai2014unsupervised}
Kevin Lai, Liefeng Bo, and Dieter Fox.
\newblock Unsupervised feature learning for 3d scene labeling.
\newblock In {\em 2014 IEEE International Conference on Robotics and Automation
  (ICRA)}, pages 3050--3057. IEEE, 2014.

\bibitem{lan2014hierarchical}
Tian Lan, Tsung-Chuan Chen, and Silvio Savarese.
\newblock A hierarchical representation for future action prediction.
\newblock In {\em ECCV}, 2014.

\bibitem{landini2022bayesian}
Federico Landini, J{\'a}n Profant, Mireia Diez, and Luk{\'a}{\v{s}} Burget.
\newblock Bayesian hmm clustering of x-vector sequences (vbx) in speaker
  diarization: theory, implementation and analysis on standard tasks.
\newblock {\em Computer Speech \& Language}, 71:101254, 2022.

\bibitem{lee-cvpr2012}
Y.~J. Lee, J. Ghosh, and K. Grauman.
\newblock Discovering important people and objects for egocentric video
  summarization.
\newblock In {\em CVPR}, 2012.

\bibitem{utego}
Y.~J. Lee, J. Ghosh, and K. Grauman.
\newblock Discovering important people and objects for egocentric video
  summarization.
\newblock In {\em Proceedings of the IEEE Conference on Computer Vision and
  Pattern Recognition (CVPR)}, 2012.

\bibitem{lee2015predicting}
Yong~Jae Lee and Kristen Grauman.
\newblock Predicting important objects for egocentric video summarization.
\newblock {\em IJCV}, 2015.

\bibitem{Lepri2012}
Bruno Lepri, Ramanathan Subramanian, Kyriaki Kalimeri, Jacopo Staiano, Fabio
  Pianesi, and Nicu Sebe.
\newblock {Connecting meeting behavior with extraversion-a systematic study}.
\newblock {\em IEEE Transactions on Affective Computing}, 3(4):443--455, 2012.

\bibitem{levenshtein1966binary}
Vladimir~I Levenshtein et~al.
\newblock Binary codes capable of correcting deletions, insertions, and
  reversals.
\newblock In {\em Soviet physics doklady}, 1966.

\bibitem{zoombie}
Cheng Li and Kris Kitani.
\newblock Model recommendation with virtual probes for ego-centric hand
  detection.
\newblock In {\em ICCV}, 2013.

\bibitem{Li2013}
Yin Li, Alireza Fathi, and James~M. Rehg.
\newblock {Learning to predict gaze in egocentric video}.
\newblock In {\em Proceedings of the IEEE International Conference on Computer
  Vision}, pages 3216--3223, 2013.

\bibitem{gtea}
Y. Li, M. Liu, and J. Rehg.
\newblock In the eye of beholder: Joint learning of gaze and actions in first
  person video.
\newblock In {\em ECCV}, 2018.

\bibitem{Li2021}
Yin Li, Miao Liu, and Jame Rehg.
\newblock {In the Eye of the Beholder: Gaze and Actions in First Person Video}.
\newblock {\em IEEE Transactions on Pattern Analysis and Machine Intelligence},
  2021.

\bibitem{li2018eye}
Yin Li, Miao Liu, and James~M Rehg.
\newblock In the eye of beholder: Joint learning of gaze and actions in first
  person video.
\newblock In {\em Proceedings of the European Conference on Computer Vision
  (ECCV)}, pages 619--635, 2018.

\bibitem{ego-exo}
Yanghao Li, Tushar Nagarajan, Bo Xiong, and Kristen Grauman.
\newblock Ego-exo: Transferring visual representations from third-person to
  first-person videos.
\newblock In {\em CVPR}, 2021.

\bibitem{bmn}
Tianwei Lin, Xiao Liu, Xin Li, Errui Ding, and Shilei Wen.
\newblock Bmn: Boundary-matching network for temporal action proposal
  generation.
\newblock In {\em Proceedings of the IEEE/CVF International Conference on
  Computer Vision}, pages 3889--3898, 2019.

\bibitem{lin2018bsn}
Tianwei Lin, Xu Zhao, Haisheng Su, Chongjing Wang, and Ming Yang.
\newblock Bsn: Boundary sensitive network for temporal action proposal
  generation.
\newblock In {\em Proceedings of the European Conference on Computer Vision
  (ECCV)}, pages 3--19, 2018.

\bibitem{lin2017fpn}
Tsung-Yi Lin, Piotr Doll{\'a}r, Ross Girshick, Kaiming He, Bharath Hariharan,
  and Serge Belongie.
\newblock Feature {{Pyramid Networks}} for {{Object Detection}}.
\newblock {\em arXiv:1612.03144 [cs]}, Apr. 2017.

\bibitem{lin2014microsoft}
Tsung-Yi Lin, Michael Maire, Serge Belongie, James Hays, Pietro Perona, Deva
  Ramanan, Piotr Doll{\'a}r, and C~Lawrence Zitnick.
\newblock Microsoft {C}{O}{C}{O}: Common objects in context.
\newblock In {\em ECCV}, 2014.

\bibitem{liu2020forecasting}
Miao Liu, Siyu Tang, Yin Li, and James~M Rehg.
\newblock Forecasting human-object interaction: joint prediction of motor
  attention and actions in first person video.
\newblock In {\em ECCV}, 2020.

\bibitem{liu2018future}
Wen Liu, Weixin Luo, Dongze Lian, and Shenghua Gao.
\newblock Future frame prediction for anomaly detection--a new baseline.
\newblock In {\em Proceedings of the IEEE Conference on Computer Vision and
  Pattern Recognition}, pages 6536--6545, 2018.

\bibitem{lotter2016deep}
William Lotter, Gabriel Kreiman, and David Cox.
\newblock Deep predictive coding networks for video prediction and unsupervised
  learning.
\newblock {\em arXiv preprint arXiv:1605.08104}, 2016.

\bibitem{lu2015personal}
Cewu Lu, Renjie Liao, and Jiaya Jia.
\newblock Personal object discovery in first-person videos.
\newblock {\em TIP}, 2015.

\bibitem{lu2013story}
Zheng Lu and Kristen Grauman.
\newblock Story-driven summarization for egocentric video.
\newblock In {\em CVPR}, 2013.

\bibitem{mahmud2017joint}
Tahmida Mahmud, Mahmudul Hasan, and Amit~K Roy-Chowdhury.
\newblock Joint prediction of activity labels and starting times in untrimmed
  videos.
\newblock In {\em Proceedings of the IEEE International Conference on Computer
  Vision}, pages 5773--5782, 2017.

\bibitem{marin2019laeo}
Manuel~J Marin-Jimenez, Vicky Kalogeiton, Pablo Medina-Suarez, and Andrew
  Zisserman.
\newblock Laeo-net: revisiting people looking at each other in videos.
\newblock In {\em Proceedings of the IEEE Conference on Computer Vision and
  Pattern Recognition}, pages 3477--3485, 2019.

\bibitem{marin2014detecting}
Manuel~Jes{\'u}s Mar{\'\i}n-Jim{\'e}nez, Andrew Zisserman, Marcin Eichner, and
  Vittorio Ferrari.
\newblock Detecting people looking at each other in videos.
\newblock {\em International Journal of Computer Vision}, 106(3):282--296,
  2014.

\bibitem{marin2011here}
Manuel~J Mar{\'\i}n-Jim{\'e}nez, Andrew Zisserman, and Vittorio Ferrari.
\newblock Here’s looking at you, kid.
\newblock {\em Detecting people looking at each other in videos. In BMVC}, 5,
  2011.

\bibitem{mathieu2015deep}
Michael Mathieu, Camille Couprie, and Yann LeCun.
\newblock Deep multi-scale video prediction beyond mean square error.
\newblock {\em arXiv preprint arXiv:1511.05440}, 2015.

\bibitem{McCowan2005}
Iain McCowan, Jean Carletta, Wessel Kraaij, Simone Ashby, Sebastien Bourban,
  Mike Flynn, Mael Guillemot, Thomas Hain, Jaroslav Kadlec, Vasilis Karaiskos,
  Melissa Kronenthal, Guillaume Lathoud, Mike Lincoln, Agnes Lisowska, Wilfried
  Post, Dennis Reidsma, and Pierre Wellner.
\newblock {The AMI meeting corpus}.
\newblock In {\em Proceedings of Measuring Behavior 2005, the 5th International
  Conference on Methods and Techniques in Behavioral Research}, pages 137--140,
  2005.

\bibitem{Mercier_2021_WACV}
Jean-Philippe Mercier, Mathieu Garon, Philippe Giguere, and Jean-Francois
  Lalonde.
\newblock Deep template-based object instance detection.
\newblock In {\em Proceedings of the IEEE/CVF Winter Conference on Applications
  of Computer Vision (WACV)}, pages 1507--1516, January 2021.

\bibitem{micheyl2008evaluation}
Christophe Micheyl, Christian Kaernbach, and Laurent Demany.
\newblock An evaluation of psychophysical models of auditory change perception.
\newblock {\em Psychological review}, 115(4):1069, 2008.

\bibitem{miech19howto100m}
Antoine Miech, Dimitri Zhukov, Jean-Baptiste Alayrac, Makarand Tapaswi, Ivan
  Laptev, and Josef Sivic.
\newblock How{T}o100{M}: {L}earning a {T}ext-{V}ideo {E}mbedding by {W}atching
  {H}undred {M}illion {N}arrated {V}ideo {C}lips.
\newblock In {\em ICCV}, 2019.

\bibitem{composition:with:context}
Ishan Misra, Abhinav Gupta, and Martial Hebert.
\newblock From red wine to red tomato: Composition with context.
\newblock In {\em CVPR}, 2017.

\bibitem{Mitsuzumi2017DEEPEC}
Yu Mitsuzumi, Atsushi Nakazawa, and Toyoaki Nishida.
\newblock Deep eye contact detector: Robust eye contact bid detection using
  convolutional neural network.
\newblock In {\em BMVC}, 2017.

\bibitem{Moltisanti2017}
Davide Moltisanti, Michael Wray, Walterio Mayol-Cuevas, and Dima Damen.
\newblock Trespassing the boundaries: Labelling temporal bounds for object
  interactions in egocentric video.
\newblock In {\em ICCV}, 2017.

\bibitem{morgadoNIPS18}
Pedro Morgado, Nono Vasconcelos, Timothy Langlois, and Oliver Wang.
\newblock Self-supervised generation of spatial audio for 360${}^\circ$ video.
\newblock In {\em NeurIPS}, 2018.

\bibitem{muller2018trackingnet}
Matthias M{\"u}ller, Adel Bibi, Silvio Giancola, Salman Alsubaihi, and Bernard
  Ghanem.
\newblock {{TrackingNet}}: {{A Large}}-{{Scale Dataset}} and {{Benchmark}} for
  {{Object Tracking}} in the {{Wild}}.
\newblock In Vittorio Ferrari, Martial Hebert, Cristian Sminchisescu, and Yair
  Weiss, editors, {\em Computer {{Vision}} \textendash{} {{ECCV}} 2018}, volume
  11205, pages 310--327. {Springer International Publishing}, {Cham}, 2018.

\bibitem{nagarajan2018grounded}
Tushar Nagarajan, Christoph Feichtenhofer, and Kristen Grauman.
\newblock Grounded human-object interaction hotspots from video.
\newblock {\em ICCV}, 2019.

\bibitem{attributes:as:operators}
Tushar Nagarajan and Kristen Grauman.
\newblock Attributes as operators: factorizing unseen attribute-object
  compositions.
\newblock In {\em Proceedings of the European Conference on Computer Vision
  (ECCV)}, pages 169--185, 2018.

\bibitem{Nagrani17}
A. Nagrani, J.~S. Chung, and A. Zisserman.
\newblock {VoxCeleb: a large-scale speaker identification dataset}.
\newblock In {\em INTERSPEECH}, 2017.

\bibitem{ecm}
Katsuyuki Nakamura, Serena Yeung, Alexandre Alahi, and Li Fei-Fei.
\newblock Jointly learning energy expenditures and activities using egocentric
  multimodal signals.
\newblock In {\em CVPR}, 2017.

\bibitem{neumann2019future}
Lukas Neumann, Andrew Zisserman, and Andrea Vedaldi.
\newblock Future event prediction: If and when.
\newblock In {\em Proceedings of the IEEE Conference on Computer Vision and
  Pattern Recognition Workshops}, pages 0--0, 2019.

\bibitem{ng2020you2me}
Evonne Ng, Donglai Xiang, Hanbyul Joo, and Kristen Grauman.
\newblock You2me: Inferring body pose in egocentric video via first and second
  person interactions.
\newblock In {\em CVPR}, 2020.

\bibitem{nikunen2014direction}
Joonas Nikunen and Tuomas Virtanen.
\newblock Direction of arrival based spatial covariance model for blind sound
  source separation.
\newblock {\em IEEE/ACM Transactions on Audio, Speech, and Language
  Processing}, 22(3):727--739, 2014.

\bibitem{egocom}
C. Northcutt, S. Zha, S. Lovegrove, and R. Newcombe.
\newblock Egocom: A multi-person multi-modal egocentric communications dataset.
\newblock {\em PAMI}, 2020.

\bibitem{owens2018audio}
Andrew Owens and Alexei~A Efros.
\newblock Audio-visual scene analysis with self-supervised multisensory
  features.
\newblock In {\em ECCV}, 2018.

\bibitem{palmero2018automatic}
Cristina Palmero, Elsbeth~A van Dam, Sergio Escalera, Mike Kelia, Guido~F
  Lichtert, Lucas~PJJ Noldus, Andrew~J Spink, and Astrid van Wieringen.
\newblock Automatic mutual gaze detection in face-to-face dyadic interaction
  videos.
\newblock {\em Measuring Behavior 2018}, 2018.

\bibitem{park2019specaugment}
Daniel~S Park, William Chan, Yu Zhang, Chung-Cheng Chiu, Barret Zoph, Ekin~D
  Cubuk, and Quoc~V Le.
\newblock Specaugment: A simple data augmentation method for automatic speech
  recognition.
\newblock {\em arXiv preprint arXiv:1904.08779}, 2019.

\bibitem{park:2016_future}
H.~S. Park, J.-J. Hwang, Y. Niu, and J. Shi.
\newblock Egocentric future localization.
\newblock In {\em CVPR}, 2016.

\bibitem{park-cvpr2016}
H.~S. Park, J.-J. Hwang, Y. Niu, and J. Shi.
\newblock Egocentric future localization.
\newblock In {\em Conference on Computer Vision and Pattern Recognition
  (CVPR)}, 2016.

\bibitem{Park2012}
Hyun~Soo Park, Eakta Jain, and Yaser Sheikh.
\newblock {3D social saliency from head-mounted cameras}.
\newblock In {\em Advances in Neural Information Processing Systems}, volume~1,
  pages 422--430, 2012.

\bibitem{park2021review}
Tae~Jin Park, Naoyuki Kanda, Dimitrios Dimitriadis, Kyu~J Han, Shinji Watanabe,
  and Shrikanth Narayanan.
\newblock A review of speaker diarization: Recent advances with deep learning.
\newblock {\em arXiv preprint arXiv:2101.09624}, 2021.

\bibitem{perrott1990minimum}
David~R Perrott and Kourosh Saberi.
\newblock Minimum audible angle thresholds for sources varying in both
  elevation and azimuth.
\newblock {\em The Journal of the Acoustical Society of America},
  87(4):1728--1731, 1990.

\bibitem{pirsiavash2012detecting}
Hamed Pirsiavash and Deva Ramanan.
\newblock Detecting activities of daily living in first-person camera views.
\newblock In {\em 2012 IEEE conference on computer vision and pattern
  recognition}, pages 2847--2854. IEEE, 2012.

\bibitem{adl}
H. Pirsiavash and D. Ramanan.
\newblock Detecting activities of daily living in first-person camera views.
\newblock In {\em Computer Vision and Pattern Recognition (CVPR)}, 2012.

\bibitem{Povey_ASRU2011}
Daniel Povey, Arnab Ghoshal, Gilles Boulianne, Lukas Burget, Ondrej Glembek,
  Nagendra Goel, Mirko Hannemann, Petr Motlicek, Yanmin Qian, Petr Schwarz, Jan
  Silovsky, Georg Stemmer, and Karel Vesely.
\newblock The {Kaldi} speech recognition toolkit.
\newblock In {\em IEEE 2011 Workshop on Automatic Speech Recognition and
  Understanding}, 2011.

\bibitem{task:driven:modular:networks:for:compositional:learning}
Senthil Purushwalkam, Maximilian Nickel, Abhinav Gupta, and Marc'Aurelio
  Ranzato.
\newblock Task-driven modular networks for zero-shot compositional learning.
\newblock In {\em Proceedings of the IEEE International Conference on Computer
  Vision}, pages 3593--3602, 2019.

\bibitem{ragusa}
F. Ragusa, A. Furnari, S. Battiato, G. Signorello, and G.~M. Farinella.
\newblock Egocentric visitors localization in cultural sites.
\newblock {\em Journal on Computing and Cultural Heritage (JOCCH)}, 2019.

\bibitem{ragusa2020meccano}
Francesco Ragusa, Antonino Furnari, Salvatore Livatino, and Giovanni~Maria
  Farinella.
\newblock The meccano dataset: Understanding human-object interactions from
  egocentric videos in an industrial-like domain.
\newblock In {\em IEEE Winter Conference on Application of Computer Vision
  (WACV)}, 2021.

\bibitem{ranftl2021vision}
Ren{\'e} Ranftl, Alexey Bochkovskiy, and Vladlen Koltun.
\newblock Vision transformers for dense prediction.
\newblock In {\em Proceedings of the IEEE/CVF International Conference on
  Computer Vision}, pages 12179--12188, 2021.

\bibitem{recasens2015they}
Adria Recasens, Aditya Khosla, Carl Vondrick, and Antonio Torralba.
\newblock Where are they looking?
\newblock In {\em Advances in Neural Information Processing Systems}, pages
  199--207, 2015.

\bibitem{redmon2018yolov3}
Joseph Redmon and Ali Farhadi.
\newblock Yolov3: An incremental improvement.
\newblock {\em arXiv preprint arXiv:1804.02767}, 2018.

\bibitem{Rehg2014}
James~M. Rehg, Agata Rozga, Gregory~D. Abowd, and Matthew~S. Goodwin.
\newblock {Behavioral Imaging and Autism}.
\newblock {\em IEEE Pervasive Computing}, 13(2):84--87, 2014.

\bibitem{ren2015faster}
Shaoqing Ren, Kaiming He, Ross Girshick, and Jian Sun.
\newblock Faster r-cnn: Towards real-time object detection with region proposal
  networks.
\newblock In {\em NeurIPS}, 2015.

\bibitem{faster:rcnn}
Shaoqing Ren, Kaiming He, Ross Girshick, and Jian Sun.
\newblock Faster r-cnn: Towards real-time object detection with region proposal
  networks.
\newblock {\em Advances in neural information processing systems}, 28:91--99,
  2015.

\bibitem{rodin2021predicting}
Ivan Rodin, Antonino Furnari, Dimitrios Mavroedis, and Giovanni~Maria
  Farinella.
\newblock Predicting the future from first person (egocentric) vision: A
  survey.
\newblock {\em Computer Vision and Image Understanding}, 2021.

\bibitem{roth2019ava}
Joseph Roth, Sourish Chaudhuri, Ondrej Klejch, Radhika Marvin, Andrew
  Gallagher, Liat Kaver, Sharadh Ramaswamy, Arkadiusz Stopczynski, Cordelia
  Schmid, Zhonghua Xi, et~al.
\newblock Ava-activespeaker: An audio-visual dataset for active speaker
  detection.
\newblock {\em arXiv preprint arXiv:1901.01342}, 2019.

\bibitem{Roth2020}
Joseph Roth, Sourish Chaudhuri, Ondrej Klejch, Radhika Marvin, Andrew
  Gallagher, Liat Kaver, Sharadh Ramaswamy, Arkadiusz Stopczynski, Cordelia
  Schmid, Zhonghua Xi, and Caroline Pantofaru.
\newblock {Ava Active Speaker: An Audio-Visual Dataset for Active Speaker
  Detection}.
\newblock In {\em ICASSP, IEEE International Conference on Acoustics, Speech
  and Signal Processing - Proceedings}, volume 2020-May, pages 4492--4496,
  2020.

\bibitem{ryoo2013first}
M.~S. Ryoo and L. Matthies.
\newblock First-person activity recognition: What are they doing to me?
\newblock In {\em IEEE Conference on Computer Vision and Pattern Recognition
  (CVPR)}, 2013.

\bibitem{sarlin2020superglue}
Paul-Edouard Sarlin, Daniel DeTone, Tomasz Malisiewicz, and Andrew Rabinovich.
\newblock Superglue: Learning feature matching with graph neural networks.
\newblock In {\em Proceedings of the IEEE/CVF conference on computer vision and
  pattern recognition}, pages 4938--4947, 2020.

\bibitem{schonberger2016structure}
Johannes~L Schonberger and Jan-Michael Frahm.
\newblock Structure-from-motion revisited.
\newblock In {\em Proceedings of the IEEE conference on computer vision and
  pattern recognition}, pages 4104--4113, 2016.

\bibitem{Senocak_2019_PAMI}
A. {Senocak}, T.-H. {Oh}, J. {Kim}, M. {Yang}, and I.~S. {Kweon}.
\newblock Learning to localize sound sources in visual scenes: Analysis and
  applications.
\newblock {\em TPAMI}, 2019.

\bibitem{Shan20}
Dandan Shan, Jiaqi Geng, Michelle Shu, and David Fouhey.
\newblock Understanding human hands in contact at internet scale.
\newblock In {\em CVPR}, 2020.

\bibitem{100doh}
Dandan Shan, Jiaqi Geng, Michelle Shu, and David Fouhey.
\newblock Understanding human hands in contact at internet scale.
\newblock In {\em CVPR}, 2020.

\bibitem{sharma2019learning}
Mohit Sharma, Kevin Zhang, and Oliver Kroemer.
\newblock Learning semantic embedding spaces for slicing vegetables.
\newblock {\em arXiv preprint arXiv:1904.00303}, 2019.

\bibitem{charades-ego}
Gunnar~A Sigurdsson, Abhinav Gupta, Cordelia Schmid, Ali Farhadi, and Karteek
  Alahari.
\newblock Charades-ego: A large-scale dataset of paired third and first person
  videos.
\newblock {\em arXiv preprint arXiv:1804.09626}, 2018.

\bibitem{silberman2012indoor}
Nathan Silberman, Derek Hoiem, Pushmeet Kohli, and Rob Fergus.
\newblock Indoor segmentation and support inference from rgbd images.
\newblock In {\em European conference on computer vision}, pages 746--760.
  Springer, 2012.

\bibitem{SileroVAD}
{Silero Team}.
\newblock Silero vad: Pre-trained enterprise-grade voice activity detector
  ({VAD}), number detector and language classifier.
\newblock \url{https://github.com/snakers4/silero-vad}, 2021.

\bibitem{Silva2018}
Michel Silva, Washington Ramos, João Ferreira, Felipe Chamone, Mario Campos,
  and Erickson~R. Nascimento.
\newblock A weighted sparse sampling and smoothing frame transition approach
  for semantic fast-forward first-person videos.
\newblock In {\em 2018 IEEE/CVF Conference on Computer Vision and Pattern
  Recognition (CVPR)}, 2018.

\bibitem{singh2016krishnacam}
Krishna~Kumar Singh, Kayvon Fatahalian, and Alexei~A Efros.
\newblock Krishnacam: Using a longitudinal, single-person, egocentric dataset
  for scene understanding tasks.
\newblock In {\em WACV}, 2016.

\bibitem{snyder2018x}
David Snyder, Daniel Garcia-Romero, Gregory Sell, Daniel Povey, and Sanjeev
  Khudanpur.
\newblock X-vectors: Robust {DNN} embeddings for speaker recognition.
\newblock In {\em 2018 IEEE International Conference on Acoustics, Speech and
  Signal Processing (ICASSP)}, 2018.

\bibitem{ucf}
Khurram Soomro, Amir~Roshan Zamir, and Mubarak Shah.
\newblock Ucf101: A dataset of 101 human action classes from videos in the
  wild.
\newblock In {\em CRCV-TR-12-01}, 2012.

\bibitem{egocart}
Emiliano Spera, Antonino Furnari, Sebastiano Battiato, and Giovanni~Maria
  Farinella.
\newblock Egocentric shopping cart localization.
\newblock In {\em International Conference on Pattern Recognition (ICPR)},
  2018.

\bibitem{straub2019replica}
Julian Straub, Thomas Whelan, Lingni Ma, Yufan Chen, Erik Wijmans, Simon Green,
  Jakob~J Engel, Raul Mur-Artal, Carl Ren, Shobhit Verma, et~al.
\newblock The replica dataset: A digital replica of indoor spaces.
\newblock {\em arXiv preprint arXiv:1906.05797}, 2019.

\bibitem{engagement}
Yu-Chuan Su and Kristen Grauman.
\newblock Detecting engagement in egocentric video.
\newblock In {\em ECCV}, 2016.

\bibitem{tao2021someone}
Ruijie Tao, Zexu Pan, Rohan~Kumar Das, Xinyuan Qian, Mike~Zheng Shou, and
  Haizhou Li.
\newblock Is someone speaking? exploring long-term temporal features for
  audio-visual active speaker detection.
\newblock {\em arXiv preprint arXiv:2107.06592}, 2021.

\bibitem{tian2018audio}
Y. Tian, J. Shi, B. Li, Z. Duan, and C. Xu.
\newblock Audio-visual event localization in unconstrained videos.
\newblock In {\em ECCV}, 2018.

\bibitem{tulving}
E. Tulving.
\newblock Episodic and semantic memory.
\newblock In E. Tulving and W. Donaldson, editors, {\em Organization of
  memory}. Academic Press, 1972.

\bibitem{Jester}
{TwentyBN}.
\newblock {The 20BN-jester Dataset V1}.
\newblock \url{https://20bn.com/datasets/jester}.

\bibitem{van2017transformation}
Joost Van~Amersfoort, Anitha Kannan, Marc'Aurelio Ranzato, Arthur Szlam, Du
  Tran, and Soumith Chintala.
\newblock Transformation-based models of video sequences.
\newblock {\em arXiv preprint arXiv:1701.08435}, 2017.

\bibitem{transformer}
Ashish Vaswani, Noam Shazeer, Niki Parmar, Jakob Uszkoreit, Llion Jones,
  Aidan~N Gomez, {\L}ukasz Kaiser, and Illia Polosukhin.
\newblock Attention is all you need.
\newblock In {\em Advances in neural in processing systems}, pages 5998--6008,
  2017.

\bibitem{vaswani2017attention}
Ashish Vaswani, Noam Shazeer, Niki Parmar, Jakob Uszkoreit, Llion Jones,
  Aidan~N Gomez, {\L}ukasz Kaiser, and Illia Polosukhin.
\newblock Attention is all you need.
\newblock In {\em Advances in neural information processing systems}, pages
  5998--6008, 2017.

\bibitem{villegas2017decomposing}
Ruben Villegas, Jimei Yang, Seunghoon Hong, Xunyu Lin, and Honglak Lee.
\newblock Decomposing motion and content for natural video sequence prediction.
\newblock {\em arXiv preprint arXiv:1706.08033}, 2017.

\bibitem{vondrick2016anticipating}
Carl Vondrick, Hamed Pirsiavash, and Antonio Torralba.
\newblock Anticipating visual representations from unlabeled video.
\newblock In {\em CVPR}, 2016.

\bibitem{wang:generative:model:human:object:interactions}
He Wang, S{\"o}ren Pirk, Ersin Yumer, Vladimir~G Kim, Ozan Sener, Srinath
  Sridhar, and Leonidas~J Guibas.
\newblock Learning a generative model for multi-step human-object interactions
  from videos.
\newblock In {\em Eurographics}, 2019.

\bibitem{tsn}
Limin Wang, Yuanjun Xiong, Zhe Wang, Yu Qiao, Dahua Lin, Xiaoou Tang, and Luc
  Van~Gool.
\newblock Temporal segment networks: Towards good practices for deep action
  recognition.
\newblock In {\em ECCV}, 2016.

\bibitem{SiamMask}
Qiang Wang, Li Zhang, Luca Bertinetto, Weiming Hu, and Philip H.~S. Torr.
\newblock Fast online object tracking and segmentation: A unifying approach,
  2019.

\bibitem{wang2016actions}
Xiaolong Wang, Ali Farhadi, and Abhinav Gupta.
\newblock Actions\~{} transformations.
\newblock In {\em CVPR}, 2016.

\bibitem{nonlocal}
Xiaolong Wang, Ross Girshick, Abhinav Gupta, and Kaiming He.
\newblock Non-local neural networks.
\newblock In {\em CVPR}, 2018.

\bibitem{wu2019detectron2}
Yuxin Wu, Alexander Kirillov, Francisco Massa, Wan-Yen Lo, and Ross Girshick.
\newblock Detectron2.

\bibitem{audiovisual-slowfast}
Fanyi Xiao, Yong~Jae Lee, Kristen Grauman, Jitendra Malik, and Christoph
  Feichtenhofer.
\newblock Audiovisual slowfast networks for video recognition.
\newblock {\em arXiv preprint arXiv:2001.08740}, 2020.

\bibitem{xingjian2015convolutional}
SHI Xingjian, Zhourong Chen, Hao Wang, Dit-Yan Yeung, Wai-Kin Wong, and
  Wang-chun Woo.
\newblock Convolutional lstm network: A machine learning approach for
  precipitation nowcasting.
\newblock In {\em Advances in neural information processing systems}, pages
  802--810, 2015.

\bibitem{msrvtt}
Jun Xu, Tao Mei, Ting Yao, and Yong Rui.
\newblock Msr-vtt: A large video description dataset for bridging video and
  language.
\newblock IEEE International Conference on Computer Vision and Pattern
  Recognition (CVPR), June 2016.

\bibitem{xu2020g}
Mengmeng Xu, Chen Zhao, David~S Rojas, Ali Thabet, and Bernard Ghanem.
\newblock G-tad: Sub-graph localization for temporal action detection.
\newblock In {\em Proceedings of the IEEE/CVF Conference on Computer Vision and
  Pattern Recognition}, pages 10156--10165, 2020.

\bibitem{Yagi_2018_CVPR}
Takuma Yagi, Karttikeya Mangalam, Ryo Yonetani, and Yoichi Sato.
\newblock Future person localization in first-person videos.
\newblock In {\em The IEEE Conference on Computer Vision and Pattern
  Recognition (CVPR)}, June 2018.

\bibitem{dyadic}
Ryo Yonetani, Kris~M. Kitani, and Yoichi Sato.
\newblock Recognizing micro-actions and reactions from paired egocentric
  videos.
\newblock In {\em CVPR}, 2016.

\bibitem{yonetani2016visual}
Ryo Yonetani, Kris~M Kitani, and Yoichi Sato.
\newblock Visual motif discovery via first-person vision.
\newblock In {\em ECCV}, 2016.

\bibitem{dla}
Fisher Yu, Dequan Wang, Evan Shelhamer, and Trevor Darrell.
\newblock Deep layer aggregation.
\newblock In {\em Proceedings of the IEEE conference on computer vision and
  pattern recognition}, pages 2403--2412, 2018.

\bibitem{zhang2018audio}
Hua Zhang, Xiaochun Cao, and Rui Wang.
\newblock Audio visual attribute discovery for fine-grained object recognition.
\newblock In {\em Proceedings of the AAAI Conference on Artificial
  Intelligence}, volume~32, 2018.

\bibitem{zhang2020span}
Hao Zhang, Aixin Sun, Wei Jing, and Joey~Tianyi Zhou.
\newblock Span-based localizing network for natural language video
  localization.
\newblock In {\em Proceedings of the 58th Annual Meeting of the Association for
  Computational Linguistics}, pages 6543--6554, Online, July 2020. Association
  for Computational Linguistics.

\bibitem{2DTAN_2020_AAAI}
Songyang Zhang, Houwen Peng, Jianlong Fu, and Jiebo Luo.
\newblock Learning 2d temporal adjacent networks formoment localization with
  natural language.
\newblock In {\em AAAI}, 2020.

\bibitem{zhao2020video}
Chen Zhao, Ali~K Thabet, and Bernard Ghanem.
\newblock Video self-stitching graph network for temporal action localization.
\newblock In {\em Proceedings of the IEEE/CVF International Conference on
  Computer Vision}, pages 13658--13667, 2021.

\bibitem{zhao2018sound}
Hang Zhao, Chuang Gan, Andrew Rouditchenko, Carl Vondrick, Josh McDermott, and
  Antonio Torralba.
\newblock The sound of pixels.
\newblock In {\em ECCV}, 2018.

\bibitem{zhou2018temporal}
Bolei Zhou, Alex Andonian, Aude Oliva, and Antonio Torralba.
\newblock Temporal relational reasoning in videos.
\newblock In {\em ECCV}, 2018.

\bibitem{zhou2021embracing}
Hao Zhou, Chongyang Zhang, Yan Luo, Yanjun Chen, and Chuanping Hu.
\newblock Embracing uncertainty: Decoupling and de-bias for robust temporal
  grounding.
\newblock In {\em Proceedings of the IEEE/CVF Conference on Computer Vision and
  Pattern Recognition}, pages 8445--8454, 2021.

\bibitem{centernet:zhou}
Xingyi Zhou, Dequan Wang, and Philipp Kr{\"a}henb{\"u}hl.
\newblock Objects as points.
\newblock {\em arXiv preprint arXiv:1904.07850}, 2019.

\bibitem{zhou-state-change}
Y. Zhou and T. Berg.
\newblock Learning temporal transformations from time-lapse videos.
\newblock In {\em ECCV}, 2016.

\bibitem{zhou2015temporal}
Yipin Zhou and Tamara~L Berg.
\newblock Temporal perception and prediction in ego-centric video.
\newblock In {\em ICCV}, 2015.

\bibitem{zhou2016learning}
Yipin Zhou and Tamara~L Berg.
\newblock Learning temporal transformations from time-lapse videos.
\newblock In {\em ECCV}, 2016.

\bibitem{zhu2021deep}
Hao Zhu, Man-Di Luo, Rui Wang, Ai-Hua Zheng, and Ran He.
\newblock Deep audio-visual learning: A survey.
\newblock {\em International Journal of Automation and Computing}, pages 1--26,
  2021.

\end{thebibliography}
}

\end{document}